\RequirePackage[l2tabu,orthodox]{nag} 
\documentclass[techreport,12pt]{gu-thesis}
\usepackage[swedish,english]{babel}
\usepackage{scrextend}

\usepackage[bbgreekl]{mathbbol}
\usepackage[utf8]{inputenc} 
\usepackage{microtype} 
\usepackage{textcomp}
\usepackage{hyperref,soul} 
\let\oldhref\href
\renewcommand{\href}[2]{\oldhref{#1}{\hrefstyle{#2}}}
\newcommand{\hrefstyle}[1]{\color{blue}{#1}}
\usepackage{authorindex}
\usepackage{multicol}

\usepackage{mathtools} 
\usepackage{amsmath,amssymb}
\usepackage{tikz} 
\usepackage{booktabs} 
\usepackage[adobe-utopia]{mathdesign}

\usepackage{overpic}
\def\myFigureScale{0.57}
\graphicspath{{figures/small/}{figures/jpg/}{figures/png/}}

\usepackage{exercise}

\numberwithin{Exercise}{chapter}

\usepackage{fancyhdr}
\usepackage{flipbook}
\usepackage[square,numbers,sort&compress]{natbib}
\usepackage{bibentry}
\usepackage{makeidx}
\makeindex

\usepackage{algorithm}\usepackage{algorithmic}

\newcommand{\ve}[1]{\ensuremath{\mbox{\boldmath$#1$}}}
\newcommand{\ma}[1]{\ensuremath{\mathbb{#1}}}

\newcommand{\eqnlab}[1]{\label{eqn:#1}}
\newcommand{\figlab}[1]{\label{fig:#1}}
\newcommand{\tablab}[1]{\label{tab:#1}}
\newcommand{\eqnref}[1]{(\ref{eqn:#1})}
\newcommand{\figref}[1]{\ref{fig:#1}}
\newcommand{\tabref}[1]{\ref{tab:#1}}

\newcommand{\chlab}[1]{\label{ch:#1}}
\newcommand{\seclab}[1]{\label{sec:#1}}

\newcommand{\chref}[1]{\ref{ch:#1}}
\newcommand{\secref}[1]{\ref{sec:#1}}

\newcommand{\Secref}[1]{Section~\ref{sec:#1}}

\usepackage{wrapfig}
\usepackage{units}
\usepackage{url}
\usepackage{upgreek}
\usepackage{lpic}
\usepackage{csquotes}
\usepackage{overpic}
\setquotestyle{swedish}

\DeclareFontFamily{OT1}{pzc}{}
\DeclareFontShape{OT1}{pzc}{m}{it}{<-> s * [1.10] pzcmi7t}{}
\DeclareMathAlphabet{\mathpzc}{OT1}{pzc}{m}{it}
\usepackage{calligra}
\newcommand{\pr}{{\mathpzc p}\,}
\newcommand{\dw}{\ve \varepsilon}
\newcommand{\rbf}{u}
\newcommand{\tomu}{^{(\mu)}}
\newcommand{\PB}{P_{\rm B}}
\newcommand{\Pd}{P_{\rm data}}
\newcommand{\prew}{p_{\rm reward}}
\newcommand{\sh}{h}
\newcommand{\sv}{v}

\newcommand{\basis}{\mathscr{B}}
\newcommand{\DKL}{D_{\rm KL}}
\newcommand{\Platent}{P_{\rm L}}
\newcommand{\pG}{P}
\newcommand{\avg}[1]{\langle #1\rangle}
\DeclareMathOperator*{\argmin}{arg\,min}

\title{Machine learning with neural networks}
\author{Bernhard Mehlig}
\documenttype{ }
\department{Department of Physics}
\copyrightyear{2021}

\acknowledgements{This textbook  is based on lecture notes for the course {\em Artificial Neural Networks} that I have given at  Gothenburg University and at Chalmers Technical University in Gothenburg, Sweden.
When I prepared my lectures, my main source was  {\em Introduction to the theory of neural computation} by Hertz, Krogh, and Palmer \cite{hertz1991introduction}. Other sources were  {\em Neural Networks: a comprehensive foundation} by Haykin \cite{Haykin}, Horner's lecture notes  \cite{Horner2001} from Heidelberg,  {\em Deep learning} by Goodfellow, Bengio \&  Courville \cite{Goodfellow},  and the online book {\em Neural Networks and Deep Learning} by Nielsen \cite{Nielsen}.

I thank  Martin \v{C}ejka for typesetting the first version of my hand-written lecture notes, Erik Werner and Hampus Linander for their help in preparing Chapter \ref{chapter:con_net}, Kristian Gustafsson for
his detailed feedback on Chapter \ref{ch:rl}, Nihat Ay for his comments on
Section \ref{sec:rbm}, and Mats Granath for discussions about autoencoders. I would also  like to thank Juan Diego Arango, Oleksandr Balabanov, Anshuman Dubey, Johan Fries,
Phillip Gr\"a{}fensteiner, Navid Mousavi, Marina Rafajlovic, Jan Schiffeler, Ludvig Storm, and Arvid Wenzel Wartenberg
for implementing algorithms described in this book. Many Figures are based on their results.
Oleksandr Balabanov, Anshuman Dubey, Jan Meibohm, and in particular Johan Fries and Marina Rafajlovic
contributed exam questions that became exercises in this book.
Finally, I would like to express my gratitude to Stellan \"Ostlund, for his encouragement and criticism.
Last but not least, a large number of colleagues and students -- past and present -- pointed out misprints and errors, and suggested improvements. I thank them all.

The present version does not contain exercises (copyright owned by Cambridge University Press). The complete book is available from \href{https://www.cambridge.org/gb/academic/subjects/physics/statistical-physics/machine-learning-neural-networks-introduction-scientists-and-engineers?format=HB}{Cambridge University Press}.
\vfill

\noindent The cover image shows an input pattern designed to maximise the  output of neurons corresponding to one feature map in a given convolution layer of  a deep convolutional neural network [129,130]. See also page \pageref{optimalpatterns}. Image by Hampus Linander. Reproduced with permission.
 }

\coverfigure{
\begin{overpic}[scale=1.28]{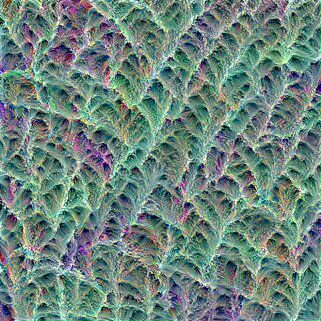}
\end{overpic}
}                 
 \covercaption{The cover image shows an input pattern designed to maximise the  output of neurons corresponding to one feature map in a given convolution layer of a deep convolutional neural network [129,130]. See also page \pageref{optimalpatterns}. Image by Hampus Linander. Reproduced with permission.}   


\usepackage{enumitem}

\begin{document}

\pagestyle{empty}
\pagenumbering{roman}

\makecoverpage
\cleardoublepage
\maketitlepage
\cleardoublepage
\makeacknowledgementspage

\cleardoublepage
\setcounter{tocdepth}{1}
\tableofcontents

\cleardoublepage
\pagenumbering{arabic}
\pagestyle{fancy}
 
\cleardoublepage
\chapter{Introduction}
\label{chap:intro}

The term {\em neural networks}  historically refers to networks of neurons in the mammalian brain.
Neurons are its fundamental units of computation, and they  are connected together in networks to process data. This can be a very complex task. The dynamics of such neural networks in response to external stimuli is therefore often  quite intricate. Inputs and outputs of each neuron vary as functions of time in the form of spike trains\index{spike train}, but also the network itself changes over time: we learn and improve our data-processing capabilities by establishing new connections between neurons. 

Neural-network algorithms for machine learning are inspired by the architecture and the dynamics of networks of neurons in the brain. The algorithms use highly idealised neuron models. Nevertheless, the fundamental principle is the same: artificial neural networks learn by changing the connections between their neurons. Such networks can perform a multitude of information-processing tasks. 

Neural networks can for instance learn to recognise structures in a 
data set and, to some extent, generalise \index{generalisation} what they have learnt. 
A {\em training set}\index{training set} contains a list of input\index{input pattern}\index{input}\index{pattern}
patterns  together with a list of corresponding labels\index{label}, or target values,\index{target}  that encode the properties of the input patterns\index{input pattern} the network is supposed to learn.  Artificial neural networks can be trained to classify such data very accurately by adjusting
the connection strengths between their neurons, and can learn to generalise the result to other data sets -- provided that the new data is not too different from the training data.
A prime example for a problem of this type is object recognition\index{object recognition} in images, for instance in the sequence of camera images taken by a self-driving car. 
Recent interest in machine learning with neural networks is driven in part by the success of  neural networks in {\em visual object recognition}.

Another task at which neural networks excel is {\em machine translation}\index{machine translation} with dynamical, or recurrent\index{recurrent network} networks. 
Such networks take sentences\index{machine translation!sentence} as inputs. As one feeds word after word, the network outputs the words in the translated sentence. Recurrent networks can be efficiently trained on large training sets of input sentences and their translations. Google translate works in this way.  Recurrent networks have also been used with considerable success to predict chaotic dynamics. These are all examples of {\em supervised} learning\index{supervised learning}, where the networks are trained to associate a certain target, or label, with each input.

Artificial neural networks are also good at analysing large sets of unlabeled, often high\--dimensional data  -- where it may be difficult to determine {\em a priori} which questions are most relevant and rewarding to ask.
{\em Unsupervised}-learning\index{unsupervised learning} algorithms 
organise the unlabeled input data in different ways. They can for instance detect familiarity\index{familiarity} and similarity (clusters)\index{cluster} of input patterns
\index{input pattern} and other structures in the input data.  Unsupervised-learning algorithms work well when there is redundancy in the input data, and they are particularly useful for large, high-dimensional data sets, where it may be a challenge  to detect clusters or other data structures by inspection.

Many problems lie between these two extremes of supervised and unsupervised learning. Consider how an agent may learn to navigate a complex environment, in order to get from one location to another as quickly as possible, or expending as little energy as possible. The method of {\em reinforcement learning} \index{reinforcement learning} allows the agent to do just that,
by optimising its behaviour in response to environmental cues in the shape  of penalties and rewards. In 
short, the agent learns to act in such a way that it receives positive feedback\index{feedback} (reward) more often than a penalty.  

\begin{figure}[t]
\centering
\begin{overpic}[scale=\myFigureScale]{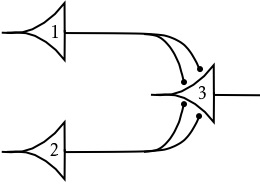}
\end{overpic}
\caption{\label{fig:MCPFig1b} Logical OR function represented by three neurons. 
Neuron 3 fires\index{neuron!firing} actively  if at least one of the neurons 1 and 2 is active\index{neuron!active}.  After Figure~1b
by McCulloch and Pitts \cite{MCP1943}.}
\end{figure}
The tools for machine learning with neural networks  were developed long ago, most of them during the second half of the last 
century. In 1943, McCulloch and Pitts \cite{MCP1943} analysed how networks of neurons can process information.
Using an abstract model for a neuron, they demonstrated how such units can be coupled together to represent logical functions (Figure \ref{fig:MCPFig1b}). Their analysis and conclusions are formulated using Carnap's logical syntax \cite{Carnap}, not in terms
of algebraic equations as we are used to today. Nevertheless, their neuron model is essentially the binary threshold unit\index{binary threshold unit}, closely related to the fundamental building block of most neural-network algorithms for machine learning to this date. In this book we therefore refer to this model as the {\em McCulloch-Pitts neuron}\index{McCulloch Pitts neuron}.
The purpose of this early research on neural networks was to explain neuro-physiological mechanisms \cite{MCP1947}. 
Perhaps the most significant advance was Hebb's learning principle, describing how neural networks learn by strengthening connections between neurons that are active simultaneously. The principle is described
in Hebb's book {\em The organization of behavior: A neuropsychological theory}  \cite{hebb1949organization}, published in 1949. 

About ten years later, research in artificial neural networks had intensified,
sparked by the influential work of Rosenblatt.
In 1958 he formulated a learning rule\index{learning rule} for the McCulloch-Pitts neuron, related to Hebb's rule, and demonstrated that the rule \index{convergence}converges to the correct solution for all problems this model can solve \cite{Rosenblatt1958}. He coined the term {\em perceptron}\index{perceptron} for layered networks\index{network!layered} of McCulloch-Pitts neurons, 
and showed that such neural networks could in principle solve tasks that a single McCulloch-Pitts neuron could not. However, there was no general 
learning rule\index{learning rule} for perceptrons at the time. The work of Minsky and Papert \cite{MinskyPapert} emphasised the geometric
aspects of learning. This allowed them to prove which kinds of problems perceptrons could solve, and which not. 
In 1969 they summarised these results in their elegant book {\em Perceptrons. An Introduction to Computational Geometry}.

Perceptrons are classifiers that output a label for each input pattern\index{input pattern}. A perceptron represents a mathematical function, an input-output mapping. A breakthrough in perceptron learning was the paper by Rumelhart {\em et al.} \cite{Rumelhart}. The authors demonstrated in 1986 that perceptrons can be trained by gradient descent.
This means that the connection strengths between the neurons are iteratively changed in small steps, to eventually minimise the output error\index{output error}. For a single McCulloch-Pitts neuron, this gives essentially Hebb's rule. The important point is that gradient descent allows to efficiently train perceptrons with many layers ({\em backpropagation} for multi-layer perceptrons)\index{backpropagation}. A second contribution of Rumelhart {\em et al.} is the idea to use local feature maps for object recognition 
with neural networks. The corresponding mathematical operation is a convolution. Therefore such architectures are now known as {\em convolutional networks}\index{convolutional network}.

The work of Hopfield popularised an entirely different approach, also based on Hebb's rule. In 1982, Hopfield analysed the properties
of a dynamical, or recurrent, network that functions as a memory \cite{Hopfield1982}: the dynamics is designed to find
stored patterns by converging to a corresponding steady state. Such {\em Hopfield networks}\index{Hopfield network}  were especially popular amongst physicists because there are close connections to the statistical physics of spin glasses that made it possible to derive a precise mathematical understanding of such artificial neural networks. Hopfield 
networks are the basis for important developments in computer science. More general recurrent networks, for example, are trained like perceptrons for language processing. Hopfield networks with hidden neurons\index{hidden neuron}, so-called Boltzmann machines\index{Boltzmann machine} \cite{Hinton_Boltzmann_3} are generative models that allow to sample from a distribution the neural network learned. The training algorithm for Boltzmann machines with many hidden layers \cite{HintonSejnowski}, 
\index{hidden layer}
published in 1986,  is one of the first algorithms for training networks with many layers,
so-called {\em deep} networks\index{deep network}.

 An important problem in behavioural psychology is to understand how we learn from experience. One hypothesis is that desirable behaviours are reinforced by positive feedback. Around the same time as researchers began to analyse perceptrons, a different line of neural-network research emerged: to find learning rules\index{learning rule}
that allow neural networks to learn by reinforcement. 
In many problems of this kind, the positive feedback or reward is not immediate but is received at some time in the future, as in a board game for example. 
Therefore it is necessary to understand 
how to estimate the future reward for a certain behaviour, and how to find strategies that optimise the future reward.  {\em Reinforcement learning}
\index{reinforcement learning} \cite{sutton1998reinforcement} is designed for this purpose. In 1995, an early application of this method demonstrated how a neural network could learn to play the board game backgammon \cite{TDGammon}.

A related research field originated from the neuro-physiological question: how do we learn to map visual or sensory stimuli to spatio-temporal patterns of neural activity in the brain? 
In 1992 Kohonen described a {\em self-organising map} \cite{kohonen1990self} 
that suggests how neurons might create meaningful geometric representations of inputs. At the same time, Kohonen's algorithm is one of the first methods for non-linear 
dimensionality reduction\index{dimensionality reduction} for large data sets.

There are many connections between neural-network algorithms
for machine learning and methods used in mathematical statistics, such as for instance Markov-chain Monte-Carlo algorithms\index{Markov chain}\index{Markov chain Monte Carlo}
and simulated-annealing methods\index{annealing, simulated}.  Certain unsupervised learning algorithms are related to principal-component analysis\index{principal component}, others to clustering algorithms such as $K$-means clustering\index{K means clustering}.  Supervised learning with deep networks is essentially regression analysis, trying to fit an input-output function to the training data.  In other words this is just function fitting -- and usually with a very large number of fitting parameters. Recent convolutional neural networks have millions of parameters. To determine so many parameters requires very large and accurate data sets.  This makes it clear that neural networks are not a {\em solution of everything}. One of the difficult problems is to understand when machine learning with neural networks is called for, and when not. Therefore we need a detailed understanding of how the algorithms work, and in particular when and how they fail.

There were some early successes of machine learning with neural networks,
but these methods were not widely used in the last century. During the past decade, by contrast, machine learning with neural networks has become increasingly successful and popular.  For many applications, neural-network based
algorithms are now regarded as the method of choice, for example for predicting how \index{protein folding}proteins fold \cite{senior2020improved}. What caused this paradigm shift? After all, the methods are essentially those developed forty or more years ago. 
A reason for the new success is perhaps that industry, in acute need of machine-learning algorithms for large data sets, has invested time and effort into generating larger and more 
accurate training sets than previously available.  
Computer hardware has improved too, so that networks with many layers containing many  neurons can now be efficiently trained, making the recent progress possible.
\begin{figure}[t]
\centering
\begin{overpic}[scale=0.8]{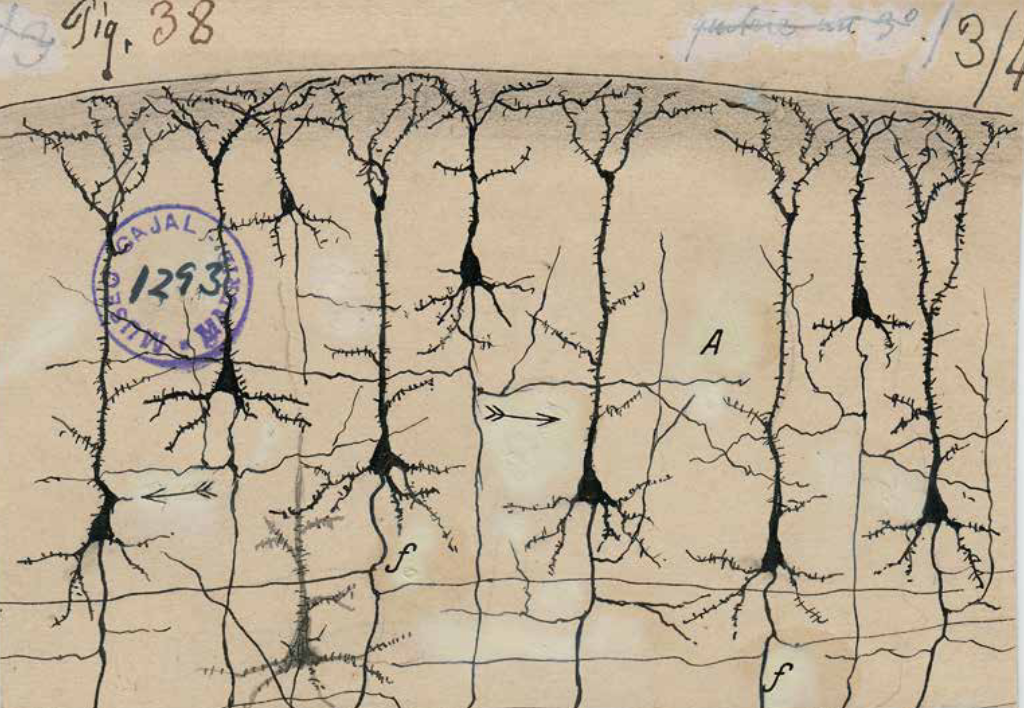}
\end{overpic}
\caption{\figlab{C1S1Neurons} Neurons in the {\em cerebral cortex}\index{cortex!cerebral}, a part of the mammalian brain. Drawing by Santiago Ram\'o{}n y Cajal,
the Spanish neuroscientist who received the Nobel Prize in Physiology and Medicine in 1906
together with Camillo Golgi \lq{}in recognition of their work on the structure of the nervous system\rq{}
\cite{nobel}.  Courtesy of the Cajal Institute, \lq{}Cajal Legacy\rq{}, Spanish National Research Council (CSIC), Madrid, Spain.}
\end{figure}

This book is based on notes for
lectures on artificial neural networks I have given for more than 15 years at
Gothenburg University and Chalmers University of Technology in Gothenburg, Sweden.
When I prepared these lectures, my primary source was {\em Introduction to the theory of neural computation} by Hertz, Krogh, and Palmer \cite{hertz1991introduction}.
The material is organised into three parts: Hopfield networks, supervised learning of labeled data, and learning for unlabeled data sets (unsupervised and reinforcement learning). One reason for starting with Hopfield networks is that there is an elegant mathematical theory that describes how these neural networks learn, making it possible to understand their strengths and weaknesses from first principles. This is not the case for most of the other algorithms discussed in this book.
The analysis of Hopfield networks sets the scene for the later parts of the book.
Part \ref{part:supervised} describes supervised learning with multilayer perceptrons and convolutional neural networks, starting from the simple geometrical picture emphasised by Minsky and Papert, and leading to the recent successes of convolutional networks in object recognition, and recurrent networks in language processing.
 Part \ref{part:unsupervised} explains what neural networks can learn about data that is not labeled, 
with particular emphasis on reinforcement learning.
 The overall goal is to explain the fundamental principles that allow neural networks to learn, emphasising ideas and concepts that are common to all three parts.

\section{Neural networks}
\index{neuron|(}
Different regions in the mammalian brain perform different tasks. 
The {\em cerebral cortex}\index{cortex!cerebral} is the outer part  of the mammalian brain,
one of its largest and most developed segments.
We can think of the cerebral cortex as a thin sheet (about 2 to 5 mm thick) that folds upon itself to form a layered structure with a large surface area that can accommodate large numbers
of nerve cells, {\em neurons}. 
The human cerebral \index{cortex!cerebral}cortex contains
about $10^{10}$ neurons. They are linked together by nerve strands
({\em axons})\index{neuron!axon} that branch and end in {\em synapses}\index{neuron!synapse}. These synapses are the connections to other neurons. The synapses connect to {\em dendrites}\index{neuron!dendrite}, branches extending from the neural \index{neuron!cell body}cell body that are designed to collect input
from other neurons in the form of electrical signals. A neuron in the human brain may have thousands of synaptic connections with other neurons. The resulting  network of connected neurons in the cerebral \index{cortex!cerebral}cortex is responsible for  processing of visual, audio, and sensory data.

Figure \figref{C1S1Neurons} shows neurons in the cerebral \index{cortex!cerebral}cortex. This drawing was made by Santiago Ram\'o{}n y Cajal more than 100 years ago.
By microscope he studied the structure of neural networks in the brain and  documented his observations by ink-on-paper drawings like
the one reproduced in Figure \figref{C1S1Neurons}.
One can distinguish the cell bodies of the neural cells, their \index{neuron!axon}axons ({\em f}), and their \index{neuron!dendrite}dendrites.
The \index{neuron!axon}axons of some neurons connect to the \index{neuron!dendrite}dendrites of other neurons, forming a neural network (see Ref.~\cite{brain} for a slightly more detailed description of this drawing).

A schematic image of a neuron is drawn in Figure \figref{C1S1Neurons1}. Information is processed from
left to right. On the left are the \index{neuron!dendrite}dendrites that receive signals and connect to the \index{neuron!cell body}cell body of the neuron where the signal is processed. The right part of the Figure shows the \index{neuron!axon}axon, through which the output is sent to \index{neuron!dendrite} other neurons. The axon connects to their dendrites via synapses.

Information is transmitted as an electrical signal. Figure \figref{C1S1SpikeTrain} shows an example of the \index{time series} time series of the electric potential for a {\em pyramidal}\index{neuron!pyramidal} neuron in fish \cite{Gabbiani1267}. The time series consists of an intermittent series of electrical-potential spikes. Quiescent periods without spikes occur when the neuron is {\em inactive}\index{neuron!inactive}, during spike-rich periods we say that the neuron is {\em active}\index{neuron!active}. 
\begin{figure}[t]
  \centering
   \begin{overpic}[scale=\myFigureScale]{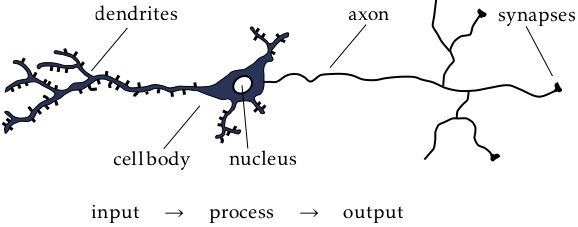}
    \end{overpic}
    \caption{\figlab{C1S1Neurons1} Schematic image of a neuron. Dendrites
\index{neuron!dendrite} receive input in the form
of electrical signals, via synapses\index{neuron!synapse}. The signals are processed in the cell body\index{neuron!cell body} of the neuron. The cell nucleus
is shown in white.  The output travels from the neural \index{neuron!cell body}cell body along the \index{neuron!axon}axon which connect via synapses \index{neuron!synapse} to other neurons.
}
\end{figure}

\section{McCulloch-Pitts neurons}
\index{McCulloch Pitts neuron|(}
\seclab{C1S2Deterministic}
In artificial neural networks, the ways in which information is processed and signals are transferred are 
highly idealised. McCulloch and Pitts \cite{MCP1943}
modelled the neuron, the computational unit of the neural network,
as  a {\em binary threshold unit}\index{binary threshold unit|textbf}.
\begin{figure}[t]
  \centering
    \begin{overpic}[scale=\myFigureScale]{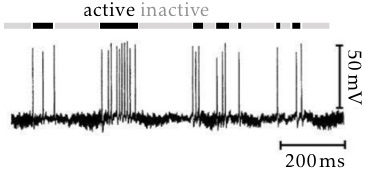}
    \end{overpic}
    \caption{\figlab{C1S1SpikeTrain} Spike train in electro-sensory pyramidal neuron of a fish. The \index{time series}time series is from Ref.~\cite{Gabbiani1267}. It is reproduced by permission of the publisher. The labels were added.}
\end{figure}
It has only two possible outputs, or {\em states}\index{neuron!state}: 
active or inactive. 
To compute the output, the unit sums the weighted inputs. If the sum exceeds a given threshold, the state of the neuron is said to be \index{state!active}active, otherwise \index{state!inactive}inactive. 
A slightly more general  model than the original one 
is illustrated in Figure \figref{C1S2MccullochPitts}.  The model performs repeated computations in discrete time steps $t=0,1,2,3,\ldots$.  The state of neuron number $j$ at time step $t$ is denoted by 
\begin{equation}
\label{eq:state_S}
    s_{j}(t) = \begin{cases} -1\quad \mbox{inactive}\,, \\ \phantom-1 \quad \mbox{active}\,.\end{cases}
\end{equation}
Given the states $s_j(t)$, neuron number $i$ computes 
\begin{equation}
\label{eq:mcp}
    s_{i}(t+1) = {\rm sgn} \big(\sum_{j=1}^N w_{ij}s_{j}(t) - \theta_{i}\big)\equiv{\rm sgn}[b_i(t)]\,.
\end{equation}
Here ${\rm sgn}(b)$ is the signum function (Figure \figref{C1S2MccullochPitts}):
\begin{equation}
\label{eq:C2S2SgnDef}
        \text{sgn}(b) = \begin{cases} -1, \quad b<0 \,,\\ +1, \quad b\geq  0 \,. \end{cases}
\end{equation}
The argument of the signum function,
\begin{equation}
\label{eq:C1S2localfield}
b_i(t)= \sum_{j=1}^N w_{ij}s_{j}(t) - \theta_{i}\,,
\end{equation}
is called the {\em local field}\index{local field|textbf}. 
\begin{figure}[t]
\centering
\begin{overpic}[scale=\myFigureScale]{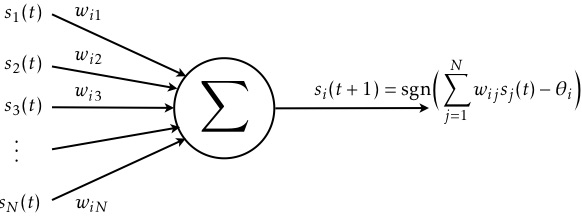}
\end{overpic}
    \caption{\figlab{C1S2MccullochPitts} Schematic diagram of a McCulloch-Pitts neuron.
The index of the neuron is $i$, it receives inputs from $N$ neurons.
The strength of the connection from  neuron $j$ to neuron $i$ is denoted by $w_{ij}$.
The {\em threshold}\index{threshold|textbf}  value for the neuron 
is denoted by $\theta_i$.  The index $t=0,1,2,3,\ldots$ labels the discrete time sequence of computation steps, and ${\rm sgn}(b)$ stands for the signum function [Figure \ref{fig:C1S2Signum} and Equation~(\ref{eq:C2S2SgnDef})].}
\end{figure}
We see that the neuron performs a weighted \index{average!weighted} average of the inputs $s_j(t)$. The parameters $w_{ij}$ are called 
{\em weights}\index{weight|textbf}. Here the first index, $i$, refers to the neuron that does the computation, and $j$ labels the  neurons that connect to neuron  $i$. 
In general weights between different pairs of neurons assume
different numerical values, reflecting different strengths of the synaptic couplings. Weights can be positive or negative,
and we say that there
is no connection when $w_{ij}=0$.

In this book we refer to the model described in Figure \ref{fig:C1S2MccullochPitts} as the {\em McCulloch-Pitts neuron}\index{McCulloch Pitts neuron}, although their original model
had additional constraints on the weights.
The \index{threshold}threshold\footnote{In the {\em deep-learning}\index{deep learning} literature \cite{Goodfellow}, 
the \index{threshold}thresholds are called {\em biases}\index{bias}, defined as 
the negative of $\theta_i$, with a plus sign in Equation (\ref{eq:C1S2localfield}).
In this book we use the convention (\ref{eq:C1S2localfield}), with the minus sign.}
for neuron $i$ is denoted by $\theta_i$.

Finally, note that the computation (\ref{eq:mcp}) is performed for all neurons $i$ in parallel, given the states $s_j(t)$ at time step $t$. The outputs $s_i(t+1)$ are the inputs to all neurons at the next time 
step. Therefore the outputs have the time argument $t+1$.
These steps are repeated many times, resulting in \index{time series}time series of the activity levels of all neurons in the network.
\begin{figure}[t]
\centering
\begin{overpic}[scale=\myFigureScale]{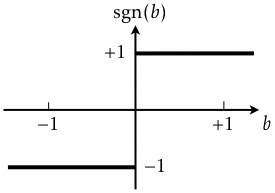}
\end{overpic}
    \caption{\figlab{C1S2Signum} Signum function [Equation (\ref{eq:C2S2SgnDef})].}
\end{figure}
\index{McCulloch Pitts neuron|)}

\section{Activation functions}
\label{sec:other:models}
The McCulloch-Pitts model\index{McCulloch Pitts neuron} approximates the patterns of spiking activity in Figure \figref{C1S1SpikeTrain} in terms of two states, $-1$ and $+1$,  
representing the inactive and active periods shown in the Figure.
For many computation tasks this is sufficient, and for our purposes it does not matter that
the dynamics of electrical signals in the \index{cortex!cerebral}cortex is quite different in detail. The aim after all is not to model the neural dynamics in the brain, but to 
construct computation models inspired by this dynamics. 

It will become apparent later that the simplest model described above must be generalised somewhat for certain tasks and questions. 
For example, the jump in the signum function at $b=0$ may cause large fluctuations in the activity levels of a network of neurons, caused
by infinitesimal changes of the local fields\index{local field} across $b=0$. To dampen this effect, one allows the neuron to respond continuously to its
inputs, replacing Eq.~(\ref{eq:mcp}) by
\begin{equation}
\label{eq:g}
        s_{i}(t+1) = g\bigg(\sum_{j} w_{ij}s_{j}(t) - \theta_{i}\bigg)\,.
    \end{equation}
Here $g(b)$ is a continuous {\em activation function}\index{activation function|textbf}. It could just be a linear function, $g(b) \propto b$. But we shall see
that many tasks require non-linear activation functions, such as $\tanh(b)$ (Figure \figref{C1S3Threshold}).  
When the activation function is continuous, the neuron states assume continuous values too, not just the discrete values $-1$ and $+1$ given in Equation (\ref{eq:state_S}). 

\begin{figure}[t]
     \centering
\begin{overpic}[scale=\myFigureScale]{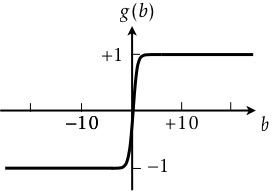}
        \end{overpic}
        \caption{\figlab{C1S3Threshold} Continuous activation function $g(b) = \tanh(b)$.}
\end{figure}
\begin{figure}[b]
        \centering
        \begin{overpic}[scale=\myFigureScale]{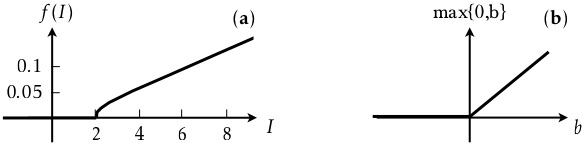}
        \end{overpic}
        \caption{  \figlab{C7S2MaxLU} ({\bf a})
Firing rate of a {\em leaky integrate-and-fire}
\index{rate!firing}\index{neuron!firing}
neuron as a function of the electrical current $I$ through the cell membrane,
Equation (\ref{eq:fI}) for $\tau=25$ and $U_{\rm c}/R=2$.
({\bf b}) Piecewise linear \index{activation function!piecewise linear}activation function, $g(b) = \mbox{max}\{0,b\}$.}
    \end{figure}
Alternatively one may use a \index{activation function!piecewise linear} piecewise linear activation function (Figure \figref{C7S2MaxLU}). This is motivated in part by 
the response curve\index{response curve} of the {\em leaky integrate-and-fire} neuron\index{neuron!leaky integrate and fire}, a model for the relation between
the electrical current $I$ through the cell membrane into the neuron cell and the membrane potential 
\index{membrane potential}
$U$. The simplest model for the dynamics of the membrane potential
represents the neuron as a capacitor. In the leaky integrate-and-fire
neuron, leakage is modelled by adding a resistor $R$ in parallel with the capacitor $C$,
so that
\begin{equation}
I = \frac{U}{R} + C \frac{{\rm d}U}{{\rm d}t}\,.
\end{equation}
For a constant current, the membrane potential grows
from zero as a function of time,
$U(t) = RI[1-\exp(-t/\tau)]$, where $\tau = RC$ is the time
constant of the model. One says that the neuron produces a {\em spike}\index{spike} when the membrane potential exceeds a critical value, $U_{\rm c}$. Immediately after, the membrane potential is set to zero (and begins to grow again).
In this model, the {\em firing rate}\index{rate!firing} $f(I)$ is thus given by $t_{\rm c}^{-1}$, where $t_{\rm c}$ is the solution
of $U(t)= U_{\rm c}$. It follows that the firing rate exhibits a threshold behaviour.
In other words,  the system works like a rectifier:
\begin{equation}
\label{eq:fI}
f(I) = \left\{
\begin{array}{ll}
0 & \mbox{for} \quad I\leq U_{\rm c}/R\,,\\
 \Big[\tau\log\Big(\frac{RI}{RI-U_{\rm c}}\Big)\Big]^{-1}
&
\mbox{for} \quad I> U_{\rm c}/R\,.
\end{array}\right .
\end{equation}
This response curve\index{response curve} is illustrated in Figure \figref{C7S2MaxLU} ({\bf a}).
The main point is that there is a \index{threshold}threshold below which the response is strictly zero.  
The response function looks qualitatively
like the piecewise linear function
\begin{equation}
\label{eq:relu}
g(b)= \mbox{max}\{0,b\}\,,
\end{equation}
 shown in panel ({\bf b}). Neurons with this activation function are called {\em rectified linear units}\index{rectified linear unit|textbf}, and the activation function (\ref{eq:relu}) is called the {\em ReLU} function\index{ReLU function|textbf}.

\section{Asynchronous updates}
Equations (\ref{eq:mcp}) and (\ref{eq:g}) are called
 {\em synchronous}\index{update rule!synchronous|textbf} update rules,
\index{update rule|textbf} 
because all neurons are updated in parallel:
at time step $t$ all inputs $s_j(t)$ are stored. Then all neurons $i$ are simultaneously updated using the stored inputs.
An alternative is to 
update only a single neuron per iteration, for example the one with index $m$:
\begin{equation}
\label{eq:s_update}
         s_{i}(t+1)= \left\{
\begin{array}{ll}
g\big(\sum_{j} w_{mj}s_{j}(t) - \theta_{m}\big) & \mbox{for $i=m$}\,,\\[2mm]
s_i(t)&\mbox{otherwise}\,.
\end{array}\right .
\end{equation}
This is called an {\em asynchronous}\index{update rule!asynchronous} update rule. 
Different schemes for choosing neurons are used in asynchronous updating. One possibility is to arrange the neurons into a two-dimensional array and to update them one by one, in a certain order.  In the
{\em typewriter scheme}\index{update rule!typewriter scheme},
for example, one updates the neurons in the top row of the array first, from left to right,
then the second row from left to right, and so forth.
 A second possibility is to choose randomly which
neuron to update.

If there are $N$ neurons, then one synchronous step corresponds to $N$ asynchronous steps, on average. This difference in time scales is not the only difference between synchronous and asynchronous updating.
The asynchronous dynamics can be shown to \index{convergence}converge to a definite state in certain cases, while the synchronous dynamics may fail to do so, 
resulting in periodic cycles that persist forever. 
\index{neuron|)}

\section{Summary}
Artificial neural networks use a highly idealised model for the fundamental computation unit: the McCulloch-Pitts neuron\index{McCulloch Pitts neuron} (Figure \figref{C1S2MccullochPitts}) is a binary threshold unit,\index{binary threshold unit} very similar to the model introduced originally by McCulloch and Pitts \cite{MCP1943}.  The units are linked together by weights $w_{ij}$,
and each unit computes a weighted average of its inputs. The network performs these computations in sequence. 
Most neural-network algorithms are built using the model described in this Chapter.

\section{Further reading}
Two accounts of the history of artificial neural networks
are especially recommended.
First, the early history of the field
is summarised in the Introduction to the second edition of {\em Perceptrons. An introduction to computational geometry} by Minsky and Papert \cite{MinskyPapert}, which came out in 1988.
This book also contains a concise bibliography 
with comments by Minsky and Papert. Second, in a short note,
Kanal \cite{Kanal} reviews the work of Rosenblatt and puts it into context.
\vfill\eject

\part{Hopfield networks}
\label{part:hopfield}

\index{Hopfield network|(}
Hopfield networks \cite{Little,Hopfield1982} are artificial neural networks that can recognise or reconstruct images,
for instance the binary images of digits shown in Figure~\ref{fig:C2S1Digits}.
The images are {\em stored}\index{pattern!stored|textbf} in the network by choosing its  weights\index{weight} $w_{ij}$ according to 
{\em Hebb's rule}~\cite{hebb1949organization}. \index{Hebb's rule} One feeds a distorted digit (Figure \figref{C2S1Image0}) to the network by assigning the initial states of its neurons to the bits of the distorted digit. The idea is that the neural-network dynamics \index{convergence}converges to the closest stored  digit. In this way the network can recognise
the input as a distorted version of the correct pattern,
it can  {\em retrieve} the correct digit.\index{pattern!retrieval} \index{pattern}
 Hopfield networks recognise patterns with many bits quite reliably, and in the past such networks were used to perform \index{pattern!recognition}pattern-recognition tasks. Today there are more efficient algorithms for this purpose (Chapter \ref{chapter:con_net}). 

Nevertheless, Hopfield networks exemplify fundamental principles of machine learning
with neural networks. For a start, most neural-network algorithms discussed in this book are built from similar building blocks and use learning rules\index{learning rule} related to Hebb's rule. 
Moreover, Hopfield networks are examples of {\em recurrent networks}\index{recurrent network}, their neurons are updated following a dynamical rule. Widely used  algorithms for machine translation and time-series prediction are based on this principle.

Furthermore, {\em restricted Boltzmann machines}\index{Boltzmann machine}
are closely related to  Hopfield networks. These
machines use hidden neurons\index{hidden neuron} to learn distributions of input patterns.
\index{input pattern}
 This makes it possible
to generate image textures and to complete partially obscured images \cite{FischerIgel}.
Generalisations of these machines, {\em deep-belief networks}\index{network!deep belief}, are examples of the first deep network architectures for machine learning. Restricted Boltzmann machines have been developed into more efficient generative models, {\em Helmholtz machines}\index{Helmholtz machine}, to sample new patterns similar to those in a given data distribution. 
The training algorithm for 
 recent generative models, {\em variational autoencoders}\index{autoencoder}, is
based on the same principles as the learning rule for Helmholtz machines.

The dynamics of Hopfield networks is closely related
to stochastic {\em Markov-chain Monte-Carlo algorithms}\index{Markov chain Monte Carlo}
used in a wide range of problems in the natural sciences.  
Finally,Hopfield networks highlight the role of stochasticity in neural-network dynamics.
A certain degree of noise, not too much, can substantially improve the performance of the Hopfield network.
In engineering problems it is usually better to avoid stochasticity, when it is due to multiplicative or additive noise that diminishes the performance of the system.  In neural-network dynamics, by contrast, stochasticity is often helpful, because it allows to explore
a wider range of configurations or actions and thus helps the dynamics to 
\index{convergence}converge to a better solution.
In general it is challenging to analyse 
the stochastic dynamics of neural networks. But for the Hopfield network much is known. 
The reason is that Hopfield networks are closely related to stochastic systems studied in statistical physics, so-called {\em spin glasses}\index{spin glass}~\cite{sherrington,sherrington1975solvable}. 
\vfill \eject

\chapter{Deterministic Hopfield networks}
\label{ch:dhn}
\begin{figure}[bt]
        \centering
        \begin{overpic}[scale=\myFigureScale]{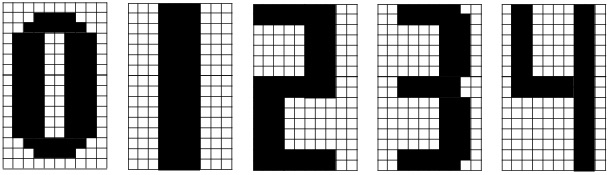}
        \end{overpic}
        \caption{\figlab{C2S1Digits} Binary representation of the digits 0 to 4. Each pattern has $10\times 16$ pixels. Adapted from Figure~14.17 in Ref.~\cite{Haykin}. The slightly peculiar shapes help the Hopfield network to distinguish
the patterns \cite{lippmann1987introduction}.
}
 \end{figure}

\section{Pattern recognition}
\label{sec:amp}
As an example for a \index{pattern!recognition} pattern-recognition task, consider $p$ images ({\em patterns})\index{pattern},  for instance the digits shown in Figure \figref{C2S1Digits}.
The different patterns are indexed by $\mu = 1,\dots,p$. Here  $p$ is the number of patterns ($p=5$ in Figure \figref{C2S1Digits}). The bits of pattern $\mu$ are denoted by $x_{i}^{(\mu)}$. The index $i=1,\ldots,N$ labels the different bits of a given pattern, and  $N$ is the number of bits per pattern ($N=160$ in Figure \figref{C2S1Digits}).  The bits are 
{\em binary}\index{bit, binary}: they can take only the values $-1$ and $+1$. 
To determine the generic properties of the algorithm, one often turns to  
{\em random patterns}\index{pattern!random|textbf}  where each bit  $x_{i}^{(\mu)}$ is chosen randomly, taking either value with probability equal to $\tfrac{1}{2}$. 
It is convenient to gather the bits of a pattern in a \index{vector!column|textbf}column vector like this
\begin{equation}
\label{eq:C2PatternVector}
    \ve x^{(\mu)} = \begin{bmatrix} 
x_1^{(\mu)} \\
x_2^{(\mu)} \\ \vdots \\ 
x_N^{(\mu)} \end{bmatrix}\,.
\end{equation}
\index{vector!notation}
In this book, vectors are written in bold math font, as in Equation (\ref{eq:C2PatternVector}).

The task of the neural network is to recognise \index{pattern!distorted|textbf}distorted patterns,
to determine for instance that the pattern on the right in  Figure \figref{C2S1Image0}
is a perturbed  version of the digit on the left in this Figure.
To this end, one {\em stores}\index{pattern!stored} $p$ patterns in the network and
presents it with a distorted version of one of these patterns, and asks the network to find the stored pattern\index{pattern!stored} that is most similar to the distorted one.

The formulation of the problem makes it necessary to define how similar 
a \index{pattern!distorted}distorted pattern $\ve x$ is to any of the stored patterns, $\ve x^{(\nu)}$ say. 
One possibility is to quantify the distance\index{distance!between patterns|textbf} $d_\nu$ between the patterns $\ve x$ and $\ve x^{(\nu)}$ in terms of the mean squared error:
\begin{align}
\label{eq:dnu}
d_\nu = \frac{1}{4N}\sum_{i=1}^{N}\big(x_i-x_{i}^{(\nu)}\big)^{2}\,.
\end{align}
The prefactor is chosen so that the distance\index{distance!between patterns} equals the fraction of bits by which  two $\pm 1 $-patterns differ.
Note that the distance\index{distance!between patterns} (\ref{eq:dnu}) does not refer to distortions by translations, rotations, or shearing. An improved measure of distance\index{distance!between patterns} might take the minimum distance between the patterns\index{distance!between patterns} subject to all possible translations, rotations, and so forth.\index{matrix!notation}
\begin{figure}[bt]
        \centering
        \begin{overpic}[scale=\myFigureScale]{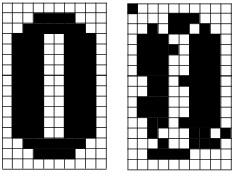}
        \end{overpic}
        \caption{\figlab{C2S1Image0} Binary image ($N=160$) of the digit 0, and a distorted version of the same image.
There are $N=160$ bits $x_i$, $i=1,\ldots,N$, and $\blacksquare$ stands for 
$x_{i}=+1$ while $\Box$ denotes $x_{i}=-1$.}
 \end{figure}

\section{Hopfield networks and Hebb's rule}
\label{sec:HN}
Hopfield networks \cite{Little,Hopfield1982} are networks of McCulloch-Pitts neurons\index{McCulloch Pitts neuron}
designed to solve the \index{pattern!recognition}pattern-recognition task described in the previous Section.
The bits of the patterns to be recognised are encoded in a particular choice of weights called Hebb's rule\index{Hebb's rule}, as explained in the following.

All possible states of the McCulloch-Pitts neurons in the        network,
\index{state!vector}
\begin{equation}
\label{eq:configuration}
\ve s =   \begin{bmatrix} s_1 \\ s_2 \\ \vdots \\ s_N \end{bmatrix}\,,
\end{equation}
form the {\em configuration space}\index{configuration space|textbf} of the network. The components of the states $\ve s$
are updated either with the synchronous  rule\index{update rule!synchronous} (\ref{eq:mcp}):
    \begin{equation}
\label{eq:C2S2Synch}
        s_{i}(t+1) = \text{sgn}\big[b_i(t)\big]\quad\mbox{with local field}\quad b_i(t) = \sum_{j=1}^N w_{ij}s_{j}(t) - \theta_{i}\,,
    \end{equation}
or with the asynchronous rule \index{update rule!asynchronous}
\begin{equation}
\label{eq:C2S2Asynch}
         s_{i}(t+1)= \left\{
\begin{array}{ll}
\mbox{sgn}\big[b_m(t)\big] & \mbox{for $i=m$}\,,\\[2mm]
s_i(t)&\mbox{otherwise}\,.
\end{array}\right .
\end{equation}
To recognise a \index{pattern!distorted}distorted pattern, one feeds its bits $x_i$ into the network, by assigning the initial states of the neurons to the pattern bits,
\begin{equation}
s_i(t=0)=x_i\,.
\end{equation}
The idea is to choose a set of weights\index{weight} $w_{ij}$ so that
the network dynamics\index{network!dynamics|textbf} (\ref{eq:C2S2Synch}) or (\ref{eq:C2S2Asynch}) \index{convergence}converges to the correct \index{pattern!stored}stored pattern. The choice of weights must depend on all $p$ patterns, $\ve x^{(1)},\ldots,\ve x^{(p)}$. We say that these patterns are {\em stored} \index{pattern!stored} in the network by assigning appropriate weights. 
For example, if \index{pattern!distorted}$\ve x$ is a distorted version of $\ve x^{(\nu)}$ (Figure \figref{C2S1Image0}), then we 
want the network to converge to this pattern: \index{convergence|textbf}
\index{convergence!criterion}
\begin{equation}
\label{eq:Sc}
\mbox{if}\quad
\ve s(t=0) = \ve x \approx \ve x^{(\nu)}\quad\mbox{then}\quad\ve s(t)\to \ve x^{(\nu)}\quad\mbox{as}\quad t\to \infty\,.
\end{equation}
Equation (\ref{eq:Sc}) means that the network corrects the errors in 
the distorted pattern\index{pattern!distorted} $\ve x$. If this works,
the stored pattern $\ve x^{(\nu)}$ is said to be an {\em attractor}\index{attractor|textbf} of the network dynamics.

But convergence is not guaranteed.  If the initial distortion is too large,
the network may \index{convergence}converge to another pattern. 
It may converge to some other state that bears no or little
relation to the \index{pattern!stored}stored patterns, or it may not converge at all. 
The region around pattern $\ve x^{(\nu)}$
in which all patterns $\ve x$ converge to $\ve x^{(\nu)}$ is called the {\em region of attraction} \index{attraction, region of} of  $\ve x^{(\nu)}$. 
The size of this region depends in an intricate way upon the ensemble of \index{pattern!stored}stored patterns, and there is no general convergence proof. 
\index{convergence!proof}

Therefore we ask a simpler question first: if one feeds one of the \index{pattern!undistorted}undistorted patterns, for instance $\ve x^{(\nu)}$, does the network recognise 
it as one of the stored, undistorted ones? The network
should  not make any changes to $\ve x^{(\nu)}$ because all bits are correct:
\begin{equation}
\label{eq:Sc2}
\mbox{if}\quad
\ve s(t=0) = \ve x^{(\nu)}\,\quad\mbox{then}\quad \ve s(t)= \ve x^{(\nu)}\quad\mbox{for all}\quad t=0,1,2,\ldots\,.
\end{equation}
How can we choose \index{weight} weights and \index{threshold}thresholds to ensure that
Equation (\ref{eq:Sc2}) holds? Let us consider 
a simple case first, where there is only one pattern, $\ve x^{(1)}$, to recognise. In this case a suitable choice of weights and \index{threshold}thresholds is 
 \begin{equation}
\label{eq:HR}
     w_{ij} = \frac{1}{N} x_i^{(1)} x_j^{(1)} \quad\mbox{and}\quad  \theta_{i} = 0\,.
\end{equation}
This {\em learning rule}\index{learning rule|textbf} reminds of a relation between activation patterns of neurons and their coupling, postulated by Hebb \cite{hebb1949organization} more than 70 years ago:\index{Hebb's rule}
\vspace*{3mm}

\hspace*{8mm}\begin{minipage}{12cm}
{\em When an axon of cell A is near enough to excite a cell B and repeatedly or persistently takes part in firing it, some growth process or metabolic change takes place in one or both cells such that A's efficiency, as one of the cells firing B, is increased.}
\end{minipage}
\vspace*{3mm}

\noindent This is a mechanism for learning through
establishing connections: the coupling between neurons tends to increase if they are active at the same time.
Equation \eqnref{C2S2HebbsRule} expresses an analogous principle. Together
with Equation (\ref{eq:Sc})  it says that the coupling $w_{ij}$  between two neurons is positive if they are both active ($s_i=s_j=1$); if their states differ, the coupling is negative. 
Therefore the rule \eqnref{C2S2HebbsRule} is called Hebb's rule\index{Hebb's rule|textbf}.
Hopfield networks are networks of of McCulloch-Pitts neurons\index{McCulloch Pitts neuron}  with weights determined by Hebb's rule.
\index{Hopfield network|textbf}  

Does a Hopfield network recognise the pattern $\ve x^{(1)}$ stored in this way?
To check that the rule (\ref{eq:HR}) does the trick, we feed the pattern to the network by assigning
$s_j(t=0) = x_j^{(1)}$, and evaluate Equation (\ref{eq:C2S2Synch}):
\label{proofp1}
\begin{equation}
\label{eq:p1}
\sum_{j=1}^N w_{ij}  x_{j}^{(1)}=\frac{1}{N}\sum_{j=1}^N x_{i}^{(1)} x_{j}^{(1)}  
x_j^{(1)}=\frac{1}{N}\sum_{j=1}^N x_i^{(1)} = x_{i}^{(1)}\,.
\end{equation}
The second equality follows because $x_{j}^{(1)}$ can only take the values $\pm 1$,
so that $[x_{j}^{(1)}]^2=1$. 
Using ${\rm sgn}(x_{i}^{(1)})=x_{i}^{(1)}$, we obtain
    \begin{equation}
\label{eq:p2}
        \text{sgn}\Big(\sum_{j=1}^Nw_{ij}x_{j}^{(1)}\Big) = x_{i}^{(1)}\,.
    \end{equation}
Comparing Equation (\ref{eq:p2}) with the update rule (\ref{eq:C2S2Synch}) shows that the bits $x_j^{(1)}$ of the pattern $\ve x^{(1)}$
remain unchanged under the update, as required by Equation~(\ref{eq:Sc2}). In other words, the network recognises the pattern as a stored one, so Equation (\ref{eq:HR}) does what we asked for. Note that
we obtain the same result if we leave out the factor of $N^{-1}$ in Equation  (\ref{eq:HR}).

But does the network correct errors?
Is the pattern $\ve x^{(1)}$ an attractor\index{attractor} [Eq.~(\ref{eq:Sc})]? This question cannot be answered in general. Yet in practice Hopfield networks work often quite well. It is a fundamental insight that neural networks may perform  well although no proof exists that their dynamics converges to the correct solution. 

\begin{figure}[bt]
    \centering
    \begin{overpic}[scale=\myFigureScale]{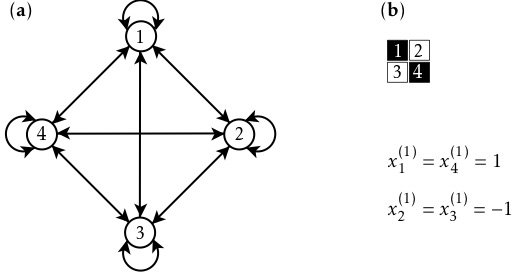}
        \end{overpic}
    \caption{\figlab{C2S2Example} Hopfield network with $N=4$ neurons.  
({\bf a}) Network layout. Neuron $i$ is represented as 
\textcircled{$i$}.
Arrows indicate symmetric connections. ({\bf b}) Pattern $\ve x^{(1)}$.}
\end{figure}
To illustrate the difficulties, consider an example, a Hopfield network with $N=4$ neurons (Figure \figref{C2S2Example}).
Store the pattern $\ve x^{(1)}$ shown in Figure \figref{C2S2Example} by assigning the weights $w_{ij}$ using Equation (\ref{eq:HR}). Now consider  a \index{pattern!distorted}distorted pattern $\ve x$ that has a non-zero \index{distance!between patterns} distance to $\ve x^{(1)}$ [Figure \figref{C2S2Reconstruction}({\bf a})]:
\begin{equation}
        d_{1} = \frac{1}{16} \sum_{i=1}^4\left(x_{i}- x_{i}^{(1)}\right)^{2}=\frac{1}{4}\,.
 \end{equation}
To feed the pattern to the network, we  set $s_j(t=0)=x_j$. Then we iterate the dynamics
using the synchronous rule\index{update rule!synchronous} (\ref{eq:C2S2Synch}).
Results for different \index{pattern!distorted}distorted patterns are shown in Figure \figref{C2S2Reconstruction}. 
We see that the first two \index{pattern!distorted}distorted patterns
\index{convergence}converge to the \index{pattern!stored}stored pattern, cases ({\bf a}) and ({\bf b}). But the third \index{pattern!distorted}distorted pattern
does not [case ({\bf c})].
\begin{figure}[b]
        \centering
        \begin{overpic}[scale=\myFigureScale]{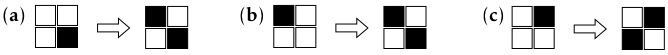}
        \end{overpic}
        \caption{\figlab{C2S2Reconstruction} Reconstruction of a \index{pattern!distorted}distorted pattern. Under synchronous updating (\ref{eq:C2S2Synch})
the first two distorted patterns ({\bf a}) and ({\bf b}) converge to the \index{pattern!stored}stored pattern $\ve x^{(1)}$, but
pattern ({\bf c}) does not.  }
\end{figure} 

To understand this behaviour, we analyse the synchronous dynamics (\ref{eq:C2S2Synch}) using the {\em weight matrix}\index{weight!matrix|textbf}
\begin{equation}
\label{eq:W}
    \ma W = \frac{1}{N} \ve x^{(1)}\ve x^{(1)\textsf T}\,.
\end{equation}
Here $\ve x^{(1)\textsf T}$ denotes the {\em transpose}\index{vector!transpose} of the column vector $\ve x^{(1)}$,
\index{vector!row}\index{vector!column}
so that $\ve x^{(1)\textsf T}$ is a row vector. 
The standard rules for matrix multiplication\index{matrix!multiplication} apply also to column and row vectors, they are just $N\times 1$ and $1\times N$ matrices. 
So the product on the r.h.s. of Equation (\ref{eq:W}) is an $N\times N$ matrix.  In the following, matrices with elements
$A_{ij}$ or $B_{ij}$ are written as $\ma A$, $\ma B$, and so forth.
The product in Equation (\ref{eq:W}) is also referred to
as an {\em outer product}\index{product!outer}.
If we change the order of $\ve x^{(1)}$ and $\ve x^{(1)\textsf T}$ in the product,
we obtain a number instead:
\begin{equation}
\label{eq:sp}
\ve x^{(1)\textsf T} \ve x^{(1)} = \sum_{j=1}^N [x_j^{(1)}]^2=N\,.
\end{equation}
The product (\ref{eq:sp}) is called {\em scalar product}\index{scalar product|textbf}. It is also denoted by 
$ \ve a \cdot  \ve b  = \ve a^{\textsf T} \ve b$ and equals $|\ve a| |\ve b|\cos\varphi$, where $\varphi$
is the angle between the vectors $\ve a$ and $\ve b$, and $|\ve a|$ is the magnitude of $\ve a$. 
We use the same notation for multiplying a transposed vector
with a matrix from the left: $ \ve x \cdot  \ma A  = \ve x^{\textsf T} \ma A$.
\begin{figure}[t]
\centering
\begin{overpic}[scale=0.3]{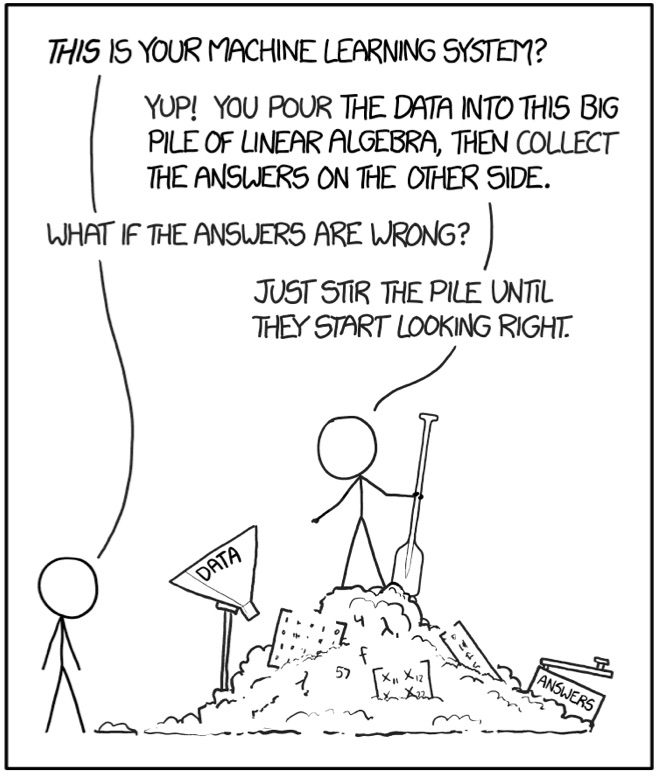}
\end{overpic}
\caption{\figlab{C13XKCD}
Reproduced from \href{https://xkcd.com/1838}{xkcd.com/1838} under the creative commons attribution-noncommercial 2.5 license.}
\end{figure}
An excellent source for those not familiar with these terms from Linear Algebra (Figure \figref{C13XKCD}) is Chapter~6 of {\em Mathematical methods of physics}  by Mathews and Walker \cite{MathewsWalker}.

Using Equation (\ref{eq:sp}) we see that $\ma W$ {\em projects}\index{matrix!projection} onto the vector $\ve x^{(1)}$,
\begin{equation}
\label{eq:W1}
    \ma W \ve x^{(1)} = \ve x^{(1)}\,.
\end{equation}
In addition, it follows from Equation (\ref{eq:sp}) that the matrix (\ref{eq:W}) is {\em idempotent}:\index{matrix!idempotent|textbf}
\begin{equation}
\label{eq:W2}
    \ma W^{t} = \ma W \quad \text{for} \quad t = 1,2,3,\dots\,.
\end{equation}
Equations (\ref{eq:W1}) and (\ref{eq:W2}) together 
with ${\rm sgn}(x_i^{(1)}) = x_i^{(1)} $ imply that the network recognises the pattern $\ve x^{(1)}$ as the stored one.
This example illustrates the general proof, 
Equations (\ref{eq:p1}) and (\ref{eq:p2}).

Now consider the \index{pattern!distorted}distorted pattern ({\bf a}) in Figure \figref{C2S2Reconstruction}. 
We feed this pattern to the network by assigning
\begin{equation}
    \ve s(t=0) = \begin{bmatrix} -1 \\-1 \\ -1 \\ 1 \end{bmatrix}\,.
\end{equation}
To compute one step in the synchronous dynamics (\ref{eq:C2S2Synch}), 
we apply $\ma W$ to $\ve s(t=0)$. This is done in two steps,
using the outer-product form (\ref{eq:W}) of the weight matrix\index{weight!matrix}. We first
multiply $\ve s(t=0)$ with $ \ve x^{(1)\textsf T}$ from the left
\begin{equation}
    \ve x^{(1)\textsf T}\ve s(t=0) =  \begin{bmatrix} 1,&-1,&-1,&1 \end{bmatrix} \begin{bmatrix} -1 \\-1 \\ -1 \\ 1 \end{bmatrix} = 2 \,,
\end{equation}
and then we multiply this result with $\ve x^{(1)}$. This  gives
\begin{equation}
\ma W \ve s(t=0) = \tfrac{1}{2} \ve x^{(1)}\,.
\end{equation}
Taking the signum of the $i$-th component of the vector $\ma W \ve s(t=0)$ yields
$s_i(t=1)$:
\begin{equation}
s_{i}(t=1) =\text{sgn}\Big(\sum_{j=1}^{N}w_{ij}s_{j}(t=0)\Big) = x_{i}^{(1)}\,.
\end{equation}
We conclude that the state of the network converges 
to the \index{pattern!stored}stored pattern, in one synchronous update. Since $\ma W$ is \index{matrix!idempotent} idempotent, the network stays there: the pattern~$\ve x^{(1)}$ is an attractor\index{attractor}. Case ({\bf b}) in Figure \figref{C2S2Reconstruction} works in a similar way. 

Now look at case ({\bf c}), where the network fails to converge to the \index{pattern!stored}stored pattern.  We  feed this pattern to the network 
by assigning $\ve s(t=0)=[-1,1,-1, -1]^{\sf T}$.
For one iteration of the synchronous dynamics we evaluate
\begin{equation}
\ve x^{(1)\textsf T}\ve s(0) =  \begin{bmatrix} 1,&-1,&-1,&1 \end{bmatrix} \begin{bmatrix} -1 \\1\\-1 \\-1 \end{bmatrix} = -2\,.
\end{equation}
It follows that 
\begin{equation}
\ma W \ve s(t=0) = -\tfrac{1}{2} \ve x^{(1)}\,.
\end{equation}
Using the update rule (\ref{eq:C2S2Synch}) we find 
\begin{equation}
\ve s(t=1) = - \ve x^{(1)}\,.
\end{equation}
Equation (\ref{eq:W2}) implies that 
\begin{equation}
\ve s(t) = - \ve x^{(1)}\quad\mbox{for}\quad t\geq 1\,.
\end{equation}
Thus the network shown in Figure \figref{C2S2Example} has two attractors,
\index{attractor}  
the pattern $\ve x^{(1)} $ as well as the {\em inverted}\index{pattern!inverted|textbf}  pattern  $-\ve x^{(1)}$. As we shall see in Section \secref{C2S3EnergyFunction}, this is a general property of Hopfield networks: if  $\ve x^{(1)}$
is an attractor, then the pattern $-\ve x^{(1)}$ is an attractor too. 
In the next Section we discuss what happens when more than one patterns are stored in the Hopfield network.

\section{The cross-talk term}
\index{cross talk term|(}
\label{sec:ctt}
When there are more patterns than just one, we need to generalise 
Equation (\ref{eq:HR}). One possibility is to simply sum Equation~(\ref{eq:HR})
over the \index{pattern!stored}stored patterns \cite{Hopfield1982}:
\begin{equation}
    \eqnlab{C2S2HebbsRule}
    w_{ij} = \frac{1}{N} \sum_{\mu=1}^{p} x_i^{(\mu)} x_{j}^{(\mu)}
\quad\mbox{and}\quad\theta_i=0\,.
\end{equation}
Equation \eqnref{C2S2HebbsRule} generalises Hebb's rule\index{Hebb's rule} to $p$ patterns.  
Because of the sum over $\mu$, the relation to Hebb's learning hypothesis is less clear, but we nevertheless refer to Equation \eqnref{C2S2HebbsRule}  as Hebb's rule.  At any rate, we see that the weights are proportional to the  
second moments  of the pattern bits. It is plausible that a neural network based the rule  \eqnref{C2S2HebbsRule}
can recognise properties of the patterns $\ve x^{(\mu)}$ that are encoded in the
\index{pattern!two point correlations|textbf}two-point correlations of their bits.

Note that the weights are symmetric\index{weight!symmetric weights}, $w_{ij} = w_{ji}$.  Also,  note that
the prefactor $N^{-1}$ in Equation \eqnref{C2S2HebbsRule} is not important. 
It is chosen to simplify the large-$N$ analysis of the model (Chapter \ref{sec:shn}).
An alternative version of Hebb's rule \cite{Hopfield1982,Haykin} sets the diagonal weights to zero:
\index{weight!diagonal|textbf}
\index{Hebb's rule}
\begin{equation}
\label{eq:HRw0}
    w_{ij} = \frac{1}{N} \sum_{\mu=1}^{p} x_i^{(\mu)} x_{j}^{(\mu)}
\quad\mbox{for}\quad i\neq j\,, \quad w_{ii}=0\,,\quad \mbox{and}\quad \theta_i=0\,.
\end{equation}
In this Section we use the form (\ref{eq:HRw0}) of Hebb's rule.        
If we assign the weights in this way,  does the network recognise the \index{pattern!stored}stored patterns? 
The question is whether
\begin{equation}
    \text{sgn} \underbrace{\Big( \frac{1}{N} \sum_{j\neq i}\sum_\mu x_i^{(\mu)} x_{j}^{(\mu)} x_{j}^{(\nu)}\Big)}_{= b_{i}^{(\nu)}}  = x_{i}^{(\nu)}
    \eqnlab{C2S2Stability}
\end{equation}
holds or not.  To check this, we repeat the calculation described in the previous
Section. As a first step we evaluate the local field
\index{local field}
\begin{equation}
\label{eq:bi}
    b_{i}^{(\nu)} = \big(1-\tfrac{1}{N}\big)x_{i}^{(\nu)} +\frac{1}{N} \sum_{j\neq i}\sum_{\mu \neq \nu} 
x_i^{(\mu)} x_{j}^{(\mu)} x_{j}^{(\nu)}\,.
\end{equation}
Here we split the sum over the patterns into two contributions. The first term corresponds
to $\mu=\nu$, where $\nu$ refers to the pattern that was fed to the network, the one we want the network to recognise. 
The second term in Equation (\ref{eq:bi}) contains the sum over the remaining 
patterns.  Condition \eqnref{C2S2Stability} is satisfied if the second term in (\ref{eq:bi}) does not affect the sign of the r.h.s. of this Equation.
This second term is called {\em cross-talk} term. \index{cross talk term|textbf}

Whether adding the cross-talk term to $\ve x^{(\nu)}$ affects $\mbox{sgn}(b_i^{(\nu)})$ or not, depends on the \index{pattern!stored}stored patterns. Since the cross-talk term contains
a sum over $\mu=1,\ldots,p$, we expect that this term  does not matter if $p$ is small enough. The fewer patterns we store in the network, the more likely it is that they are all recognised.
Furthermore, by analogy with
the example described in the previous Section, it is plausible that the \index{pattern!stored}stored patterns are then also attractors, \index{attractor}
so that slightly \index{pattern!distorted}distorted patterns converge to the correct stored pattern. 

For a more quantitative analysis of the effect of the cross-talk term we store patterns
with random bits ({\em random patterns})\index{pattern!random}.
Different bits 
are assigned 
$\pm 1$ independently with equal probability:
\begin{equation}
\label{eq:prob}
    \text{Prob}(x_{i}^{(\nu)} = \pm 1) = \tfrac{1}{2}\,.    
\end{equation}
This means in particular that $\langle x_j^{(\mu)}\rangle =0$, and that the {\em covariance}\index{covariance} of different bits vanishes:
\begin{equation}
\label{eq:avg_patterns}
\langle x_i^{(\mu)} x_j^{(\nu)}\rangle = \delta_{ij} \delta_{\mu\nu}\,.
\end{equation}
Here $\langle \cdots\rangle$ denotes an average over many realisations of random patterns, and $\delta_{ij}$ is the {\em Kronecker delta}\index{Kronecker delta|textbf}, equal to unity if $i=j$ but zero otherwise. In summary, different bits are uncorrelated\index{pattern!uncorrelated}.

Given an ensemble of random patterns, what is the probability that the cross-talk term changes $\mbox{sgn}(b_i^{(\nu)})$? 
In other words, what is the probability that the network produces a wrong bit in
one asynchronous update, if all bits were initially correct? 
The magnitude of the cross-talk term does not matter when it has the same
sign as $x_{i}^{(\nu)}$. If it has a different sign, then the cross-talk term 
may matter. To determine when this is the case,
one defines
\begin{equation}
\label{eq:def_cross_talk_term}
    C_{i}^{(\nu)} \equiv - x_{i}^{(\nu)} \underbrace{\frac{1}{N}\sum_{j\neq i}\sum_{\mu \neq \nu} x_i^{(\mu)} x_{j}^{(\mu)} x_{j}^{(\nu)}}_{\text{cross-talk term}}\,.
\end{equation}
If $C_{i}^{(\nu)} < 0$ then the cross-talk term has same sign as $x_{i}^{(\nu)}$, so that the cross-talk term does not make a difference: adding this term does not change
the sign of $x_i^{(\nu)}$.
If~$0<C_{i}^{(\nu)} < 1$ it does not matter either, only when $C_{i}^{(\nu)} > 1$. 
The network produces
an error in bit $i$ of pattern $\nu$ if $C_{i}^{(\nu)}>1$ (we approximated $1-1/N \approx 1$ in Equation~(\ref{eq:bi}), assuming that $N$ is large).
\index{cross talk term|)}

\section{One-step error probability}
\index{error probability|(}
\label{sec:osep}
The one-step {\em error probability}\index{error probability}
$P_{\rm error}^{t=1}$
is defined as the probability that an error occurs in one attempt to update a bit, given that initially all bits were correct: 
\begin{equation}
\label{eq:P_error_t1_def}
    P_{\text{error}}^{t=1} = \text{Prob}(C_{i}^{(\nu)} > 1)\,.
\end{equation}
Since patterns and bits are identically distributed, $\text{Prob}(C_{i}^{(\nu)} > 1)$
does not
depend on $i$ or $\nu$. Therefore $P_{\rm error}^{t=1}$ does not carry any indices.  

How does $P_{\text{error}}^{t=1}$ depend on the parameters of the problem,
$p$ and $N$?  When both $p$ and $N$ are large, we can use the 
{\em central-limit theorem} \cite{MathewsWalker,Feller} to answer this question.  Since different bits/patterns are independent, we can think of $C_i^{(\nu)}$ as a sum of $M=(N-1)(p-1)$ independent random numbers $c_m$ that take the values $-1$ and $+1$ with equal probabilities, 
\begin{equation}
\label{eq:C_clt}
    C_{i}^{(\nu)} = -\frac{1}{N}\sum_{j\neq i}\sum_{\mu \neq \nu} x_i^{(\mu)} x_{j}^{(\mu)} x_{j}^{(\nu)} x_{i}^{(\nu)} = - \frac{1}{N} \sum_{m=1}^{M} \hspace*{-3mm}c_m 
\,.
\end{equation}
There are $M=(N-1)(p-1)$ terms in the sum on the r.h.s. because terms with $\mu=\nu$ are excluded, and also those with $j=i$ [Equation (\ref{eq:HRw0})].
If we use the rule \eqnref{C2S2HebbsRule} instead, then there is a correction
to Equation (\ref{eq:C_clt}) from the diagonal weights. For $p\ll N$ this 
\index{weight!diagonal}
correction is small.

When $p$ and $N$ are large, the sum $\sum_{m=1}^M c_m$ contains a large number of independently identically distributed random numbers with mean zero and variance unity. 
It follows from the central-limit theorem 
that $\sum_{m=1}^M c_m$ is Gaussian distributed with mean zero and variance
$M$. 

Since the central-limit theorem\index{central limit theorem|textbf} plays an important role in the analysis of neural-network algorithms, it is 
worth discussing this theorem in a little more detail. To begin with, note that the sum $\sum_{m=1}^M c_m$
equals $2k-M$, where $k$ is
the number of occurrences of $c_m=+1$ in the sum.  Choosing $c_m$ randomly to equal either $-1$ or $+1$
is called a {\em Bernoulli trial}\index{Bernoulli trial} \cite{Feller}, and 
the probability $P_{k,M}$ of drawing $k$ times $+1$ and $M-k$ times $-1$ is
given by the {\em binomial} distribution \index{binomial distribution}\cite{Feller}.
In our case the probability of $c_m=\pm 1$ equals $\tfrac{1}{2}$, so that
\begin{equation}
\label{eq:Pk}
    P_{k,M} = \binom{M}{k}\,\,\, \big(\tfrac{1}{2}\big)^{k} \big(\tfrac{1}{2}\big)^{M-k}\,.
\end{equation}
Here
$\binom{M}{k} = {M!}/[{k!\,(M-k)!}]$
denotes the number of ways in which $k$ occurrences of $+1$ can be distributed over $M$ places.

We want to show that $P_{k,M}$ approaches a Gaussian distribution for large $M$, 
with mean zero and with variance $M$. Since the variance diverges as $M\to\infty$,
it is convenient to use the variable
$z= (2k-M)/\sqrt{M}$. The central-limit theorem implies that $z$ is Gaussian with mean zero and unit variance in the limit of large $M$. To prove that this is the case, we substitute $k = \frac{M}{2}+ \frac{\sqrt{M}}{2}z$ into Equation (\ref{eq:Pk}) and take
the limit of large $M$ using Stirling's approximation
\begin{equation}
n! \approx  {\rm e}^{n\log n - n + \tfrac{1}{2}\log 2\pi n}\,.
\end{equation}
Expanding $P_{k,M}$ to leading order in $M^{-1}$
assuming that $z$ remains of order unity gives $P_{k,M} = $ $\sqrt{2/(\pi M)}\exp{(-z^2/2)}$.
Now one changes variables from $k$ to $z$. This stretches local neighbourhoods ${\rm d}k$ to ${\rm d}z$. Conservation of probability
implies that $P(z) {\rm d}z = P(k) {\rm d}k$. It follows that $P(z) = (\sqrt{M}/2) P(k)$,
so that $P(z) = (2\pi)^{-1/2}$ $\exp(-z^2/2)$. In other words, the distribution of $z$ is Gaussian with zero mean and unit variance, as
we intended to show.
\begin{figure}[bt]
        \centering
        \begin{overpic}[scale=\myFigureScale]{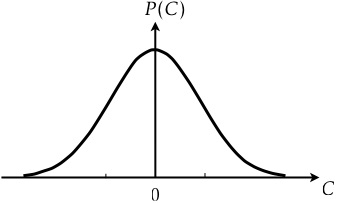}
        \end{overpic}
        \caption{ \figlab{C2S2Perror} Gaussian distribution of the quantity $C$
defined in Equation~(\ref{eq:def_cross_talk_term}).  }
\end{figure}

Returning to Equation (\ref{eq:C_clt}), we conclude that $C_i^{(\nu)}$ 
is Gaussian distributed 
\begin{equation}
P(C) = (2\pi \sigma_C^2)^{-1/2} \exp[-C^2/(2\sigma_C^2)]\,,
\end{equation}
with zero mean, as illustrated in Figure~\figref{C2S2Perror}, 
 and with variance
\begin{equation}
\label{eq:sigma_C}
\sigma^2_C = \frac{M}{N^2} \approx \frac{p}{N}\,.
\end{equation}
Here we used $M\approx Np$ for large $N$ and $p$.

Another way to compute this variance is to square Equation (\ref{eq:C_clt})
and to average over random patterns:
\begin{equation}
\sigma^2_C = \frac{1}{N^2} \Big \langle\Big(\sum_{m=1}^M c_m\Big)^2\Big\rangle
= \frac{1}{N^2} \sum_{n=1}^M\sum_{m=1}^M \langle c_n c_m\rangle \,.
\end{equation}
Here $\langle \cdots\rangle$ denotes the average over random realisations of $c_m$. 
Since the random numbers $c_m$ are independent for different indices and
because $\langle c_m^2\rangle=1$, we have that $\langle c_n c_m\rangle = \delta_{nm}$.
So only the diagonal terms in the double sum contribute, summing to $M\approx Np$. 
This yields Equation (\ref{eq:sigma_C}).

To determine $P_{\rm error}^{t=1}$ [Equation (\ref{eq:P_error_t1_def})] we must integrate the distribution 
of $C$ from $1$ to $\infty$: 
\begin{equation}
    P_{\text{error}}^{t=1} = \frac{1}{\sqrt{2\pi}\sigma_C} \int_{1}^{\infty}\hspace{-6pt}{\rm d}C\,  \,
{\rm e}^{-\frac{C^{2}}{2\sigma_C^{2}}}  = \frac{1}{2} \left[1-\text{erf}\left(\sqrt{\frac{N}{2p}}\right)\right]\,.
\eqnlab{C2S2Perror}
\end{equation}
Here ${\rm erf}$ is the {\em error function}\index{error!function}  defined as \cite{errorfunction}
\begin{equation}
\text{erf}(z) = \frac{2}{\sqrt{\pi}} \int_{0}^{z}\text{d}x\, \,{\rm e}^{-x^2}\,.
\end{equation}
Since erf$(z)$ increases monotonically as $z$ increases, we conclude that 
$P_{\text{error}}^{t=1}$ increases as $p$ increases, or as $N$ decreases.
This is expected: it is more difficult for the network to distinguish \index{pattern!stored}stored patterns when there are more of them.
On the other hand, it is easier to differentiate stored patterns if they have more bits. 
We also
see that the one-step error probability depends on $p$ and $N$ only through the combination
\begin{equation}
\alpha \equiv \frac{p}{N}\,.
\end{equation}
The parameter $\alpha$ is called the {\em storage capacity}\index{storage capacity} of the network.
Figure \figref{C2S2Perror2} shows how $P_{\text{error}}^{t=1}$ depends 
on the storage capacity.
For $\alpha      = 0.2$ for example,
the one-step error probability is slightly larger than 1\%.
\begin{figure}
\centering
\begin{overpic}[scale=\myFigureScale]{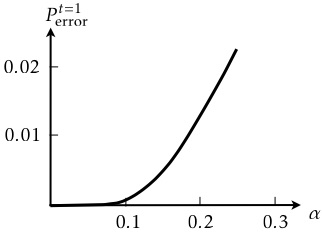}
\end{overpic}
\caption{\figlab{C2S2Perror2} Dependence of the one-step
error probability on the storage capacity $\alpha$ 
according to Equation \eqnref{C2S2Perror}.}
\end{figure}

In the derivation of Equation \eqnref{C2S2Perror} 
we assumed that the \index{pattern!stored}stored patterns are random with independent bits. Realistic patterns are not random. We nevertheless expect that $P^{t=1}_{\text{error}}$ describes the typical one-step error probability of the Hopfield network when $p$ and $N$ are large.  However, it is straightforward
to construct counter examples. Consider for example
{\em orthogonal patterns}:\index{pattern!orthogonal|textbf}
    \begin{equation}
         \ve x^{(\mu)}\cdot \ve x^{(\nu)} = 0 \quad \mbox{for}\quad
\mu \neq \nu\,.
    \end{equation}
\index{cross talk term} 
For such patterns, the cross-talk term vanishes in the limit of large $N$ (Exercise 2.2), so that $P^{t=1}_{\rm error}=0$.

More importantly, the error probability defined in this Section refers only
    to the initial update, the  first iteration.  
What happens in the next iteration, and after many iterations? Numerical experiments show that the error probability can be much higher in later iterations, because one error tends to increase the probability of making another error later on. So the estimate $P_{\rm error}^{t=1}$ is only a lower bound
for the probability of observing errors in the long run.
\index{error probability|)}

\section{Energy function} \seclab{C2S3EnergyFunction}
\index{energy function|(}
Consider the long-time limit $t\to\infty$. 
Does the Hopfield dynamics \index{convergence|textbf}converge, as required by Equation (\ref{eq:Sc})? This is an important question in the analysis of neural-network algorithms, because an algorithm that does not converge to a meaningful solution is useless. 

The standard way of analysing \index{convergence!energy function}convergence of neural-network algorithms is to define an {\em energy function} 
$H(\ve s)$ that has a minimum at the desired solution, 
$\ve s = \ve x^{(1)}$ say. 
We monitor how the energy function changes as we iterate, and keep track of the
smallest values of $H$ encountered, to find the minimum.
If we store only one pattern, $p=1$, then a suitable energy function
is
\begin{equation}
\label{eq:Halt}
    H=-\frac{1}{2N} \bigg(\sum_{i=1}^Ns_{i}x_i^{(1)}\bigg)^2 \,.
\end{equation}
This function is minimal when $\ve s = \ve x^{(1)}$, {\em i.e.}, when $s_i = x_i^{(1)}$ for all $i$. It is customary to insert the factor $1/(2N)$, this does not change the fact that $H$ is minimal at $\ve s = \ve x^{(1)}$. 

A crucial point is that the asynchronous McCulloch-Pitts dynamics (\ref{eq:C2S2Asynch})
\index{convergence}{\em converges} to a minimum of $H$ \cite{Hopfield1982}. This follows from the fact that $H$ cannot increase under the update rule (\ref{eq:C2S2Asynch}). To prove this important property,
we begin by evaluating the expression on the r.h.s. of Equation (\ref{eq:Halt}):
\begin{equation}
    H= -\frac{1}{2} \sum_{ij}^{N}\bigg(\frac{1}{N}x_i^{(1)} x_{j}^{(1)}\bigg) s_{i}s_{j}\,.
\end{equation}
Using Hebb's rule (\ref{eq:HR}) we find that 
the energy function (\ref{eq:Halt}) becomes
\begin{equation}
\label{eq:H}
H = -\frac{1}{2}\sum_{ij}w_{ij}s_{i}s_{j}\,.
\end{equation}
This function has the same form as the energy function 
(or {\em Hamiltonian}\index{Hamiltonian}) 
for certain physical models of magnetic systems consisting of interacting
spins \cite{Kadanoff}, where the interaction energy between spins $s_i$ and $s_j$ is $\tfrac{1}{2}(w_{ij}+w_{ji}) s_i s_j$. 
Note that Hebb's rule (\ref{eq:HR}) yields symmetric weights,\index{weight!symmetric weights}
$w_{ij}= w_{ji}$, and $w_{ii}>0$. Note also that setting the diagonal weights
\index{weight!diagonal}
to zero
does not change the fact that $H$ is minimal
at $\ve s = \ve x^{(1)}$ because $s_i^2=1$. The diagonal weights just give a constant contribution to $H$, independent of $\ve s$. 

The second step is to show that $H$ cannot increase under the 
asynchronous McCulloch-Pitts dynamics (\ref{eq:C2S2Asynch}). In this case we 
say that the energy function is a {\em Lyapunov function}\index{Lyapunov!function|textbf}, or {\em loss function}\index{loss function}. To demonstrate that the energy function is a Lyapunov function, choose a neuron $m$ and update it 
according to Equation (\ref{eq:C2S2Asynch}).
We denote the updated state of neuron $m$ by $s_m'$:
\begin{equation}
\label{eq:update}
s'_{m} = \text{sgn}\bigg(\sum_{j} w_{mj} s_{j} \bigg)\,.
\end{equation}
All other neurons remain unchanged. 
There are two possibilities, 
either $s'_{m}=s_{m}$ or $s'_{m}=- s_{m}$.  In the first case $H$ remains 
the same, $H'=H$. Here $H'$ refers to the value of the energy
function after the update (\ref{eq:update}).
 When $s'_{m} =-s_{m}$, by contrast, the energy function changes
by the amount
\begin{align}
    H'-H &= - \frac{1}{2} \sum_{j\neq m} 
(w_{mj}+w_{jm}) (s_{m}'s_{j} - s_{m}s_{j}) -\frac{1}{2} w_{mm} 
(s_{m}'s_{m}' - s_{m}s_{m})\nonumber\\
          &= \sum_{j\neq m} (w_{mj}+w_{jm}) s_{m}s_{j}  \,.
\end{align}
The sum goes over all neurons $j$ that are connected to the 
neuron $m$, the one to be updated in Equation (\ref{eq:update}).
Now if the weights are symmetric\index{weight!symmetric weights}, $H'-H$ equals
\begin{equation}
\label{eq:deltaH}
 H'-H = 2\sum_{j\neq m} w_{mj} s_{m}s_{j} 
=2\sum_j w_{mj} s_{m}s_{j} -2w_{mm}\,.
\end{equation}
Since the sign of $\sum_j w_{mj} s_j$ is that of $s_{m}'=- s_m$ and if  
$w_{mm}\geq 0$, it follows that
\begin{equation}
H'-H < 0\,.
\end{equation}
In other words,the value of $H$  must decrease when the state of neuron $m$ changes, $s_m'\neq s_m$.
In summary,\footnote{The derivation outlined here did not use the
specific form of Hebb's rule (\ref{eq:HR}), only that the weights are \index{weight!symmetric weights}symmetric, and that $w_{mm}{\geq}0$. However, 
the derivation fails when $w_{mm}<0$. In this case it is still true that $H$ assumes a minimum at $\ve s = \ve x^{(1)}$,
but $H$ can increase under the update rule, so that \index{convergence}convergence is not guaranteed.  We therefore require that the diagonal weights are not negative.}
 $H$ either remains constant under
the asynchronous McCulloch-Pitts dynamics ($s'_m=s_m$), or its value decreases ($s_m'\neq s_m$).
Note that this does not hold for the \index{update rule!synchronous} synchronous dynamics (\ref{eq:C2S2Synch}), see Exercise 2.9.
Since the energy function cannot increase under the asynchronous McCulloch-Pitts dynamics, 
it must converge to a minimum of the energy function. For the energy function (\ref{eq:Halt}) 
this implies that the dynamics must either converge to the stored pattern or to its inverse. \index{pattern!inverse}
Both are attractors. \index{attractor}

We assumed the \index{threshold}thresholds to vanish, but the proof also works when the thresholds are not zero, in this case for the energy function
\begin{equation}
\label{eq:H_with_thresholds}
H = -\frac{1}{2}\sum_{ij}w_{ij}s_{i}s_{j} + \sum_i \theta_i s_i
\end{equation}
in conjunction with the update rule $s_m' = {\rm sgn}(\sum_j w_{mj}s_j-\theta_m)$.
\index{update rule}

Up to now we considered only one \index{pattern!stored}stored pattern, $p=1$. If we store more than one pattern [Hebb's rule \eqnref{C2S2HebbsRule}], the proof that (\ref{eq:H})  cannot increase under the McCulloch-Pitts dynamics works in the same way
because no particular form of the weights $w_{ij}$ was assumed, only that
they must be \index{weight!symmetric weights}symmetric, and that the diagonal weights must not be negative.
\index{weight!diagonal}
Therefore it follows in this case too that
the minima of the energy function
must correspond to attractors\index{attractor}, as illustrated schematically in Figure \figref{C2S3EnergyMinima}.
\begin{figure}[tb]
        \centering
        \begin{overpic}[scale=\myFigureScale]{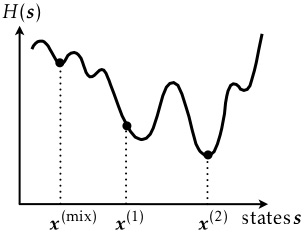}
        \end{overpic}
        \caption{\figlab{C2S3EnergyMinima} Minima of the energy function are attractors in configuration space, the space of all network states. Not all minima correspond to \index{pattern!stored}stored patterns
($\ve x^{({\rm mix})}$ is a mixed state\index{state!mixed}, see text), and stored patterns need not correspond to minima. }
\end{figure}
The configuration space\index{configuration space} of the network, corresponding to all possible choices of $\ve s= [s_1,\ldots s_N]^{\sf T}$, is drawn as a single axis, the $x$-axis. But when $N$ is large, the configuration space is really very high dimensional. 

However, some \index{pattern!stored}stored patterns may not be attractors when $p>1$. 
This follows from our analysis of the \index{cross talk term} cross-talk term in Section \ref{sec:HN} .  If the \index{cross talk term} cross-talk term causes errors for a certain stored pattern, then  this pattern is not located at a minimum of the energy function. Another way to see this is to combine 
Equations  \eqnref{C2S2HebbsRule} and (\ref{eq:H}) to give:
\begin{equation}
\label{eq:Halt2}
    H=-\frac{1}{2N} \sum_{\mu=1}^p\bigg(\sum_{i=1}^Ns_{i}x_i^{(\mu)}\bigg)^2 \,.
\end{equation}
While the energy function
defined in Equation (\ref{eq:Halt}) has a minimum at $\ve x^{(1)}$,
Equation (\ref{eq:Halt2}) need not have a minimum at $\ve x^{(1)}$ (or at any other \index{pattern!stored}stored pattern), because
a maximal value of   $\,\big(\sum_{i=1}^Ns_{i}x_i^{(1)}\big)^2$ may be compensated by terms stemming from other patterns.  
This happens rarely when $p$ is small (Section
\ref{sec:HN}).

Conversely there may be minima that do not correspond to stored
patterns. Such states are referred to as {\em spurious states}\index{spurious state|textbf}. The network may \index{convergence!spurious state}converge to spurious states. This is undesirable but it occurs  even when there is only one \index{pattern!stored}stored pattern, 
as we saw in Section \ref{sec:HN}: the McCulloch-Pitts dynamics
may \index{convergence!inverted pattern}converge to the inverted pattern\index{pattern!inverted}. 
This follows also from Equation (\ref{eq:Halt2}): if $\ve s = \ve x^{(1)}$ is a \index{local minimum}local minimum of $H$, then so is 
$\ve s = -\ve x^{(1)}$.  This is a consequence of the invariance of $H$ under  $\ve s \to -\ve s$.\index{energy function!invariance}  
There are other types of spurious states besides inverted patterns. 
An example are {\em mixed states}\index{state!mixed|textbf}, {\em superpositions}\index{superposition} of an odd number $2n+1$ of patterns \cite{hertz1991introduction}. For $n=1$, for example,
the bits of a mixed state read:
\begin{equation}
\label{eq:mix}
    x_{i}^{\text{(mix)}} = \text{sgn}(\pm x_{i}^{(1)} \pm x_{i}^{(2)} \pm x_{i}^{(3)})\,.
\end{equation}
The number of mixed states increases as $n$ increases.  There are 
$2^{2n+1} {p\choose{2n+1}}$ mixed states that are superpositions\index{superposition} of $2n+1$ out of $p$ patterns, for $n=1,2,\ldots$ (Exercise 2.4).
Mixed states such as (\ref{eq:mix}) are sometimes recognised by the network
(Exercise 2.5), 
therefore it may happen that the network \index{convergence!mixed state}converges to these states.
Finally, there are spurious states that are not related in
any way to the \index{pattern!stored}stored patterns $x_{j}\tomu$. Such {\em spin-glass}\index{state!spin glass} states are discussed in detail in Refs.~\cite{Amit1985,AmitGutfreund1987,sherrington1975solvable}, and also by Hertz, Krogh and Palmer \cite{hertz1991introduction}. 
\index{energy function|)}

\section{Summary}
Hopfield networks are networks 
of McCulloch-Pitts neurons
that recognise, or {\em retrieve},\index{pattern!retrieval}  patterns 
(Algorithm \ref{detalg}).
Their layout is defined by connection strengths, or weights,  chosen according to Hebb's rule.  The  weights $w_{ij}$ are \index{weight!symmetric weights}symmetric, and the network is in general fully connected.  Hebb's rule ensures that \index{pattern!stored}stored patterns  are recognised, at least most of the time if the number of patterns is not too large.  Convergence of the McCulloch-Pitts dynamics is analysed in terms of an energy function, which cannot increase under this dynamics.

A single-step estimate for the error probability of the network dynamics
was derived  in Section \ref{sec:HN}. If one iterates several steps, the error probability is usually much larger, but it is difficult
to evaluate in general. For stochastic Hopfield networks the steady-state error probability can be estimated more easily, because the dynamics converges to a 
steady state. \index{steady state}

\begin{algorithm}[t]
\caption{\label{detalg} \index{pattern!recognition} pattern recognition with deterministic Hopfield network}
\begin{algorithmic}
\STATE store patterns $\ve x\tomu$ using Hebb's rule;
\STATE feed distorted pattern $\ve x$ into network by assigning $\ve s(t=0) \leftarrow \ve x$;
\FOR {$t=1,\ldots,T$}
\STATE choose a value of $m$ and update $s_m(t) \leftarrow  \mbox{sgn}\big(\sum_{j=1}^N w_{mj} s_j(t-1)\big)$;\label{update_step}
\ENDFOR
\STATE read out pattern $\ve s(T)$;
\end{algorithmic}
\end{algorithm}

\vfill\eject

\chapter{Stochastic Hopfield networks}
\index{dynamics!stochastic|(}
\label{sec:shn}
Two related problems became apparent in the previous Chapter. First, the Hopfield dynamics may get stuck
in \index{spurious minimum}spurious minima. In fact, if there is a \index{local minimum}local minimum downhill from a given initial state, between this state
and the correct attractor\index{attractor}, then the dynamics arrests in the \index{local minimum}local minimum, so that the algorithm fails
to \index{convergence!to attractor}converge to the correct attractor. Second, the energy function usually is a strongly varying function over
a high-dimensional configuration space\index{configuration space}. Therefore it is difficult to predict the first \index{local minimum}local minimum encountered by the down-hill dynamics of the network.

Both problems are solved by introducing an element of stochasticity into the dynamics. This is a trick that works
for many neural-network algorithms. 
In general, however, it is quite challenging to analyse the stochastic dynamics. For \index{dynamics!stochastic}
the Hopfield network, by contrast, much is known. The reason is that the stochastic Hopfield network is closely
related to systems studied in statistical mechanics, so-called 
\index{spin glass} spin glasses. Like these systems
-- and many other physical systems -- the stochastic Hopfield network exhibits an {\em order-disorder transition}\index{order disorder transition}.
This transition becomes sharp in the limit of a large number of neurons.
This has important consequences. Suppose that the network produces satisfactory results
for a given number of patterns with a certain number of bits. If one tries to store just one more pattern, the network may 
fail to recognise anything. The goal of this Chapter is to explain why this occurs, and how it can be avoided.

\section{Stochastic dynamics}
\label{sec:nd}
The asynchronous update rule\index{update rule!asynchronous} (\ref{eq:C2S2Asynch}) is called {\em deterministic}\index{update rule!deterministic}, because a given set of states $s_j$ determines the 
outcome of the update of neuron $m$. To introduce noise, 
one replaces the rule (\ref{eq:C2S2Asynch})
by  an asynchronous {\em stochastic} rule \cite{hinton1983optimal}:
\index{dynamics!stochastic|textbf}
\index{update rule!stochastic|textbf}
\begin{subequations}
    \eqnlab{C3S2StochasticRule}
\begin{equation}
\label{eq:31a}
    s_{m}' = \begin{cases} +1\quad \text{  with probability  } \quad \pr(b_{m})\,, \\ -1 \quad \text{  with probability  } \quad 1-\pr(b_{m})\,.\end{cases}
\end{equation}
A neuron with update rule (\ref{eq:31a}) is called {\em binary stochastic neuron}.\index{neuron:binary stochastic}
As before,   $b_m = \sum_j w_{mj} s_j-\theta_m$ is the local field\index{local field},  and the probability $\pr(b)$ is given by:    
\begin{equation}
    \pr(b) = \frac{1}{1+{\rm e}^{-2\beta b}}\,.
    \eqnlab{C3S1Sigmoid}
\end{equation}
\end{subequations}
The function $\pr(b)$
is plotted in Figure  \figref{C3S1Sigmoid}. 
\begin{figure}[tb]
        \centering
        \begin{overpic}[scale=\myFigureScale]{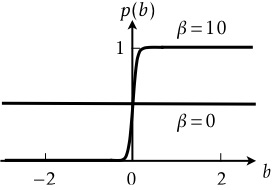}      
        \end{overpic}
        \caption{\figlab{C3S1Sigmoid}  Probability function \eqnref{C3S1Sigmoid} used in the definition of the stochastic rule \eqnref{C3S2StochasticRule}, plotted for $\beta=10$ and $\beta=0$.}
\end{figure} 
The parameter $\beta$ is the noise parameter. 
When $\beta$ is large, the {\em noise level} \index{noise level|textbf} is small. 
As $\beta$ tends to infinity, the function $\pr(b)$ approaches 
zero if $b$ is negative, and it tends to unity if $b$ is positive. 
So for $\beta\to\infty$, the stochastic update rule 
\index{update rule!stochastic}
\eqnref{C3S2StochasticRule} converges to the \index{deterministic limit}\index{rule!deterministic}deterministic rule (\ref{eq:C2S2Asynch}).
In the opposite limit, when $\beta=0$, the update probability $\pr(b)$ simply equals~$\tfrac{1}{2}$. In this case $s_i$ is updated to $-1$ or $+1$ randomly, with equal probability. The dynamics does not depend upon the \index{pattern!stored}stored patterns contained in the local field $\ve b$. \index{local field}

The idea is to keep a small but finite noise level.\index{noise level} Then the network dynamics is very similar to the deterministic Hopfield dynamics analysed in the previous Chapter. But the noise allows the system 
to escape \index{spurious minimum}spurious minima.  However, since the dynamics is stochastic, we must 
rephrase the convergence criterion (\ref{eq:Sc}). This is discussed next.
\index{convergence!criterion}

\section{Order parameters}
If we feed one of the \index{pattern!stored}stored patterns, $\ve x^{(1)}$ for example, then we want the stochastic dynamics to\index{dynamics!stochastic} 
stay in the vicinity of $\ve x^{(1)}$. This can only work if the noise is weak enough, and even then it is not guaranteed.
At time step $t$, bit $i$ is correct if $s_i(t) x_i^{(1)}=1$. All bits are correct when
$\sum_{i=1}^N s_i(t) x_i^{(1)}=N$, otherwise the sum takes a value smaller than $N$. 
One measures success by averaging $\tfrac{1}{N}\sum_{i=1}^N s_i(t) x_i^{(1)}$ over the 
 asynchronous stochastic  dynamics of the network
from $t=0$ to $t=T$, for
given bits $x_i^{(\mu)}$:
\begin{subequations}
\label{eq:cma}
\begin{align}
m_\mu(T)=\frac{1}{T}\sum_{t=1}^T\bigg(\frac{1}{N} \sum_{i=1}^{N}s_{i}(t)x_i^{(\mu)}\bigg)\,.
\end{align}
\begin{figure}[bt]
        \centering
        \begin{overpic}[scale=\myFigureScale]{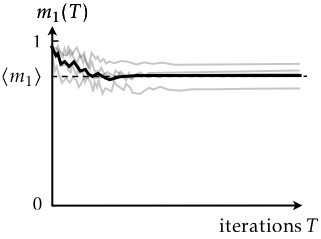}
        \end{overpic}
        \caption{\figlab{C3S1OrderParam} Illustrates how the 
finite-time \index{time average}average $m_1(T)$ depends upon the total iteration time $T$. The light gray lines show 
results for $m_1(T)$ for
different realisations 
of random patterns\index{pattern!random} \index{pattern!stored}stored in the network, at a large but finite value of $N$. The black line is the average of $m_1(T)$ over the different realisations of random patterns.}
\end{figure} 
If we feed pattern $\ve x^{(1)}$ to the network, 
we have $m_1(t\!=\!0)=1$ initially. We want that $m_1(t)$ remains close to unity, so that the network 
recognises the pattern $\ve x^{(1)}$. In practice, 
the quantity $\frac{1}{N} \sum_{i=1}^{N}s_{i}(t)x_i^{(1)}$
settles into a {\em steady state}\index{steady state|textbf}, where it fluctuates around a mean value with a definite distribution that becomes independent of the iteration number $t$.  If the network works well, the \index{time average}finite-time average $m_1(T)$ \index{convergence!order parameter}converges to a value of order unity after an initial {\em transient}\index{transient} (Figure~\figref{C3S1OrderParam}).
The limiting value,
\begin{equation}
\label{eq:m1limit}
m_1 \equiv \lim_{T\to\infty} m_1(T)\equiv \frac{1}{N} \sum_{i=1}^{N}\langle s_{i}\rangle x_i^{(1)} \,,
\end{equation}
\end{subequations}
is called the {\em order parameter}\index{order parameter|textbf}.
Since there is noise, the \index{order parameter}order parameter $m_1$ is usually smaller than unity.
The last equality in Equation (\ref{eq:m1limit}) defines the \index{time average}time average $\langle s_i\rangle$ over the stochastic network dynamics.\index{network!dynamics}

Figure \figref{C3S1OrderParam} also illustrates a subtlety. For finite values of $N$, the \index{order parameter}order parameter $m_1$ depends upon the \index{pattern!stored}stored patterns. Different realisations $\ve x^{(1)},\ldots, \ve x^{(p)}$ 
of random patterns\index{pattern!random}  yield different values of $m_1$.
In the limit of $N\to\infty$ this problem does not occur, 
the \index{order parameter}order parameter $m_1$ becomes independent of
the \index{pattern!stored}stored patterns. We say that the system is {\em self averaging}
\index{self averaging} in this limit.
To obtain a 
definite value  
for the \index{order parameter}order parameter for finite values of $N$, one usually averages 
$m_1$ over different realisations of random patterns\index{pattern!random} 
\index{pattern!stored}stored in the network (thick solid line in Figure \figref{C3S1OrderParam}). The dashed line in Figure  \figref{C3S1OrderParam} shows this average, $\langle m_1\rangle$.

The other components, $m_\mu = \lim_{T\to\infty} m_\mu(T)$ for $\mu > 1$, are expected to be small.  
This is certainly true for \index{pattern!random} random patterns with many independent bits.
If $s_i(t) \approx x_i^{(1)}$, the individual terms in
the sumer over $i$ in Equation (\ref{eq:m1limit}) cancel approximately
upon summation, because the bits of the patterns $\ve x^{(2)}$ to $\ve x^{(p)}$ are independent from those of $\ve x^{(1)}$.
 In summary,  if we feed pattern $\ve x^{(1)}$ and if the network works well, 
we expect in the limit of large $N$:
\begin{equation} m_{\mu} \approx \begin{cases} 1 \quad \mbox{if}\quad \mu =1, \\ 0 \quad \text{otherwise.}
\end{cases}  
\label{eq:m1_assumption}
\end{equation}
Whether this is the case or not depends on 
 the values of $p$, $N$, and $\beta$. In the next Sections
we determine how $m_{1}$ depends on these parameters.

\section{Mean-field theory}
\index{mean field theory|(}
\label{sec:mft}
The \index{order parameter}order parameter is defined as an average over the 
stochastic dynamics \index{dynamics!stochastic} of the network in its steady state (Figure \figref{C3S1OrderParam}). 
It is a challenging task to compute this average 
because all neurons interact with each other in a nonlinear fashion.
Consider neuron number $i$. 
The fate of $s_{i}$ is determined by its local field\index{local field} $b_{i}$,
through Equation  \eqnref{C3S2StochasticRule}. The difficulty is
that the local field in turn depends on the states $s_{j}$ of all other neurons in the network:\footnote{We set the \index{threshold}thresholds to zero, as assumed in Hebb's rule (\ref{eq:HRw0}).}
\begin{equation}
 b_{i}(t) = \sum_{j=1}^{N}w_{ij}s_{j}(t)\,.
\end{equation}
When $N$ is large, we may assume that $b_i(t)$ remains essentially constant
in the steady state\index{steady state}, independent of $t$, because fluctuations of $s_j(t)$ average out when summing over $j$:
\begin{equation}
\label{eq:mfa0}
 b_{i}(t) = \avg{b_{i}} + \mbox{fluctuations}\,.
\end{equation}
Since $b_i(t)$ is given by a sum over many random  numbers, we appeal to the central-limit theorem\index{central limit theorem} 
and argue that the fluctuations of $b_i(t)$  are of order $N^{-1/2}$. Since $\langle b_i(t) \rangle \sim {1}$, 
we ignore the fluctuations in the limit of large $N$
and write
\begin{equation}
\label{eq:mfa}
b_{i}(t) \approx  \avg{ b_i} = \sum_{j=1}^{N}w_{ij}\langle s_{j}\rangle=
\frac{1}{N}\sum_{\mu}\sum_{j\neq i} x_i^{(\mu)} x_{j}^{(\mu)}
\avg{ s_{j}}\,,
\end{equation}
using Hebb's rule  (\ref{eq:HRw0}) for given patterns $\ve x^{(\mu)}$. 
The time-averaged local field\index{local field}\index{local field!time averaged} $\avg{b_{i}}$ is called the {\em mean field}\index{mean field}.
Theories that neglect the fluctuations in Equation (\ref{eq:mfa0}) are called {\em mean-field theories}\index{mean field theory}.
They require a self-consistent solution, because 
the average $\avg{ s_{j}}$
on the r.h.s. of Equation (\ref{eq:mfa}) depends on the mean field. Using the stochastic 
update rule \eqnref{C3S2StochasticRule} we find:
\begin{align} 
\avg{s_i} & = \text{Prob}(s_{i} = +1)-\text{Prob}(s_{i} = -1)
=\pr(\langle b_i\rangle )-[1-\pr(\langle b_i\rangle)]
= \tanh(\beta \langle b_{i}\rangle)\,.  
\eqnlab{C3S4Output} 
\end{align} 
Equations (\ref{eq:mfa}) and \eqnref{C3S4Output}  yield a 
set of $N$ non-linear self-consistent equations for
$\avg{s_i}$,
\begin{equation}
\label{eq:mfe}
    \avg{ s_{i} } = \tanh(\beta\avg{ b_{i}})
\quad\mbox{with}\quad
    \avg{ b_{i}} = \frac{1}{N}\sum_{\mu}\sum_{j\neq i} x_i^{(\mu)} x_{j}^{(\mu)}
\avg{ s_{j}}\,.
\end{equation}
Recall that the averages $\langle \cdots\rangle$ are \index{time average}time averages, evaluated for given patterns $\ve x^{(\mu)}$. 

An equivalent yet slightly different derivation of the \index{mean field equation|textbf} mean-field equations (\ref{eq:mfe}) is this: 
suppose we average $s_i$ over the dynamics \eqnref{C3S2StochasticRule}
at fixed $s_j\neq s_i$, and then we average all $s_{j}$ over the dynamics. This gives
$\avg{ s_{i} } = \langle \tanh(\beta b_{i})\rangle$. Comparing with Equation
(\ref{eq:mfe}), we see that the mean-field approximation corresponds to approximating $\langle \tanh(\beta b_{i})\rangle\approx \tanh(\beta\avg{ b_{i}})$.

Now, in order to calculate the \index{order parameter}order parameters (\ref{eq:def_m_mu}),
\begin{equation}
\label{eq:def_m_mu}
m_\mu =  \frac{1}{N}\sum_{j=1}^{N} \avg{s_{j}}   x_{j}^{(\mu)} \,,
\end{equation} 
we must solve the \index{mean field equation} mean-field equations (\ref{eq:mfe}) to obtain the \index{time average}time averages $\langle s_i\rangle$ in Equation (\ref{eq:def_m_mu}). 
To this end
we express the mean field $\langle b_i\rangle$ in terms of the \index{order parameter}order parameters $m_\mu$:
\begin{equation}
\label{eq:bsum}
\begin{split}
    \avg{ b_{i}} &=  \frac{1}{N}\sum_{\mu=1}^p\sum_{j\neq i} x_i^{(\mu)} x_{j}^{(\mu)}\avg{s_{j}}  \approx \sum_{\mu=1}^p x_i^{(\mu)} m_{\mu} \,.
\end{split}
\end{equation}
The last equality is only approximate because the $j$-sum in the definition
of $m_\mu$ contains the term $j=i$. Whether or not
to include this term makes only a small difference in the limit of large $N$.

Let us first calculate $m_1$ assuming Equation (\ref{eq:m1_assumption}), neglecting terms with $\mu\neq1$ in Equation (\ref{eq:bsum}).
To make sure  that these small $\mu\!\neq\!1$-terms 
do not add up to a substantial correction
to the first term, 
the storage capacity must be small enough. For large values of $N$
we assume
\begin{equation}
\label{eq:alphacondition}
\alpha    {\ll 1}\,.
\end{equation} 
In this case it is sufficient to keep only the first term on the r.h.s. of Equation (\ref{eq:bsum}) \cite{MullerReinhardt}. This approximation yields together
with Equation (\ref{eq:mfe}):
\begin{equation}
\avg{ s_i} = \tanh(\beta\avg{b_{i}}) \approx \tanh(\beta m_1 x_i^{(1)}) \,.
\end{equation}
Applying the definition (\ref{eq:def_m_mu}) of the \index{order parameter}order parameter, 
one finds
\begin{equation}
   m_{1} = \frac{1}{N} \sum_{i=1}^{N} \tanh\left(\beta m_{1}  x_{i}^{(1)}\right)  x_{i}^{(1)}\,.
\end{equation}
Using that $\tanh(z) =-\tanh(-z)$ as well as the fact that the bits  $x_i^{(\mu)}$
can only assume the values $\pm 1$, one obtains:
\begin{equation}
\label{eq:m1}
   m_{1} = \tanh(\beta m_{1})\,.
\end{equation}
This is a self-consistent equation for $m_1$. 
For $\beta\to 0$, it has the solution $m_1=0$. This is not the desired solution 
because $m_1=0$ means that $\ve x^{(1)}$ is not recognised.
For $\beta\to\infty$, by contrast, there are three solutions,
$m_1=0,\pm 1$. Figure \figref{C3S3OrderOnNoise} 
shows results of the numerical evaluation of Equation (\ref{eq:m1})
for intermediate values of $\beta$. 
For $\beta$ larger than the critical value,
\index{noise level!critical}
\begin{figure}[bt]
        \centering
        \begin{overpic}[scale=\myFigureScale]{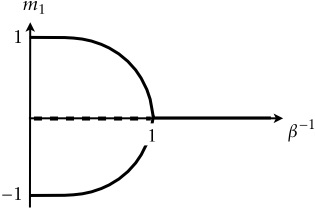}
        \end{overpic}
        \caption{ \figlab{C3S3OrderOnNoise} Solutions of the \index{mean field equation} mean-field equation (\ref{eq:m1}), solid lines. The critical noise level is $\beta_c=1$. The dashed line corresponds to an unstable solution.}
\end{figure} 
\begin{equation}
\beta_c=1\,,
\end{equation}
the three solutions persist.  The
solution $m_1=0$ is {\em unstable}
\index{solution!unstable}. This can be shown by computing
the derivatives of the {\em free energy}\index{free energy} of the Hopfield network \cite{hertz1991introduction}. In other words,  if we start with an initial condition that corresponds to $m_1=0$, the network dynamics
does not stay there.
The other two solutions are {\em stable}\index{solution!stable}: when the network is
initialised close to $\ve x^{(1)}$,
then it \index{convergence!order parameter}converges to  $m_1=O(1)$,
as long as $\beta > \beta_{\rm c}$.

The symmetry\index{symmetry} of the problem dictates that there must also be a solution 
with $-m_1$. This solution corresponds to the 
inverted pattern\index{pattern!inverted} $-\ve x^{(1)}$ (Section \secref{C2S3EnergyFunction}).
If we start in the vicinity of  $\ve x^{(1)}$, then the network is unlikely to 
\index{convergence!order parameter}converge to $-\ve x^{(1)}$, provided
that $N$ is large enough. The probability of the dynamical transition $\ve x^{(1)} \to -\ve x^{(1)}$ vanishes very rapidly as $N$ increases
and as the noise level decreases. If this transition happens in a simulation in this limit, the network then stays near $-\ve x^{(1)}$ for a very long time.
Consider the limit where  $T$ tends to $\infty$ at a finite but  large
value of $N$. Then the network  jumps back and
forth between $\ve x^{(1)}$ and $-\ve x^{(1)}$ at a very small rate. As a result,  the \index{order parameter}order parameter averages to zero. This shows that the limits of large $N$ and large $T$
do not commute:
\begin{equation}
\lim_{T\to\infty}\lim_{N\to\infty} m_1(T)
\neq 
\lim_{N\to\infty}\lim_{T\to\infty} m_1(T) \,.
\end{equation}
In practice the interesting limit is the left one, that of a large network run
for a time $T$ much longer than the initial transient, but not infinite.
This is precisely where the mean-field theory\index{mean field theory} applies. It corresponds to
taking the limit $N\to\infty$ first, at finite but large $T$.
This describes simulations where the transition $\ve x^{(1)} \to -\ve x^{(1)}$ does not occur.

In summary, Equation (\ref{eq:m1}) predicts that the \index{order parameter}order parameter \index{convergence!order parameter}converges to a definite value, $m_1$, independent of the \index{pattern!stored}stored patterns in the limit $N\to\infty$.
When $N$ is finite, the limiting value of the order parameter  does depend on the \index{pattern!stored}stored patterns (Figure \figref{C3S1OrderParam}).
In this case one averages also  over different realisations 
of the \index{pattern!stored}stored patterns, as mentioned above. 
The value of this average,  $\langle m_1\rangle $, determines the average 
\index{error probability}
error probability $P_{\rm error}^{t=\infty}$ in the steady state, the average fraction of wrong bits. 
The steady-state average number of correct bits is given by
\begin{equation}
\label{eq:Ncorrect0}
   \begin{split}
        \Big\langle \frac{1}{2} \sum_{i=1}^{N}\Big(1+\avg{s_i}x_i^{(1)}\Big) \Big\rangle 
 = \frac{N}{2} (1+\langle m_{1}\rangle)\,,
   \end{split}
\end{equation}
because $\tfrac{1}{2}(1+s_i x_i^{(1)})=1$ if $x_i^{(1)}$ is correct, and equal to zero otherwise. 
The outer average is over different realisations of \index{pattern!random} random patterns (the inner average is over the network dynamics).  The second equality follows from Equation (\ref{eq:m1limit}). 
Since the l.h.s. of Equation (\ref{eq:Ncorrect0}) equals $N$ times
$1-P_{\rm error}^{t=\infty}$, we deduce that
\begin{equation}
\label{eq:Pe_infty_alpha_0}
P_{\rm error}^{t=\infty} = \tfrac{1}{2} (1-\langle m_1\rangle)\,.
\end{equation}
Since $m_{1}\to 1$ as $\beta \to \infty$, the steady-state error probability tends
to zero in this limit. 
This is expected since the \index{pattern!stored}stored patterns $\ve  x^{(\mu)}$ are recognised for small enough values of $\alpha$ in the \index{deterministic limit}deterministic limit, when the \index{cross talk term} \index{cross talk term} cross-talk term is negligible. 
But note that the stochastic dynamics {\em slows down}\index{slowing down} as the noise
level tends to zero. The lower the noise level\index{noise level}, the longer the network remains stuck in \index{local minimum}local minima,
so that it takes longer time to reach the steady state\index{steady state}, and to sample the steady-state statistics
of $H$.  
In the opposite limit, $\beta \to 0$, the steady-state error probability tends to $\tfrac{1}{2}$, because $m_1\to 0$.
\index{error probability}
In this noise-dominated limit the stochastic network ceases to function. If one 
were to assign $N$ bits entirely randomly, then half of them would be correct, on average,
$P_{\rm error}^{t=\infty} = \tfrac{1}{2}$.

It is important to note that noise can help in another way:
it may prevent the network dynamics from converging to mixed states (Section  \secref{C2S3EnergyFunction}).
This can be seen as follows \cite{Amit1985,hertz1991introduction}.
To derive the above mean-field result we assumed $m_1\approx 1$ and  $m_\mu \approx 0$ for $\mu\neq 1$.
Mixed states correspond to solutions where an odd number of components
of $\ve m$ is non-zero, for example: \index{order parameter!mixed state}
\begin{equation}
\label{eq:mfe_mixed}
\ve m^{({\rm mix})} = \begin{bmatrix} m\\m\\m\\0\\\vdots \end{bmatrix}\,.
\end{equation}
Neglecting the \index{cross talk term} \index{cross talk term} cross-talk term, the \index{mean field equation} mean-field equation reads 
\begin{equation}
\langle s_i\rangle = 
\mbox{tanh}\Big(\beta \sum_{\mu=1}^p m_\mu^{({\rm mix})} x_i^{(\mu)}\Big)\,.
\end{equation}
In the limit of $\beta\to\infty$, the averages $\avg{s_i}$
\index{convergence!mixed state}converge to the \index{state!mixed} mixed states (\ref{eq:mix}) when
$\ve m^{({\rm mix})}$ is given by Equation (\ref{eq:mfe_mixed}). 
Averaging over the bits of the \index{pattern!random} random patterns 
one finds:
\begin{equation}
\label{eq:mft_final}
m_\mu^{({\rm mix})} = \Big\langle x_i^{(\mu)} \mbox{tanh}\Big(\beta \sum_{\nu=1}^p m_\nu^{({\rm mix})} x_i^{(\nu)}\Big)\Big\rangle\,.
\end{equation}
The numerical solution of Equation (\ref{eq:mft_final}) shows that
there is a non-zero solution for $\beta^{-1} < \beta_{\rm c}^{-1}=1$. Yet
this solution is unstable for 
 $0.46 < \beta^{-1} < 1$ \cite{Amit1985}. 
In other words, the \index{state!mixed}mixed states have a lower critical noise level than the \index{pattern!stored}stored patterns,
equal to $0.46$.
For noise levels\index{noise level} larger than that, but still smaller than unity, the network can recognise the \index{pattern!stored}stored patterns, and it does not \index{convergence!mixed state}converge to \index{state!mixed}mixed states. 

However, these results were obtained assuming that only one (or a few) \index{order parameter}order parameters are not zero.
This corresponds to the limit of $\alpha=p/N\to 0$, where the \index{cross talk term} cross-talk term 
(Section \ref{sec:ctt}) is negligible. The next Section describes a \index{mean field theory} mean-field theory that remains accurate for larger values of $\alpha$.
\index{mean field theory|)}

\section{Critical storage capacity}
\label{sec:sc_mft}
The analysis in the preceding Section replaced the sum (\ref{eq:bsum}) by its first
term, $x_i^{(1)}m_1$. This can only work when $p/N$ is small enough.
Now we discuss how to proceed when $p/N$ is not small.

Note that the analysis in Section \ref{sec:HN} did not assume that $p/N$ is small, but it yielded
only the one-step \index{error probability} error probability $P_{\rm error}^{t=1}$, and we discussed the \index{storage capacity} storage capacity
$\alpha=p/N$ in relation to the one-step error probability. As the network dynamics is iterated, however, the number of errors
tends to increase, at least when $\alpha$ is large enough so that the \index{cross talk term} cross-talk term matters.
Now we describe how to compute $P_{\rm error}^{t=\infty}$ for general values 
of the storage capacity $\alpha$, in order to 
demonstrate how the errors multiply as one iterates, causing the network to fail. 

As before, we store $p$ patterns in the network using Hebb's rule (\ref{eq:HRw0}) and feed
pattern $\ve x^{(1)}$ to the network. The aim is to determine the \index{order parameter}order parameter $m_1$ and the corresponding
error probability in the
steady state\index{steady state} for $p \sim N$, so that $\alpha$ remains finite as $N\to \infty$. 
In this case we can no longer approximate
the sum in Equation (\ref{eq:bsum}) just by its first term, because the other terms for $\mu>1$ may sum up to a contribution that is of the same 
order as $m_1$. Instead we must evaluate all $m_\mu$ to compute the mean field $\avg{b_i}$.

The relevant calculation is summarised in Chapter 4 of Ref.~\cite{geszti1990physical}. It is also outlined in Section 2.5 of Hertz, Krogh and Palmer~\cite{hertz1991introduction}. The remainder of this Section follows this outline quite closely.
One starts by rewriting the mean-field equations (\ref{eq:mfe}) 
\index{mean field equation}
in terms of the \index{order parameter}order parameters $m_\mu$.
Using 
\begin{equation}
    \avg{s_{i}} = \tanh(\beta \sum_{\mu} x_i^{(\mu)} m_{\mu})
\end{equation}
we find
\begin{equation}
\label{eq:mfe_m}
    m_{\nu} = \frac{1}{N}\sum_{i} x_{i}^{(\nu)} \avg{s_{i}} = \frac{1}{N} \sum_{i} x_{i}^{(\nu)} \tanh\bigg(\beta \sum_{\mu} x_i^{(\mu)}  m_{\mu}\bigg)\,.
\end{equation}
This coupled set of $p$ non-linear equations is equivalent to the mean-field equations (\ref{eq:mfe}).

Now feed pattern $\ve x^{(1)}$ to the network.
We assume that the network stays close to the pattern $\ve x^{(1)}$ in the steady state\index{steady state}, so that $m_{1}$ remains of order unity.
The other $m_\mu$ remain small.  When $p$ is large, however, we cannot simply approximate the sum over $\mu$ on the r.h.s. of Equation (\ref{eq:mfe_m}) by its first term only, because the sum of the remaining (small) terms might not
be negligible. Therefore we need to estimate these terms, the other \index{order parameter}order parameters  $m_{\mu}$ for $\mu\neq 1$.

The trick is to assume that the pattern
bits are random, uncorrelated with mean zero [Equations (\ref{eq:prob}) and (\ref{eq:avg_patterns})].
\index{pattern!random} In this case the order parameters $m_\mu$, $\mu=2,\ldots,p$, become random numbers that fluctuate around zero 
with variance $\langle m_\mu^2\rangle$ (this average is over \index{pattern!random} random patterns).
We use Equation (\ref{eq:mfe_m}) to compute the variance approximately.

In the $\mu$-sum on the r.h.s of Equation (\ref{eq:mfe_m}) we must treat the term $\mu=\nu$ separately, because the index $\nu$ appears also on the l.h.s. of this equation. Also the term $\mu=1$ must
be treated separately, as before, because $\mu=1$ is the index of the pattern that is fed
to the network. 
As a consequence, the calculations of $m_1$ and $m_\nu$ for $\nu\neq 1$
proceed slightly differently. 
We begin with the first case. 
Using that  $x_{i}^{(\mu)} = \pm 1$, and that $\mbox{tanh}(z)$ is an odd function, Equation (\ref{eq:mfe_m})
simplifies to:
\begin{equation}
\label{eq:m1_sum}
    m_1 = \frac{1}{N} \sum_{i}  \tanh \bigg(\beta m_{1} + \beta\sum_{\mu \neq 1}
 x_{i}^{(\mu)}x_{i}^{(1)} m_{\mu}\bigg) \,.
\end{equation}
The next steps are similar to the analysis of the \index{cross talk term} cross-talk term in Section \ref{sec:HN}. One assumes that the patterns are random, that
their bits $x_i^{(\mu)}=\pm 1$ are independently and identically 
distributed. In the limit of large $N$ and $p$, the sums 
in Equation (\ref{eq:m1_sum}) can then be estimated using the central-limit theorem \index{central limit theorem}. 
For \index{pattern!random} random patterns, the variable
\begin{equation}
\label{eq:zdef}
z\equiv \sum_{{\mu \neq 1}}  x_{i}^{(\mu)}  x_{i}^{(1)} m_{\mu}
\end{equation}
is a sum of many independent, identically distributed random numbers
 with mean zero and finite variance. The variable $z$ is therefore approximately Gaussian distributed, with mean zero. 
As a consequence, the distribution of $z$ is entirely 
determined by its variance $\sigma_{\!z}^2$, and it is indepedent of $i$.

Returning to Equation (\ref{eq:m1_sum}), 
one approximates the sum $\frac{1}{N} \sum_{i}$ as an average over the Gaussian distributed
variable $z$. This yields:
\begin{equation} 
\label{eq:m1_sc} 
m_{1} = \int \frac{{\rm d}z}{\sqrt{2\pi\sigma_{\!z}^{2}}} e^{-\frac{z^{2}}{2\sigma_{\!z}^{2}}} \tanh(\beta m_{1} + \beta z)\,.  
\end{equation} 
Equation (\ref{eq:m1_sc}) is the desired result, a self-consistent equation for $m_1$ 
replacing the mean-field equation (\ref{eq:m1}). 

In order to determine $m_1$, we need to estimate the variance $\sigma_{\!z}^2$
featuring in Equation (\ref{eq:m1_sc}). To this end, one squares Equation (\ref{eq:zdef}) and averages the resulting double sum. 
Since the bits $x_i^{(\mu)}$ and $x_i^{(\mu')}$ are independent
when $\mu\neq \mu'$, only the diagonal terms in this double sum contribute to the average:
\begin{equation}
\label{eq:sm2}
 \sigma_{\!z}^{2} = \sum_{{\mu \neq 1}} \langle m_ \mu^{2}\rangle \approx p \langle m_{\mu}^{2}\rangle \quad\mbox{for any}\quad \mu\neq 1\,.
\end{equation}
Here we assumed that $p$ is large, and approximated $p-1 \approx p$.
To evaluate the variance further, it is necessary to estimate the remaining order parameters.
One starts again from Equation (\ref{eq:mfe_m})
 and writes for $\nu\neq 1$
\begin{equation}
\label{eq:startmnu}
\begin{split}
    m_{\nu} &= \frac{1}{N} \sum_{i} x_{i}^{(\nu)} \tanh\bigg(\beta  x_{i}^{(1)} m_{1} + \beta  x_{i}^{(\nu)} m_{\nu} + \beta \sum_{\substack{\mu \neq 1 \\ \mu \neq \nu}} x_{i}^{(\mu)} m_{\mu}  \bigg) \\
    & = \frac{1}{N} \sum_{i}  x_{i}^{(\nu)}   x_{i}^{(1)} \tanh\bigg(\underbrace{\beta m_{1}}_{\textcircled{1}} + \underbrace{\beta  x_{i}^{(1)}  x_{i}^{(\nu)} m_{\nu} }_{\textcircled{2}} + \underbrace{\beta\sum_{\substack{\mu \neq 1\\ \mu \neq \nu}} x_{i}^{(\mu)}  x_{i}^{(1)} m_{\mu}}_{\textcircled{3}} \bigg)\,.
\end{split}
\end{equation}
Consider the three terms in the argument of $\tanh(\dots)$.
The term \textcircled{1} is of order unity, it is independent of $N$.
The term \textcircled{3} may be of the same order, because
the sum over $\mu$ contains $\sim p N$ terms.
The term \textcircled{2}, by contrast, is small for large values of $N$.
Therefore it is a good approximation to Taylor-expand tanh(\dots) as follows:
\begin{equation}
\tanh\Big(\textcircled{1}+\textcircled{2}+\textcircled{3}\Big) \approx \tanh\Big(\textcircled{1}+\textcircled{3}\Big)
+ \textcircled{2} \frac{{\rm d}}{{\rm d}x}\tanh \bigg|_{\textcircled{1}+\textcircled{3}} + \ldots\,.
\end{equation}
Using $\frac{{\rm d}}{{\rm d}x} \tanh(x) = 1-\tanh^{2}(x)$ one obtains
\begin{equation}
\begin{split}
m_{\nu} = \frac{1}{N}\sum_i  x_{i}^{(\nu)}  x_{i}^{(1)} \tanh \bigg(\underbrace{\beta m_{1}}_{\textcircled{1}} + \underbrace{\beta\sum_{\substack{\mu \neq 1\\ \mu \neq \nu}}x_i^{(\mu)}  x_{i}^{(1)} m_{\mu}}_{\textcircled{3}} \bigg) \\
+ \frac{1}{N} \sum_{i} x_i^{(\nu)}  x_{i}^{(1)} \underbrace{\beta  x_{i}^{(1)}  x_{i}^{(\nu)} m_{\nu} }_{\textcircled{2}} \bigg[1-\tanh^{2}\bigg(\beta m_{1}+\beta\sum_{\substack{\mu \neq 1\\ \mu \neq \nu}} x_{i}^{(\mu)}   x_{i}^{(1)} m_{\mu} \bigg)\bigg] \,.
\end{split}
\end{equation}
Using the fact that $x^{(\mu)} = \pm 1$ and thus $[ x_{i}^{(\mu)}]^{2} = 1$, this expression simplifies:
\begin{equation}
\label{eq:mnu}
\begin{split}
    m_{\nu} = \frac{1}{N} \sum_{i}  x_{i}^{(\nu)}  x_{i}^{(1)} \tanh \bigg(\beta m_{1} + \beta\sum_{\substack{\mu \neq 1\\ \mu \neq \nu}}
 x_{i}^{(\mu)}x_{i}^{(1)} m_{\mu}\bigg) + \\
    + \beta m_{\nu} \frac{1}{N} \sum_{i} \bigg[1-\tanh^{2}\bigg(\beta m_{1}+ \beta\sum_{\substack{\mu \neq 1\\ \mu \neq \nu}} x_{i}^{(\mu)}  x_{i}^{(1)} m_{\mu}\bigg) \bigg]\,.
\end{split}
\end{equation}
The goal is now to solve for $m_\nu$.
Approximating the sum $\frac{1}{N} \sum_{i}$ in the second line as an average over the Gaussian distributed
variable $z$ [Equation (\ref{eq:zdef})] gives:
\begin{equation}
\label{eq:mnuint}
\beta m_{\nu} \int_{-\infty}^{\infty} \!\!\!{\rm d}z 
\frac{1}{\sqrt{2 \pi} \sigma_{\!z}} {\rm e}^{-\frac{z^{2}}{2\sigma_{\!z}^{2}}} \left[1-\tanh^{2}\left(\beta m_{1} + \beta z\right)\right]\,.
\end{equation}
Defining the parameter $q$
\begin{equation}
\label{eq:defq}
q \equiv \int_{-\infty}^{\infty} \!\!\!{\rm d}z
\frac{1}{\sqrt{2 \pi} \sigma_{\!z}} {\rm e}^{-\frac{z^{2}}{2\sigma_{\!z}^{2}}} \tanh^{2}\left(\beta m_{1} + \beta z\right)\,,
\end{equation}
one can write Equation (\ref{eq:mnuint}) as
\begin{equation}
\beta m_{\nu}\bigg[1- \int_{-\infty}^{\infty} \!\!\!{\rm d}z
\frac{1}{\sqrt{2 \pi} \sigma_{\!z}} {\rm e}^{-\frac{z^{2}}{2\sigma_{\!z}^{2}}} \tanh^{2}\left(\beta m_{1} + \beta z\right) \bigg] \equiv \beta m_{\nu}(1-q) \,.\eqnlab{C3S6IntegralGives}
\end{equation}
Returning to Equation (\ref{eq:mnu}), we see that it takes the form
\begin{equation}
m_{\nu} = \frac{1}{N} \sum_{i}x_i^{(\nu)} x_{i}^{(1)} \tanh\bigg(\beta m_{1} + \beta\sum_{\substack{\mu \neq 1\\ \mu \neq \nu}} x_{i}^{(\mu)}  x_{i}^{(1)} m_{\mu}\bigg) + (1-q)\beta m_{\nu}\,.
\end{equation}
Solving for $m_\nu$ one finds for $\nu \neq 1$:
\begin{equation}
\label{eq:mnu_sc}
m_{\nu} = \frac{ \frac{1}{N} \sum_{i}x_i^{(\nu)}  x_{i}^{(1)} \tanh\bigg(\beta m_{1} + \beta\sum_{\substack{\mu \neq 1\\ \mu \neq \nu}} x_{i}^{(\mu)}  x_{i}^{(1)} m_{\mu}\bigg)}{ 1-  \beta(1-q)}  \,.
\end{equation}

This expression allows us to compute the variance 
$\sigma_{\!z}$, defined by Equation (\ref{eq:sm2}). Equation (\ref{eq:mnu_sc}) shows that the average $\langle m_\nu^2\rangle$ contains a double sum over the 
bit index, $i$. Since the bits are independent, only the diagonal terms contribute, so that
\begin{equation}
\langle m_\nu^2 \rangle \approx \frac{\frac{1}{N^2} \sum_i \tanh^2\bigg(\beta m_{1} + \beta\sum_{\substack{\mu \neq 1\\ \mu \neq \nu}} x_{i}^{(\mu)}  x_{i}^{(1)} m_{\mu}\bigg)}{[1-  \beta(1-q)]^2}
\end{equation}
for $\nu \neq 1$, but otherwise independent of $\nu$.
The numerator is just $q/N$, from Equation (\ref{eq:defq}). So the variance evaluates to
\begin{equation}
\label{eq:varz}
 \sigma_{\!z}^{2} = \frac{\alpha q}{[1- \beta(1-q)]^{2}}\,.
\end{equation}
In summary there are three coupled equations, for $m_1$, $q$, and 
$\sigma_{\!z}$, Equations  (\ref{eq:m1_sc}), \eqnref{C3S6IntegralGives}, and (\ref{eq:varz}). They must be solved
together to determine how $m_1$ depends on $\beta$ and $\alpha$. 

In order to compare with the results described in Section \ref{sec:HN}, we must take the \index{deterministic limit}deterministic limit, $\beta\to\infty$.
In this limit, $q$ approaches unity, 
yet $\beta(1-q)$ remains finite \cite{hertz1991introduction}. Setting $q=1$ in Equation~(\ref{eq:varz})
but retaining $\beta(1-q)$ one finds:
\begin{subequations}
\label{eq:sc3}
\begin{align}
 \sigma_{\!z}^{2} & =  \frac{\alpha }{[1- \beta(1-q)]^{2}}\,.
\end{align}
The \index{deterministic limit}deterministic limits of Equations
\eqnref{C3S6IntegralGives} and (\ref{eq:m1_sc}) become \cite{hertz1991introduction}:
\begin{align}
 \beta(1-q) &= \sqrt{\frac{2}{\pi \sigma_{\!z}^2}} {\rm e}^{-\frac{m_{1}^{2}}{2\sigma_{\!z}^2}}\,, \\
\label{eq:m12}
        m_{1} &= \text{erf}\left(\frac{m_{1}}{\sqrt{2 \sigma_{\!z}^2}}\right)\,. 
\end{align}
\end{subequations}
\index{error probability|(}
Recall expression (\ref{eq:Pe_infty_alpha_0})  for the steady-state error probability. 
Inserting Equation (\ref{eq:m12}) for $m_1$ into this expression we find
in the same  limit:
\begin{equation}
\label{eq:Pe_infty}
P_{\rm error}^{t=\infty} =  \frac{1}{2} \left[1-\text{erf}\left(\frac{m_{1}}{\sqrt{2\sigma_{\!z}^2}}\right)\right]\,.
\end{equation}
Compare this with Equation \eqnref{C2S2Perror} for the one-step error probability in the \index{deterministic limit}deterministic limit.
That equation was derived for only one step of the network dynamics,
while Equation (\ref{eq:Pe_infty})
describes the limit of many steps, the long-time or steady-state limit.  

Yet it turns out that Equation (\ref{eq:Pe_infty}) reduces to \eqnref{C2S2Perror} as  $\alpha\to 0$. 
To see this, one solves the set of Equations (\ref{eq:sc3}) by introducing
the variable $y=m_1/\sqrt{2\sigma_{\!z}^2}$. One
obtains the following one-dimensional equation for $y$ \cite{hertz1991introduction,AmitGutfreund1987}:
\begin{equation}
\label{eq:y}
y(\sqrt{2\alpha}+ (2/\sqrt{\pi})\,\, {\rm e}^{-y^2}) = \mbox{erf}(y)\,.
\end{equation}
The relevant solutions are those satisfying $0 \leq {\rm erf}(y) \leq 1$,
because the \index{order parameter}order parameter is restricted to this range (transitions
to $-m_1$ do not occur in the limit $N\to\infty$). 
Figure \figref{C3S4ErrorProbabilitySteadyState} shows the steady-state error probability obtained from 
Equations (\ref{eq:Pe_infty}) and (\ref{eq:y}). Also shown is the one-step
error probability 
\begin{equation}
\nonumber
    P_{\text{error}}^{t=1} = \frac{1}{2} \left[1-\text{erf}\left(\frac{1}{\sqrt{2\alpha}}\right)\right]
\end{equation} 
derived in Section \ref{sec:HN}. 
As stated above,  $P_{\rm error}^{t=\infty}$ approaches $P_{\rm error}^{t=1}$
for small $\alpha$.  We conclude: in this limit, for small $\alpha$, the error probability does not increase
significantly as one iterates the network dynamics\index{network!dynamics}. Errors in earlier iterations have little effect on the probability that later errors occur. 
\begin{figure}[t]
        \centering
        \begin{overpic}[scale=\myFigureScale]{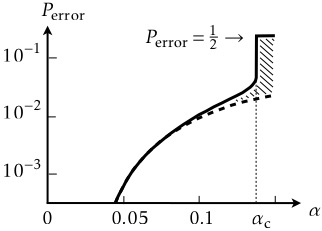}
        \end{overpic}
\caption{\figlab{C3S4ErrorProbabilitySteadyState} Error probability as a function of the \index{storage capacity}storage capacity $\alpha$ in the \index{deterministic limit}deterministic limit. The one-step error probability $P_{\rm error}^{t=1}$ [Equation \eqnref{C2S2Perror}] is shown as a dashed line, the steady-state error probability $P_{\rm error}^{t=\infty}$ [Equation (\ref{eq:Pe_infty})] is shown as a solid line. In the hashed region, error avalanches\index{error!avalanche} increase the error probability. After Figure 1 in Ref.~\cite{AmitGutfreund1987}.}
\end{figure}

The situation is different at larger values of $\alpha$. In that case, $P_{\rm error}^{t=1}$ significantly underestimates the steady-state error probability. In the hashed region, errors in the dynamics increase the probability of errors in subsequent steps, giving rise to {\em error avalanches}\index{error!avalanche|textbf}.  
Figure \figref{C3S4ErrorProbabilitySteadyState}  illustrates
that the steady-state error probability tends to $\tfrac{1}{2}$
as the parameter $\alpha$ increases beyond a critical value, $\alpha_c$.
Equation (\ref{eq:y}) yields
\begin{equation}
\label{eq:alphac_mft}
\alpha_c \approx 0.1379
\end{equation}
for the critical storage capacity \index{storage capacity!critical}$\alpha_{\rm c}$.
When $\alpha>\alpha_c$, the network produces just noise. When $\alpha< \alpha_{\rm c}$, by contrast, the network works well. The smaller
the storage capacity $\alpha$, the better the network performs.

 Figure \figref{C3S4ErrorProbabilitySteadyState} shows that the steady-state error probability changes very abruptly near $\alpha_c$. Suppose we store 137 patterns with 1000 bits in a Hopfield network. In this case the network can retrieve the patterns 
with a comparatively small error probability.
However, if we try to store one or two more patterns, the network fails to produce
output meaningfully related to the \index{pattern!stored}stored patterns. This rapid change
is an example of a {\em phase transition}\index{phase transition}. In many physical systems one observes similar transitions between ordered and disordered phases \cite{Kadanoff}.

What happens at higher \index{noise level} noise levels? The numerical solution of Equations \eqnref{C3S6IntegralGives}, (\ref{eq:m1_sc}), and (\ref{eq:varz}) shows that the \index{storage capacity!critical} critical storage capacity $\alpha_c$ decreases as the noise level increases (smaller values of $\beta$). This is shown schematically in Figure \figref{C3S4Phase}. Below the 
solid line the error probability is smaller than $\tfrac{1}{2}$, so that the network operates reliably (although less so as one approaches the phase-transition boundary). Outside this region the the error probability equals $\tfrac{1}{2}$. In this region the network fails. In the limit of small $\alpha$ the critical noise level is $\beta_c=1$. In this regime the network is described by the theory explained in Section \ref{sec:mft}, Equation (\ref{eq:m1}). 
\index{error probability|)}

Alternatively these two different phases of the Hopfield network are characterised in terms of the \index{order parameter}order parameter $m_1$. We see that  $m_1\neq 0$ below the solid line, while $m_1=0$ above it, in the limit of large $N$.
\begin{figure}[t]
        \centering
        \begin{overpic}[scale=\myFigureScale]{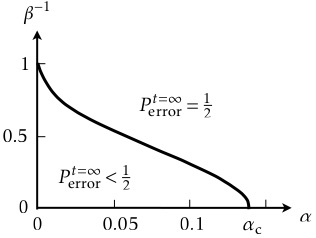}
        \end{overpic}
        \caption{\figlab{C3S4Phase} Phase diagram of the Hopfield network in the limit of large $N$ (schematic).
The region with $ P_{\rm error}^{t=\infty} < \tfrac{1}{2}$ is the ordered phase, the region with $ P_{\rm error}^{t=\infty} = \tfrac{1}{2}$ is the disordered phase. After Figure 2 in Ref.~\cite{AmitGutfreund1987}.}
\end{figure}

\section{Beyond mean-field theory}
\label{sec:bmt}
The theory  summarised in this Chapter rests on a \index{mean field theory} mean-field approximation for the local field\index{local field}, Equation (\ref{eq:mfa}). 
The main result is the phase diagram shown in Figure \figref{C3S4Phase},
derived in the limit $N\to\infty$. 
For smaller values of $N$ one expects the transition to be less sharp, so that $m_1$ is non-zero also for values of $\alpha$ larger than the critical
storage capacity $\alpha_{\rm c}$. 

But even for large values of $N$, 
the question remains how reliable the mean-field theory really is. 
\index{mean field theory}
To answer this question, one uses a more accurate theory, based on the
so-called
{\em replica trick}\index{replica!trick}.   
One starts from the steady-state distribution\index{distribution!steady state} of $\ve s$ for fixed patterns $\ve x^{(\mu)}$. 
In Chapter \ref{chapter:so} we will see that 
the steady-state distribution for the McCulloch-Pitts dynamics is the
{\em Boltzmann distribution}\index{Boltzmann distribution}
\begin{equation}
\label{eq:C3S4BoltzmannDistribution}
\PB(\ve s) = Z^{-1} {\rm e}^{-\beta H(\ve s)}
\end{equation}
(the proof in Chapter \ref{chapter:so} assumes that the diagonal weights are set to zero\index{weight!diagonal}). 
The normalisation factor $Z$ is called the {\em partition function}\index{partition function}
\begin{equation}
Z = \sum_{\ve s}{\rm e}^{-\beta H(\ve s)}\,.
\end{equation}
In  order to compute the \index{order parameter}order parameter, one adds a \index{threshold}threshold term to the energy function (\ref{eq:H})
\begin{equation}
\label{eq:H5}
H = -\frac{1}{2}\sum_{ij}w_{ij}s_{i}s_{j}+\sum_\mu \lambda_\mu\sum_i x_i^{(\mu)}s_i\,.
\end{equation}
Then the \index{order parameter}order parameter $m_\mu$ is obtained by taking a derivative w.r.t $\lambda_\mu$:
\begin{equation}
m_\mu = \Big\langle \frac{1}{N} \sum_i x_i^{(\mu)} \avg{n_i}
\Big\rangle
= -\frac{1}{N\beta} \frac{\partial}{\partial \lambda_\mu} \big\langle \log Z\big\rangle\,.
\end{equation}
The outer average is over different realisations of random patterns.\index{pattern!random} 
The logarithm of $Z$ is averaged using the replica trick. 
The idea is to represent the average of the logarithm as
\begin{equation}
\langle \log Z \rangle = \lim_{n\to 0} \tfrac{1}{n} (\langle Z^n\rangle-1)\,,
\end{equation}
The function $Z^n$ looks like the partition function\index{partition function}  of $n$ copies of the system, hence the name {\em replica}\index{replica} trick. If one assumes that all copies yield the same \index{order parameter}order parameter, one obtains the mean-field solution
described in Section \ref{sec:sc_mft}. If one allows different copies to have different \index{order parameter}order parameters ({\em replica-symmetry breaking}\index{replica!symmetry, breaking of}), \index{replica!symmetry}
\index{symmetry}
one obtains a more accurate solution for the \index{storage capacity!critical} critical storage capacity \cite{Steffan},
\begin{equation}
\label{eq:alphac_rsb}
\alpha_{\rm c} = 0.138187\,.
\end{equation}
The mean-field result (\ref{eq:alphac_mft}) differs only slightly from Equation (\ref{eq:alphac_rsb}).
The most precise Monte-Carlo simulations (Section \ref{sec:mcs}) for
finite values of $N$ \cite{Volk} yield upon extrapolation to $N=\infty$
\begin{equation}
\alpha_{\rm c} =0.143 \pm 0.002\,.
\end{equation}
This is close to, yet significantly different from the best theoretical estimate, Equation (\ref{eq:alphac_rsb}), and also different from the mean-field result (\ref{eq:alphac_mft}). 
\index{mean field theory}

To put these results into context, note that for other systems mean-field theories
tend to give results much worse than here.  Usually, mean-field theories yield at best a 
qualitative description of a phase transition. For the Hopfield network, by contrast, the mean-field theory works very \index{mean field theory}
well because every neuron is connected with every other neuron. This helps to average out the fluctuations in Equation (\ref{eq:mfa}).
In physical systems with local interactions, mean-field theories tend to work better in higher dimensions, because there are more neighbours to average over (Exercise~3.5).

\section{Correlated and non-random patterns}
\index{pattern!correlations}
\label{sec:cp1}
In the two previous Sections we assumed 
that the stored patterns \index{pattern!random} are random with independently identically distributed bits. This allowed us to calculate the storage capacity of the Hopfield network using the central-limit theorem. The hope is that the result describes what happens for typical, non-random patterns, or for \index{pattern!random} random patterns with correlated bits.
Correlations affect the distribution of the \index{cross talk term} cross-talk term, and thus the storage capacity of the Hopfield network.
It has been argued that
the storage capacity increases when the patterns are more strongly correlated, 
while others have claimed that the capacity decreases in this limit (see Ref.~\cite{Lowe} for a discussion).

For a set of definite patterns (no randomness to average over), the situation seems to be even more challenging. Yet there is
a way of modifying Hebb's rule to deal with this problem, at
least when the patterns are linearly independent.
 The recipe is explained by Hertz, Krogh,
and Palmer \cite{hertz1991introduction}. One simply
incorporates the overlaps\index{pattern!overlap}
\label{page:Q}
\begin{equation}
Q_{\mu\nu}= \frac{1}{N} \ve  x^{(\mu)} \cdot \ve  x^{(\nu)}
\end{equation} 
into Hebb's rule. To this end one defines the $p\times p$ {\em overlap matrix}\index{overlap matrix|textbf} $\ma Q$ with elements $Q_{\mu\nu}$ and writes:
\begin{equation}
w_{ij}= \frac{1}{N} \sum_{\mu\nu} x_i^{(\mu)} \left(\ma Q^{-1}\right)_{\mu\nu}  x^{(\nu)}_{j}\,.
\eqnlab{C5S1WeightsFormula}
\end{equation}
For orthogonal patterns\index{pattern!orthogonal} ($Q_{\mu\nu} =\delta_{\mu\nu}$), this modified Hebb's rule is identical to
Equation  \eqnref{C2S2HebbsRule}. For non-orthogonal patterns, the rule \eqnref{C5S1WeightsFormula} ensures that all patterns are recognised. 
Equation \eqnref{C5S1WeightsFormula} requires that the matrix $\ma Q$ is invertible: its columns must be linearly independent (and this implies that the rows are linearly independent too). This limits the number of patterns one can store with the rule \eqnref{C5S1WeightsFormula}, because $p>N$ implies linear dependence. 
\index{pattern!linearly dependent}

For linearly independent patterns one can find the weights $w_{ij}$ iteratively, by successive improvement from an arbitrary starting point. We can say that the network learns the task through a sequence of weight changes.  This is the idea used to solve \index{classification!task} classification tasks with perceptrons (Part \ref{part:supervised}). 

\section{Summary}
In this Chapter we analysed the stochastic dynamics of Hopfield networks. 
We asked under which circumstances the network dynamics can reliably recognise \index{pattern!stored}stored patterns. If the stored patterns are random, the performance of the Hopfield network depends on their number, on the number of bits per pattern, and upon the noise level.  The storage capacity $\alpha$ equals the ratio of the number of \index{pattern!stored}stored patterns to the number of bits per pattern. The network operates reliably when this ratio is small, and provided the noise level is not too large. A mean-field analysis of the $N\to\infty$-limit shows that there is a phase transition \index{phase transition} in the parameter plane of the Hopfield network (Figure \figref{C3S4Phase}): when $\alpha$ exceeds the \index{storage capacity!critical} critical storage capacity $\alpha_{\rm c}$, the network ceases to function. 

Hopfield networks share many properties with the networks discussed later on in this book. The most important point is perhaps that introducing noise in the dynamics allows to study the \index{convergence!Hopfield network}convergence and performance of the network: in the presence of noise there is a well-defined steady state\index{steady state} that can be analysed. Without noise, in the \index{deterministic limit}deterministic limit, the network dynamics arrests  in \index{local minimum}local minima of the energy function, and may not reach the \index{pattern!stored}stored patterns.  Naturally the noise must be small enough for the network to function accurately.  Finally, the building blocks of Hopfield networks are McCulloch-Pitts neurons and Hebb's rule for the weights.  \index{McCulloch Pitts neuron} Many of the algorithms discussed in the coming Chapters use these elements in some form.

\section{Further reading}
The statistical mechanics of Hopfield networks is explained in {\em Introduction to the theory of neural computation} by Hertz, Krogh, 
and Palmer \cite{hertz1991introduction}. Starting from the Boltzmann distribution, Chapter 10 in this book summarises how to compute the \index{order parameter}order parameters, and how to evaluate
the stability of the corresponding solutions.  For more details on the 
replica trick\index{replica!trick}, see 
the books by M\"uller, Reinhard and Strickland~\cite{MullerReinhardt}
and by Engel and van den Broeck \cite{engel2001statistical}, as well as the review article \cite{watkin1993statistical}.

\vfill\eject

\chapter{The Boltzmann distribution}

\label{chapter:so}
In Chapter \ref{ch:dhn}
we saw that the deterministic dynamics (\ref{eq:C2S2Asynch}) of Hopfield networks admits
the Lyapunov function\index{Lyapunov!function}
\begin{equation}
\label{eq:HB}
H = -\frac{1}{2} \sum_{ij} w_{ij} s_i s_j+\sum_i \theta_i s_i\,,
\end{equation}
if the weights $w_{ij}$ are \index{weight!symmetric weights}symmetric,
and $w_{ii}>0$.
In this Chapter\footnote{In this Chapter we set the
diagonal weights to zero\index{weight!diagonal}.} we show that the asynchronous stochastic
McCulloch-Pitts dynamics \eqnref{C3S2StochasticRule} \index{convergence!stochastic dynamics}converges to a steady state
where the state vector\index{state!vector} $\ve s$ follows the Boltzmann distribution\index{Boltzmann distribution|textbf}
\begin{align}
\label{eq:SvDist}
\PB(\ve s) = Z^{-1} {\rm e}^{-\beta H(\ve s)}
\quad\mbox{with normalisation}\quad Z = \sum_{\ve s} {\rm e}^{-\beta H(\ve s)}\,.
\end{align}
The stochastic dynamics \eqnref{C3S2StochasticRule} is closely related to that of {\em Markov-chain Monte-Carlo}\index{Markov chain Monte Carlo} algorithms, designed to efficiently sample
from the Boltzmann distribution. 
\index{dynamics!stochastic}
We also discuss how to solve optimisation tasks by Monte-Carlo simulation: one assigns a suitable energy $H$ to each configuration $\ve s$, 
so that the function $H(\ve s)$  \index{energy function} has global minimum for the optimal configuration $\ve s_{\rm min}$.  
The stochastic dynamics finds low-energy configurations (but not necessarily $\ve s_{\rm min}$), in particular
if one iteratively decreases the noise level\index{noise level} by increasing $\beta$ ({\em simulated annealing}\index{annealing, simulated} \cite{Kirkpatrick}).

Last but not least we look at {\em Boltzmann machines}\index{Boltzmann machine} \cite{HintonSejnowski,Hinton_Boltzmann_2,Hinton_Boltzmann_3,Hinton_Boltzmann,MacKay}, stochastic Hopfield networks with {\em hidden neurons}\index{hidden neuron|textbf} that are neither used for input nor for output. 
Boltzmann machines can be {\em trained}\index{training} to learn the properties of a distribution $\Pd(\ve x)$ of binary input patterns\index{input pattern!binary} $\ve x$. The idea is to iteratively change the weights in Equation (\ref{eq:HB}) until the Boltzmann distribution represents the \index{input distribution}input distribution. This idea,\index{Boltzmann distribution}  
to iterate the weights until the network learns the \index{input distribution}input distribution $\Pd$, is used in a slightly different form in {\em supervised learning}\index{supervised learning} (Part \ref{part:supervised}).
Boltzmann machines are closely related to Hopfield networks. Without {\em hidden neurons}, both models learn
\index{Hopfield network}
to represent \index{pattern!two point correlations}two-point correlations $\langle x_i^{(\mu)} x_j^{(\mu)}\rangle$
of pattern bits. 

When important information about the inputs is
encoded in higher-order correlations, one can use  {\em hidden neurons} to represent 
\index{pattern!correlations}
these correlations.\index{hidden neuron}
Generally Boltzmann machines are hard to train, in particular if they have many hidden neurons. {\em Restricted Boltzmann machines} \index{restricted Boltzmann machine} are neural networks with hidden neurons, but with fewer connections: 
only those between {\em visible}\index{neuron!visible} and hidden neurons are allowed. These neural networks can be fairly efficiently trained and can solve a number of different tasks. Apart from learning a distribution of input patterns, they can for instance be trained to recognise incomplete input patterns, and to classify inputs~\cite{FischerIgel}.

\section{Convergence of the stochastic dynamics}
\label{sec:C4conv}
We begin by showing that the \index{dynamics!stochastic} stochastic dynamics \eqnref{C3S2StochasticRule} has a steady state where $\ve s$ is distributed according to the Boltzmann 
distribution (\ref{eq:SvDist}). To this end, we consider  an
alternative yet equivalent formulation of the network dynamics.\index{network!dynamics}
It consists of two parts. First, choose a neuron randomly, number $m$ say.
Second, change  $s_{m}$ to $s_{m}' \neq s_{m}$ with probability
\label{nd}
\begin{subequations}
\label{eq:move}
    \begin{equation}
        \text{Prob}(s_{m} \to s'_{m}) = \frac{1}{1+e^{\beta \Delta    H_{m}}}\,,
    \end{equation}
    with  
    \begin{equation}
        \Delta  H_{m} = H(\ldots,s'_{m},\ldots) - H(\ldots,s_{m},\ldots)\,.
    \end{equation}
\end{subequations}
To explore the relation between the stochastic rules (\ref{eq:move}) and \eqnref{C3S2StochasticRule},  
we use that 
\begin{equation}
\label{eq:dh9}
\Delta H_{m} = -b_{m}(s'_{m} - s_{m})
\end{equation}
 with local field $b_m = \sum_j w_{mj} s_j-\theta_m$. \index{local field}
To derive Equation (\ref{eq:dh9}), we assumed that the weights are \index{weight!symmetric weights}symmetric, and that the diagonal weights vanish.
The result is obtained with a calculation similar to the one leading to Equation (\ref{eq:deltaH}), except that we have non-zero 
\index{threshold}thresholds here. 
To proceed,  \index{weight!diagonal}
we break the rule (\ref{eq:move}) up into different cases.
The state of neuron $m$ changes with probability
\begin{subequations}
\label{eq:cases2}
\begin{align}
\text{if } ~ s_{m} & = -1 & \text{obtain} ~  s'_{m} & =  1 ~ \text{with prob.} & \frac{1}{1+e^{-2\beta b_{m}}} & = \pr(b_{m})\,, \\
\text{if } ~ s_{m} & =  1 & \text{obtain} ~  s'_{m} & = -1 ~ \text{with prob.} & \frac{1}{1+e^{ 2\beta b_{m}}} & = 1-\pr(b_{m})\,. 
\end{align}
In the second row we used that  $1-\pr(b) = 1- \frac{1}{1+ {\rm e}^{{-2\beta b}}} = \frac{1+ {\rm e}^{-2\beta b}-1}{1+ {\rm e}^{-2\beta b}} = \frac{1}{1+ {\rm e}^{{2\beta b}}}$.
The state remains unchanged with 
probability:
\begin{align}
\text{if } ~ s_{m} & = -1 &  \text{obtain } ~  s_{m}' & = -1  ~ \text{ with prob.} &  1-\pr(b_{m})& =  \frac{1}{1+e^{2\beta b_{m}}}\,,\\
\text{if } ~ s_{m} & =  1 &  \text{obtain } ~  s_{m}' & =  1  ~ \text{ with prob.} &    \pr(b_{m})& =  \frac{1}{1+e^{-2\beta b_{m}}}\,.
\end{align}
\end{subequations}
Comparing with Equation \eqnref{C3S2StochasticRule} we conclude  that the two schemes \eqnref{C3S2StochasticRule} and  (\ref{eq:move}) 
are equivalent under the assumptions made ($w_{ij}= w_{ji}$ and $w_{ii}=0$).
Note that Equation (\ref{eq:move}) is more general than the stochastic Hopfield dynamics, because
it does not require the energy function\index{energy function}  to be of the form (\ref{eq:HB}).  In particular it is neither needed that
the weights are \index{weight!symmetric weights}symmetric, nor that the diagonal weights vanish.
\index{weight!diagonal}
Equations \eqnref{C3S2StochasticRule} and (\ref{eq:move}) are not equivalent if these conditions are not satisfied (Exercise~4.1).

The rule (\ref{eq:move}) defines
a {\em Markov chain}\index{Markov chain|textbf} 
of states 
\begin{align}
\label{eq:markov}
\ve s_{t=0} \to \ve s_{t=1} \to \ve s_{t=2} \to \ldots
\end{align}
As before, the index $t$ counts the iteration steps.
A Markov chain is a {\em memoryless}\index{sequence, memoryless} random sequence of states defined by 
{\em transition probabilities}\index{transition probability|textbf} 
$p(\ve s'|\ve s)$
from state $\ve s$~to~$\ve s'$  \cite{VanKampen2007}.
The \index{transition probability}transition probability $p(\ve s'|\ve s)$ connects arbitrary states. 
One distinguishes between {\em local moves}\index{Markov chain Monte Carlo!local move|textbf} where only one neuron
may change, as above, and {\em global moves}\index{Markov chain Monte Carlo!global move|textbf} where many neurons may  change their states in a single step.

In both cases, an update consists of two parts.
First, a new state $\ve s'$ is suggested with probability $q(\ve s'|\ve s)$. 
Second, the new state
$\ve s'$  is accepted with {\em acceptance probability}\index{acceptance probability} 
\begin{equation} \label{eq:plk}
p_{\rm a}(\ve s'|\ve s) = \frac{1}{1+{\rm e}^{\beta\Delta H}}\quad\mbox{with}\quad 
\Delta H = H(\ve s')-H(\ve s)\,.
\end{equation}
As result, the \index{transition probability}transition probability is given by a product
of two factors
\begin{align}
\label{eq:qpa}
p(\ve s'|\ve s)=q(\ve s'|\ve s) p_{\rm a}(\ve s'|\ve s)\,.
\end{align}
These steps are repeated many times, creating the chain of states (\ref{eq:markov}). 

The Markov chain  defined by the \index{transition probability}transition probability (\ref{eq:qpa})
has the Boltzmann distribution (\ref{eq:SvDist}) as a steady-state distribution if \index{Boltzmann distribution}\index{distribution!steady state}
the {\em detailed-balance}\index{detailed balance} condition is satisfied:
\begin{equation}
\label{eq:dbc}
p(\ve s'|\ve s) \PB(\ve s)  = p(\ve s|\ve s')\PB(\ve s')\,.
\end{equation}
Note that this is a sufficient condition, not a necessary one \cite{sokal1997monte}.
There are Markov chains that do not satisfy detailed balance
but still have a steady state (Exercise 4.4).
Usually detailed balance implies not
only that the Markov chain
has $\PB(\ve s)$ as a steady-state distribution,
but also that the distribution of states generated by the sequence (\ref{eq:markov})
\index{convergence!Markov chain}converges to $\PB(\ve s)$, see Ref.~\cite{VanKampen2007} for details.

To prove that the detailed-balance  condition (\ref{eq:dbc}) holds for the \index{transition probability}transition probability 
(\ref{eq:qpa}), assume that a single neuron is picked randomly 
with uniform probability 
\begin{equation}
 \label{eq:suggest}
 q=N^{-1}\,,
 \end{equation}
where $N$ is the number of neurons in the network. Since $q$ does
not depend on either $\ve s$ or $\ve s'$, the probability
of suggesting a new state is clearly symmetric.
Equations (\ref{eq:SvDist}), (\ref{eq:plk}) then imply:
\begin{align}
\frac{q{\rm e}^{-\beta H(\ve s)}}{ 1+{\rm e}^{\beta [H(\ve s')-H(\ve s)]}}
&=\frac{q}{ {\rm e}^{\beta H(\ve s')} +{\rm e}^{\beta H(\ve s)}}=
\frac{q{\rm e}^{-\beta H(\ve s')} }{ 1+{\rm e}^{\beta [H(\ve s)-H(\ve s')]}}\,.
\end{align}
This demonstrates that the Boltzmann distribution is a steady state\index{steady state} of the Markov chain
defined by (\ref{eq:plk}), (\ref{eq:qpa}), and (\ref{eq:suggest}).
As a consequence, the Boltzmann distribution is a steady state of the Markov chain.
If the simulation \index{convergence!Markov chain}converges to the steady state\index{steady state} (as it usually does), then states visited by the Markov chain are distributed according to the Boltzmann distribution. 
This also means that the steady-state distribution for the Hopfield model is the Boltzmann distribution, as stated in Section~\ref{sec:bmt}.

It is important to stress that Equation (\ref{eq:dbc}) is  
a condition for the \index{transition probability}transition probability
$p(\ve s'|\ve s)=q(\ve s'|\ve s)p_{\rm a}(\ve s'|\ve s)$, 
not just for the acceptance probability $p_{\rm a}(\ve s'|\ve s)$.  
For the \index{Markov chain Monte Carlo!local move}local moves discussed above,
$q$ is a constant, so that $p(\ve s'|\ve s)\propto p_{\rm a}(\ve s'|\ve s)$. In this case it is sufficient
to check the detailed-balance condition for the acceptance probability. In general,
and in particular for \index{Markov chain Monte Carlo!local move}global moves, it is necessary to include
$q(\ve s'|\ve s)$
in the detailed-balance check \cite{Mehlig92}.

\section{Monte-Carlo simulation}
\index{Markov chain Monte Carlo|(}
\label{sec:mcs}
The Markov chain described in the previous Section is the basis for the {\em Markov-chain Monte-Carlo}\index{Markov chain Monte Carlo} algorithm. 
This method is widely used in statistical physics and in mathematical statistics. It is therefore important to understand the connections between the different formulations.

The \index{Boltzmann distribution} Boltzmann distribution describes the probabilities of observing configurations of a large class of physical systems in their steady states \cite{Kadanoff}.  The statistical mechanics of systems with energy function\index{energy function} (also
called {\em Hamiltonian}\index{Hamiltonian}) $H$ shows that their configurations are distributed according to the Boltzmann distribution in thermodynamic equilibrium at a given 
temperature\index{temperature|textbf} $T$ (in this context $\beta^{-1} = k_{\rm B}T$ where $k_{\rm B}$ is the Boltzmann constant), and free from any other constraints.  If we denote the configuration of a system by the vector $\ve s$, then the Boltzmann distribution 
takes the form  (\ref{eq:SvDist}). 
The normalisation factor $Z = \sum_{\ve s} {\rm e}^{-\beta H(\ve s)}$ is also called {\em partition function}\index{partition function}.  For systems with a large number of interacting degrees of freedom, the partition function 
can be very expensive to compute, because the sum over $\ve s$ contains many terms.  
Therefore, instead of computing the distribution directly one generates
a Markov chain of states with a suitable \index{transition probability}transition probability, for instance (\ref{eq:move}).

In practice one often uses a slightly different form of the \index{transition probability}transition probability ({\em Metropolis algorithm} \cite{metropolis1953equation})\index{Metropolis algorithm}. Assuming that $q$ is constant, one takes:
\begin{equation}
\label{eq:plk2}
p(\ve s'|\ve s) =  q
 \left\{\begin{array}{ll}
 {\rm e}^{-\beta\Delta H} & \mbox{when}\quad \Delta H>0\,,\\
 1                        & \mbox{when}\quad \Delta H\leq 0\,,
 \end{array}\right .
\end{equation}
with  $\Delta H = H(\ve s')-H(\ve s)$ as before. 
That the Metropolis rates obey 
the detailed-balance condition (\ref{eq:dbc}) can be seen as follows:
\begin{align}
\nonumber
p(\ve s'|\ve s)\PB(\ve s) 
 &=qZ^{-1} {\rm e}^{-\beta H(\ve s)} \left \{ \begin{array}{ll}
                                       {\rm e}^{-\beta[H(\ve s')-H(\ve s)]} & \mbox{if}\quad H(\ve s')> H(\ve s)\\
                                       1 & {\rm otherwise}
                                       \end{array}\right .\\
                            & =qZ^{-1} {\rm e}^{-\beta {\rm max}\{H(\ve s),H(\ve s')\}}\\
\nonumber &= qZ^{-1}{\rm e}^{-\beta H(\ve s')}\left \{ \begin{array}{ll}
                                       {\rm e}^{-\beta[H(\ve s)-H(\ve s)]} & \mbox{if}\quad H(\ve s)> H(\ve s')\\
                                       1 & {\rm otherwise}
                                       \end{array}\right .\\
\nonumber &=   p(\ve s|\ve s')\PB(\ve s')\,.
\end{align}
The Metropolis algorithm is summarised in Algorithm \ref{metropolis}.
It provides an elegant way of computing the average $\langle A\rangle$ of an 
\index{observable} observable $A(\ve s)$ 
over the Boltzmann distribution of~$\ve s$:\index{Boltzmann distribution}
\begin{equation}
\label{eq:sumA7}
\langle A\rangle=Z^{-1} \sum_{\ve s}\, A(\ve s)\,{ {\rm e}^{-\beta H(\ve s)}}
 \approx \frac{1}{T} \sum_{t=1}^{T} A(\ve s_t)\,.
\end{equation}
This particular way of evaluating the average $\langle A\rangle$ is a special case of the more general
method of {\em importance sampling}\index{importance sampling} \cite{Binder}.
The central-limit theorem implies that the error of this estimate for $\langle A\rangle$ decreases $\propto T^{-1/2}$ as $T$ increases. The prefactor is determined by the correlations between subsequent terms in the sum (\ref{eq:sumA7}): the states in the sequence (\ref{eq:markov}) are correlated\index{state!correlations}, in particular when the moves are local\index{Markov chain Monte Carlo!local move}, because then subsequent configurations are similar. Generating many quite strongly correlated samples from a distribution is not a very efficient way of sampling this distribution. Sometimes it may be more efficient to suggest global moves instead, in order to avoid that subsequent
\index{Markov chain Monte Carlo!global move}
states in the Markov chain are similar.
But it is not guaranteed that global moves lead to weaker correlations. For global moves, $\Delta H$ may be more likely to assume large positive values, so that fewer suggested moves are accepted. As a consequence the Markov chain may stay in certain states for a long time, increasing correlations in the sequence. Usually a compromise is most efficient, moves that are neither local nor global.
In summary, the \index{convergence!Monte Carlo sampling}convergence of Monte-Carlo sampling is quite slow. This motivated Sokal
to begin his lecture notes on Monte-Carlo simulation with the warning \cite{sokal1997monte}
\vspace*{3mm}

\hspace*{8mm}\begin{minipage}{12cm}
{\em Monte Carlo is an extremely bad method; it should be used only when all alternative methods are worse.}
\end{minipage}
\vspace*{3mm}

\noindent 
Monte-Carlo algorithms are very widely used, and the original reference for the Metropolis algorithm \cite{metropolis1953equation} is generally considered one of the most significant scientific papers in computational physics.
Sokal's point is of course that many problems cannot be solved in any other way, so that Monte-Carlo simulation
is the only option.  But we should be aware of the shortcomings of the method. The same caution applies more generally to the topic of this book, machine-learning algorithms with neural networks.
\begin{algorithm}[t]
\caption{\label{metropolis} Metropolis algorithm for symmetric $q(\ve s'|\ve s)$}
\begin{algorithmic}
\STATE initialise $\ve s = \ve s_0$;
\FOR {$t=1,\ldots, T$}
\STATE suggest a new  state $\ve s'$ with probability $q(\ve s'|\ve s)$      ;
\STATE compute $\Delta H = H(\ve s') - H(\ve s)$;
\IF {$\Delta H \leq 0$}
    \STATE accept the new state: $\ve s = \ve s'$;
\ELSE
   \STATE draw a random number $r$ uniformly distributed in $[0,1]$;
   \IF {$ r< \exp(-\beta\Delta H)$}
      \STATE accept the new state: $\ve s = \ve s'$;
   \ELSE
      \STATE reject $\ve s'$;
   \ENDIF
\ENDIF
\STATE sample $\ve s_{\!t} = \ve s$ and $A_t = A(\ve s_t)$;
\ENDFOR
\end{algorithmic}
\end{algorithm}
\index{Markov chain Monte Carlo|(}

\section{Simulated annealing}
\label{sec:cop}
\begin{figure}[b!]
        \centering
        \begin{overpic}[scale=\myFigureScale]{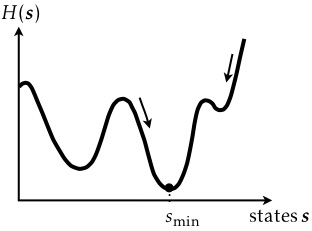}
        \end{overpic}
\vspace{7pt}
        \caption{\figlab{C4S1EnergyFunction}
 Schematic. Simulated annealing (arrows) tends to reduce the energy function\index{energy function}. Noise helps to avoid
that the dynamics arrests in a \index{local minimum}local minimum.}
\end{figure}

Combinatorial optimisation problems admit $2^{k}$ or $k!$ configurations - 
\index{combinatorial optimisation|textbf}
too many to find the optimal one by complete enumeration when $k$ is large.
\index{complete enumeration} 
An alternative strategy is to assign an energy $H(\ve s)$ to each configuration $\ve s$ so that $H$ is minimal at the optimal configuration
$\ve s_{\rm min}$. One minimises $H(\ve s)$ by \index{Markov chain Monte Carlo} Monte-Carlo simulation,
using that the Monte-Carlo dynamics tends to decrease $H$ when the \index{temperature}temperature
$k_{\rm B}T=\beta^{-1}$  is low, 
Figure \figref{C4S1EnergyFunction}. A common strategy is to lower the \index{temperature}temperature on the fly. In the beginning of the simulation, the \index{temperature}temperature is high, so that the dynamics first explores the rough features of the energy landscape\index{energy landscape|textbf}. When the \index{temperature}temperature is lowered, the dynamics perceives finer and finer features of $H(\ve s)$. The hope is that it ends up in the global minimum
$H_{\rm min} = H(\ve s_{\rm min})$ at zero \index{temperature}temperature,
where $\PB(\ve s)=0$ when $H(\ve s)>H_{\rm min}$ and $\PB(\ve s)>0$ only for $H(\ve s)=H_{\rm min}$.
This method is called {\em simulated annealing}\index{annealing, simulated|textbf} \cite{Kirkpatrick}, see also Section~10.9 in {\em Numerical Recipes} \cite{Recipes}.
Slowly lowering the \index{temperature}temperature during the simulation mimics the slow cooling of a physical system. It passes through a sequence of quasi-equilibrium Boltzmann distributions with lower and lower \index{temperature}temperatures, until the system hopefully finds the global minimum $H_{\rm min}$.\index{Boltzmann distribution}

For a number of combinatorial optimisation problems one can write
\index{combinatorial optimisation}
down energy functions\index{energy function} that have the same form as Equation (\ref{eq:H_with_thresholds}) with \index{weight!symmetric weights}symmetric weights \cite{HopfieldTank}. 
Since $s_{ij}^2=1$, one can always assume that the diagonal weights
\index{weight!diagonal}
vanish, because they make only a constant contribution to $H$. In short, 
one can use the Hopfield dynamics \eqnref{C3S2StochasticRule} to minimise $H$. 
The {\em travelling-salesman} problem has been solved in this way \cite{HopfieldTank,hertz1991introduction}, gradually reducing the noise level as one iterates
\index{noise level}
the stochastic dynamics. \index{dynamics!stochastic}

It is by no means necessary to use a Hopfield model for this purpose. 
Instead we can just use the stochastic dynamics (\ref{eq:move}) or the Metropolis  algorithm (\ref{eq:plk2}) to solve combinatorial optimisation problems by simulated annealing. 
\index{simulated annealing}
\index{combinatorial optimisation}
Nevertheless, a crucial step is to find a suitable energy function. 

As an example, consider
the {\em double-digest problem}\index{double digest problem}. It arose when sequencing the human genome \cite{Waterman,lander2001initial}.\index{human genome sequence}
The human genome sequence was first assembled by piecing together overlapping 
\index{DNA segment}DNA segments in the right order by making sure that overlapping segments share the same \index{DNA sequence}DNA sequence.
To this end it is necessary to uniquely identify the DNA segments. The actual DNA sequence of a segment is a unique identifier. But it is  sufficient and more efficient to identify a
DNA segment by a {\em fingerprint}\index{fingerprint}, for example the sequence of {\em restriction sites}\index{restriction!site}.
These are short subsequences (four or six base pairs long) that are recognised by \index{restriction!enzyme}enzymes that cut  ({\em digest}\index{digest}) the DNA strand precisely at these sites. A DNA segment is identified by the types and locations of \index{restriction!site} restriction sites that it contains, the so-called {\em restriction map}\index{restriction!map}.
\begin{table}[t]
\centering
\begin{tabular}{l}
$L=10000$\\
\small $a=$ \tt\tiny   [5976, 1543, 1319, 1120, 42]\\
\small $b=$ \tt\tiny   [4513, 2823, 2057, 607]\\
\small $c=$ \tt\tiny   [4513, 1543, 1319, 1120, 607, 514, 342, 42]\\[0.2cm] $L=20000$\\
\small $a=$ \tt\tiny    [8479, 4868, 3696, 2646, 169, 142]\\
\small $b=$ \tt\tiny [11968, 5026, 1081, 1050, 691, 184]\\
\small $c=$ \tt\tiny [8479, 4167, 2646, 1081, 881, 859, 701, 691, 184, 169, 142]\\[0.2cm] $L=40000$\\
\small $a=$ \tt\tiny  [9979, 9348, 8022, 4020, 2693, 1892, 1714, 1371, 510, 451]\\
\small $b=$ \tt\tiny  [9492, 8453, 7749, 7365, 2292, 2180, 1023, 959, 278, 124, 85]\\
\small $c=$ \tt\tiny  [7042, 5608, 5464, 4371, 3884, 3121, 1901, 1768, 1590, 959, 899, 707, 702, 510, 451, 412, \\ \phantom{\small $c=$}\phantom{\tt\tiny  [} \tt\tiny 278, 124, 124, 85]
\end{tabular}
\caption{\tablab{C4S4Segments} Example configurations for the double-digest problem \cite{Waterman} for three different chromosome lengths $L$. For each example, three ordered \index{restriction!fragment}fragment sets are given, corresponding to the result of digestion with A, with B, and with both A and B.}
\end{table}

\index{restriction!fragment}
When a  \index{DNA segment}DNA segment is cut by two different \index{restriction!enzyme}enzymes one can  experimentally determine the  lengths of the resulting \index{restriction!fragment}fragments.  Is it possible to determine  how the cuts were ordered in the \index{DNA sequence}DNA sequence of the segment from the \index{restriction!fragment length}fragment lengths, to find the \index{restriction!map}restriction map? This is the double-digest problem \cite{Waterman}. 
In a double-digest experiment,  a given DNA sequence is first digested by one \index{restriction!enzyme}enzyme ($A$ say). Assume that this results in $n$ \index{restriction!fragment}fragments with lengths $a_i$ ($i=1,\ldots,n$).  Second, the DNA sequence  is digested  by another \index{restriction!enzyme}enzyme, $B$. In this case $m$ \index{restriction!fragment}fragments are found, with lengths  $b_1, b_2,\ldots, b_m$.
Third, the DNA sequence is digested with both \index{restriction!enzyme}enzymes $A$ and $B$, yielding $l$ \index{restriction!fragment}fragments with lengths $c_1,\ldots,c_l$, see Table  \tabref{C4S4Segments} for examples.
The task is now to determine all possible orderings of the $a$- and $b$-cuts
that result in  $l$ \index{restriction!fragment}fragments with lengths $c_1,c_2,\ldots,c_l$.\index{restriction!fragment}
Since the solutions of the double-digest problem are degenerate, an important 
question is \index{solution!degenerate}
to determine how many distinct solutions there are (Exercise 4.5).

\index{energy function}
To write down an energy function, denote the ordered set of \index{restriction!fragment length}fragment lengths  produced by digesting with \index{restriction!enzyme}enzyme $A$ by $a=\{a_1,\ldots,a_n\}$, where $a_1\ge a_2\ge \ldots\ge a_n\ge 1$. Similarly $b=\{b_1,\ldots,b_m\}$ ($b_1\ge b_2\ge \ldots\ge b_m\ge 1$) for  \index{restriction!fragment length}fragment lengths produced by \index{restriction!enzyme}enzyme $B$, and $c=\{c_1,\ldots,c_l\}$ ($c_1\ge c_2\ge \ldots\ge c_l\ge 1$) for \index{restriction!fragment length}fragment lengths produced by digesting first with $A$ and then with $B$.  Permutations $\sigma$ and $\mu$  of the sets $a$ and $b$ result in a set of $c$-fragments that we call $\hat c(\sigma,\mu)$. Solutions
of the double-digest problem correspond to  permutations  $[\sigma,\mu]$
that  yield $\hat c(\sigma,\mu)=c$. A suitable energy function is therefore
\begin{align}
\label{eq:DDPH}
H(\sigma,\mu) = \sum_j c_j^{-1} [c_j-\hat c_j(\sigma,\mu)]^2\,,
\end{align}
and configuration space\index{configuration space} is the space of all permutation pairs $\ve s= [\sigma,\mu]$.
Local moves in \index{cross talk term} configuration space correspond 
to inversions of short subsequence of $\sigma$ and/or $\mu$. 
One can show that the corresponding $q(\ve s'|\ve s)$ is symmetric 
(Exercise 4.5).
As mentioned above, this is necessary for the stochastic dynamics to 
\index{convergence!stochastic dynamics}converge in its simplest form,
Equation (\ref{eq:plk2}) and Algorithm \ref{metropolis}. \index{dynamics!stochastic}

For the simulation one chooses a larger \index{temperature}temperature $k_{\rm B}T = \beta^{-1}$ to begin with, so that the stochastic dynamics explores the rough features of the \index{energy landscape}energy landscape at first. As the simulation proceeds,
the \index{temperature}temperature is gradually reduced. This allows the dynamics to learn finer features of the landscape, as described above.

\section{Boltzmann machines}
\index{Boltzmann machine|(}
\label{sec:bm1}
Boltzmann machines are generalised Hopfield networks
that can learn to approximate data distributions of binary input patterns. 
\index{input pattern!binary}
Boltzmann machines differ from Hopfield networks
\index{Hopfield network}
in two essential ways. First, instead of using Hebb's rule, the weights are adjusted until the Boltzmann 
machine approximates the data distribution precisely. The weights are iteratively refined to minimise the difference
between the data distribution and the model (the Boltzmann distribution). Nevertheless, this procedure is closely related
to Hebb's rule, as we shall see. Second, to represent higher-order correlations between bits of input patterns, Boltzmann machines\index{input pattern}  
employ hidden neurons.\index{hidden neuron}

\begin{figure}[t]
  \centering
    \begin{overpic}[scale=\myFigureScale]{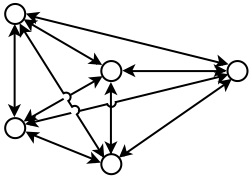}
    \end{overpic}
\caption{\figlab{BS1Boltzmann} Boltzmann machine\index{Boltzmann machine} with
five neurons. All weights are symmetric,
the \index{weight!diagonal} diagonal weights are set to zero. The states of the neurons are denoted by $s_i = \pm 1$. This neural network has no hidden units. It looks like a Hopfield network, but the weights are not given by Hebb's rule.}
\end{figure}
We begin with Boltzmann machines without
hidden neurons (Figure \figref{BS1Boltzmann}),
because they are simpler to analyse. Then 
we discuss why hidden neurons are necessary 
to learn the properties of general \index{input distribution}input distributions $\Pd(\ve x)$
of binary inputs $\ve x$. The training algorithm\index{training!algorithm}
for Boltzmann machines with hidden neurons is described in Section \ref{sec:rbm}. 

The goal of the training algorithm is to find weights so that the 
Boltzmann distribution \index{Boltzmann distribution}
\begin{align}
\label{eq:approx}
\PB(\ve s = \ve x) = Z^{-1} {\rm exp}\Big({\tfrac{1}{2}\sum_{i\neq j} w_{ij} x_i x_j}\Big)
\end{align}
approximates the distribution $\Pd(\ve x)$ as precisely as possible. Here and in the 
remainder of this Chapter we set $\beta=1$. 
 The input patterns have $N$ binary bits \index{input pattern!binary}
[Equation (\ref{eq:C2PatternVector})] with values $\pm 1$. The \index{weight!matrix}weight matrix $\ma W$ is symmetric, $w_{ij}=w_{ji}$, and its 
diagonal elements are set to zero, $w_{ii}=0$.\index{weight!diagonal} In this Section we also set the \index{threshold}thresholds to zero.

The Boltzmann machine is trained by iteratively adjusting the weights $w_{ij}$, using a sequence  of input patterns\index{input pattern|textbf}  $\ve x^{(\mu)}$ ($\mu=1,\ldots,p$), independently sampled from the data distribution $\Pd(\ve x)$. 
This is achieved by maximising the {\em likelihood}\index{likelihood}
$\mathscr{L}=\prod_{\mu=1}^p \PB(\ve s=\ve x^{(\mu)})$
that the Boltzmann machine produces the sequence $\ve x^{(1)},\ldots,\ve x^{(p)}$ of input patterns. Any pattern may appear more than once in the sequence, with frequency proportional to $\Pd(\ve x)$. Maximising $\mathscr{L}$
therefore corresponds to approximating the data distribution as accurately as possible. 
Usually one maximises the logarithm of the likelihood,
the {\em log-likelihood}\index{likelihood} function
\begin{align}
\label{eq:ll}
\log \mathscr{L} =\log \prod_{\mu=1}^p \PB(\ve s=\ve x^{(\mu)})
 =\sum_{\mu=1}^p \log\PB(\ve s=\ve x^{(\mu)})\,.
\end{align}
The logarithm is a monotonic function, so the log-likelihood has its maximum at the same weight values as the likelihood. Taking the logarithm simplifies 
the analysis of the learning algorithm, 
because $\log \PB(\ve s=\ve s^{(\mu)})$ is simply a quadratic function
of $x_j^{(\mu)}$. Also, a learning algorithm based on the log-likelihood 
is usually more stable numerically.

A different reasoning behind maximising the log-likelihood starts from
the {\em Kullback-Leibler divergence},
defined as \index{Kullback Leibler divergence|textbf}
\begin{align}
\label{eq:KLD}
\DKL = \sum_{\mu=1}^p \Pd(\ve x^{(\mu)}) \log [\Pd(\ve x^{(\mu)})/\PB(\ve s=\ve x^{(\mu)})]\,.
\end{align}
Terms in the sum with $\Pd(\ve x^{(\mu)}) {=0}$ are set to zero, and $\DKL$ is defined to equal infinity when
there are patterns for which $\PB=0$ but $\Pd\neq 0$. The Kullback-Leibler divergence is
a measure of the difference between the two distributions: $\DKL$ is non-negative, and it assumes its global minimum $\DKL=0$ for $\Pd(\ve x^{(\mu)})=\PB(\ve s=\ve x^{(\mu)})$, see Exercise 4.6. We infer from Equation (\ref{eq:KLD}) that minimising $\DKL$ corresponds to maximising $\log\mathscr{L}$.

To find the global maximum of the log-likelihood, we use {\em gradient ascent}\index{gradient ascent}:
we repeatedly change the weights by adding increments
\index{weight increment}
\begin{equation}
\label{eq:grad_asc}
w_{mn}' = w_{mn}+\delta\! w_{mn}\quad\mbox{with}\quad
\delta\! w_{mn} = \eta \frac{\partial \log\mathscr{L}}{\partial w_{mn}} \,.
\end{equation}
The small parameter $\eta>0$ is the {\em learning rate}\index{learning rate|textbf}.
The gradient points in the steepest uphill direction of  $\mathscr{L}$.
The idea is to take many small uphill steps until one hopefully (but not necessarily) reaches the global maximum. 
Since the likelihood is a product of many possibly quite small factors,
$\mathscr{L}$ can become very small. This can lead to numerical instabilities.
Maximising $\log\mathscr{L}$ instead of $\mathscr{L}$ can be more stable because it
yields an additional factor $\mathscr{L}^{-1}$ in the gradient: 
$\partial \log\mathscr{L}/\partial w_{mn} = \mathscr{L}^{-1} \partial \mathscr{L}/\partial w_{mn}$. 

To evaluate the gradient of $\mathscr{L}$ we start from Eq.~(\ref{eq:ll})
\index{log likelihood!gradient}
\begin{equation}
\label{eq:split4}
\log\mathscr{L} = \sum_{\mu=1}^p \Big[ -\log Z + \tfrac{1}{2}\sum_{i\neq j} w_{ij} x_i^{(\mu)} x_j^{(\mu)}\Big]\,.
\end{equation} 
\index{weight!diagonal}
This expression assumes that the diagonal weights vanish, just
like Equation (\ref{eq:approx}).
The first step is to evaluate the derivative of 
\begin{align}
\log Z = \log \sum_{s_1=\pm 1,\ldots,s_N=\pm 1} {\rm exp}\Big( \tfrac{1}{2}\sum_{i\neq j} w_{ij} s_i s_j\Big)\,.
\end{align}
To compute $\partial \log Z/\partial w_{mn}$ we use the chain rule together with
\begin{equation}
\label{eq:Kronecker_w_0}
\frac{\partial w_{ij}}{\partial w_{mn}} = \delta_{im} \delta_{jn}+\delta_{jm} \delta_{in}\,.
\end{equation}
This relation is
valid for symmetric weights and provided that $i\neq j$ and $m\neq n$.  In Equation (\ref{eq:Kronecker_w_0}), $\delta_{kl}$ is the Kronecker delta\index{Kronecker delta}, $\delta_{kl}=1$ if $k=l$ and zero otherwise (Chapter \ref{ch:dhn}). In particular, $\delta_{im}\delta_{jn}=1$ only if $i=m$ and $j=n$. Otherwise the product of Kronecker deltas equals zero.
Equation (\ref{eq:Kronecker_w_0}) is illustrated by the following story (a modification of a well-known maths joke):
\vspace*{3mm}

\hspace*{8mm}\begin{minipage}{12cm}
{\em The linear function, $x$, and the constant function are going for a walk. When they suddenly see the derivative approaching, the constant function gets worried. "I'm not worried" says the function $x$ confidently, "I'm not put to zero by the derivative."
When the derivative comes closer, it says "Hi! I'm $\partial/\partial \!y$. How are you?"}
\end{minipage}
\vspace*{3mm}

\noindent The moral is: since $x$ and $y$ are independent variables,
${\partial}x/{\partial}y=0$. Equation (\ref{eq:Kronecker_w_0}) reflects the same principle: the weights $w_{ij}$ and $w_{mn}$ are independent variables unless
their indices agree. Equation (\ref{eq:Kronecker_w_0}) is valid for  \index{weight!off diagonal}off-diagonal weights, and 
there are two terms on the r.h.s. because the weights are symmetric. 
\index{weight!symmetric weights}

Returning to the derivative of $\log Z$ with respect to $w_{mn}$, one finds using Equation~(\ref{eq:Kronecker_w_0}):
\begin{align}
\label{eq:Zderiv}
 \frac{\partial \log Z}{\partial w_{mn}} = \sum_{s_1=\pm 1,\ldots,s_N=\pm 1} s_m s_n \PB(\ve s)
\equiv \langle s_m s_n\rangle_{\rm model}\,.
\end{align}
The last equality defines the \index{pattern!two point correlations}two-point correlations of the model, $\langle s_m s_n\rangle_{\rm model}$, 
computed using the steady-state distribution (\ref{eq:approx}) of the Boltzmann machine. \index{distribution!steady state}
Evaluating the derivative of the
second term in Equation (\ref{eq:split4}) gives:
\begin{equation}
\frac{\partial}{\partial w_{mn}}  \tfrac{1}{2} \sum_{i\neq j} w_{ij} x_i^{(\mu)} x_j^{(\mu)}
= x_m^{(\mu)} x_n^{(\mu)}\,.
\end{equation}
In summary,
\begin{align}
\frac{\partial \log\mathscr{L}}{\partial w_{mn}}
= \sum_{\mu=1}^p \big( x_m^{(\mu)} x_n^{(\mu)}
-\langle s_m s_n\rangle_{\rm model}\big)= 
p \big( \langle x_m x_n\rangle_{\rm data} -\langle s_m s_n\rangle_{\rm model}\big)\,.
\end{align}
Here $\langle x_m x_n\rangle_{\rm data}=p^{-1}\sum_{\mu=1}^p  x_m^{(\mu)} x_n^{(\mu)}$ is the \index{pattern!two point correlations}two-point correlation of the input data.
Using (\ref{eq:grad_asc}), the {\em learning rule}\index{learning rule|textbf} becomes:
\begin{align}
\label{eq:Bl1}
\delta w_{mn} = \eta \big( \langle x_m x_n\rangle_{\rm data} -\langle s_m s_n\rangle_{\rm model}\big)\,,
\end{align}
where we dropped a factor of $p$ that only affects the numerical value of the learning rate\index{learning rate} $\eta$. 
The \index{weight increment|textbf}weight increments are determined by
the \index{pattern!two point correlations}two-point pattern correlations, just like Hebb's rule
\eqnref{C2S2HebbsRule}. 
The first term on the r.h.s. of Eq.~(\ref{eq:Bl1}) has precisely 
the same form as Equation~\eqnref{C2S2HebbsRule},
a sum over \index{pattern!two point correlations}two-point correlations of the input patterns\index{input pattern}.
The second average is over the steady-state distribution
(\ref{eq:approx}) of the Boltzmann machine. 
The learning rule takes the form of the difference between \index{pattern!two point correlations} two-point correlations
because the task is to minimise the difference between two distributions.
It is plausible that the learning rule  may converge
\index{convergence!learning rule}
because the \index{weight increment}weight increments vanish when the model correlations equal
the data correlations.
\index{weight increment}

The average $\langle s_ms_n\rangle_{\rm model}$  can be approximated by numerical simulation of the McCulloch-Pitts
dynamics
\begin{equation}
    s_i' = 
\begin{cases} \phantom-1\quad \text{  with probability  } \quad \pr(b_{i})\,, \\ -1 \quad \text{  with probability  } \quad 1-\pr(b_{i})\,,\end{cases}
    \eqnlab{BS1StochasticRule}
\end{equation}
with $b_i = \sum_j w_{ij} s_j$ and $\pr(b_i) = \frac{1}{1+{\rm e}^{-2 b_i}}$. 
One must iterate Equation~\eqnref{BS1StochasticRule}  until  the
system has reached its steady state, long enough so that 
any initial transient becomes negligible. 
\index{transient}

The training algorithm\index{training!algorithm|textbf} can be summarised as follows. One initialises all weights
and computes $\langle x_m x_n\rangle_{\rm data}$ from 
the given sequence of input patterns. One estimates 
$\langle s_m s_n\rangle_{\rm model}$ by numerical simulation of the dynamics of the Boltzmann machine,
and changes  the weights using (\ref{eq:Bl1}). This step is iterated,
either with a sequence of new inputs, or with the same inputs but in permuted sequence. In each iteration one
must compute $\langle s_m s_n\rangle_{\rm model}$ again, because
the weights changed.
This procedure is quite slow, because it usually takes long simulations to estimate $\langle x_m x_n\rangle_{\rm model}$ accurately, in each iteration
of the learning algorithm. 

There is a more fundamental problem \cite{MacKay}.
\index{Hebb's rule} Like Hebb's rule, the learning rule (\ref{eq:Bl1}) relies entirely upon \index{pattern!two point correlations}two-point correlations of the input bits. This means that the Boltzmann machine cannot learn higher-order correlations between inputs. However, \index{pattern!two point correlations}two-point correlations may not be sufficient to represent the information encoded in the input data. To illustrate this point, consider the Boolean \index{XOR function} XOR function (Exercise 2.13 and Chapter \chref{C5Chapter}). It can be encoded in the four patterns $[-1,-1,-1]$, $[1,1,-1]$, $[-1,1,1]$, and $[1,-1,1]$. The first two components represent the input to the XOR function.
The third component represents the output, which depends on both input variables as prescribed by the XOR function. Let us
define an \index{input distribution}input distribution that reflects these three-point correlations by assigning
$\Pd=\tfrac{1}{4}$ to the four patterns, and setting $\Pd=0$ otherwise. 
A Boltzmann machine with three neurons cannot represent this \index{input distribution}input distribution, because there is no energy function of the form (\ref{eq:HB}) that has four minima at these patterns. 
So the three-point correlations encoded
in the four patterns cannot be represented in terms of a Boltzmann machine in its simplest form. 

Also Hopfield networks fail for the XOR function\index{XOR function}: the four states are not attractors of a Hopfield network with three neurons (Exercise 2.13).
One could consider neural networks with
third- or higher-order interactions \cite{MacKay}, 
\begin{equation}
H=-\tfrac{1}{2}\sum_{ij} w_{ij}^{(2)} s_i s_j - \tfrac{1}{6} \sum_{ijk} w_{ijk}^{(3)} s_i s_j s_k + \ldots
\end{equation}
(Exercise 2.7).  But the number of weights proliferates as the order increases, rendering the training very slow. \index{training}

An alternative is to use Boltzmann machines with 
{\em hidden}\index{hidden neuron} neurons, that are neither input nor 
output units.\index{output unit}
The idea is that the hidden neurons can learn to represent such correlations \cite{MacKay}. The 
learning rule for the Boltzmann machines with hidden neurons 
is very similar to Equation (\ref{eq:Bl1}), but when the number
of hidden neurons is large, the Boltzmann machine is very slow to train.
It is more efficient to remove all weights between visible neurons, and between hidden neurons. This is described in the next Section.

\section{Restricted Boltzmann machines}
\index{restricted Boltzmann machine|(}
\label{sec:rbm}
\begin{figure}[t]
  \centering
    \begin{overpic}[scale=\myFigureScale]{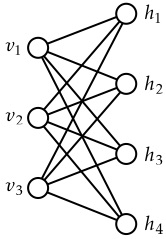}
    \end{overpic}
\caption{\figlab{BRestrBM} Restricted Boltzmann machine\index{restricted Boltzmann machine} with three visible 
neurons, $v_j$, and four hidden neurons, $h_i$.}
\end{figure}
{\em Restricted Boltzmann machines}\index{restricted Boltzmann machine} \cite{smolensky1987information}  consist of visible and hidden neurons arranged in an 
undirected \index{undirected bipartite graph}bipartite graph (Figure~\figref{BRestrBM}): the only connections are between neurons of different kinds, there are no connections between visible neurons, no connections between hidden neurons either. So the energy function for
a restricted Boltzmann machine 
for $N$ visible neurons $v_j$ and $M$ hidden neurons $h_i$ takes the form
\index{restricted Boltzmann machine!energy function}
\index{restricted Boltzmann machine|textbf}
\begin{equation}
\label{eq:Hrbm}
H = -\sum_{i=1}^M\sum_{j=1}^N w_{ij} \sh_i \sv_j 
 + \sum_{j=1}^N \theta^{({\rm v})}_j \sv _j + 
 \sum_{i=1}^M \theta^{({\rm h})}_i \sh_i\,,
\end{equation}
 with weights $w_{ij}$ and \index{threshold}thresholds $\theta_j^{({\rm v})}$ and $\theta_i^{({\rm h})}$.
The McCulloch-Pitts dynamics reads
\begin{subequations} \label{eq:MPDRBM}
\begin{equation}
\label{eq:MPDRBMh}
    \sh_i' =
    \begin{cases} 
       \phantom-1\quad \text{with probability}  \quad \pr(b_{i}^{({\rm h})}) \\ 
      -1 \quad \text{with probability} \quad 1-\pr(b_{i}^{({\rm h})})
    \end{cases}\quad\mbox{with}\quad b_i^{({\rm h})} = \sum_{j=1}^N w_{ij}\sv_j-\theta^{({\rm h})}_i\,,
\end{equation}
and
\begin{equation}
\sv_j' =
    \begin{cases} 
       \phantom-1\quad \text{with probability}  \quad \pr(b_{j}^{({\rm v})})\\
      -1 \quad \text{with probability} \quad 1-\pr(b_{j}^{({\rm v})})
    \end{cases} \quad\mbox{with}\quad b_j^{({\rm v})} = \sum_{i=1}^M h_i w_{ij}- \theta^{({\rm v})}_j\,.
\end{equation}
\end{subequations}
The diagonal weights are assumed to vanish, but the \index{weight!matrix}weight matrix is not required to be symmetric (Exercise 4.9). Since most often $M\gg N$, it is usually not even a square matrix.

The learning rule\index{learning rule} for the weights of the restricted Boltzmann machine is derived using gradient ascent on the log-likelihood
for a single pattern $\ve x^{(\mu)}$:
\begin{equation}
\log P(\ve x^{(\mu)}) = \log \sum_{h_1=\pm 1,\ldots, h_M = \pm 1} \PB(\ve \sv=\ve x^{(\mu)},\ve \sh)\,.
\end{equation}
\begin{algorithm}[p]
\caption{\label{cdk} contrastive divergence CD-$k$ for $\pm 1$ neurons}
\begin{algorithmic}
\STATE initialise weights and thresholds;
\FOR {$\nu=1,\ldots, \nu_{\rm max}$} 
\STATE sample $p_0$ patterns from the data distribution ($p_0 \leq p$);
\FOR {$\mu=1,\ldots,p_0$}
  \STATE initialise $\ve v(0)\leftarrow \ve x^{(\mu)}$;\\ 
\STATE update all hidden neurons: $\ve b^{({\rm h})}(0) \leftarrow \ma W \ve v(0)-\ve\theta^{({\rm h})}$;
  \FOR {$i=1,\ldots,M$}
  \STATE $h_i(0)\leftarrow +1$ with probability $\pr\big(b_i^{({\rm h})}(0)\big)$ otherwise $h_i(0)\leftarrow -1$;
\ENDFOR
  \FOR {$t=1,\ldots,k$}
      \STATE update all visible neurons: $\ve b^{({\rm v})}(t-1) \leftarrow \ve h(t-1)\cdot \ma W-\ve\theta^{({\rm v})}$;
      \FOR {$j=1,\ldots,N$}
      \STATE $v_j(t)\leftarrow +1$ with probability $\pr\big(b_j^{({\rm v})}(t-1)\big)$ otherwise $v_j(t)\leftarrow -1$;
      \ENDFOR
      \STATE update all hidden neurons: $\ve b^{({\rm h})}(t) \leftarrow \ma W \ve v(t)-\ve\theta^{({\rm h})}$;
      \FOR {$i=1,\ldots,M$}
      \STATE $h_i(t)\leftarrow +1$ with probability $\pr\big(b_i^{({\rm h})}(t)\big)$ otherwise $h_i(t)\leftarrow -1$;
      \ENDFOR
  \ENDFOR
  \STATE compute weight and theshold increments:\\
$\delta w_{mn}^{{(\mu)}} \leftarrow \eta \Big[
\tanh\big(b_m^{({\rm h})}(0)\big) v_n(0)-\tanh\big(b_m^{({\rm h})}(k)\big) v_n(k)\Big]$;\\
$\delta\theta_n^{({\rm v}{, \mu})} \leftarrow -\eta[v_n(0)-v_n(k)]$;\\
$\delta\theta_m^{({\rm h}{, \mu})} \leftarrow -\eta\big[{\rm tanh}\big(b_m^{({\rm h})}(0)\big)-{\rm tanh}\big(b_m^{({\rm h})}(k)\big)\big]$;
\ENDFOR
\STATE adjust weights and thresholds  using $\sum_{\mu=1}^{p_0} \delta w_{mn}^{(\mu)}$, $\sum_{\mu=1}^{p_0}\delta\theta_n^{({\rm v},\mu)}$, and $\sum_{\mu=1}^{p_0}\delta\theta_m^{({\rm h},\mu)}$;
\ENDFOR
\end{algorithmic}
\end{algorithm}
Proceeding as in the previous Section one finds:
\begin{align}
\label{eq:Bl3}
\delta w_{mn}^{(\mu)} = \eta \big( \langle \sh_m x_n^{(\mu)}\rangle_{\rm data} -\langle \sh_m \sv_n\rangle_{\rm model}\big)\,.
\end{align}
The first average,
\begin{align}
\label{eq:CD_1}
\langle \sh_m x_n^{(\mu)}\rangle_{\rm data} &= \sum_{\sh_1 = \pm 1,\ldots,\sh_M = \pm 1}\!\!\!\! h_m x_n^{(\mu)}\Bigg[\prod_{i=1}^M P(h_i |\ve \sv = \ve x^{(\mu)})\Bigg]\,,
\end{align}   
can be evaluated further, using the fact that there are no connections between the
hidden units. Making use of  the update rule\index{update rule} (\ref{eq:MPDRBMh}) we find
\begin{align}
\sum_{h_m=\pm 1} h_m P(h_m |\ve \sv = \ve x^{(\mu)})  
=  p(b_m^{({\rm h})}) -[1-p(b_m^{({\rm h})})]=\mbox{tanh}(b_m^{({\rm h})})\,,
\end{align}
just like Equation (\ref{eqn:C3S4Output}). For the other sums in Equation (\ref{eq:CD_1}) we use the normalisation condition  $1=\sum_{h_k=\pm 1}  P(h_k |\ve \sv = \ve x^{(\mu)})$ to obtain:    
\begin{align}
\label{eq:CD_1}
\langle \sh_m x_n^{(\mu)}\rangle_{\rm data}
&= \tanh(b_m^{({\rm h})}) \,x_n^{(\mu)} =\tanh\Big(\sum_{j=1}^N w_{mj}x_j^{(\mu)}- \theta^{({\rm h})}_m \Big) \,x_n^{(\mu)}\,.
\nonumber
\end{align}
The second average on the r.h.s. of Equation (\ref{eq:Bl3})
simplifies to
\begin{equation}
\label{eq:CD_2}
\langle \sh_m \sv_n\rangle_{\rm model} = \Big\langle \tanh\Big(\sum_{j=1}^N w_{mj} v_j- \theta^{({\rm h})}_m\Big) \,\sv_n\Big\rangle_{\rm model}\,.
\end{equation}
The average $\langle \cdots\rangle_{\rm model}$ is computed by \index{Markov chain Monte Carlo}Monte-Carlo sampling,
using the McCulloch-Pitts dynamics  
(\ref{eq:MPDRBM}) to generate the Markov chain
\begin{equation}
\ve \sv_{t=0}\to\ve\sh_{t=0}\to \ve\sv_{t=1}\to \ve\sh_{t=1} \to \ve\sv_{t=2}\to\cdots\,.
\end{equation}
In the limit $t\to\infty$, the steady state of this sequence is
distributed according to the model distribution, the Boltzmann distribution
\index{Boltzmann distribution} 
with energy function (\ref{eq:Hrbm}). In general only the asynchronous\index{update rule!asynchronous} McCulloch-Pitts dynamics can be proven to 
\index{convergence!stochastic dynamics}converge (Sections \secref{C2S3EnergyFunction} and  \ref{sec:C4conv}). Here, however, the Markov chain can be generated more efficiently by updating all hidden neurons $\ve \sh_{t}$ 
at the same time, given $\ve \sv_t$, because the components of $\ve \sh_t$ are independent from each other since there are no connections between them. In the same way the visible neurons $\ve \sv_t$  are updated in parallel. 
To speed up the computation further, one usually only
iterates for a finite number of steps, up to $t=k$ say, and initialises
the chain with $\ve v_{t=0}=\ve x^{(\mu)}$. 
After $k$ steps one approximates
\begin{equation} 
\label{eq:avap}
\Big\langle \tanh\Big(\sum_{j=1}^N w_{mj} v_j- \theta^{({\rm h})}_m\Big) \,\sv_n\Big\rangle_{\rm model}
\approx \tanh\Big(\sum_{j=1}^N w_{mj} v_{j,t=k}- \theta^{({\rm h})}_m\Big) \,\sv_{n,t=k}\,.
\end{equation}
This algorithm is called {\em contrastive-divergence} or CD-$k$ algorithm (Algorithm \ref{cdk}). Since the average over the model distribution
is approximated [Equation (\ref{eq:avap})], this algorithm  does not precisely correspond to gradient ascent.

In summary,
\begin{align}
\label{eq:Bl4}
\delta\! w_{mn} &\!=\! \eta 
 \Big[\tanh\Big(\sum_j w_{mj}v_{\!j,t\!=0}- \theta^{({\rm h})}_m\Big) v_{\!n,t\!=0}\!-\!
\tanh\Big(\sum_j w_{mj}v_{\!j,t\!=\!k}- \theta^{({\rm h})}_m\Big) v_{\!n,t\!=\!k}\Big]\,.
\end{align}
The analogous learning rules for the \index{threshold} thresholds read:
\begin{subequations}
\label{eq:update_th_rbm}
\begin{align}
\delta\theta_n^{(\rm v)} &= -\eta  \big( \sv_{n,t=0} -\sv_{n,t=k}\big)\,,\\
\delta \theta_m^{(\rm h)} &= - \eta \Big[\tanh\Big(\sum_j w_{mj}v_{j,t=0}- \theta^{({\rm h})}_m\Big) 
-\tanh\Big(\sum_j w_{mj}v_{j,t=k}- \theta^{({\rm h})}_m\Big)\Big]\,.
\end{align}
\end{subequations}
The derivation of Equations (\ref{eq:update_th_rbm}) is left as an exercise (Exercise 4.10).
Restricted Boltzmann machines may have 0/1 neurons with state values $0$ and $1$ instead of $-1$ and $1$. For 0/1 neurons, 
the CD-$k$ algorithm is slightly different (Exercise 4.11).

\begin{figure}[t]
  \centering
    \begin{overpic}[scale=\myFigureScale]{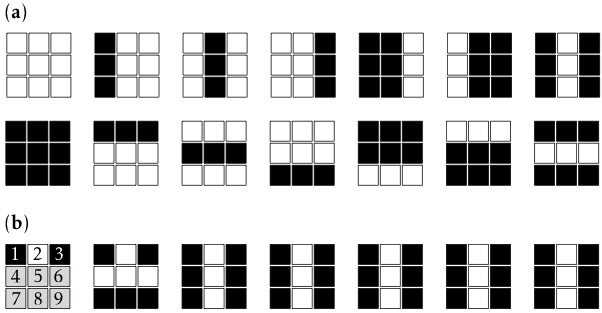}
    \end{overpic}
\caption{\figlab{BAS} Pattern completion for \index{bars and stripes data set} bars-and-stripes 
data set \cite{MacKay}.  ({\bf a}) All patterns in the $3\times 3$ bars-and-stripes data set, $\Box$ corresponds to $-1$, $\blacksquare$ to $+1$.
 ({\bf b}) The three visible units $[v_1,v_2,v_3]$ corresponding to the first row are \index{input!clamp}clamped  to $[+1,-1,+1]$ and remain
fixed to these values. The remaining units are initially set to $0$ (gray bits), and their
states are allowed to change 
 while sampling from the restricted Boltzmann machine using
After a short transient of the McCulloch-Pitts dynamics, the pattern is correctly completed.
Schematic, after Figure~7 in Ref.~\cite{FischerIgel}.}
\end{figure}
Figure \ref{fig:BAS} illustrates how a restricted Boltzmann machine can learn to complete patterns,
using the \index{bars and stripes data set|textbf}bars-and-stripes data set \cite{MacKay,FischerIgel} as an example.  To begin with, the restricted Boltzmann machine 
is trained using the CD-$k$ algorithm. Then consider a partially obscured pattern. Assume for instance that only
the upper row of its bits is known: $v_1 = +1$ ($\blacksquare$),
$v_2=-1$ ($\Box$), and $v_3=+1$ ($\blacksquare$).
The remaining bits $v_4,\ldots,v_9$ are obscured, their
states are set to zero as shown in Figure~\ref{fig:BAS}({\bf b}). 
To complete the pattern, one samples from the Boltzmann 
distribution $\PB(v_4,\ldots,v_9|v_1=+1,v_2=-1,v_3=+1)$ keeping $v_1=+1,v_2=-1,v_3=+1$ fixed
 ({\em clamping}\index{input!clamp|textbf} these neurons), and iterates 
the McCulloch-Pitts dynamics for the remaining
ones. Panel ({\bf b}) shows how the machine
outputs the correct completed pattern.

\begin{figure}[bt]
	\centering
	\begin{overpic}[scale=\myFigureScale]{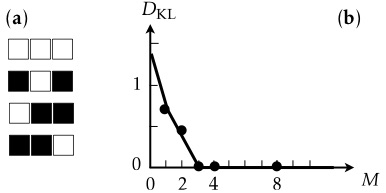}
	\end{overpic}
\caption{\label{fig:KLXOR} Restricted-Boltzmann-machine learning for the XOR problem [panel ({\bf a})], see Section \ref{sec:bm1}.  
Panel ({\bf b}) shows numerical estimates of $\DKL$ versus the number $M$ of \index{hidden neuron} hidden neurons, in comparison with the upper bound (\ref{eq:DKL_ub}). Schematic, 
based on simulations performed by Arvid Wenzel Wartenberg using the CD-$k$ algorithm.} 
\end{figure}

This requires \index{hidden neuron} hidden neurons, because 
the data distribution has non-zero three-point correlations 
\cite{MacKay} (Exercise 2.13). In general a restricted Boltzmann machine can approximate a distribution $\Pd$ of binary input data
better with more hidden neurons. How many are needed \cite{leroux2008representational,leroux2010deep}?  The answer is not known in general, but it is plausible  $M\sim 2^N$
hidden neurons are sufficient, because each hidden neuron can encode one of the binary 
\index{input pattern!binary} input patterns ({\em winning neuron}\index{winning neuron}, Section \ref{sec:hmhl}). More precisely,
it can be shown that $M = 2^{N-1}-1$ hidden neurons are sufficient 
to reach arbitrarily small Kullback-Leibler divergence 
\cite{montufar2011refinements}.  
An upper bound for the Kullback-Leibler divergence was derived in Refs.~\cite{,montufar2011expressive,montufar2013maximal}:
\begin{align}
\label{eq:DKL_ub}
\DKL \leq \log (2)\, \,
\begin{cases} 
 N-\lfloor \log_2 (M +1)\rfloor-\frac{M+1}{2^{\lfloor \log_2(M+1)\rfloor}} &
 M < 2^{N-1}-1\,,\\
0 & M\geq 2^{N-1}-1\,.\\  
\end{cases} 
\end{align}
Here $\lfloor \cdots\rfloor$ denotes the integer part.
Figure \ref{fig:KLXOR} illustrates 
this result. It demonstrates 
how well a restricted Boltzmann machine approximates the 
\index{XOR function}
XOR distribution introduced in Section \ref{sec:bm1}. 
The Figure shows how the Kullback-Leibler divergence depends on
the number of hidden neurons (Exercise 4.7). 
In this example there are $N=3$ inputs.
 We see that three hidden neurons are sufficient to allow 
the restricted Boltzmann machine to approximate the data distribution very precisely, consistent with Equation (\ref{eq:DKL_ub}). 
In general, however, the CD-$k$ algorithm is not guaranteed to \index{convergence!CD k}converge to the optimal solution corresponding to the estimate (\ref{eq:DKL_ub}).

Restricted Boltzmann machines are {\em generative models}\index{generative model|textbf}, they can be used to sample from a distribution the machine has learned \cite{FischerIgel}.
In this way, the machine can complete missing information, as illustrated in Figure \ref{fig:BAS}.
Restricted Boltzmann machines can also learn to classify patterns, by learning a distribution of binary inputs together with their labels.
To this end one splits the visible neurons into input neurons and output neurons\index{output neuron} with labels or targets\index{target}\index{label}. This is a supervised-learning task,
the subject of Part \ref{part:supervised}.
Recently, restricted Boltzmann machines were used to represent and analyse ground-state wave functions of quantum many-body systems \cite{carleo2017solving}.

\index{Boltzmann machine|)}
\index{restricted Boltzmann machine|)}

\section{Summary}
This Chapter dealt with the Boltzmann distribution.  Two main points are, first, that the stochastic McCulloch-Pitts dynamics (\ref{eqn:C3S2StochasticRule}) has the Boltzmann distribution as a steady state. Second, the update rule (\ref{eqn:C3S2StochasticRule}) is a special case of the \index{Markov chain Monte Carlo}Markov-chain Monte-Carlo algorithm, for Hopfield models with the energy
function  (\ref{eq:H}). Since this algorithm tends to decrease the energy function, it can be used to solve complex optimisation problems. In simulated annealing one gradually reduces the noise level as the simulation proceeds.  This mimics the slow cooling of a physical system, usually an efficient way of bringing the system into its global optimum.  

Boltzmann machines are generalisations of Hopfield networks that can learn distributions of binary data by iteratively changing the \index{weight} weights and \index{threshold}thresholds until the corresponding Boltzmann distribution approximates the data distribution. The learning rule is derived using gradient ascent on a target function, in this case the log-likelihood. A related idea is used for training\index{training!deep networks} deep neural networks with stochastic gradient descent (Part \ref{part:supervised}). To learn general \index{input distribution}input distributions  of binary patterns requires hidden neurons, also this is a central topic of Part \ref{part:supervised}. Since Boltzmann machines with many hidden neurons are hard to train, one removes connections that are not needed. Restricted Boltzmann machines have connections only between visible and hidden neurons.

\section{Further reading}
\label{sec:fr4}
Older but still good references 
for Monte-Carlo methods in statistical physics are the book {\em Monte Carlo methods in Statistical Physics} edited by Binder \cite{Binder}, and Sokal's lecture notes \cite{sokal1997monte}. Some historical notes
are found in Ref.~\cite{gubernatis2005marshall}.

For a concise introduction to Boltzmann machines, refer to {\em Information theory, inference and learning algorithms} by  MacKay \cite{MacKay}, or 
to {\em Machine learning: a probabilistic perspective} by Murphy \cite{Murphy}.  Ref.~\cite{fischer2012introduction} is a more mathematical review of restricted Boltzmann machines.   

How many hidden neurons should one allow for in a restricted Boltzmann machine? Little is known apart from the 
upper bound (\ref{eq:DKL_ub}) for the Kullback-Leibler divergence, and 
simulations \cite{montufar2011refinements} show that one can get very precise approximations of $\Pd$ with less hidden neurons than stipulated by Equation (\ref{eq:DKL_ub}). 

{\em Deep-belief networks} consist
of layers of restricted Boltzmann machines \cite{Haykin}. Contrastive-divergence \index{training}training for such deep architectures (networks with many layers) is one of the first examples of deep-learning\index{deep learning} algorithms \cite{bengio2009learning} (Chapter \ref{chap:dl}).

{\em Helmholtz machines}\index{Helmholtz machine|textbf} \cite{dayan1995helmholtz,dayan1996varieties} are generalisations of Boltzmann machines designed as more efficient generative models\index{generative model}.  They consist of two networks, encoder and decoder, just like variational autoencoders (Section \ref{sec:ae}).  The encoder (called recognition model in Ref.~\cite{dayan1996varieties}) generates a compressed representation of the data distribution, and the decoder (generative model) generates patterns from the compressed representation.

\vfill\eject

\cleardoublepage
\part{Supervised learning}
\label{part:supervised}

The Hopfield network described in Part \ref{part:hopfield} recognises
patterns stored using Hebb's rule. Its neurons act as inputs and outputs. 
After feeding a distorted pattern into the network, 
the network dynamics runs until it reaches a steady state\index{steady state} which hopefully
corresponds to the stored pattern closest to the distorted one. 
In this case, the network classifies the distorted pattern by associating it with the closest one amongst the stored patterns.

Part \ref{part:supervised} describes {\em supervised learning}\index{supervised learning|textbf}, 
a different way of solving {\em classification tasks}\index{classification!task|textbf}
with neural networks using labeled data sets. 
The {\em machine-learning repository}      \cite{UCI} at the University of California Irvine\index{machine learning repository}
contains a number of such data sets.  An example is the {\em iris} data set\index{iris data set|textbf} which lists certain properties  of 150 iris plants. For each plant, four attributes are given (Figure \figref{C5S1Iris}): its sepal length, sepal width, petal length, and petal width. 
Each entry in the data set contains a {\em label}\index{label|textbf} (or {\em target}\index{target}) that says
which class the plant belongs to: {\em iris setosa}, {\em iris versicolor}, or {\em iris virginica}.
This data set was described by the geneticist R. A. Fisher \cite{Fisher}.

The machine-learning task is to adjust weights and \index{threshold}\index{weight}thresholds of a neural network so that it correctly
 determines the class of each plant from its attributes. 
To this end one uses a {\em training} set\index{training set} of labeled data.
Each set of attributes is an input pattern\index{input pattern} to the network. The neural network is supposed
to output the correct label (or target\index{target}), in this case whether the plant is an {\em iris setosa}, {\em iris versicolor}, or {\em iris virginica}. 
One compares the network output with the corresponding \index{target}target, for all input patterns in the training set,  \index{training set}
and changes the weights and \index{threshold}\index{weight}thresholds until the network computes the correct output for each input pattern.       
The crucial question is whether the trained network can {\em generalise}\index{generalisation}: does it find the correct \index{label}labels  for 
an input pattern not contained in the training set?\index{training set}

The networks used for supervised learning are called {\em perceptrons}\index{perceptron|textbf} \cite{Rosenblatt1958}. They consist of layers
of McCulloch-Pitts neurons: usually some  layers
\index{McCulloch Pitts neuron}
of {\em hidden}\index{hidden neuron} neurons, and an output layer. 
We briefly discussed the idea of hidden neurons in connection with restricted Boltzmann machines (Section \ref{sec:rbm}),
but perceptrons have different layouts, and they are trained in a different way.
The layers are usually arranged from the left (input) to the right (output). All connections are one-way,
from neurons in one layer to neurons in the layer immediately to the right. 
There are no connections between neurons in a given layer, or  back to layers on the left. This arrangement ensures \index{convergence!training}convergence of the 
\index{training!algorithm}training algorithm ({\em stochastic gradient descent})\index{gradient descent!stochastic}. During training with this algorithm, the network parameters 
are changed iteratively. In each step, an input is applied, and weights and thresholds of the network are updated to reduce the output error\index{output error}. Loosely speaking, each step corresponds to adding a little bit of Hebb's rule to the weights. This is repeated until the network classifies the \index{training set}training set correctly.

Stochastic gradient descent for multilayer perceptrons has received much attention  recently, after it was realised that networks with many hidden layers\index{hidden layer} can be trained to reliably recognise and classify image data ({\em deep learning})\index{deep learning|textbf}. 

\chapter{Perceptrons}\chlab{C5Chapter}
\begin{figure}[t]
  \centering
    \raisebox{-12mm}{
    \begin{overpic}[scale=\myFigureScale]{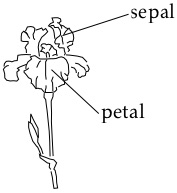}
    \end{overpic}}\hspace*{1cm}
\begin{tabular}{lllll}
\hline\hline
\small sepal  & & \small petal &  & \small classification\\
\small length & \small width & \small length & \small width & \\\hline
        6.3&    2.5&    5.0 &   1.9 &   \small virginica\\
        5.1 &   3.5 &   1.4 &   0.2 &   \small setosa\\
        5.5 &   2.6  &  4.4  &  1.2   &  \small versicolor\\
        4.9 &   3.0 &   1.4 &   0.2 &   \small setosa\\
        6.1  &  3.0  &  4.6  &  1.4  &   \small versicolor\\
        6.5&    3.0 &   5.2 &   2.0 &    \small virginica\\
\hline\hline
\end{tabular}
    \centering \caption{\figlab{C5S1Iris} Left: petals and sepals of the iris flower.
Right: six entries of the iris data set \cite{UCI}. All lengths in cm. The whole data set contains 150 entries.}
\end{figure}

In 1958 Rosenblatt \cite{Rosenblatt1958} suggested to connect McCulloch-Pitts neurons \index{McCulloch Pitts neuron}
into layered {\em feed-forward} networks\index{network!layered|textbf}\index{feed forward network} to process information. He referred to these networks as {\em perceptrons}\index{perceptron}.  The layout is illustrated in 
 in Figure \figref{C5S1SingleLayerNetwork}. 
The leftmost layer consists of input terminals,\index{input terminal|textbf}
drawn in black in Figure \figref{C5S1SingleLayerNetwork}.
To the right follow two layers of McCulloch-Pitts neurons. The rightmost layer consists of output neurons.\index{output neuron} 
The intermediate layer is a {\em hidden} layer\index{hidden layer|textbf}. 
The states of its 
neurons are not read out. 
\begin{figure}[tb]
  \centering
    \begin{overpic}[scale=\myFigureScale]{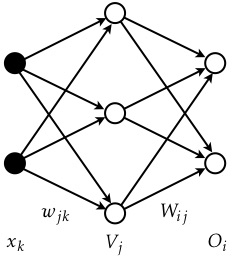}
    \end{overpic}
    \centering \caption{\figlab{C5S1SingleLayerNetwork}
Feed-forward network with one hidden layer. The input terminals\index{input terminal} are coloured black.
We use the notation of Ref.~\cite{hertz1991introduction}: $W_{ij}$ for the weights connecting to the output neuron\index{output neuron} $O_i$ 
(with \index{threshold}threshold $\Theta_i$), and $w_{jk}$ for the weights connecting to the 
hidden neuron\index{hidden neuron} $V_j$ (with \index{threshold}threshold $\theta_j$).}
\end{figure} 
All connections are one-way: every neuron 
feeds {\em forward}, only to neurons in the layer immediately to the right. There are no connections within layers, no back connections, no connections that skip a layer.
There are $N$ input terminals.\index{input terminal} As in Part \ref{part:hopfield}, we denote the input patterns
\index{input pattern|textbf}
by
\begin{equation}
    \ve x^{(\mu)} = \begin{bmatrix} x_1^{(\mu)} \\x_2^{(\mu)} \\ \vdots \\ x_N^{(\mu)} \end{bmatrix}\,.
\end{equation}
The index $\mu=1,\ldots,p$ 
labels the different input patterns.
The hidden neurons compute
\begin{equation}
\label{eq:Vj}
V_{j} = g\left(b_{j}\right) \quad\mbox{with}\quad b_{j}= \sum_{k} w_{jk} x_{k}-\theta_j\,, 
\end{equation}
with weights $w_{jk}$ and \index{threshold}\index{weight}thresholds $\theta_j$.
The function $g(b)$ is an activation function, and its argument is called local field\index{local field} (Section \secref{C1S2Deterministic}).
\index{local field}  
The output neurons\index{output neuron} of the network shown  in Figure  \figref{C5S1SingleLayerNetwork} perform the computation
\begin{equation} 
\label{eq:Oi}
O_{i} = g\left(B_{i}\right)\quad\mbox{with}\quad B_{i} = \sum_{j} W_{ij} V_{j} - \Theta_{i}\,.
\end{equation}
The index $i=1,\ldots,M$ labels the output neurons\index{output neuron} with weights $W_{ij}$, and with \index{threshold}\index{weight}thresholds $\Theta_i$.

A classification problem is given by a training set\index{training set|textbf}  of 
input patterns $\ve x^{(\mu)}$ and the corresponding 
\index{target!vector} {\em target} vectors \index{target|textbf}
\begin{equation}
    \ve t^{(\mu)} = \begin{bmatrix} t_1^{(\mu)} \\t_2^{(\mu)} \\ \vdots \\ t_M^{(\mu)} \end{bmatrix}\,.
\end{equation}
The idea is to choose all weights and thresholds so that the network produces the desired output:
\begin{figure}[t]
  \centering
    \begin{overpic}[scale=\myFigureScale]{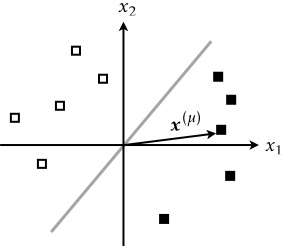}
    \end{overpic}
\caption{\figlab{C5S1CT1} Classification problem with two-dimensional real-valued inputs and
\index{target} targets equal to $\pm 1$. The gray solid line is the decision boundary. Legend:
$\blacksquare$ corresponds to  $t\tomu = 1$, and $\Box$ to  $t\tomu = -1$.
}
\end{figure}
\begin{equation}
O_{i}^{(\mu)} =  t_{i}^{(\mu)} \quad\mbox{for all}\quad i\quad\mbox{and}\quad \mu\,.
\eqnlab{C5S2ONotEqualToZeta}
\end{equation} 
In the Hopfield networks described in Part \ref{part:hopfield}, the weights were assigned using Hebb's rule  (\ref{eq:HRw0}).  Perceptrons, by contrast, are trained by iteratively updating their weights and thresholds until Equation 
(\ref{eqn:C5S2ONotEqualToZeta}) is satisfied. This is achieved by repeatedly adding small multiples of Hebb's rule to the weights (Section \ref{sec:ila}).
An alternative approach is to define an energy function\index{energy function}, a function of the weights of the network, that has a global minimum when
Equation (\ref{eqn:C5S2ONotEqualToZeta}) is satisfied. The network is trained by taking small steps in weight space that reduce the energy function
(gradient descent\index{gradient descent}, Section \ref{sec:gdl}). 

\section{A classification problem}
\index{decision boundary|(}
\label{sec:act}
To illustrate how perceptrons can solve classification problems, consider the simple example shown in Figure \figref{C5S1CT1}. There are ten patterns, each has two real-valued components:
\begin{equation}
\ve x^{(\mu)} = \begin{bmatrix}  x_{1}^{(\mu)} \\\ x_{2}^{(\mu)} \end{bmatrix}\,.
\end{equation}
In Figure \figref{C5S1CT1} the patterns are drawn as points in
the $x_1$-$x_2$ plane, the {\em input plane}\index{input!plane}. There are two classes of patterns,
with \index{target} targets  $\pm 1$:
\begin{equation}
t^{(\mu)} = 1 \quad\mbox{for}\quad \blacksquare \quad\mbox{and}\quad
t^{(\mu)} =-1  \quad\mbox{for}\quad \Box\,.
\end{equation}
A single neuron suffices to classify these patterns, a binary threshold unit
\index{binary threshold unit} with activation function $g(b) = \text{sgn}(b)$, consistent with the possible \index{target} target values.
Since there is only one neuron, we can arrange the weights into a weight vector
\begin{equation}
\ve w = \begin{bmatrix}  w_{1} \\ w_{2}\end{bmatrix}\,.
\end{equation} 
The network performs the computation
\begin{equation}
\label{eq:pO}
O = {\rm sgn}(w_1x_1 + w_2x_2-\theta) = {\rm sgn}(\ve w \cdot \ve x-\theta)\,.
\end{equation}
Here $\ve w \cdot \ve x=w_1x_1 + w_2x_2$ is the scalar product\index{scalar product}  between the vectors $\ve w$ and $\ve x$ (Chapter~\ref{ch:dhn}).

This example allows us to find a geometrical interpretation of the
classification problem. We see in Figure \figref{C5S1CT1} that the patterns
fall into two clusters: $\Box$ to the left  and $\blacksquare$ to the right. We can
classify the patterns by drawing a line that separates the two clusters, so that
everything to the right of the line has $t={+}1$, while the patterns to the left
of the line have $t=-1$. This line is called the {\em decision boundary}\index{decision boundary|textbf}. 
To find the geometrical significance of Equation (\ref{eq:pO}), let us put the \index{threshold}threshold to zero for a moment, so that
\begin{equation}
O = {\rm sgn}(\ve w \cdot \ve x)\,.
\end{equation}
\begin{figure}[t]
  \centering
    \begin{overpic}[scale=\myFigureScale]{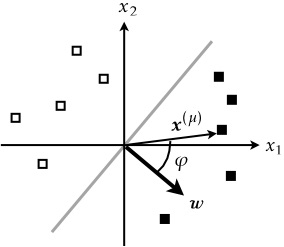}
    \end{overpic}
\caption{\figlab{C5S1CT3} The perceptron classifies the patterns correctly for the weight vector $\ve w$ shown, orthogonal to the decision boundary (gray solid line).  Legend: $\blacksquare$ corresponds to  $t\tomu = 1$, and $\Box$ to  $t\tomu = -1$.}
\end{figure}
Then the classification problem takes the form
\begin{equation} 
\text{sgn}\left(\ve w\cdot \ve x\tomu\right) = t\tomu \,.
\end{equation}
To evaluate the scalar product, we write the vectors as
\begin{equation}
\ve w = |\ve w|\,\begin{bmatrix} \cos\alpha \\ \sin\alpha \end{bmatrix}      
\quad\mbox{and}\quad
\ve x= |\ve x|\,\begin{bmatrix} \cos\beta\\ \sin\beta \end{bmatrix}\,.
\end{equation} 
Here $|\ve w|=\sqrt{w_1^2+w_2^2}$ denotes the norm of the vector $\ve w$, and $\alpha$ and $\beta$
are the angles of the vectors with the $x_1$-axis. 
Then  $\ve w\cdot \ve x =|\ve w||\ve x|\cos(\alpha-\beta)=|\ve w||\ve x|\cos\varphi$,
where $\varphi$ is the angle between the two vectors. When $\varphi$  is between $-\pi/2$ and $\pi/2$, the scalar product is positive, otherwise negative. As a consequence, the network classifies the patterns
in Figure  \figref{C5S1CT1} correctly if the weight vector is orthogonal to the decision boundary, as shown in Figure \figref{C5S1CT3}.

What is the role of the \index{threshold|textbf}threshold $\theta$? Equation (\ref{eq:pO}) implies that the decision boundary is parameterised by $\ve w\cdot\ve x=\theta$, or
\begin{equation}
x_2 = -(w_1/w_2)\,x_1 + \theta/w_2\,.
\end{equation}
Therefore the \index{threshold}threshold determines the intersection of the decision boundary with the $x_2$-axis (equal to $\theta/w_2$).
This is illustrated in Figure \figref{C5S1WeightPlane}.
\begin{figure}
  \centering
    \begin{overpic}[scale=\myFigureScale]{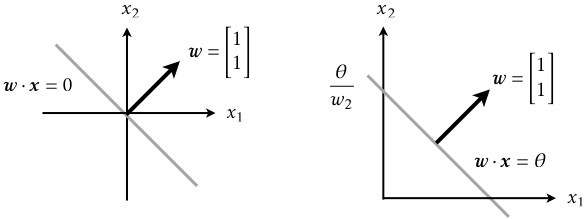}
    \end{overpic}
    \caption{\figlab{C5S1WeightPlane}Decision boundaries without and with \index{threshold}threshold.}
\end{figure}

The decision boundary -- the straight line  orthogonal to $\ve w$ -- should divide inputs with positive and negative \index{target} targets.
If such a line can be found, then the problem can be solved with a single neuron. 
We say that the problem is {\em linearly separable}\index{linear separability|textbf}.
Conversely, if no such line exists, the problem not linearly separable.  
This can occur only when $p>N$.
Figure \figref{C5S1SeparableNonseparable} shows two problems. The left one is linearly separable, the right one is not.
\begin{figure}[b]
\centering
    \begin{overpic}[scale=\myFigureScale]{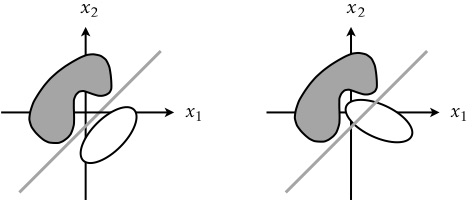}
    \end{overpic}
    \caption{\figlab{C5S1SeparableNonseparable}Linearly separable and non-separable data in two-dimensional input space\index{input space}.}
\end{figure}

Other examples are {\em Boolean functions}\index{Boolean function|textbf}.
A Boolean function takes $N$ binary inputs and has one binary output.
The Boolean AND function (two inputs) is illustrated in Figure  \figref{C5S1ANDFunction}.
\index{Boolean function!AND}
The value table of the function is shown on the left. The graphical
representation is shown on the right of the Figure ($\Box$ corresponds
to $t=-1$ and $\blacksquare$ to $t=+1$). Also shown is the decision boundary
of a binary threshold unit\index{binary threshold unit}
and its weight vector $\ve w$.
\begin{figure}
    \centering
        \begin{tabular}{lccc}
        &&& \\\hline\hline
        $x_{1}$  & $x_{2}$ & $ t $\\
        \hline
        0 &  0  & -1  \\
        0 &  1  & -1 \\
        1 &  0  & -1 \\
        1 &  1  & +1 \\
        \hline
        \hline
        \end{tabular}
\hspace*{10mm}
\raisebox{-2cm}{
    \begin{overpic}[scale=\myFigureScale]{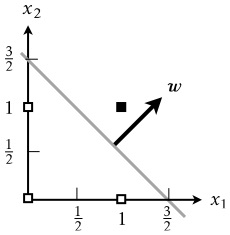}
    \end{overpic}}
    \caption{\figlab{C5S1ANDFunction} 
Boolean AND function: value table (left)
and  geometrical representation in the input plane (right).
 Legend:
$\blacksquare$ corresponds to  $t\tomu = 1$, and $\Box$ to  $t\tomu = -1$.}
\end{figure}
It is important to note that the decision boundary is not unique, neither are the weight vector\index{weight!vector} and \index{threshold}threshold
value that solve the problem. The norm of the weight vector, in particular, is arbitrary. 
Figure  \figref{C5S1XORFunction} illustrates that the Boolean XOR function is 
\index{XOR function|textbf}
\index{Boolean function}
not linearly separable \cite{MinskyPapert}.
\index{XOR function!not linearly separable}
There are 16 different Boolean functions of two variables. 
Only two are not linearly separable (Exercise 5.2), 
XOR (Figure~\figref{C5S1XORFunction}) and XNOR.
\index{Boolean function!XNOR}
\begin{figure}[b]
    \centering
        \begin{tabular}{lccc}
        &&& \\\hline\hline
        $x_{1}$  & $x_{2}$ & $ t $\\
        \hline
        0 &  0  & -1 \\
        0 &  1  & +1 \\
        1 &  0  & +1 \\
        1 &  1  & -1 \\
        \hline
        \hline
        \end{tabular}
\hspace*{1cm}
    \raisebox{-2cm}{
\begin{overpic}[scale=\myFigureScale]{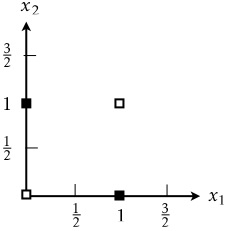}
    \end{overpic}}
    \caption{\figlab{C5S1XORFunction} The Boolean XOR function is not linearly separable.  Legend: \index{XOR function}
$\blacksquare$ corresponds to  $t\tomu = 1$, and $\Box$ to  $t\tomu = -1$.} 
\end{figure}

Up to now we discussed only one single neuron. If the classification problem requires several output neurons\index{output neuron}, each has its own weight vector $\ve w_i$ and \index{threshold}threshold $\theta_i$.
We can group the weight vectors into a \index{weight!matrix}weight matrix $\ma W$ as in Part \ref{part:hopfield}, so that 
the row vectors $\ve w_i^{\sf T}$ are the rows of 
the weight matrix $\ma W$. 

\section{Iterative learning algorithm}
\label{sec:ila}
In the previous Section we determined the weights and \index{threshold}\index{weight}threshold for the 
Boolean AND function by inspection\index{Boolean function}
(Figure~\figref{C5S1ANDFunction}). Now we discuss an algorithm that finds the weights iteratively. 
It is illustrated in Figure \figref{C5S1CT5}.
In panel ({\bf a}), the pattern $\ve x^{(8)}$ ($t^{(8)}=1$) is on the wrong side of the 
decision boundary. \index{decision boundary}
\begin{figure}
\centering
\begin{overpic}[scale=\myFigureScale]{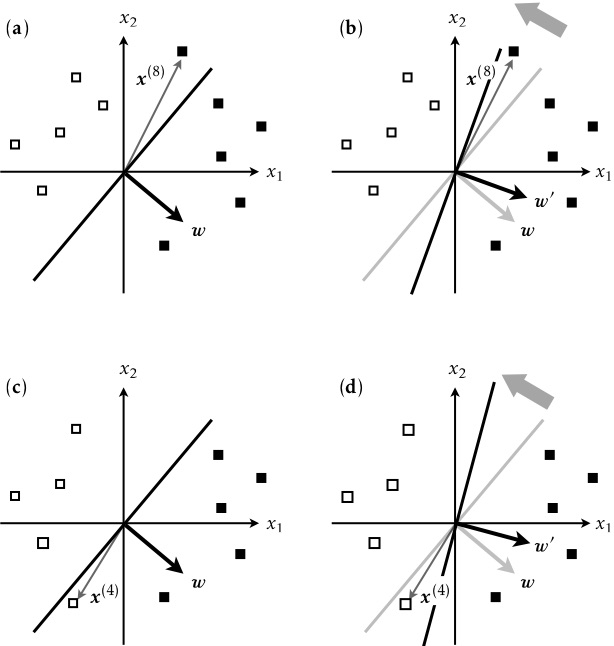}
\end{overpic}
\caption{\figlab{C5S1CT5} Illustration of the learning algorithm. 
In panel ({\bf a}) the $t={+}1$ pattern $x^{(8)}$ is on the 
wrong side of the decision boundary (solid red line). To correct the error
the weight must be rotated anti-clockwise [panel ({\bf b})].
In panel ({\bf c}) the $t=-1$ pattern $x^{(4)}$ is on the 
wrong side of the decision boundary. To correct the error
the weight must be rotated anti-clockwise [panel ({\bf d})].}
\end{figure}
In order to correct this error, one turns the decision boundary anti-clockwise.
To this end,  one {\em adds} a small multiple of the pattern vector $\ve x^{(8)}$ to the weight vector
\begin{equation}
\ve w' = \ve w + \delta\!\ve w \quad\mbox{with}\quad \delta\! \ve w = \eta \ve x^{(8)}\,.
\end{equation}
The parameter $\eta>0$ is called the {\em learning rate}\index{learning rate|textbf}. It must be small, so that the decision boundary
is not rotated too far. The result is shown in panel ({\bf b}). Panel ({\bf c}) shows another case,
where pattern $\ve x^{(4)}$ ($t^{(4)}=-1$) is on the wrong side of the decision boundary. In order to turn the decision
boundary in the right way, anti-clockwise, one {\em subtracts} a small multiple of $\ve x^{(4)}$:
\begin{equation}
\ve w' = \ve w + \delta\!\ve w \quad\mbox{with}\quad \delta\! \ve w = -\eta \ve x^{(4)}\,.
\end{equation}
These two {\em learning rules}\index{learning rule} combine to
the learning rule of Rosenblatt \cite{Rosenblatt1958}:
\begin{equation}
\ve w' = \ve w + \delta\!\ve w\tomu \quad\mbox{with}\quad \delta\! \ve w\tomu = \eta  t^{(\mu)}\ve x^{(\mu)}\,.
\end{equation}
For more than one neuron, the rule reads
\begin{equation}
\label{eq:HReta}
 w_{ij}' = w_{ij} + \delta\! w_{ij}\tomu \quad\mbox{with}\quad 
\delta\!  w_{ij}\tomu = \eta  t_i^{(\mu)} x_j^{(\mu)}\,.
\end{equation}
This rule is reminiscent of Hebb's rule (\ref{eq:HR}), except that
here inputs and outputs are associated with distinct units. Therefore
we have $t_i^{(\mu)} x_j^{(\mu)}$ instead of $x_i^{(\mu)} x_j^{(\mu)}$.
One applies (\ref{eq:HReta}) iteratively for a sequence of randomly chosen patterns $\mu$, until the problem is solved. This corresponds to adding a little bit of Hebb's rule in each iteration.
To ensure that the algorithm stops when the problem is solved, one can use
the learning rule \cite{hertz1991introduction}
\begin{equation} 
\label{eq:HReta2}
\delta\!  w_{ij}\tomu = \eta  (t_i^{(\mu)}- O_i\tomu ) x_j\tomu \,.
\end{equation}
\index{decision boundary|)}

\section{Gradient descent  for linear units}
\index{gradient descent|(}
\label{sec:gdl}
In this Section, the learning algorithm (\ref{eq:HReta2}) is derived in a different way,
by minimising an energy function using gradient descent. 
This requires differentiation, therefore
we must choose a differentiable activation function. The simplest choice is 
a linear activation function\index{activation function!linear|textbf}, $g(b) = b$. We set $\theta=0$,
so that the network computes:
\begin{equation} O_{i}\tomu = \sum_{k}w_{ik} x_{k}\tomu\,. \eqnlab{C5S5Output}
\end{equation}
A neuron with a linear activation function is called a 
{\em linear unit}\index{unit!linear|textbf}.
The outputs $O_{i}\tomu$ assume continuous values, 
but not necessarily the \index{target} targets $t_{i}^{(\mu)}$.
For linear units\index{unit!linear}, the classification problem
\begin{equation} O_{i}\tomu = t_{i}^{(\mu)} \quad\mbox{for}\quad i=1,\ldots,N\quad
\mbox{and}\quad \mu=1,\ldots,p
\eqnlab{C5S5OutputInequality} 
\end{equation} 
has the formal solution
\begin{equation} w_{ik} = \frac{1}{N} \sum_{\mu\nu} t_{i}^{(\mu)} \Big(\ma Q^{-1}\Big)_{\mu\nu} x_{k}^{(\nu)} \,.
\eqnlab{C5S5GradientWeights} 
\end{equation} 
This can be verified by inserting Equation \eqnref{C5S5GradientWeights} into 
\eqnref{C5S5Output}.
Here $\ma Q$ is the overlap matrix with elements\index{overlap matrix} 
\begin{equation} Q_{\mu\nu} = \tfrac{1}{N} \ve x^{(\mu)} \cdot \ve x^{(\nu)}  \eqnlab{C5S5GradientQ}  
\end{equation}
(Section \ref{sec:cp1}).  
For the solution \eqnref{C5S5GradientWeights} to exist, 
the matrix $\ma Q$ must be invertible.\index{matrix!inverse}
As mentioned in Section \ref{sec:cp1}, this requires that $p \leq N$, because otherwise the input-pattern vectors are {\em linearly dependent}\index{input pattern!linearly dependent}, and thus also the columns (and rows) of $\ma Q$. If the matrix $\ma Q$ has linearly dependent columns or rows, it cannot be inverted. \label{ldp}

Let us assume that the input patterns\index{input pattern}  are linearly independent, 
so that the solution \eqnref{C5S5GradientWeights}
exists.  In this case we can find the solution iteratively. 
To this end one defines the energy function
\index{energy function|textbf}
\begin{equation}
\label{eq:Hw}
H= \frac{1}{2} \sum_{i\mu}\Big(t_i^{(\mu)} - O_{i}\tomu\Big)^{2} \,.
\end{equation}
This function is non-negative, and it vanishes when all ouputs equal the corresponding targets, for all patterns.

The energy function (\ref{eq:Hw}) is regarded as a function of the weights $w_{ij}$, unlike the energy function
in Part \ref{part:hopfield} which is a function of the state-variables of the neurons.
The goal is now to find weights that minimise $H$. If the input patterns are linearly independent, $H$ vanishes at the global miminum, corresponding to the desired solution of the problem (Exercise 5.1).
Let us use {\em gradient descent}\index{gradient descent}
to minimise $H$,
\begin{equation} 
\label{eq:HReta0}
w_{mn}' = w_{mn}+\delta\! w_{mn}\quad\mbox{with weight increments}\quad
\delta\! w_{mn} = -\eta \frac{\partial H}{\partial w_{mn}} \,.
\end{equation}
\index{weight increment[textbf}
with learning rate\index{learning rate} $\eta>0$. 
This is analogous to Equation~(\ref{eq:grad_asc}), apart from the minus sign. 
In Section \ref{sec:bm1} the goal was to maximise the target function, here we want to minimise $H$
by
taking
many downhill steps in search of the global minimum.
The derivatives in Equation (\ref{eq:HReta0}) are evaluated with the chain rule, together with 
Equation (\ref{eq:Kronecker_w_0}) which takes the form
\begin{equation}
\label{eq:Kronecker_w_1}
\frac{\partial w_{ij}}{\partial w_{mn}} = \delta_{im} \delta_{jn}
\end{equation}
for \index{weight!asymmetric weights}asymmetric weights.
This  yields the weight increments
\begin{equation}
\label{eq:HReta3}
\delta\! w_{mn} = \eta \sum_{\mu}\left(t_m^{(\mu)} - O_{m}\tomu\right) x_{n}\tomu  \,.
\end{equation}
This learning rule\index{learning rule}  is very similar to Equation (\ref{eq:HReta2}). 
One difference is that Equation (\ref{eq:HReta3}) contains a sum over all patterns. It is important to keep in mind also that
the activations functions are different, while Equation (\ref{eq:HReta2}) was derived for $g(b)={\rm sgn}(b)$,
the learning rule (\ref{eq:HReta3})  was derived for $g(b)=b$.
An advantage of the rule (\ref{eq:HReta3}) is that it is derived from an energy function. 
\index{energy function}
This helps to analyse the \index{convergence!gradient descent}convergence of the algorithm, as we have seen in Chapter~\ref{ch:dhn}.
\index{gradient descent|)}

\label{lu}
Linear units\index{unit!linear} [Equation \eqnref{C5S5Output}] are special. 
The Boolean AND problem (Figure \figref{C5S1ANDFunction}) does
\index{Boolean function}
not admit the solution \eqnref{C5S5GradientWeights}, 
even though the problem is linearly separable. Since the pattern vectors $\ve x^{(\mu)}$ are linearly dependent, the solution
\eqnref{C5S5GradientWeights} does not exist. Shifting the patterns or introducing a \index{threshold}threshold does not change this fact.  

In Section \ref{sec:mlp} we discuss how to solve problems that are not linearly separable using a hidden layer of neurons with non-linear activation functions. Note that introducing hidden layers with linear units does not help, 
\index{hidden layer}
because the resulting input-output mapping is still linear if all neurons have linear activation functions, so that only problems with $p \leq N$ can be solved. This is the main reason for using hidden layers with
non-linear activation functions. 

There are four points to keep in mind. First, 
if the the patterns are linearly independent, then we can use gradient descent to determine suitable weights (and \index{threshold}\index{weight}thresholds) of linear units. Second, in general
hidden layers with non-linear units are required, because  a single neuron with a continuous non-linear and monotonous activation function can only solve 
problems with linearly independent patterns (Exercise 5.11).
Third, for gradient descent for non-linear units we must require
that the activation function $g(b)$ is differentiable, or at least piecewise differentiable. Fourth, in
this case we calculate the gradients using the chain rule, resulting in factors of derivatives 
 $\tfrac{{\rm d}}{{\rm d}b}g(b)$. 
This is the origin of
the {\em vanishing-gradient problem}\index{vanishing gradient}
 (Chapter~\ref{chap:dl}).

\section{Classification capacity}
\label{sec:ct}
In Chapter \ref{sec:shn} we analysed the storage capacity
of Hopfield networks. The analogous question for the classification problem 
described in Section \ref{sec:act} is: how many patterns can a single neuron with activation
function $g(b) = \mbox{sgn}(b)$ classify? As in the case of Hopfield networks,
one can find a general answer for random binary classification problems.
\begin{figure}[t]
  \centering
    \begin{overpic}[scale = \myFigureScale]{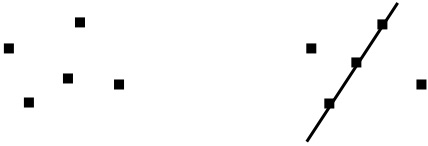}
    \end{overpic}
   \centering \caption{\figlab{C10S1GeneralPosition} Left: Five points in general position
in the plane. Right: these points are not in general position because three points lie
on a straight line.}
\end{figure}

Consider $p$ points with coordinate vectors $\ve x^{(\mu)}$ in $N$-dimensional 
input space\index{input space}, and assign random \index{target!random} targets:
\begin{equation}
    t^{(\mu)} = \begin{cases} +1 \quad \text{with probability }\frac{1}{2}\,,\\
    -1 \quad \text{with probability }\frac{1}{2}\,. \end{cases}
\end{equation}
This random classification problem is {\em homogeneously}\index{homogeneously linearly separable} linearly separable if we can
find an $N$-dimensional weight vector  $\ve w$,
so that $\ve w\cdot\ve x=0$ is a valid decision boundary that goes
\index{decision boundary}
through the origin:
\begin{equation}
    \ve w\cdot \ve x^{(\mu)} > 0 \quad \text{if} \quad t^{(\mu)} = 1\quad
\mbox{and} \quad  \ve w\cdot \ve x^{(\mu)} < 0 \quad \text{if} \quad t^{(\mu)} =- 1\,.
\end{equation}
So {\em homogeneously linearly separable} problems are binary classification
problems that are linearly separable by a hyperplane that contains
the origin. Problems with this property can be solved by a binary threshold unit\index{binary threshold unit} with \index{threshold}threshold $\theta=0$.
\begin{figure}[t]
  \centering
    \begin{overpic}[scale = \myFigureScale]{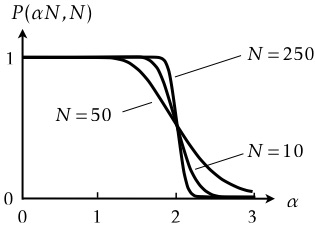}               
    \end{overpic}
   \centering \caption{\figlab{C10S1LambdaThreshold} Probability (\ref{eq:Ppm})
of separability as a function of $\alpha=p/N$ for three different values of the dimension $N$ of input space.\index{input space}  
Note the pronounced threshold near $\alpha = 2$, for large values of $m$.}
\end{figure}

Now assume that the points (including the origin) are in {\em general position} (Figure \figref{C10S1GeneralPosition}).
In this case {\em Cover's theorem} \cite{Cover} gives an expression for the probability that the random binary classification problem of $p$ patterns in dimension $N$ is homogeneously linearly separable:
\begin{equation}
\label{eq:Ppm}
\begin{split}
P(p,N) = \begin{cases} \left(\frac{1}{2}\right)^{p-1}\sum_{k=0}^{N-1} \binom{p-1}{k} &\text{for} ~  p> N\,, \\
                        1 & \text{otherwise}\,.
\end{cases}
\end{split}
\end{equation}
Here $\binom{l}{k} = \frac{l!}{(l-k)!k!}$ are the binomial coefficients,
for $l\geq k\geq 0$.  
Equation (\ref{eq:Ppm}) is proven by recursion, starting from a set of $p-1$ points in general position.  Assume that the number $C(p-1,N)$ of homogeneously linearly separable classification problems given these points is known.  After adding one more point, one can compute the $C(p,N)$ in terms of $C(p-1,N)$, and recursion yields Equation (\ref{eq:Ppm}).
Figure \figref{C10S1LambdaThreshold} shows this result as a function of $\alpha=p/N$ for different
values of $N$. For $p\leq N$, any random classification problem is homogeneously linearly separable. In this case the pattern vectors are linearly independent, so that the problem can also be solved by a linear unit\index{unit!linear} 
(Section \ref{sec:gdl}).
But a neuron with activation function $\mbox{sgn}(b)$ can classify problems
with more than $N$ patterns.
In the limit of $N\to\infty$, the function $P(\alpha N,N)$ approaches 
a step function $\theta_{\rm H}(2-\alpha)$ (Exercise 5.12). In this limit the maximal classification capacity is therefore $\alpha_{\rm max}=2$.

\begin{figure}[tb]
    \centering
\begin{overpic}[scale=\myFigureScale]{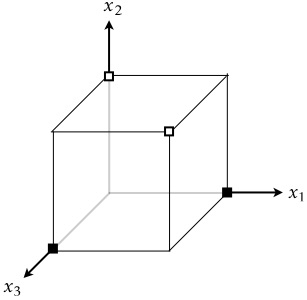}
    \end{overpic}
\vspace*{2mm}
\caption{\figlab{lift} The \index{XOR function} XOR problem can be solved by 
\index{embedding}embedding
into a three-dimensional input space\index{input space}.}
\end{figure}

What is the expected classification
capacity for finite values of $N$? To answer this question, consider a random sequence of patterns $\ve x^{(1)}, \ve x^{(2)}, \ldots$ and \index{target} targets $t^{(1)},t^{(2)},\ldots$ and ask \cite{Cover}: what is the distribution of the largest integer so that the problem $\ve x^{(1)}, \ve x^{(2)}, \ldots,\ve x^{(n)}$ is separable in dimension $N$, but $\ve x^{(1)}, \ve x^{(2)}, \ldots,\ve x^{(n)},\ve x^{(n+1)}$ is not? $P(n,N)$ is the probability that $n$ patterns
are linearly separable in $N$-dimensional input space\index{input space}.
We can write $P(n+1,N) = q(n+1|n) P(n,N)$ where $q(n+1|n)$ is
the conditional probability that $n+1$ patterns are linearly separable if
the $n$ patterns are. 
Then the probability that $n+1$ patterns are not separable
(but $n$ patterns are) reads $[{1}-q(n+1|n)]P(n,N) = P(n,N)-P(n+1,N)$.
We can interpret the right-hand side of this Equation  as a distribution $p_n$ of the random variable $n$,
the maximal number of separable patterns in dimension $N$:
\begin{equation}
    p_n = P(n,N) - P(n+1,N)\nonumber 
=\left(\frac{1}{2}\right)^{n} \binom{ n-1}{N-1} \quad\mbox{for}\quad  n = 0,1,2,\dots \,.
\end{equation}
It follows that the expected maximal number of separable patterns is
\begin{equation}
\label{eq:sum_npn}
\langle n\rangle  = \sum_{n=0}^{\infty} n p_n = 2N\,.
\end{equation}
So the expected classification capacity 
is twice the input  dimension:
\begin{align}
\langle\alpha_{\rm max}\rangle = 2\,.
\end{align}
This quantifies the notion that it is easier to separate patterns in 
higher-dimensional input space. 
As an illustration, consider the \index{XOR function} XOR problem which is not linearly separable in two-dimensional input space\index{input space}. The problem becomes separable when we {\em embed}\index{embedding|textbf} the points in three-dimensional space, for instance  by assigning $x_3=0$ to the  $t=+1$ patterns and $x_3=1$ to the $t=-1$ patterns (Figure \figref{lift}). 

\section{Multi-layer perceptrons}
\index{decision boundary|(}
\index{hidden layer|(}
\label{sec:mlp}
\begin{figure}[t]
      \centering  
\begin{overpic}[scale=\myFigureScale]{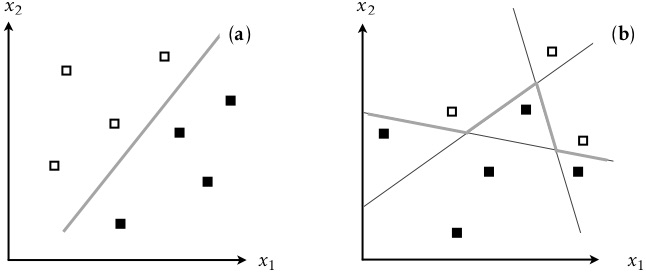}
        \end{overpic}
 \caption{\label{fig:sns} ({\bf a}) Linearly
separable problem. ({\bf b}) Problems that are not linearly separable can be solved by a piecewise linear decision boundary\index{decision boundary!piecewise linear}.  Legend: $\blacksquare$ corresponds to  $t\tomu = 1$, and $\Box$ to  $t\tomu = -1$.}
\end{figure}
In Sections \ref{sec:act} and \ref{sec:ila}  we discussed how to solve linearly separable problems [Figure \ref{fig:sns}({\bf a})]. The aim of this Section is to show that non-separable problems like the one    in Figure \ref{fig:sns}({\bf b}) can be solved by a perceptron with one hidden layer.
A network that does the trick for the classification problem in Figure \ref{fig:sns}({\bf b}) is depicted in Figure \figref{C5S4HiddenNetwork}.
\begin{figure}[b!]
  \centering
    \begin{overpic}[scale=\myFigureScale]{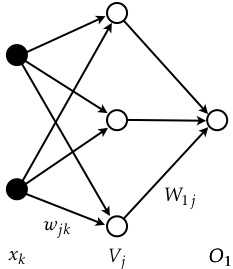}
    \end{overpic}
    \caption{\figlab{C5S4HiddenNetwork} Hidden-layer perceptron to solve the problem shown in Figure \ref{fig:sns} ({\bf b}). }
\end{figure} 
As in the previous Section, all neurons have the signum function as activation function, 
with possible outputs $\pm 1$:
\begin{equation}
\begin{split}
V_{j}\tomu &= {\rm sgn}\left(b_{j}\tomu\right) \hspace{1pt} \quad\mbox{with}\quad b_{j}\tomu = \sum_{k} w_{jk} x_{k}\tomu -\theta_j\,,\\
O_{1}\tomu &= \mbox{sgn}\left(B_{1}\tomu\right) \quad\mbox{with}\quad   B_{1}\tomu = \sum_{j}  W_{1j} V_{j}\tomu -\Theta_1\,.
\end{split} \eqnlab{C6S1VJO1}
\end{equation} 
Each of the three neurons in the hidden layer has its own decision boundary. 
\index{decision boundary}
The idea is to choose the weights $w_{jk}$ and the thresholds $\theta_j$ in such a way
that the three decision boundaries partition the input plane into distinct regions, so that each region
contains either only $t=-1$ patterns or $t={+}1$ patterns \cite{Horner2001}.

How this construction works is shown in Figure \figref{C5S4RoofPointsFilled}.
\begin{figure}[bt]
  \centering
\raisebox{-2.7cm}{
    \begin{overpic}[scale=\myFigureScale]{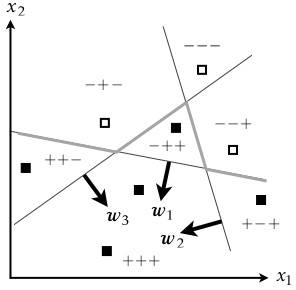}
    \end{overpic}}
\hspace*{15mm}
\begin{tabular}{lcccc}
&&& \\\hline\hline
 $V_{1}$ &  $V_{2}$ &  $V_{3}$ & target\\
\hline
  $-$  & $-$  &  $-$&$-1$\\
  $+$  & $-$  &  $-$&-\\
  $-$  &$+$   &  $-$&$-1$\\
  $-$  & $-$  &  $+$&$-1$\\
  $+$  & $+$  &  $-$&$+1$\\
  $+$  & $-$  &  $+$&$+1$\\
  $-$  & $+$  &  $+$&$+1$\\
  $+$  & $+$  &  $+$&$+1$\\
\hline
\hline
\end{tabular}
    \caption{\figlab{C5S4RoofPointsFilled}Left: decision boundaries 
[Figure \ref{fig:sns}({\bf b})], regions,
\index{decision boundary}
and the corresponding binary codes determined by the states of the hidden neurons.
\index{hidden neuron}  
 Legend: $\blacksquare$ corresponds to  $t\tomu = 1$, and $\Box$ to  $t\tomu = -1$.
Right: encoding of the regions and corresponding \index{target} targets. The region $+--$ does not exist.}
\end{figure}
\begin{figure}[bt]
  \centering
    \begin{overpic}[scale=\myFigureScale]{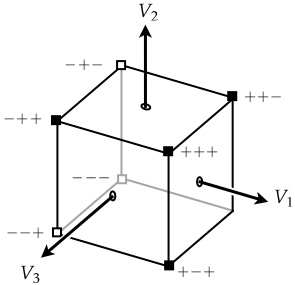}
    \end{overpic}
    \caption{\figlab{C5S4V1V2V3}Graphical representation of the output problem for the classification problem
shown in Figure \figref{C5S4RoofPointsFilled}.     }
\end{figure}
The left part of the Figure shows the  three decision boundaries
with their weight vectors, and how they divide the input plane into different regions
which contain either only $\Box$ or only $\blacksquare$. Each region bears a three-digit code made out of the symbols $+$ and $-$. The codes are determined
by the states of the hidden neurons.
A $+$ sign in the $j$-th entry of the code means that
$V_j= +1$. So the region in question is on the weight-vector side of the decision boundary $j$. A $-$ sign, by contrast, corresponds to $V_j=-1$. In this case the region is
on the other side of the decision boundary, the one opposite the weight vector.  The value table shows the \index{target} targets associated with each region, together with the code of the region. 

The weights $W_{1j}$ and the threshold $\Theta_j$ of the output neuron\index{output neuron} are chosen so
that it associates the correct target value with each region.
A graphical representation of the output problem is shown in Figure \figref{C5S4V1V2V3}.  This problem is linearly separable (Exercise 5.3).
\begin{figure}[t]
\centering
\begin{overpic}[scale=\myFigureScale]{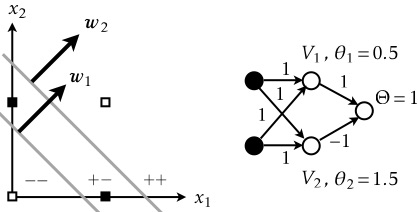}
\end{overpic}
\hspace*{15mm}
 \raisebox{18mm}{\begin{tabular}{lccc} 
\hline\hline $V_{1}$ & $ V_{2}$&$t$\\
        \hline
          $-$  & $-$  & -1\\
          $+$  & $-$ & +1\\
          $-$  & $+$ & -\\
          $+$  & $+$ & -1\\
        \hline
        \hline
        \end{tabular}}
    \caption{\figlab{C5S4XORWithHiddenGraph} Boolean XOR function: geometrical representation, network layout, and value table for the output neuron\index{output neuron}. The region $-\,+$ does not exist. \index{XOR function}
All neurons assume two possible states, $+1$ or $-1$.  Legend for the geometrical representation:
$\blacksquare$ corresponds to  $t\tomu = 1$, and $\Box$ to  $t\tomu = -1$.}
\end{figure}
The following function computes the correct output for each region:
\begin{equation}
    O_{1}\tomu = \mbox{sgn}\left(V_{1}\tomu + V_{2}\tomu + V_{3}\tomu \right)\,.
    \eqnlab{C6S1O1}
\end{equation}
This solves the binary classification problem\index{classification!problem, binary|textbf}  described in Figure~\figref{C5S4RoofPointsFilled}, but
note that the solution is not unique. There is a range of different
weights and \index{threshold}\index{weight}thresholds that solve the problem, and there are other solutions
based on different network layouts. Nevertheless, the solution illustrates how
non-linearly separable classification problems can be solved 
by adding a hidden layer to the network layout. The neurons in the hidden layer define segments of a piecewise linear decision boundary. 
More hidden neurons are needed if the decision boundary is very wiggly. 

Figure \figref{C5S4XORWithHiddenGraph} 
\index{XOR function}
shows another example, how to solve the Boolean XOR problem with a perceptron 
that has two neurons in a hidden layer, with activation functions $\mbox{sgn}(b)$, thresholds $\tfrac{1}{2}$
and $\tfrac{3}{2}$, and all weights equal to unity. The output neuron\index{output neuron} 
has weights $+1$ and $-1$ and unit threshold:
\begin{equation}
O_1 = \mbox{sgn}(V_1-V_2-1)\,.
\end{equation}
Minsky and Papert \cite{MinskyPapert} proved in 1969 that all Boolean functions can be represented by multilayer perceptrons, but that at least one hidden neuron must be connected
\index{hidden neuron}
to {\em all} input terminals.\index{input terminal}  
This means that not all neurons in the network are {\em locally} 
connected\index{local connection} (the neurons have only a few incoming weights). 
Since fully connected networks are much harder to train than locally connected ones, 
this was considered a shortcoming at the time.
Now, almost 50 years later, the perspective has changed.
Convolutional networks (Chapter \ref{chapter:con_net}) have only local connections to the inputs and can be trained to recognise objects in images with high accuracy.

In summary, perceptrons are trained on a training set $[\ve x^{(\mu)},\ve t^{(\mu)}]$, $\mu=1,\ldots,p$,
by moving the decision boundaries into the correct positions. This is achieved by repeatedly applying Hebb's rule
to adjust all weights. A related learning rule is obtained by gradient-descent \index{gradient descent} on the energy function (\ref{eq:Hw}).
Also, we have not discussed how to update the {\em thresholds}\index{threshold} yet, 
but it is clear that they too can be updated with gradient-descent learning.

Once all decision boundaries are in the right places we must ask: 
\index{decision boundary}
what happens if we apply the trained network to a new data set? Does it classify the new inputs correctly? In other words, can the network {\em generalise}?
An example is shown in Figure \figref{C5S5Validation}. Panel ({\bf a}) shows the result of training \index{training}
the network on a \index{training set}training set. The decision boundary separates $t=-1$ patterns from $t={+}1$ patterns, 
so that the network classifies all patterns in the training set correctly. In panel ({\bf b}) 
the trained network is applied to patterns in a {\em validation set}\index{validation!set}. 
We see that most patterns are correctly classified, save for one error. This means that
the energy function (\ref{eq:Hw}) is not exactly zero for the \index{validation!set}validation set. Nevertheless, the network 
does quite a good job. Usually it is not a good idea to try to precisely classify all patterns near the decision
boundary, because real-world data sets are subject to noise 
(Section \ref{sec:crossvalidation}).
\begin{figure}[bt]
  \centering
    \begin{overpic}[scale=\myFigureScale]{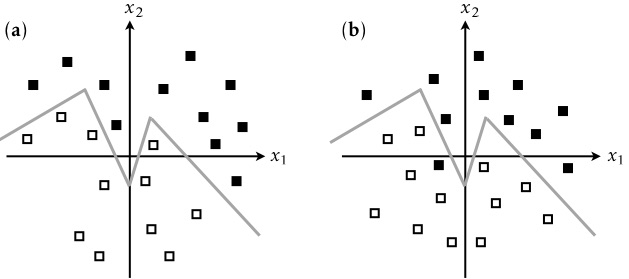}
        \end{overpic}
   \caption{\figlab{C5S5Validation} ({\bf a}) Result of \index{training}training the network on
a training set.  Legend:
$\blacksquare$ corresponds to  $t\tomu = 1$, and $\Box$ to  $t\tomu = -1$.
({\bf b}) Classification of a \index{validation!set}validation set.  One 
pattern is wrongly classified.}
\end{figure}
\index{hidden layer|)}

\section{Summary}
Perceptrons are layered feed-forward networks that can learn to classify data in a \index{training set}training set $[\ve x\tomu,\ve t\tomu]$.  For each input pattern $\ve x\tomu$, \index{input pattern} the network finds the correct \index{target} target vector  $\ve t\tomu$. We discussed the learning algorithm for a simple example: real-valued patterns with just two components, and one binary target.  This allowed us to represent the classification problem graphically, and to see how linearly separable classification problems can be solved by a simple perceptron.  There are three different ways of understanding how the perceptron learns. First, geometrically, the perceptron learn by moving decision boundaries into the correct locations. Second, this can be achieved by repeatedly adding a little bit of Hebb's rule. Third, these rules are similar to the learning rule derived from
gradient descent on the energy function (\ref{eq:Hw}). Cover's theorem quantifies the capacity of a simple perceptron to separate patterns with binary targets.  Finally we discussed how to solve non-linearly separable classification problems with perceptrons with a hidden layer.

\section{Further reading}
As mentioned in the Introduction, a short account of the history of perceptron research is the review by Kanal \cite{Kanal}. The remarkable book by Minsky and Papert explains the geometry of perceptron learning in great depth, 
and in a very elegant fashion. 
For a proof of Cover's theorem see Ref.~\cite{Sompolinsky}. 
\vfill\eject

\chapter{Stochastic gradient descent} \chlab{C6Chapter}
In Chapter \chref{C5Chapter} we discussed how a hidden layer helps to classify problems
that are not linearly separable. We explained how the decision boundary
\index{decision boundary}
in Figure  \figref{C5S4RoofPointsFilled} is represented in terms of the weights and thresholds of the hidden neurons\index{hidden neuron}, and introduced a 
\index{training!algorithm}training algorithm based on gradient descent.  In this Section, the training algorithm is discussed in more detail. 

Figure \figref{C5S1SingleLayerNetwork} shows the layout of the network to be trained.  There are $p$ input patterns\index{input pattern} $\ve x^{(\mu)}$ with $N$ components each, as before.  The output of the network has $M$ components:
\begin{equation}
    \ve O^{(\mu)} = \begin{bmatrix} O_1^{(\mu)} \\O_2^{(\mu)} \\ \vdots \\ O_M^{(\mu)} \end{bmatrix}\,,
\end{equation}
to be matched to the \index{target!vector}target vector $\ve t^{(\mu)}$. 
 The network shown in Figure \figref{C5S1SingleLayerNetwork}
computes
\begin{subequations}
\begin{align}
V_{j}\tomu &= g\left(b_{j}\tomu\right)\quad\mbox{with}\quad  b_{j}\tomu = \sum_{k=1}^N w_{jk}  x_{k}\tomu-\theta_j\,,\\
O_{i}\tomu &= g\left(B_{i}\tomu\right)\quad\mbox{with}\quad B_{i}\tomu = \sum_{j} W_{ij} V_{j}^{(\mu)}-\,\Theta_i\,.
\label{eq:62b}
\end{align}
\eqnlab{C6S2MultiLayerFlow}
\end{subequations}
Equation \eqnref{C6S2MultiLayerFlow} shows that the outputs are obtained in terms of nested activation functions.
They must be differentiable (or at least piecewise differentiable). Apart from that there is no need to specify them further at this point. 

\section{Chain rule and error backpropagation}
\index{backpropagation|(}
\label{sec:chainrule}
The network in Figure \figref{C5S1SingleLayerNetwork} is trained by gradient-descent\index{gradient descent} learning in the same
way as in Section \ref{sec:gdl}. The weight increments are
given by:\index{weight increment}
\begin{equation}
\label{eq:weightupdate6}
\delta\! W_{mn} = -\eta\frac{\partial H}{\partial W_{mn}} \quad\mbox{and}
\quad
\delta\! w_{mn} = -\eta\frac{\partial H}{\partial w_{mn}}\,,
\end{equation}
with energy function\index{energy function}
\begin{equation}
\label{eq:H3}
H = \frac{1}{2} \sum_{\mu i}\left(t_{i}^{(\mu)} - O_{i}\tomu\right)^2 \,.
\end{equation}
The small parameter $\eta>0$ in Equation (\ref{eq:weightupdate6}) is the learning rate\index{learning rate},
as in  Section \ref{sec:gdl}.
The derivatives of the energy function are evaluated with the 
{\em chain rule}\index{chain rule|textbf}. For the weights connecting
to the output layer we apply the chain rule once
\begin{subequations}
\begin{align} \frac{\partial H}{\partial W_{mn}} &= -\sum_{\mu i}\Big(t_{i}^{(\mu)}-O_{i}\tomu\Big)
\frac{\partial O_i\tomu}{\partial W_{mn}}\,,
\end{align}
and then once more, using Equation (\ref{eq:Kronecker_w_1}):
\begin{align}
\frac{\partial O_i\tomu}{\partial W_{mn}} 
&= g'(B_i\tomu) \delta_{im} V_n\tomu\,.
\end{align}
\end{subequations}
Here $g'(B)={\rm d}g/{\rm d}B$ is the derivative of the activation function with respect to the local field $B$, \index{local field} 
and $\delta_{im}$ is the Kronecker delta\index{Kronecker delta}: $\delta_{im}$=1 if $i=m$ and zero otherwise.

An important point is that the states $V_j$ of the neurons in the 
\index{hidden layer} hidden layer do not depend on $W_{mn}$, because these neurons do not have incoming connections with these weights, a consequence of the {\em feed-forward layout}\index{feed forward network} of the network.
In summary we obtain for the \index{weight increment}increments of the weights connecting to the output layer:
\begin{subequations}
\begin{equation}
    \delta\! W_{mn} = -\eta\frac{\partial H}{\partial W_{mn}} =
    \eta \sum_{\mu=1}^p \left(t_m\tomu-O_{m}\tomu\right)g'\left(B_{m}\tomu\right)
V_n^{(\mu)} \equiv \eta \sum_{\mu=1}^p \Delta_{m}\tomu V_{n}\tomu\,.
\eqnlab{C6S2SynapticWeightsOutput}
\end{equation}
The quantity 
\begin{equation}
\label{eq:def_output_error}
\Delta_m\tomu = (t_m\tomu-O_{m}\tomu)g'(B_{m}\tomu)
\end{equation}
\end{subequations}
is a weighted output {\em error} \index{output error|textbf}: it vanishes when $O_{m}\tomu = t_m\tomu$. 
The weights connecting to the \index{hidden layer} hidden layer are adjusted in a similar fashion, by
applying the chain rule four times:
\begin{subequations}
\begin{align} \frac{\partial H}{\partial w_{mn}} &= -\sum_{\mu i}\Big(t_{i}^{(\mu)}-O_{i}\tomu\Big)
\frac{\partial O_i\tomu}{\partial w_{mn}}\,,\\
\label{eq:Oderiv}
\frac{\partial O_i\tomu}{\partial w_{mn}} 
 &= \sum_l \frac{\partial O_i\tomu}{\partial V_l^{(\mu)}} \frac{\partial V_l\tomu}{\partial w_{mn}}\,,\\
\frac{\partial O_i\tomu}{\partial V_l^{(\mu)}} & = g'(B_i\tomu) W_{il}\,,\\
\frac{\partial V_l\tomu}{\partial w_{mn}} &= g'(b_l\tomu) \,\delta_{lm} x_n\tomu\,.
\end{align}
\end{subequations}
Here we used Equation (\ref{eq:Kronecker_w_1}). 
With the definition of the output error\index{output error}, $\Delta_i\tomu$, Equation (\ref{eq:weightupdate6}) yields:
\begin{equation}
    \delta\! w_{mn}
    = \eta \sum_{\mu}\sum_{i}\Delta_i\tomu W_{im}g'\left(b_{m}\tomu\right) x_n\tomu
\equiv\eta\sum_\mu \delta_{m}\tomu x_{n}\tomu\,.
\eqnlab{C6S2SSynapticWeightsHidden}
\end{equation}
The last equality defines weighted errors\index{error},
\begin{equation}
    \delta_{m}\tomu = \sum_{i} \Delta_{i}\tomu W_{im} g'\left(b_{m}\tomu\right)\,,
    \eqnlab{C6S2Delta}
\end{equation}
\index{hidden layer}
associated with the hidden layer.
Note that the $\delta_m^{(\mu)}$ vanish when the output errors $\Delta_i^{(\mu)}$ are zero.\index{output error}
Equation \eqnref{C6S2Delta} shows that the errors are determined recursively. 
The neuron states are also updated recursively, Equation \eqnref{C6S2MultiLayerFlow}, but there is
an important difference between Equations \eqnref{C6S2Delta} and \eqnref{C6S2MultiLayerFlow}.
The feed-forward structure of the layered network\index{network!layered} implies that the neurons are updated from left to right. Equation \eqnref{C6S2Delta}, by contrast, says that the errors\index{error!backpropagation|textbf}  are updated from right to left, from the output layer to the hidden layer.  The term {\em backpropagation}\index{backpropagation|textbf} refers to this difference: the neurons are updated forward, the errors are updated backwards.

In terms of the errors $\Delta_m\tomu$ and $\delta_m\tomu$,
\index{error}
the \index{weight increment}weight increments have the same form for both layers:
\begin{equation}
 \delta\! W_{mn} = \eta \sum_{\mu=1}^p \Delta_{m}\tomu V_{n}\tomu\quad\mbox{and}\quad
    \delta\! w_{mn} = \eta \sum_{\mu=1}^p \delta_{m}\tomu x_{n}\tomu \,.
\eqnlab{C6S2GeneralRule}
\end{equation}
The rule \eqnref{C6S2GeneralRule} is also called $\delta$-rule\index{delta rule} \cite{hertz1991introduction}. 
The thresholds\index{threshold} are adjusted in a similar way:
\begin{subequations}
 \eqnlab{C6S2BackpropThreshold}
    \begin{align}  \label{eq:Theta_output}
        \delta \Theta_{m} &= -\eta \frac{\partial H}{\partial \Theta_{m}} = \eta\sum_{\mu=1}^p\left(t_m^{(\mu)} - O_{m}\tomu\right)\left[-g'\left(B_{m}\tomu\right)\right] 
=  -\eta\sum_{\mu=1}^p\Delta_m^{(\mu)}\,,
\end{align}
\begin{align}
        \delta \theta_{m} &= -\eta \frac{\partial H}{\partial \theta_{m}}  
=  \eta \sum_{\mu=1}^p\sum_{i} \Delta_{i}\tomu W_{im}\left[-g'\left(b_{m}\tomu\right)\right] = -\eta\sum_{\mu=1}^p\delta_m^{(\mu)}\,.
    \end{align}
\end{subequations}
So, the general form for the \index{threshold}threshold increments is analogous to Equation \eqnref{C6S2GeneralRule}
\begin{equation}
\label{eq:update_th}
\delta\Theta_m = -\eta \sum_{\mu=1}^p\Delta_m^{(\mu)}\quad\mbox{and}\quad
\delta\theta_m = -\eta \sum_{\mu=1}^p\delta_m^{(\mu)}\,,
\end{equation}
but without the state variables of the neurons (or the inputs). 
A way to remember the difference between Equations \eqnref{C6S2GeneralRule} and (\ref{eq:update_th}) is to note
that the formula for the \index{threshold!increments}threshold increments looks like the one for the \index{weight increment}weight increments if one
sets the \index{state!value} state values of the neurons to $-1$. This follows from  Equation \eqnref{C6S2MultiLayerFlow}.

The backpropagation rules \eqnref{C6S2GeneralRule} and (\ref{eq:update_th}) contain sums over patterns. This corresponds to feeding
all patterns at the same time to compute the increments of weights and thresholds ({\em batch} training\index{batch training}). Alternatively one may choose a single pattern, update the weights by backpropagation, and then continue to iterate these training steps many times
({\em sequential} training\index{training!sequential}).
One iteration corresponds to feeding a single pattern, $p$ iterations are called one {\em epoch}\index{epoch|textbf} (in batch training\index{batch training}, one iteration corresponds to one epoch).
If one chooses the patterns randomly, then sequential \index{training!sequential} training results in {\em stochastic gradient descent}\index{stochastic gradient descent}:
\begin{subequations}
\begin{align}
\label{eq:update_sgd}
 \delta\! W_{mn} &= \eta  \Delta_{m}\tomu V_{n}\tomu\quad\mbox{and}\quad
    \delta\! w_{mn} = \eta  \delta_{m}\tomu x_{n}\tomu \,,\\
\delta\Theta_m &= -\eta \Delta_m^{(\mu)}\quad\mbox{and}\quad
\delta\theta_m = -\eta \delta_m^{(\mu)}\,.
\end{align}
\end{subequations}
\index{backpropagation|)}
Since the sum
over pattern is absent, the steps do not necessarily decrease the energy function. Their directions fluctuate, but the average \index{weight increment}weight increment (averaged over all patterns) points downhill.
The result is a {\em stochastic path}\index{stochastic path} through parameter space,
less prone to getting stuck in \index{local minimum}local minima (but see Section \ref{sec:further_reading_7}).

\section{Stochastic gradient-descent algorithm}
\seclab{C6S2BPAlgo}
\index{hidden layer|(}
The stochastic-gradient descent formulae derived in the previous Section 
were derived for a network with one hidden layer. 
This Section describes the details of the stochastic-gradient algorithm for 
deep networks with many hidden layers. 
To this end we need to adapt our notation, as described in Figure \figref{C6S2PartOfMultilayerNetwork}.
 We label
the layers by the index $\ell$. The layer of input terminals\index{input terminal} has label $\ell=0$, while layer $\ell=L$
denotes the layer of output neurons\index{output neuron}.  The state variables for the neurons in layer $\ell $ are $V_j^{(\ell)}$, the weights connecting into these neurons from the left are $w_{jk}^{(\ell)}$, the errors associated with layer $\ell$ are 
denoted by $\delta_k^{(\ell)}$. In this notation, Equations \eqnref{C6S2MultiLayerFlow} read:
\begin{align} 
V_{j}^{(\ell)} = g\Big(\sum_{k}w_{jk}^{(\ell)}V_{k}^{(\ell-1)}-\theta_j^{(\ell)}\Big)\,.
\end{align}
\begin{figure}[p]
  \centering
\begin{overpic}[scale=\myFigureScale]{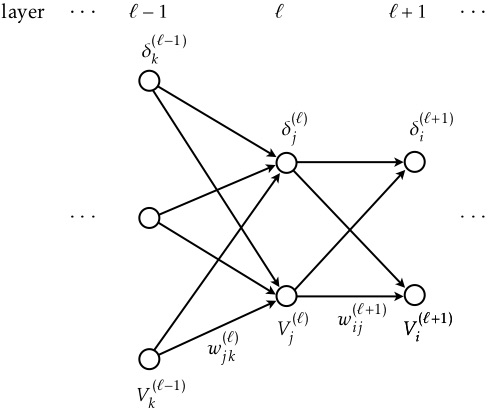}
    \end{overpic}
   \centering \caption{\figlab{C6S2PartOfMultilayerNetwork}
Illustrates the notation used in Algorithm \ref{bp_algorithm}.}
\end{figure}
\begin{algorithm}[p]
\caption{\label{bp_algorithm} stochastic gradient descent}
\begin{algorithmic}
\STATE initialise  weights $w_{mn}^{(\ell)}$ to random numbers, thresholds to zero, $\theta_m^{(\ell)}=0$;
\FOR {$\nu=1,\ldots,\nu_{\rm max}$}
   \STATE choose a value of $\mu$ and  apply pattern $\ve x\tomu$ to input layer, $\ve V^{(0)} \leftarrow \ve x\tomu$;
   \FOR {$\ell=1,\ldots,L$}
      \STATE  propagate forward:  $V_{j}^{(\ell)} \leftarrow  g\Big(\sum_{k}w_{jk}^{(\ell)}V_{k}^{(\ell-1)}-\theta_j^{(\ell)}\Big)$;
\label{step:propagate forward}
   \ENDFOR
   \STATE compute errors for output layer: $\delta_{i}^{(L)} \leftarrow  g'(b_{i}^{(L)})(  t_i- V_{i}^{(L)})$;
   \FOR{$\ell =L,\ldots,2$}
       \STATE propagate backward: $\delta_{j}^{(\ell-1)} \leftarrow  \sum_{i}\delta_{i}^{(\ell)} w_{ij}^{(\ell)}
g'(b_{j}^{(\ell-1)})$;
     \ENDFOR
\FOR{$\ell=1,\ldots,L$}
       \STATE change weights and thresholds: $w_{mn}^{(\ell)} \leftarrow  w_{mn}^{(\ell)} + \eta \delta_{m}^{(\ell)} V_{n}^{(\ell-1)}$
and $\theta_{m}^{(\ell)} \leftarrow  \theta_{m}^{(\ell)} -  \eta \delta_{m}^{(\ell)}$;
   \ENDFOR
\ENDFOR
\end{algorithmic}
\end{algorithm}
Repeating the steps outlined in the previous Section, we arrive at the update formulae
\begin{align}
\label{eq:w_update_general}
\delta w_{mn}^{(\ell)} = \eta \delta_{m}^{(\ell)} V_{n}^{(\ell-1)}\quad\mbox{and}\quad
\delta\theta_{m}^{(\ell)} = -  \eta \delta_{m}^{(\ell)}\,,
\end{align}
with errors
\begin{align}
\label{eq:delta_gradients}
\delta_{j}^{(\ell-1)}= \sum_i\big(t_i-V_i^{(L)}\big) \frac{\partial V_i^{(L)}}{\partial V_j^{(\ell-1)}} g'(b_j^{(\ell-1)})\,,
\end{align}
where $b_j^{(\ell)} = \sum_k w_{jk}^{(\ell)} V_k^{(\ell-1)} -\theta_j^{(\ell)}$
is the local field\index{local field}  of $V_j^{(\ell)}$. It involves
the matrix-vector product between the \index{weight!matrix}weight matrix $\ma W^{(\ell)}$
and the vector $\ve V^{(\ell-1)}$.
Evaluating the gradients ${\partial V_i^{(L)}}/{\partial V_j^{(\ell-1)}}$ with  the chain rule, 
one obtains the recursion
\begin{equation} 
\label{eq:delta_rec_2}
\delta_{j}^{(\ell-1)} =  \sum_{i}\delta_{i}^{(\ell)} w_{ij}^{(\ell)}
g'(b_{j}^{(\ell-1)})\,,
\end{equation}
with initial condition $\delta_i^{(L)} = (t_i-V_i^{(L)}) g'(b_i^{(L)})$.
For one hidden layer, Equation (\ref{eq:delta_rec_2}) is equivalent to \eqnref{C6S2Delta}.
The result of the recursion (\ref{eq:delta_rec_2}) is  a vector $\ve \delta^{(\ell-1)}$ with components $\delta_{j}^{(\ell-1)}$, 
obtained by component-wise multiplication of 
$[{\ma W^{(\ell)}}^{\sf T} \ve \delta^{(\ell)}]_j$ with $g'(b_{j}^{(\ell-1)})$. 
Component-wise multiplication of vectors is sometimes 
called Schur or Hadamard product \index{product!Schur}\cite{Greub}\index{product!Hadamard}, denoted by $\ve a \odot \ve b = [a_1b_1,\ldots, a_N b_N]^{\small \sf T}$. It does not have a geometric meaning like the scalar product or the cross product of vectors, and therefore there is little point in using it. Also,  note that 
the vector $\ve\delta^{(\ell)}$ is multiplied by the transpose of \index{matrix!transpose}\index{weight!matrix, transpose}
the \index{weight!matrix}weight matrix, ${\ma W^{(\ell)}}^{\sf T}$, rather than by the \index{weight!matrix}weight matrix itself.
We return to this point in Section \ref{sec:rbp}.

The stochastic-gradient algorithm is summarised in Algorithm \ref{bp_algorithm}.
One feeds an input $\ve x^{(\nu)}$, updates the weights using (\ref{eq:w_update_general}), and iterates these steps
until the energy function (\ref{eq:Hw}) is deemed sufficiently small.
Note that the resulting weights and \index{threshold}\index{weight}thresholds are not unique.  In Figure \figref{C5S4XORWithHiddenGraph} all weights for the Boolean XOR function are equal to $\pm 1$. But the \index{training!algorithm}training algorithm  \index{XOR function}
\index{Boolean function}
\eqnref{C6S2GeneralRule}
corresponds to repeatedly adding \index{weight increment}weight increments. This may cause the weights to grow.

In practice, the stochastic gradient-descent dynamics may be too noisy. In this case it is
better to average over a small number of randomly chosen patterns. Such
a set is called {\em mini batch}\index{mini batch|textbf}, of size $m_B$ say. 
In {\em stochastic gradient descent with mini batches}  one replaces Equations \eqnref{C6S2GeneralRule} and (\ref{eq:update_th}) by
\begin{align}
\label{eq:minibatch}
    \delta\! W_{mn} &= \eta \sum_{\mu=1}^{m_B} \Delta_{m}\tomu V_{n}\tomu \quad\mbox{and}\quad \delta\! \Theta_{m} = -\eta \sum_{\mu=1}^{m_B} \Delta_{m}\tomu\,,\\
    \delta\! w_{mn} &= \eta \sum_{\mu=1}^{m_B} \delta_{m}\tomu x_{n}\tomu \quad\mbox{and}\quad \delta\! \theta_{m} = -\eta \sum_{\mu=1}^{m_B} \delta_{m}\tomu \,.
\nonumber
\end{align}
Sometimes the mini-batch rule is quoted with prefactors of $m_B^{-1}$ before the sums. 
The factors  $m_B^{-1}$ can just be absorbed in the learning rate\index{learning rate}, but
when comparing learning rates for different implementations one needs to check whether or not there are factors of $m_B^{-1}$ in front of the sums in Equation (\ref{eq:minibatch}).

How does one select which inputs to include in a \index{mini batch}mini batch? This is discussed below, in Section \ref{sec:pprid}: at the beginning of each epoch, \index{epoch}
one randomly {\em shuffles}\index{shuffle inputs} the sequence of the input patterns\index{input pattern} in the \index{training set}training set. Then
the first \index{mini batch}mini batch contains patterns $\mu=1,\ldots,m_B$, and so forth.

Common choices for the activation functions $g(b)$ are   
the {\em sigmoid} function\index{activation function!sigmoid|textbf} or tanh:
\index{activation function!tanh|textbf}
\begin{subequations}
\label{eq:activation_functions}
\begin{align}
\label{eq:sigmoid}
    g(b) &= \frac{1}{1+e^{-b}}\equiv\sigma(b)\,,\\
\label{eq:tanh}
    g(b) &= \tanh\left(b\right)\,.
\end{align}
\end{subequations}
\begin{figure}[t]
  \centering
    \begin{overpic}[scale=\myFigureScale]{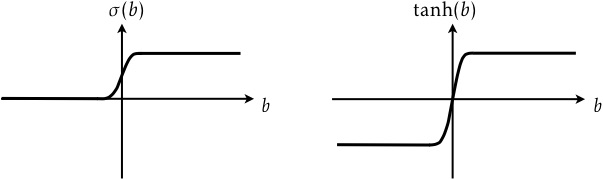}
    \end{overpic}
 \caption{\figlab{C6S1Saturation} Saturation of the activation 
\index{activation function!saturation}
functions (\ref{eq:activation_functions}). The derivatives 
$g'(b) \equiv \tfrac{{\rm d}}{{\rm d}b}g(b)$ of both activation functions tend to zero for large values of $|b|$.}
\end{figure}
In both cases, the derivatives 
can be expressed in terms of the function itself:
    \begin{align}
\label{eq:dsigma}
\tfrac{{\rm d}}{{\rm d}b}  \sigma(b) &= \sigma(b)[1-\sigma(b)]\,,\quad
        \tfrac{{\rm d}}{{\rm d}b} \tanh\left(b\right) = \left[1-\tanh^{2}\left( b\right)\right]\,.
    \end{align}
The second equality was used in Section \ref{sec:sc_mft}. The following short-hand notation for the derivative of the activation function $g(b)$ is common:  $g'(b) \equiv \tfrac{{\rm d}}{{\rm d}b}g(b)$.
\index{activation function!derivative of}

As illustrated in Figure \figref{C6S1Saturation}, the activation functions (\ref{eq:activation_functions}) saturate at large values of $|b|$: their derivatives $g'(b)$ tend to zero. Since the backpropagation rule
(\ref{eq:delta_gradients})
contains factors of $g'(b)$, this implies that the algorithm slows down if $|b|$ becomes too large.
For this reason, the initial weights and thresholds\index{weight!initial}\index{threshold!initial}  should be chosen 
so that the local fields\index{local field} $b$ are not too large in magnitude, to avoid that $g'(b)$ becomes too small.
A standard procedure is to take all weights to be initially randomly distributed, for example Gaussian with 
zero mean, and with a suitable variance. The performance of networks with many hidden layers ({\em deep} networks\index{network}\index{deep network|textbf}) can be
sensitive to the initialisation of the weights
(Section \ref{sec:vg}).

It is sometimes argued that the initial values of the \index{threshold!initial} thresholds are not so critical.
The idea is that they are learned more rapidly than the weights, at least initially, 
and a common choice is to initialise the thresholds to zero. Section \ref{sec:vg}
summarises a mean-field argument that comes to a different conclusion.
\index{hidden layer|)}

\section{Preprocessing the input data}
\label{sec:pprid}
\index{input!preprocessing|(}
It can be useful to preprocess the input data, although any preprocessing may remove information from the data.  Nevertheless, it is usually advisable to shift the data so the mean of each component over all $p$ patterns vanishes:
\begin{equation}
\label{eq:subtract_mean}
    \langle x_k\rangle=\frac{1}{p}\sum_{\mu = 1}^{p} x_{k}^{(\mu)} = 0\,.
\end{equation}
There are several reasons for this. First, large mean values can cause 
large gradients in the energy function\index{energy function} (Exercise 6.9)\index{energy function!large gradient} that are difficult to navigate with gradient descent. 
Different input-data variances in different directions have a similar effect. Therefore one {\em scales} the inputs \index{input}\index{input!scaling} so that the input-data distribution has the same variance in all directions (Figure \figref{C6S3ShiftScale}), equal to unity for instance:
\begin{equation}
\label{eq:sig_unity}
\sigma^2_{\!k} = \frac{1}{p}\sum_{\mu=1}^p\big(x_k^{(\mu)}-\langle x_k\rangle\big)^2=1\,.
\end{equation}
Second, to avoid that the neurons connected to the inputs saturate, their local fields\index{local field}  must not be too large (Section \secref{C6S2BPAlgo}). If one initialises the weights to Gaussian random numbers with mean zero and unit variance, large activations are quite likely if the distribution
of input patterns has a large mean or a large variance.
Third, enforcing zero input mean by shifting the input data avoids that the weights of the neurons in the first hidden layer must decrease or increase together \cite{LeCun1998}. Equation (\ref{eq:minibatch}) shows that the components of $\delta\!\ve w_m\propto \delta_m \ve x$ into 
\index{hidden neuron} hidden neuron $m$ are likely to have the same signs if the input data has a large mean.
This makes it difficult 
for the network to learn to differentiate.
In summary, it is advisable to shift and scale the input-data distribution so that it has mean zero and unit variance, as illustrated in Figure \figref{C6S3ShiftScale}. 
\begin{figure}[bt!]
  \centering
\begin{overpic}[scale = \myFigureScale]{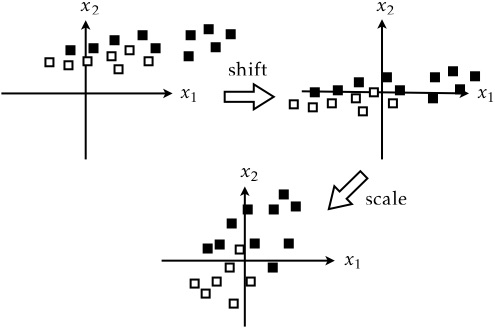}
\end{overpic}
\caption{\figlab{C6S3ShiftScale} Shift and scale the input data to achieve zero mean and unit variance.}
\end{figure}
\begin{figure}[b]
\centering
    \begin{overpic}[scale = \myFigureScale]{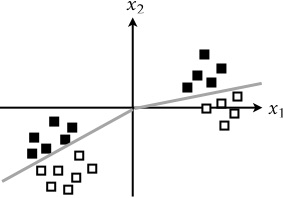}
    \end{overpic}
\caption{\figlab{C6S3Shuffle} When the input data falls into clusters as shown in this Figure, one should randomly pick
data from either cluster. The decision
boundary is shown as a solid gray line. It has different slopes for the two clusters.}
\end{figure}
The same transformation (using the mean values and scaling factors determined for the \index{training set}training set) should be applied
to any new data set that the network is supposed to classify after it has been trained on the training set.

Figure \figref{C6S3Shuffle} shows a distribution of inputs that falls into two distinct clusters. The difference between the clusters is sometimes
called {\em covariate shift}\index{covariate shift}, here {\em covariate} is just another term for input. Imagine feeding just inputs from one
of the clusters to the network. It will learn local properties
of the decision boundary,\index{decision boundary} instead of its global features. Such 
global properties are efficiently learned if the network is more frequently confronted with unfamiliar data.  
For sequential training\index{training!sequential} (stochastic gradient descent\index{stochastic gradient descent}) this is not a problem, 
because the sequence of input patterns presented to the network is random.  
However, if one trains with mini batches\index{mini batch}, the mini batches should contain randomly chosen patterns
 in order to avoid covariate shifts. To this end one randomly {\em shuffles}\index{input pattern!shuffle} 
the sequence of the input patterns in the training set\index{training set}, at the beginning of each epoch.
\index{epoch}

It is also recommended \cite{LeCun1998} to observe the output errors\index{output error} during training\index{training}. If the errors are similar for a number of subsequent learning steps,  the corresponding inputs appear familiar to the network. Larger errors correspond to unfamiliar inputs, and Ref.~\cite{LeCun1998} suggests to feed such inputs more often.

\index{principal component|(}
When the input data is very high dimensional, many input terminals\index{input terminal} are needed. This usually means that one should use  many neurons in the hidden layers. 
This can be problematic because it increases the risk of overfitting the input data. 
To avoid this as far as possible, one can reduce the dimensionality of the input data by {\em principal-component analysis}.
\index{principal component}  This method  allows to project high-dimensional  data to a lower dimensional subspace (Figure \figref{C6S3PCA}).
\begin{figure}[bt!]
  \centering
    \begin{overpic}[scale = \myFigureScale]{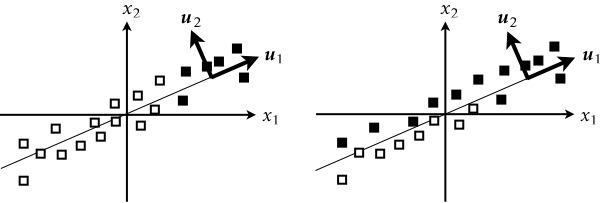}
    \end{overpic}
   \centering \caption{\figlab{C6S3PCA} 
Principal-component analysis (schematic). The data set on the left can
be classified keeping only the principal component $\ve u_1$ of the data. This is not true 
for the data set on the right.}
\end{figure}

The data shown on the left of Figure \figref{C6S3PCA} falls approximately onto a straight line, 
the principal direction $\ve u_1$.
We see that the coordinate orthogonal to the principal direction is not useful in classifying the data.
Consequently this coordinate can be disregarded, reducing the dimensionality of the data set. 
The idea of principal component analysis
is to rotate the basis in input space\index{input space} 
so that the variance of the data along the first axis of the new coordinate
system, $\ve u_1$, is maximal. One keeps the input components corresponding
to $\ve u_1$, discarding those corresponding to $\ve u_2$ (Figure~\figref{C6S3PCA}).

To determine the maximal-variance direction, consider the data variance along a unit direction vector $\ve v$ ($|\ve v|=1$):
\begin{equation}
\sigma^2_{\!\! v} = \langle(\ve x\cdot \ve v)^2\rangle - \langle\ve x\cdot \ve v\rangle^2 
= \ve v \cdot \ma C \ve v\,.
\end{equation}
Here 
\begin{equation}
\label{eq:dcm}
\ma C = \langle \delta\! \ve x\, \delta\!\ve x^{\sf T}\rangle
\quad \mbox{with}\quad \delta\!\ve x = \ve x -\langle \ve x\rangle
\end{equation}
is the data {\em covariance matrix}\index{covariance matrix|textbf}.
The variance $\sigma^2_{\!\! v}$ is maximal when
$\ve v$ points in the direction of the leading  \index{eigenvector!leading} eigenvector\index{eigenvector|textbf}
of the \index{covariance matrix}covariance matrix $\ma C$. This can be seen as follows.
The covariance matrix\index{covariance matrix} is symmetric, therefore its  \index{eigenvector}eigenvectors $\ve u_1,\ldots,\ve u_N$ form an 
orthonormal basis\index{basis!orthonormal} of input space\index{input space}. This allows us to express the  matrix
$\ma C$ as
\begin{equation}
\label{eq:decom}
\ma C = \sum_{\alpha=1}^N \lambda_\alpha \ve u_\alpha \ve u_\alpha^{\sf T}\,.
\end{equation}
The \index{eigenvalue|textbf}eigenvalues $\lambda_\alpha$ are non-negative. This 
follows from Equation (\ref{eq:dcm}) and the \index{eigenvalue}eigenvalue equation
$\ma C\ve u_\alpha = \lambda_\alpha \ve u_\alpha$. 
We arrange the \index{eigenvalue}eigenvalues by magnitude, $\lambda_1 \geq \lambda_2\geq \ldots \geq \lambda_N\geq 0$. Using Equation (\ref{eq:decom}) we can write for the variance  
\begin{equation}
\label{eq:var_decom}
\sigma^2_{\!\! v} = \sum_{\alpha=1}^N \lambda_\alpha v_\alpha^2
\end{equation}
with $v_\alpha = \ve v\cdot \ve u_\alpha$. We want to show that $\sigma^2_{\!\! v}$ is maximal  for $\ve v=\pm  \ve u_1$
subject to the constraint that $\ve v$ is normalised to  unity,
\begin{equation}
\label{eq:sum_sigma_constraint}
\sum_{\alpha=1}^N v_\alpha^2 = 1\,.
\end{equation} 
To ensure that this constraint is satisfied as the $v_\alpha$ are varied, one
introduces a \index{Lagrange multiplier}{\em Lagrange multiplier} 
$\lambda$ (Exercises 6.10 and 6.11). The constraint (\ref{eq:sum_sigma_constraint}) is multiplied with $\lambda$ and added to the 
target function (\ref{eq:var_decom}). The function to maximise reads
\begin{equation}
\label{eq:mathscrL}
\mathscr{L} = \sum_\alpha \lambda_\alpha v_\alpha^2 - \lambda\big(1-\sum_\alpha v_\alpha^2\big)\,.
\end{equation}
To find the maximum of $\mathscr{L}$, we determine its singular points,
defined by $\partial\mathscr{L}/\partial v_\beta=0$. This yields
$v_\beta(\lambda_\beta+\lambda)=0$. 
The maximum of $\mathscr{L}$ is obtained for $\lambda=-\lambda_1$, where $\lambda_1$ is the maximal \index{eigenvalue!maximal}eigenvalue of $\ma C$ with
\index{eigenvector}   eigenvector $\ve u_1$.
We conclude that all components $v_\beta$ must vanish, except one which
must equal unity. 
This shows that the variance $\sigma^2_{\!\! v}$ is maximised by the principal direction. 

In more than two dimensions there is commonly 
more than one direction along which the data varies significantly.
These $k$ principal directions correspond to the $k$  \index{eigenvector}eigenvectors
of $\ma C$ with the largest \index{eigenvalue}eigenvalues. This can be shown recursively. One projects the data to the subspace orthogonal to $\ve u_1$
by applying the \index{matrix!projection} projection matrix $\ma P_1 = \ma 1-\ve u_1 \ve u_1^{\sf T}$\index{projection}. Then one repeats the procedure outlined above, and finds
that the data varies maximally along $\ve u_2$. Upon iteration, one
obtains the $k$ principal directions $\ve u_1,\ldots, \ve u_k$.
Often there is a gap between the $k$ largest \index{eigenvalue}eigenvalues and the small ones (all close to zero). Then one can safely project
the data onto the subspace spanned by the $k$ principal directions.
If there is no gap then it is less clear what to do.

The data set shown on the right of 
Figure \figref{C6S3PCA} illustrates another problem. This data set is much
harder to classify if we use only the principal component alone. In this case we lose important information by projecting the data on its principal component.
\index{principal component|)}
\index{input!preprocessing|)}         

\section{Overfitting and cross validation}
\index{cross validation|(}
\label{sec:crossvalidation}
\begin{figure}[t]
  \centering
    \begin{overpic}[scale=\myFigureScale]{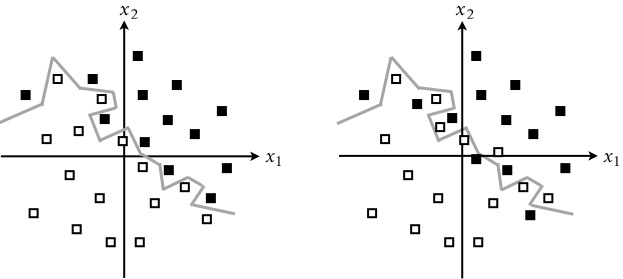}
    \end{overpic}
\caption{\figlab{C6S3CT10} Overfitting.\index{overfitting} Left: accurate representation of the decision boundary in the \index{training set}training set,\index{decision boundary}
for a network with a single hidden layer with 15 neurons. Right: this new data set differs from the first one just by a little bit of noise. The points in the vicinity of the decision boundary are not correctly classified.  Legend:
$\blacksquare$ corresponds to  $t\tomu = 1$, and $\Box$ to  $t\tomu = -1$.}
\end{figure}
The goal of supervised learning\index{supervised learning} is to generalise from a training set to new data.
Only general properties of the training set are of interest, not specific ones that are particular to the training set in question.  A neural network with more neurons may classify the input data better, because it more accurately
represents all specific features of the given data set. 
But a different set of patterns from the same
input distribution can look
quite different in detail, in which case the decision boundary may
not classify the new data very well (Figure \figref{C6S3CT10}). 
In other words, the network may fit too fine details (for instance noise in the training set) that have no general meaning. This problem,
illustrated in Figure \figref{C6S3CT10}, is called
 {\em overfitting}\index{overfitting|textbf}. 
The tendency to overfit is larger
for networks with more neurons. In general, we should look for
a compromise, reducing the tendency of the network to overfit 
at the expense of training accuracy.

One way of avoiding
overfitting\index{overfitting} is to use {\em cross validation}\index{cross validation}
and {\em early stopping}\index{early stopping}.
One splits the data into two sets:
a {\em training set}\index{training set|textbf} and a {\em validation set}\index{validation!set|textbf}. The idea is that these sets share the general features
to be learnt. But although training and validation sets\index{training set}\index{validation!set} are drawn
from the same distribution, they may differ in details that are not of interest.

While the network is trained on the \index{training set}training set, one monitors
not only the energy function\index{energy function} for the training set, but
also the energy function evaluated using the validation data. As long
as the network learns general features of the \index{input distribution}input distribution, 
both \index{training!energy}training and \index{validation!energy}validation energies decrease. But when the network starts to learn specific features of the training set, then the validation energy saturates, or may start to increase. At this point the
\index{training}training is stopped. The scheme is illustrated in Figure \figref{C6S3ErrorPlot}.
\begin{figure}[t]
  \centering
    \begin{overpic}[scale = \myFigureScale]{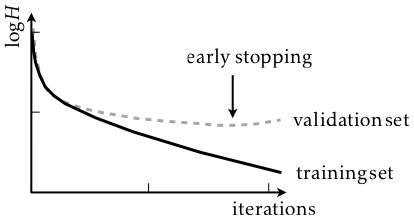}
    \end{overpic}
\vspace*{2mm}
   \centering \caption{\figlab{C6S3ErrorPlot} 
Progress of \index{training!error}training and \index{validation!error}validation errors. 
The plot is schematic, and the data is smoothed.
Based on simulations performed by Oleksandr Balabanov.
Shown is the natural logarithm of the energy functions
for the training set (solid line) and the validation\index{validation!set} 
set (dashed line) as a function of the number of training iterations.
The training is stopped when the \index{validation!energy}validation energy begins to increase. }
\end{figure}

Often the possible state values of the output neurons\index{output neuron} are continuous while the \index{target}targets assume only
discrete values. In this case one may also monitor the {\em classification error}
\index{classification error|textbf} of
the \index{validation!set}validation set.
The definition of the \index{classification error}classification error depends on the type of the classification problem.
For one single output neuron\index{output neuron} with \index{target}targets $t=0/1$, the \index{classification error}classification error is defined as
\begin{subequations}
\begin{equation}
\label{eq:Ce}
C=\frac{1}{p}\sum_{\mu=1}^p\big|t^{(\mu)} -\theta_{\rm H}(O^{(\mu)}-\tfrac{1}{2})\big|\,.
\end{equation}
If, by contrast, the \index{target}targets take the values $t=\pm 1$, then the \index{classification error}classification error reads:       
\begin{equation}
\label{eq:Cesgn}
C=\frac{1}{2p}\sum_{\mu=1}^p\big|t^{(\mu)} -\mbox{sgn}(O^{(\mu)})\big|\,.
\end{equation}
\end{subequations}
As a third example,  consider a classification problem where inputs must be classified into $M$ mutually exclusive classes,
such as the \href{http://yann.lecun.com/exdb/mnist/}{MNIST}\index{MNIST data set} data set of hand-written digits
(Section \ref{sec:MNIST}) where $M=10$. 
Another example is given in Table \ref{tab:2problems}, with $M=3$ classes.
In both cases, one of the \index{target}targets equals unity while all others equal zero.  As a consequence, the
\index{target}targets sum to unity: $\sum_i^M t_i^{(\mu)}=1$. 
Now assume that the network has sigmoid outputs\index{activation function!sigmoid}, $O_i^{(\mu)}=\sigma(b_i^{(\mu)})$.
To classify input $\ve x^{(\mu)}$ from the network outputs $O_i^{(\mu)}$ we define 
\begin{subequations}
\label{eq:Cey}
\begin{equation}
y_i^{(\mu)} = \begin{cases}
1 & \mbox{if $O_i^{(\mu)}$ is the largest of all outputs $i=1,\ldots,M$},\\
0 & \mbox{otherwise}.
\end{cases}
\end{equation}
Then the \index{classification error}classification error can be computed as
\begin{equation}
C = \frac{1}{2p}\sum_{\mu=1}^p\sum_{i=1}^M \big|t_i^{(\mu)}-y_i^{(\mu)}\big|\,.
\end{equation}
\end{subequations}
In all cases, the {\em classification accuracy}\index{classification!accuracy|textbf} is defined as
$(1-C)\,100\%$, it is usually quoted in percent.

The \index{classification error}classification error determines the fraction of
inputs that are classified wrongly. However, it contains less information
than the energy function\index{energy function}, which is in fact a mean-squared \index{output error} error of the outputs. 
This is illustrated in Table \ref{tab:2problems}. All three inputs are classified correctly,
but there is a substantial mean-squared error. This indicates that the classification is
not very reliable. 
\begin{table}[t]
 \begin{tabular}{lllll}
 \hline\hline
 \small $\mu$ &\small output $\ve O^{(\mu)}$          &\small target $\ve t^{(\mu)}$  &classification& \small correct?\\
 \hline
 $1$&\small  [0.4, 0.5, 0.4] & [0,1,0] & \small  versicolor & yes \\
 $2$&\small  [0.4, 0.3, 0.5] & [0,0,1] & \small  setosa & \small  yes \\
 $3$&\small  [0.6, 0.5, 0.4] & [1,0,0] & \small
virginica  & \small  yes \\
 \hline\hline
 \end{tabular}
 \caption{\label{tab:2problems}
Illustrates the
difference between energy function\index{energy function} and classification error.\index{classification error}  
The table shows network outputs for three different inputs from the iris data set, as well as the correct classifications.
All inputs are classified correctly, but the difference
between outputs and \index{target}targets  is substantial. }
 \end{table}
\index{cross validation|(}

\section{Adaptation of the learning rate}
\label{sec:alr}
It is tempting to choose larger learning rates\index{learning rate}, because they enable the network to escape more efficiently from shallow minima. But this can lead to problems when the energy function varies rapidly,\index{energy function}
causing the training to fail. To avoid this, one uses an {\em adaptive} 
\index{learning rule!adaptive}  learning rule, such as:
\begin{equation}
\label{eq:inertia}
\delta\! w_{mn}^{(t)}= -\eta \frac{\partial H}{\partial w_{mn}}\Bigg|_{\{w_{ij}\}=\{w_{ij}^{(t)}\}} + \alpha \delta\! w^{(t-1)}_{mn}\,.
\end{equation}
Here $t=1,2,\ldots,T$ labels the iteration number.
 We see that the increment at step $t$ depends not only on the instantaneous gradient,
but also on the weight change $\delta\! w^{(t-1)}_{mn}$ of the previous iteration. We say that the dynamics
becomes {\em inertial}\index{inertia}, the weights gain {\em momentum}\index{momentum|textbf}. The parameter $\alpha\geq 0 $ is called 
momentum constant\index{momentum!constant}. It determines how strong
the inertial effect is. We see that $\alpha=0$ corresponds to the usual backpropagation rule. 
When $\alpha$ is positive, then how does inertia change the learning process? Iterating Equation (\ref{eq:inertia}) yields 
\begin{equation}
\label{eq:momentum_3}
\delta\! w_{mn}^{(T)} = -\eta\sum_{t=0}^T \alpha^{T-t} \frac{\partial H}{\partial w_{mn}^{(t)}} \,.
\end{equation}
Here and in the following we use the short-hand notation 
\begin{equation*} 
\frac{\partial H}{\partial w_{mn}^{(t)}}\equiv \frac{\partial H}{\partial w_{mn}}\Bigg|_{\{w_{ij}\}=\{w_{ij}^{(t)}\}}\,.
\end{equation*}
Equation (\ref{eq:momentum_3}) shows that $\delta\! w_{mn}^{(T)}$ is a weighted average of the gradients encountered during training.
Now assume that the training is stuck in a shallow minimum.
Then the gradient ${\partial H}/{\partial w_{mn}^{(t)}}$ remains roughly constant through many time steps. To illustrate what happens, let us assume
that  
${\partial H}/{\partial w_{mn}^{(t)}}={\partial H}/{\partial w_{mn}^{(0)}}$ for $t=1,\ldots,T$. In this case we can write
\begin{equation}
\label{eq:alphasum}
\delta\! w_{mn}^{(T)} \approx -\eta \frac{\partial H}{\partial w_{mn}^{(0)}}  \sum_{t=0}^T \alpha^{T-t} = -\eta \frac{\alpha^{T+1}-1}{\alpha-1} \frac{\partial H}{\partial w_{mn}^{(0)}} \,.
\end{equation}
In this situation, \index{convergence!accelerated}convergence is accelerated when $\alpha$ is close to unity.
We also see that it is necessary that $\alpha < 1$ for the sum 
in Equation (\ref{eq:alphasum}) to converge.

The other limit to consider is that the gradient changes rapidly from iteration to iteration. How is the learning rule  modified in this case? As an example, let us assume that the gradient remains of
the same magnitude, but that its sign oscillates,
${\partial H}/{\partial w_{mn}^{(t)}}=(-1)^t{\partial H}/{\partial w_{mn}^{(0)}}$ for $t=1,\ldots,T$. 
Inserting this into Equation (\ref{eq:momentum_3}),  we obtain:
\begin{equation}
\delta\! w_{mn}^{(T)} \approx -\eta \frac{\partial H}{\partial w_{mn}^{(0)}} 
\sum_{t=0}^T (-1)^t\alpha^{T-t} = -\eta \frac{\alpha^{T+1}+(-1)^{T}}{\alpha+1} \frac{\partial H}{\partial w_{mn}^{(0)}} \,.
\end{equation}
Here the increments are much smaller compared with those
in Equation (\ref{eq:alphasum}). This shows that introducing inertia can substantially accelerate \index{convergence!accelerated}convergence without
sacrificing accuracy. The disadvantage is, of course, that there is yet another parameter to choose, namely the \index{momentum!constant}momentum constant $\alpha$.
\begin{figure}[t]
\centering
\begin{overpic}[scale=\myFigureScale]{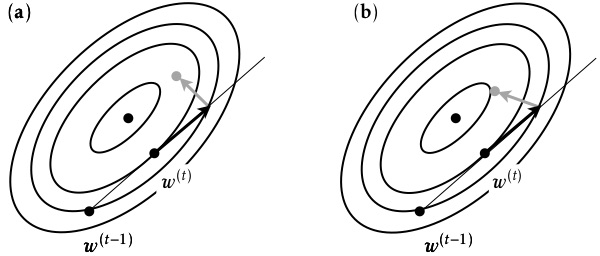}
\end{overpic}
\caption{\figlab{C6S3Nesterov} ({\bf a}) Momentum method (\ref{eq:inertia}). 
The gray arrow represents the increment $-\eta (\partial\! H/\partial \!w_{mn})|_{\{w_{ij}^{(t)}\}}$.
({\bf b}) Nesterov's accelerated gradient
method (\ref{eq:nesterov}).  
The gray arrow
represents $-\eta ({\partial \!H}/{\partial \!w_{mn}})|_{ \{w_{ij}^{(t)}+\alpha_{t-1}\delta\! w_{ij}^{(t-1)}\}}$. 
The location of $\ve w^{(t+1)}$ (gray point) is closer to the minimum (black point) than in panel ({\bf a}).  }
\end{figure}

Nesterov's {\em accelerated gradient}\index{gradient!accelerated, Nesterov} method \cite{Nesterov1983}
is another way of implementing \index{momentum}momentum. The algorithm was developed for smooth optimisation problems, 
but it has been suggested to use the method when \index{training!deep networks}training deep neural networks with gradient descent \cite{SutskeverPhD}:
\begin{align}
\label{eq:nesterov}
\delta\! w_{mn}^{(t)} &= 
-\eta \frac{\partial H}{\partial w_{mn}}\Big|_{ \{w_{ij}^{(t)}+\alpha_{t-1}\delta\! w_{ij}^{(t-1)}\}}
+ \alpha_{t-1} \delta\! w^{(t-1)}_{mn}\,.
\end{align}
A suitable sequence of coefficients $\alpha_t$ 
is defined by recursion \cite{SutskeverPhD}. The coefficients $\alpha_t$ approach unity from below as $t$ increases. 

Nesterov's accelerated-gradient method is more accurate  than the simple \index{momentum}momentum method, because 
the accelerated-gradient method evaluates the gradient at an extrapolated point, not 
at the initial point. Figure \figref{C6S3Nesterov} illustrates a situation where Nesterov's method
\index{convergence!accelerated}converges more rapidly.
Nesterov's method is not much
more difficult to implement than Equation (\ref{eq:inertia}), and it is not much more expensive in terms of computational cost.

There are other ways of adapting the learning rate\index{learning rate} during training, described in Section 4.10 in Haykin's book \cite{Haykin}. Finally, the learning rate\index{learning rate} need not be the same for all neurons. If the weights of neurons in different layers change 
at very different speeds (Section \ref{sec:vg}), one could define a layer-dependent learning\index{learning rate} rate $\eta_\ell$ that is larger for neurons with smaller gradients.

\section{Summary}
Backpropagation is an efficient algorithm for stochastic gradient-descent on the energy function (\ref{eq:H3}) in weight space, because it refers only to quantities that are local to the weight to be updated. Networks with many \index{hidden neuron} hidden neurons have many free parameters (their weights and thresholds). This increases the risk of overfitting\index{overfitting}, which
reduces the power of the network to generalise. 
Deep networks with many hidden layers are particularly prone to overfitting (Chapter \ref{chap:dl}).
The tendency of networks to overfit can be reduced  by \index{cross validation}cross validation.

\section{Further reading}
The backpropagation algorithm is explained in Section 6.1 of Hertz, Krogh and Palmer \cite{hertz1991introduction}, and in Chapter 4 of Haykin's book \cite{Haykin}. The paper \cite{LeCun1998}
by LeCun {\em et al.} predates deep learning\index{deep learning}, but it is still a very nice collection of recipes for making
backpropagation more efficient. 

One of the first papers on error backpropagation is the one by Rumelhart {\em et al.} \cite{Rumelhart} from 1986. The authors provide an elegant explanation and summary of the backpropagation algorithm.
They also describe results of different numerical experiments, and one of them
introduces convolutional networks (Chapter \ref{chapter:con_net}) to learn to tell the difference
between the letters T and C (Figure \figref{C6S5TC}).
\begin{figure}[tb]
\centering
\begin{overpic}[scale=\myFigureScale]{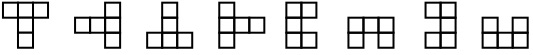}
\end{overpic}
\caption{\figlab{C6S5TC} Patterns detected by the convolutional network of Ref.~\cite{Rumelhart}. After Fig. 13 in Ref.~\cite{Rumelhart}.}
\end{figure}
\vfill\eject

\chapter{Deep learning}      
\index{deep learning|(}
\label{chap:dl}
\begin{figure}
\centering
\begin{overpic}[height=3cm]{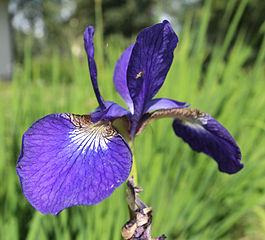}
\end{overpic}
\begin{overpic}[height=3cm]{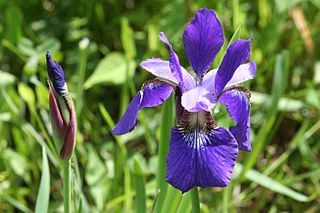}
\end{overpic}
\begin{overpic}[height=3cm]{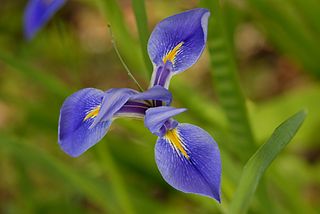}
\end{overpic}
\caption{\figlab{C7S1Iris} Images of iris flowers. From left
to right: iris setosa (copyright T. Monto), iris versicolor (copyright R. A. Nonenmacher), 
and iris virginica (copyright A. Westermoreland). All images are copyrighted under
the creative commons license.}
\end{figure}

\section{How many hidden layers?}
\index{hidden layer|(}
\label{sec:hmhl}
\begin{figure}[bt]
  \centering
    \begin{overpic}[scale=\myFigureScale]{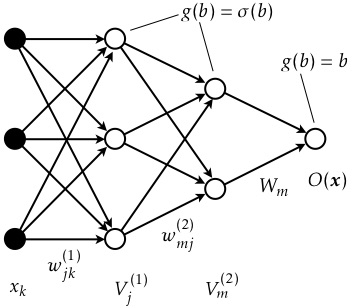}
    \end{overpic}
   \centering \caption{\figlab{C7S1ExampleNetwork} Multi-layer perceptron for function approximation.}
\end{figure}
In Chapter \ref{ch:C5Chapter} 
we saw why it is sometimes necessary to have a hidden layer: in order to solve problems that are not linearly separable. Under which circumstances is one hidden layer sufficient? Are there problems that require more than one hidden layer? Even if not necessary, may additional hidden layers improve the performance of the network? 

To understand how many hidden layers suffice,  it is useful to view the 
\index{hidden layer}
classification problem\index{classification!task} as an {\em approximation problem}\index{learning!approximation problem} \cite{Hornik}. Consider
the classification problem $[\ve x^{(\mu)},t^{(\mu)}]$ for $\mu=1,\ldots,p$.
This problem defines a {\em target function}\index{target!function} $t(\ve x)$.
Training a network to solve this task corresponds to approximating
the target function $t(\ve x)$ by the output function 
$O(\ve x)$ of the network, from $N$-dimensional input space to one-dimensional 
outputs.
\index{function!approximation}

How many hidden layers are necessary or sufficient to approximate a given set of functions to a certain accuracy,
by choosing weights and thresholds?  The answer depends on the nature of the target function. Is it real-valued, perhaps continuous, or does it assume only discrete values? 

We start with real-valued inputs and a single output \cite{lapedes1987how,hertz1991introduction}.
Consider the network drawn in Figure \figref{C7S1ExampleNetwork}. 
The neurons in the hidden layers have sigmoid activation functions\index{activation function!sigmoid} 
$\sigma(b) = (1+{\rm e}^{-b})^{-1}$.
The output is a linear unit\index{unit!linear}, with linear activation function $g(b)=b$. 
\index{activation function!linear}
With two hidden layers one tries to approximate the function $t(\ve x)$  by
\begin{equation}
\label{eq:approximation_network}
    O(\ve x) = \sum_{m} W_{m} g\Bigg(\sum_{j} w_{mj}^{(2)} g\Big(\sum_{k} w_{jk}^{(1)} x_{k} - \theta_{j}^{(1)}\Big)  
- \theta_{m}^{(2)}\Bigg) - \Theta\,.
\end{equation}
\begin{figure}
\centering
\begin{overpic}[scale=\myFigureScale]{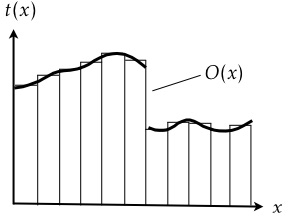}
\end{overpic}
\caption{\figlab{C7S1Approximation} The neural-network output $O(x)$ approximates
the \index{target!function} target function $t(x)$.}
\end{figure}
\begin{figure}[b]
\vspace*{4mm}
  \centering
    \begin{overpic}[scale=\myFigureScale]{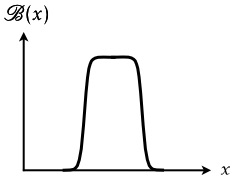}
    \end{overpic}
\caption{\figlab{C7S1BasisFunction} Basis function 
used to approximate a one-dimensional target function $t(x)$.}
\end{figure}
In the simplest case the inputs are one-dimensional  (Figure \figref{C7S1Approximation}).  
 The \index{training set}training set consists of pairs $[x^{(\mu)},t^{(\mu)}]$ that encode
the \index{target!function} target function $t({x})$. 
The task is then to approximate $t(x)$ by the network output $O(x)$: 
\begin{equation}
O(x)\approx t(x)\,.
\end{equation}
\index{basis!function|(}
To this end one uses linear combinations of the basis functions $\basis(x)$ shown in Figure \figref{C7S1BasisFunction}.
Any reasonable real-valued function $t(x)$ can be approximated by  sums of such basis functions, each suitably shifted and scaled.
Furthermore, these basis functions can be expressed as scaled differences of sigmoid activation functions \index{activation function!sigmoid}
\begin{equation}
\basis(x)= W [\sigma(w^{(1)}x-\theta_1^{(1)})-\sigma(w^{(1)}x-\theta_2^{(1)})]\,.
\end{equation}
Comparing  with Equation (\ref{eq:approximation_network}) shows that one hidden layer is sufficient
to construct the function $O(x)$ in this way.
\begin{figure}[t]
  \centering
    \begin{overpic}[scale=\myFigureScale]{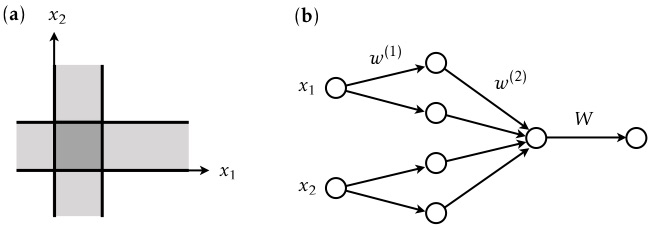}
    \end{overpic}
\caption{\label{fig:C7S1BasisFunction2D}  Two-dimensional
basis functions. ({\bf a}) To make a localised basis function with two
inputs,  one needs two hidden layers of neurons with sigmoid activation functions. 
\index{activation function!sigmoid}
One layer determines the lightly shaded cross
in terms of a linear combination of four sigmoid outputs. The second layer localises the final output
to the darker square [Equation (\ref{eq:basis2D})]. ({\bf b}) Network layout for one basis function,
after Figure~8 in Ref.~\cite{lapedes1987how}.}
\end{figure}

Now consider two-dimensional inputs. In this case, a suitable basis function  is
(Figure \ref{fig:C7S1BasisFunction2D}):
\begin{equation}
\label{eq:basis2D}
\mathscr{B}(\ve x)\!=\!  W \sigma\big\{w^{(2)} [\sigma(w^{(1)}x_{1}) - \sigma(w^{(1)}x_{1}\!-\!\theta_{2}^{(1)}) + \sigma(w^{(1)}x_{2}) -\sigma(w^{(1)}x_{2} -\theta_{4}^{(1)})]\! -\!\theta^{(2)}\big\}\,.
\end{equation}
So two hidden layers are sufficient for two input dimensions.
For each basis function we require four neurons in the first hidden layer and one neuron in the second hidden layer. 
The construction is analogous for more than two input dimensions. For each basis
function 
we need $2N$ 
neurons in the first and one neuron in second layer.
In conclusion,  two hidden layers are sufficient to approximate a real-valued target function.
\index{basis!function|)}
\index{hidden layer}

Yet it is not always necessary to have two layers for real-valued functions. For continuous functions, one
hidden layer is sufficient. This is ensured by the {\em universal approximation theorem}\index{universal approximation theorem} \cite{Haykin}.  
\index{learning!approximation problem}  It says any continuous function\index{function!continuous}  can be approximated to arbitrary accuracy by a network with
a single hidden layer, for sufficiently many neurons in the hidden layer. 

In Chapter \ref{ch:C5Chapter} we considered discrete Boolean functions. 
\index{Boolean function}
Any Boolean function with $N$-dimensional
inputs can be represented by a network with one hidden layer, using $2^N$ neurons in the hidden layer. An example for such a network  is discussed in Ref.~\cite{hertz1991introduction}: 
\begin{align}
    &x_{k} \in \{+1,-1\}  & k = 1,\dots,N \quad\mbox{inputs}\nonumber\\
    &V_{j}   & j = 0,\dots,2^N-1 \quad\mbox{hidden neurons}\nonumber\\
    &g(b) = {\rm sgn}(b) & \text{activation function of hidden neurons} \\
    &g(b) = \text{sgn}(b) & \text{activation function of output neuron\index{output neuron}} \nonumber
\end{align}
\index{activation function!tanh}
\index{hidden neuron|(}
A difference compared with the Boolean-function representations in Section \ref{sec:mlp} is that here the inputs take the values $\pm 1$. The reason 
is that this simplifies the proof, which is by construction \cite{hertz1991introduction}. For each hidden neuron one assigns the weights as follows
\begin{equation}
\label{eq:consprinc}
    w_{jk} = \begin{cases}
            \delta &\text{if the }  k^{\text{th}} \text{digit of binary representation of $j$ is $1$}\,, \\
            - \delta &\text{otherwise,}
            \end{cases}
\end{equation}
with $\delta > 1$ (see below). The \index{threshold}thresholds $\theta_j$ of all hidden neurons are the same, equal to $N(\delta-1)$. The idea
is that each input pattern\index{input pattern} turns on exactly one neuron in the hidden layer, the {\em winning neuron})\index{winning neuron}.
The weights feeding into the output neuron\index{output neuron} are determined as follows. If the output for the pattern represented by neuron $V_j$ is $+1$, let $W_{1j}=+1$, otherwise $W_{1j}=-1$. The \index{threshold}threshold is set to $\Theta=\sum_j W_{1j}$.

To show how this construction works, consider the Boolean XOR function as an example (Figure \figref{C7S3XOR}).\index{XOR function}
\index{Boolean function}
\begin{figure}[bt!]
\centering
        \begin{tabular}{lccc}
        &&& \\\hline\hline
        \small $x_{1}$  & \small $x_{2}$ & \small $t $\\
        \hline
        \small-1 &  \small -1  & \small -1  \\
        \small +1 &  \small -1  & \small +1 \\
        \small -1 &  \small +1  & \small +1 \\
        \small +1 &  \small+1  & \small -1 \\
        \hline
        \hline
        \end{tabular}
\hspace*{.5cm}\raisebox{-2.0cm}{
\begin{overpic}[scale=\myFigureScale]{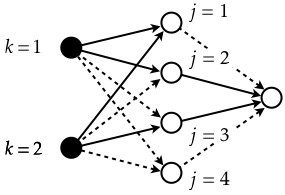}
            \put(-71,67){({\bf a})}
            \put(0,67){({\bf b})}
            \put(112,67){({\bf c})}
        \end{overpic}}\hspace*{0.5cm}
        \begin{tabular}{lccc}
        &&& \\\hline\hline
        \small $j$  &  \small  digit $1$ &  \small  digit $2$ \\
        \hline
         \small  0 &  \small   0  & \small   0  \\
         \small  1 &   \small  0  &  \small  1 \\
         \small 2 &   \small  1  &  \small 0 \\
         \small  3 &   \small  1  &  \small  1 \\
        \hline
        \hline
        \end{tabular}
     \caption{\figlab{C7S3XOR} \index{XOR function} Boolean XOR function. ({\bf a}) Value table, ({\bf b}) Network layout.  
For the weights feeding into the hidden layer, dashed lines correspond to $w_{jk}=-\delta$ , solid lines to $w_{jk}=\delta$.  
For the weights feeding into the output neuron\index{output neuron}, dashed lines correspond to $W_{1j}=-\gamma$, and solid lines to $W_{1j}=\gamma$. Panel ({\bf c}) summarises 
the binary representation of $j$ used to determine the weights $w_{jk}$ of the hidden layer [Equation (\ref{eq:consprinc})]. }
\end{figure}
To confirm that only the corresponding winning neuron gives a positive signal, consider pattern $\ve x^{(1)} = [ -1, -1]^{\sf\small T}$. It activates the first neuron in
the hidden layer ($j=0$). To see this, compute the local fields of the
hidden neurons:
\begin{eqnarray} 
    b_0^{(1)}&=& 2\delta -2(\delta -1) =2   \,, \\
    b_1^{(1)}&=&   -2(\delta -1) =2-2\delta  \nonumber\,, \\
    b_2^{(1)}&=&  -2(\delta -1) =2-2\delta  \nonumber\,,\\
    b_3^{(1)}&=&  -2\delta-2(\delta -1)=2-4\delta\,.  \nonumber
\end{eqnarray}
If we choose $\delta  > 1$ then the output of the first hidden neuron gives a positive output ($V_0>0$), the other neurons produce
negative outputs, $V_j<0$ for $j=1,2,3$. In conclusion, output neuron\index{output neuron} $1$ is the winning neuron for this pattern.
Now consider
$\ve x^{(3)} = [ -1,+1]^{\sf\small T}$. In this case
\begin{eqnarray}
  b_0^{(3)}&=& -2(\delta -1) =2-2\delta\,, \\
  b_1^{(3)}&=&  -2 \delta -2(\delta -1)=2-4\delta\,, \nonumber \\
  b_2^{(3)}&=&  2 \delta -2(\delta -1) =2\,,  \nonumber\\
  b_3^{(3)}&=&  -2(\delta -1)=2-2\delta\,.  \nonumber
\end{eqnarray}
Now the third hidden neuron gives a positive output, while the others yield negative values.
It works in the same way for the other two patterns, $\ve x^{(2)}$ and $\ve x^{(4)}$.
In summary, there is a unique winning neuron for each pattern.\footnote{That pattern $\mu=k$ gives the winning
neuron $j=k-1$ is of no importance, it is just a consequence of how the patterns are ordered in the value table in 
Figure~\figref{C7S3XOR}} 
Figure \ref{fig:C7S1InputPlaneXOR} shows how the four 
decision boundaries corresponding to $V_j$ \index{decision boundary}
partition the input plane.
\begin{figure}[t]
\centering
\begin{overpic}[scale=\myFigureScale]{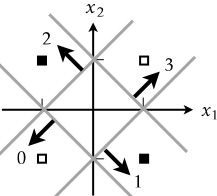}
\end{overpic}
\caption{\figlab{C7S1InputPlaneXOR} Shows how the XOR network depicted in Figure \ref{fig:C7S3XOR} partitions the input plane. Target values are encoded as in Figure \figref{C5S1XORFunction}: $\Box$ corresponds to $t=-1$ and $\blacksquare$ to $t=+1$).}
\end{figure}
\begin{figure}[tbp]
\centering
  \begin{overpic}[scale=\myFigureScale]{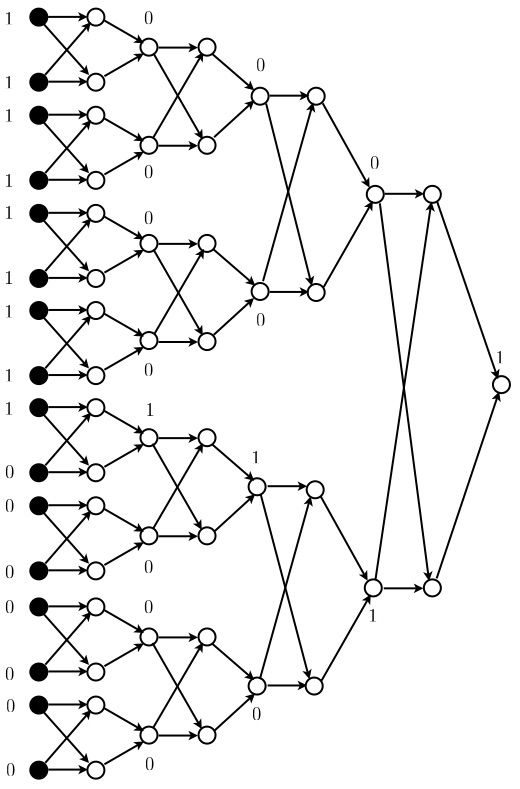}
  \end{overpic}
\caption{\figlab{C7S1Parity} Solution of the parity problem
for $N$-dimensional inputs. The network is built from
XOR units\index{XOR function} (Figure \figref{C5S4XORWithHiddenGraph}, here with 0/1 neurons). Each XOR unit has a hidden layer with two neurons. Only the states of the inputs and outputs of the XOR units are shown, not those of the hidden neurons. In total, the whole network has only $O(N)$ neurons. After Figure~2 in Ref.~\cite{Franco}.}
\end{figure}

According to the scheme outlined above, the output neuron\index{output neuron} computes
\begin{equation}
O_1 = \mbox{sgn}(-V_1 + V_2 +  V_3 - V_4) 
\end{equation}
with $\Theta=\sum_j W_{1j}=0$.  For $\ve x^{(1)}$ and  $\ve x^{(4)}$ we find the correct result $O_1=-1$.
The same is true for $\ve x^{(2)}$ and  $\ve x^{(3)}$, in this case we obtain $O_1=1$.
In summary, this example illustrates how an $N$-dimensional Boolean function is represented by a network with one hidden layer, with $2^N$ neurons.  The problem is of course that this network is expensive\index{Boolean function}
to train for large values of $N$ because the number of hidden neurons is very large.
\index{hidden neuron|)}

There are more efficient layouts if one uses more than one hidden layer. As an example, consider the {\em parity function}\index{parity function}
for $N$-dimensional binary inputs with bits equal to 0 or 1. It measures the parity of sequences of input bits. The function  evaluates to unity if there is an odd number
of ones in the input, otherwise to zero. A construction similar to the above yields a network layout with $2^N$ neurons in the hidden layer. 
If one instead wires together the XOR networks\index{XOR function}, 
one can solve the parity problem with $O(N)$ neurons
\cite{Franco} (Figure \figref{C7S1Parity}). When $N$ is a power of two, this network has $3(N-1)$ neurons. 
To see this, set the number of inputs to $N=2^k$. Figure
 \figref{C7S1Parity} shows that the number
$\mathscr{N}_k$ of neurons satisfies
the recursion $\mathscr{N}_{k+1} = 2\mathscr{N}_k+3$ with 
$\mathscr{N}_{1}=3$. The solution of this recursion is 
$\mathscr{N}_{k}=3(2^k-1)$.

This example also illustrates a second reason why it may be useful to have more than one hidden layer. To design a neural network for a certain task it is 
often convenient to build the network from well-studied building blocks. One wires them together, often in a hierarchical fashion.  In Figure \figref{C7S1Parity} there is only one building block, the XOR network from Figure \figref{C5S4XORWithHiddenGraph}.  Other examples are convolutional networks for image analysis. \index{convolutional network}
Here the fundamental building blocks are so-called feature maps\index{feature map}, they
recognise different geometrical features in the image, such as edges or corners (Chapter\ref{chapter:con_net}). \index{hidden layer|)}

\section{Vanishing and exploding  gradients}
\index{vanishing gradient|(}
\index{exploding gradient|(}
\label{sec:vg}
This Section describes an inherent instability in the 
\index{training!instability}training of \index{instability}
deep networks\index{network}\index{deep network} with stochastic gradient descent,
the vanishing- or exploding-gradient problem.

In Chapter \ref{ch:C6Chapter} we saw that learning slows down when the factors $g'(b)$ 
in the recursion (\ref{eq:delta_rec_2})
become small.  When the network has several hidden layers, like the\index{hidden layer} 
one shown in Figure \ref{fig:C7S2DeepNet}, potentially small factors of $g'(b)$ are multiplied, aggravating the problem. As a consequence, the 
weights of hidden neurons close to the input layer 
change only by small amounts, 
the smaller the more hidden layers the network has. 
This is the {\em vanishing-gradient problem}\index{vanishing gradient|textbf}.
\begin{figure}[t]
\centering
\begin{overpic}[scale=\myFigureScale]{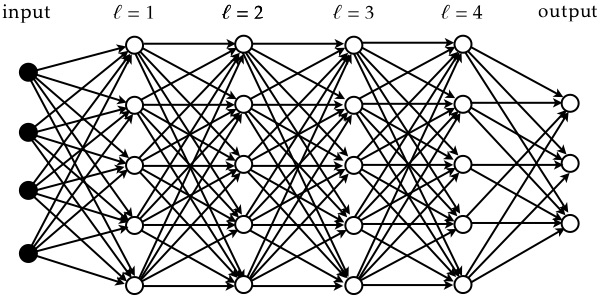}
\end{overpic}
\caption{\figlab{C7S2DeepNet} Fully connected deep network\index{network}\index{deep network} with four hidden
layers.}
\end{figure}

\begin{figure}[tb]
\centering
\begin{overpic}[scale=\myFigureScale]{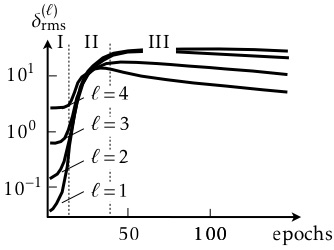}
\end{overpic}
\caption{\figlab{C7S2VG} Vanishing-gradient problem for a network with four 
fully connected hidden layers.\index{layer!fully connected}
The Figure illustrates schematically how the r.m.s. error
$\delta_{\rm rms}^{(\ell)}$ in layer $\ell$
depends on the number of \index{training!epoch}training epochs.\index{epoch}
During phase I, the vanishing-gradient problem is severe, during phase II the network starts to learn, phase III is the \index{convergence!phase}convergence phase where the errors decline. Schematic, based on simulations
performed by Ludvig Storm, training a network with four hidden layers and $N=30$ neurons per layer
on the \href{http://yann.lecun.com/exdb/mnist/}{MNIST} \index{MNIST data set}data set.}
\end{figure}

Figure \figref{C7S2VG} quantifies the problem. The Figure shows that the r.m.s. errors averaged over different realisations of
random initial weights, $\delta_{\rm rms}^{(\ell)}\equiv\big(\langle N^{-1} \sum_{j=1}^{N} [\delta^{(\ell)}_j]^2\rangle\big)^{1/2}$, 
tend to be very small during initial training. 
To explain this phenomenon, consider the simple example 
discussed in Ref.~\cite{Nielsen}: a long chain of neurons with only one neuron
per layer (Figure \figref{C7S2DeepNet1}).
The output $V^{(L)}$ is given by nested activation functions
\begin{equation}
\label{eq:Odef} V^{(L)} \!=\! g\Big(w^{(L)} g\Big(w^{(L-1)} \cdots g\big(w^{(2)} g(w^{(1)} x-\theta^{(1)})
-\theta^{(2)}\big)\ldots \!-\!\theta^{(L-1)}\Big)-\theta^{(L)}\Big)\,.
\end{equation}
Let us compute the errors $\delta^{(\ell)}$ \index{error} using Equation (\ref{eq:delta_gradients}). The partial derivative 
in (\ref{eq:delta_gradients}) is
evaluated using the chain rule:
\begin{eqnarray}
\frac{\partial V^{(L)}}{\partial V^{(L-1)}} &=& g'(b^{(L)})w^{(L)}\nonumber\,,\\
\frac{\partial V^{(L)}}{\partial V^{(L-2)}} &=&
\frac{\partial V^{(L)}}{\partial V^{(L-1)}} \frac{\partial V^{(L-1)}}{\partial V^{(L-2)}}
=  g'(b^{(L)})  w^{(L)}g'(b^{(L-1)})w^{(L-1)}\nonumber\,,\\
&\vdots&
\end{eqnarray}
where $b^{(k)} = w^{(k)} V^{(k-1)} -\theta^{(k)}$ is the \index{local field}local field for neuron $k$.
This yields the following expression for ${\partial V^{(L)}}/{\partial V^{(\ell)}} $:
\begin{equation}
\label{eq:prod}
\frac{\partial V^{(L)}}{\partial V^{(\ell)}}=
\prod_{k=L}^{\ell+1} [g'(b^{(k)})w^{(k)}]\,.
\end{equation}
Inserting this expression into Equation (\ref{eq:delta_gradients}), we find:
\begin{equation}
\label{eq:delta_prod}
\delta^{(\ell)} = [t-V^{(L)}(x)]g'(b^{(L)}) \prod_{k=L}^{\ell+1} [w^{(k)} g'(b^{(k-1)})]\,.
\end{equation}
One can also obtain this result by applying the recursion
from Algorithm \ref{bp_algorithm}, $\delta^{(\ell)} = \delta^{(\ell+1)} w^{(\ell+1)} g'(b^{(\ell)})$.
\begin{figure}[t]
\centering
\begin{overpic}[scale=\myFigureScale]{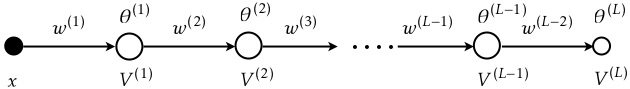}
\end{overpic}
\caption{\figlab{C7S2DeepNet1} Chain of neurons used
to illustrate the vanishing-gradient problem \index{vanishing gradient} \cite{Nielsen},
with neurons $V^{(\ell)}$, weights $w^{(\ell)}$, and \index{threshold}thresholds $\theta^{(\ell)}$.}
\end{figure}

Now consider the early stages of \index{training!early stages of}training 
\cite{Nielsen}.
For the activation functions (\ref{eq:activation_functions}), the maximum of $g'(b)$ is $\tfrac{1}{{4}}$ and $1$, respectively,
and $g'(b)$ becomes exponentially small  if $|b|$ is large. 
If one initialises the weights as described in Section
\secref{C6S2BPAlgo},  to Gaussian random variables
with mean zero and  variance $\sigma^2_w=1$ say,
then the factors $w^{(k)}g'(b^{(k-1)})$ tend to be smaller than unity. 
In this case, Equation (\ref{eq:prod}) implies
that the error or gradient $\delta^{(\ell)}$
\index{error} vanishes quickly as $\ell$ decreases. The reason is simply that the number of small factors in the product (\ref{eq:prod}) increases  when $\ell$ becomes smaller, and
multiplying many small numbers gives a very small product. As a result, 
the \index{training!slow down}training slows down.
As mentioned above, this is the {\em vanishing-gradient problem}\index{vanishing gradient|textbf}
(phase I in Figure \figref{C7S2VG}).

What happens at later times?  
Figure \figref{C7S2VG}
indicates that the network continues to learn slowly. 
For the particular example shown in Figure \figref{C7S2VG}, the effect persists
for about 20 epochs. Then the first layers begin to learn faster (phase II). 
There is to date no mathematical theory describing how this transition occurs. Much later in \index{training}training, the errors decay as the learning converges (phase III in Figure \figref{C7S2VG}).

There is a second, equivalent, point of view \cite{Nielsen}: the learning is slow
in a layer far from the output because the output is not very sensitive to the state of these neurons. 
The effect of a given neuron on the output is measured by Equation (\ref{eq:prod}), which describes
how the output of the network changes when changing the {\em state} of a neuron in a particular layer.
At any rate, Equation (\ref{eq:delta_prod}) demonstrates that
hidden layers far from the output learn slowly, 
at least initially when the weights are still random.

Suppose we try to combat the vanishing-gradient problem by increasing the weight variance $\sigma_w^2$. 
The problem is that this may cause the
factors $w^{(k)}g'(b^{(k-1)})$ to become larger than unity.
As a consequence,
the gradients increase
exponentially instead ({\em exploding gradients}\index{exploding gradient}).
In conclusion, the \index{training!instability}training dynamics is fundamentally unstable. 
This is due to the multiplicative nature of the recursion for the errors.\index{error}
Taking the logarithm of the product in Equation (\ref{eq:prod}) and assuming
that the weights are independently distributed random numbers, the {\em central-limit theorem}\index{central limit theorem} (Chapter \ref{ch:dhn}) implies 
that the distribution of $\log \delta^{(\ell)}$ is Gaussian. In other words,  the distribution of the errors is lognormal\index{distribution!log normal},\index{error!distribution} implying that
very small and very large values of $\delta^{(\ell)}$ occur with high probability.

In networks like the one
shown in Figure \figref{C7S2DeepNet} the principle is the same. 
Assume that all layers $\ell=1,\ldots,L$ have the same number $N$ of neurons.
When $N>1$, one multiplies $N\times N$ matrices, instead of numbers.
The product (\ref{eq:prod}) of random numbers becomes a product of 
random matrices. \index{matrix!random}
Using the chain rule\index{chain rule} we find:
\begin{eqnarray}
\label{eq:sum_prod}
\frac{\partial V^{(L)}_i}{\partial V^{(\ell)}_j} &=& \sum_{m=1}^N\sum_{n=1}^N\cdots\sum_{p=1}^N
\frac{\partial V^{(L)}_i}{\partial V^{(L-1)}_m} \frac{\partial V^{(L-1)}_m}{\partial V^{(L-2)}_n} \cdots \frac{\partial V^{(\ell+1)}_p}{\partial V^{(\ell)}_j}\,.
\end{eqnarray}
With the update rule
\begin{equation}
\label{eq:update_rule}
V^{({k})}_m = g\Big(\sum_{j=1}^N w_{ij}^{({k})} V_j^{({k}-1)}-\theta_i^{({k})}\Big)
\end{equation}
we can evaluate each factor:
\begin{equation}
\frac{\partial V^{({k})}_m}{\partial V^{({k}-1)}_n} = g'(b^{({k})}_m)w^{({k})}_{mn}\,.
\end{equation}
Substituting this result into Equation (\ref{eq:sum_prod}), we see that the partial derivatives $\partial V^{(L)}_i/\partial V^{(\ell)}_j$
can be computed in the form of a matrix product. The matrix $\ma J'_{L-\ell}$ with elements  $[\ma J'_{L-\ell}]_{ij}=\partial V^{(L)}_i/\partial V^{(\ell)}_j$ is given by:
\begin{equation}
\label{eq:maJ}
\ma J_{L-\ell}^{\prime}= 
\ma D^{(L)} \ma W^{(L)}\ma D^{(L-1)} \ma W^{(L-1)}\cdots \ma D^{(\ell+1)}\ma W^{(\ell+1)}\,.
\end{equation}
Here $\ma W^{({k})}$ is the matrix of weights feeding into layer ${k}$, and
\begin{equation}
\ma D^{({k})} = \begin{bmatrix}
g'(b_1^{({k})})&&\\
&\ddots&\\
&&g'(b_N^{({k})}) 
\end{bmatrix}
\end{equation}
is the diagonal matrix with entries $D_{jj}^{(k)} = g'(b_j^{(k)})$.
The matrix product  (\ref{eq:maJ}) determines the error dynamics, just 
\index{error!dynamics}
like Equation (\ref{eq:delta_prod}):
\begin{align} 
 \ve \delta^{(\ell)\sf T} 
= \ve \delta^{(L)\sf T}\ma J_{L-\ell}\quad\mbox{with}\quad
\ma J_{L-\ell} = [\ma D^{(L)}]^{-1}\ma J^\prime _{L-\ell}\ma D^{(\ell)}\,.
\end{align}
Does the magnitude of  $\big|\ve \delta^{(\ell)}\big|^2={\ve \delta^{(\ell)}}^{\sf T}\ve \delta^{(\ell)}$ of the errors shrink or grow as they propagate through the layers?
This is determined by the  \index{eigenvalue}eigenvalues of the {\em left Cauchy-Green} matrix
\index{Cauchy Green matrix|textbf}
$\ma J_{{p}}\ma J_{{p}}^{\sf \small T}$, with ${p} =L-\ell$.
This matrix is symmetric,  and its \index{eigenvalue!non negative}eigenvalues are non-negative. Their square roots are the {\em singular values} of $\ma J_{{p}}$:
\begin{align}
\label{eq:singular_values}
\Lambda^{({p})}_1 \geq \Lambda^{({p})}_2 \geq \cdots \geq \Lambda^{({p})}_N \geq 0\,.
\end{align}
It is customary to sort the singular values by their magnitudes, as in Equation (\ref{eq:singular_values}).
When there are many layers,
the number ${p}=L-\ell$ of factors in Equation (\ref{eq:maJ}) is  large.
In this case, the maximal singular value either decreases or increases exponentially
as a function of ${p}$ \cite{Crisanti}. The corresponding rate 
\begin{equation}
\lambda_1 = \lim_{{p}\to\infty}\frac{1}{{p}} \log \Lambda_1^{({p})}
\end{equation}
is called the maximal {\em Lyapunov exponent}\index{Lyapunov!exponent|textbf}. 
A negative maximal Lyapunov exponent indicates that the errors vanish exponentially.
The  \index{eigenvector}eigenvectors of $\ma J_{{p}} \ma J_{{p}}^{\sf \small T}$
are called {\em backward Lyapunov vectors}\index{Lyapunov!vector}. 
They describe how the errors change as they propagate through the network.  
\index{error}
How small differences between the inputs  change, is determined by the {\em forward Lyapunov vectors}\index{Lyapunov!vector}, the  \index{eigenvector}eigenvectors of $\ma J_{{p}}^{\sf \small T}\ma J_{{p}}$. Since $\ma J_{{p}}^{\sf \small T} \ma J_{{p}}$ and $\ma J_{{p}} \ma J_{{p}}^{\sf \small T}$ have the 
same \index{eigenvalue}eigenvalues,
the rate of decay or increase of 
the magnitude of input differences is the same as that of the error magnitudes. 

The concept of a maximal Lyapunov exponent is borrowed from chaos theory\index{chaos theory} \cite{ChaosBook,Ruelle,strogatz2000chaos}, where 
$\lambda_1>0$ implies that small perturbations of the initial conditions grow exponentially 
as a function of time.
The iterated map (\ref{eq:Odef}) 
is a dynamical system. 

The transition in Figure \figref{C7S2VG} is triggered
by a change of the Lyapunov exponent from negative values to $\lambda_1\approx 0$ \cite{Storm}.
In summary, the unstable-gradient problem in deep networks\index{network}\index{deep network} is due to the fact that the maximal singular value $\Lambda_1^{({p})}$
either increases or decreases exponentially as one moves away from the output layer, depending on whether the maximal Lyapunov exponent
 is negative or positive.

Pennington {\em et al.} \cite{Pennington2017} suggested to combat the unstable-gradient problem by initialising
weights and thresholds in such a way that
the maximal Lyapunov exponent is close to zero, in order to make sure that the errors neither grow nor shrink exponentially. 
Consider the network shown in Figure \figref{C7S2DeepNet}, with $N$ neurons
per hidden layer, and initialise\index{weight!initial}
the weights to independent Gaussian random numbers with mean zero and variance $\sigma^2_w$. The \index{threshold!initial}thresholds are initialised in the same way, with variance
$\sigma_\theta^2$. In the limit of $N\to\infty$ one can use a mean-field theory\index{mean field theory} \cite{Pennington2017}, just as in Chapter \ref{sec:shn}, to estimate the maximal Lyapunov exponent\index{Lyapunov!exponent}.

Following Ref.~\cite{Pennington2017}, the first step is to compute how the errors\index{error}  propagate through the network.
We assume uncorrelated \index{pattern!random} random input patterns\index{input pattern!random}\index{pattern!correlations}, Equation (\ref{eq:prob}), and random weights
with mean zero and variance  $\langle w_{ij} w_{kl}\rangle
=\sigma_w^2 \delta_{ij}\delta_{kl}$.
When $N\to\infty$, the errors are sums of many random numbers [Equation (\ref{eq:delta_rec_2})]. 
Invoking the central-limit theorem (Chapter \ref{ch:dhn}), one concludes that the errors
are approximately Gaussian distributed, with mean zero and with variance
\index{error!variance}
\begin{equation} 
\label{eq:ds10}
\langle [\delta_j^{(\ell-1)}]^2\rangle
=  \big\langle \sum_{i,k=1}^N \delta_i^{(\ell)} \delta_k^{(\ell)} \,
 w_{ij}^{(\ell)} w_{kj}^{(\ell)}\, [g'(b_j^{(\ell-1)})]^2\big\rangle
\approx \sigma_w^2 \sum_{i=1}^N \langle[\delta_i^{(\ell)}]^2\rangle \langle  [g'(b_j^{(\ell-1)})]^2\rangle\,.
\end{equation}
The last approximation neglects
possible correlations with the local fields. \index{local field}
The variance $\langle[\delta_i^{(\ell)}]^2\rangle$ does not depend on $i$,
so that the sum just gives a factor of $N$.

In the limit of large $N$, the central-limit theorem ensures that the local
fields $b_j^{(\ell)}$ are Gaussian distributed too, 
with mean zero
and variance 
\begin{align}
\label{eq:sl12}
\sigma_{\ell}^2 = \frac{1}{N} \sum_{j=1}^{N} [b_j^{(\ell)}]^2\,.
\end{align}
This allows us to estimate
\begin{equation}
\langle [{g'}(b_j^{(\ell)}]^2\rangle\sim  \int\!\! {\rm d}z_\ell  \,
\frac{{\rm e}^{-z^2/2\sigma_{\ell}^2}} {\sqrt{2\pi\sigma_{z_\ell}^2}}
[g'(z_\ell)]^2\equiv F(\sigma_{\ell})\,.
\end{equation}
Equation (\ref{eq:sl12}) describes how the distribution of \index{local field}local fields $b_j^{(\ell)}$ narrows or broadens as one iterates.  
For $g(b) = {\rm tanh}(b)$, it was shown in Ref.~\cite{Pennington2017}
that $\sigma_\ell$ approaches a fixed point, $\sigma^\ast=\lim_{\rm \ell\to \infty} \sigma_\ell$, under certain conditions
on the variances of the weights and thresholds, $\sigma_w^2$ and $\sigma_\theta^2$.
If $\sigma_\ell$ is well approximated by $\sigma^\ast$, 
Equation (\ref{eq:ds10}) simplifies to:
$\delta_{\rm rms}^{(\ell-1)} \approx  \delta_{\rm rms}^{(\ell)}
\sqrt{\sigma_w^2 N F(\sigma^\ast)}$.
This results in a mean-field estimate of the
the maximal Lyapunov exponent,
\begin{equation} 
\label{eq:l1_estimate}
\lambda_1 \sim\log\big|\delta^{(\ell-1)}/\delta^{(\ell)}\big|\approx\tfrac{1}{2}\log [\sigma_w^2 N F(\sigma^\ast)]\,.
\end{equation}
The network parameters should be adjusted so that this exponent is as close
to zero as possible. This means, in particular, that one should
take 
\begin{equation}
\sigma_w^2 \propto N^{-1}\,,
\end{equation}
see also Refs.~\cite{SutskeverMartensDahlHinton,GlorotBengio2010}.
But we must keep in mind that Equation (\ref{eq:l1_estimate}) relies on taking
the limit $N\to \infty$. It is expected that the assumptions
underlying Equation (\ref{eq:l1_estimate})  break down when $N$ is finite, causing
the mean-field theory to fail. The tails
of the error distribution, for example, are expected to become heavier as $N$ decreases, as indicated by the results for $N=1$ described above.
\index{error!distribution}
Note also that $\ma J_p$ assumes rank zero with a small but non-zero probability when $N$ is finite. In this case $\lambda_1=-\infty$.

There are a number of other tricks that help to cope with unstable gradients in practice, to some extent at least. First, it is sometimes 
argued that activation functions which do not saturate at large $b$, such as the ReLU function \index{ReLU function}, help against the vanishing-gradient problem.   Second, batch normalisation\index{batch normalisation} (Section \ref{sec:bn}) may reduce the unstable-gradient problem. Third, introducing connections that
skip layers ({\em residual networks}\index{residual network}) can alleviate the unstable-gradient problem.
This is discussed in Section  \ref{sec:resnets}.

Finally, there is an important aspect of the problem 
that we did not discuss: unstable gradients limit the extent to which information can propagate through the network in a meaningful way.
This is explained in Ref.~\cite{schoenholz2016deep}.

In this Section we assumed all along that the weights are random numbers. When the network starts to learn, this is no longer the case. The question is how the singular values of $\ma J_{p}$ change when correlations between different factors in the product (\ref{eq:maJ}) develop. 
\index{vanishing gradient|)}
\index{exploding gradient|)}

\section{Rectified linear units}
\index{rectified linear unit|(}
\label{sec:ReLU}
Glorot {\em et al.} \cite{GlorotBordesBengio2011} suggested to use a piecewise
activation function, the ReLU function\index{ReLU function|(}\footnote{Since the derivative of the ReLU function
is discontinuous at $b=0$, a common convention
is to set the derivative to zero at $b=0$.} 
 $\max\{0,b\}$ (Chapter \ref{chap:intro}).
What is the point of using ReLU neurons? When training
\index{training!deep networks}
a deep network\index{network}\index{deep network} with  ReLU activation functions, many of the 
\index{hidden neuron} hidden neurons produce output zero. This means that the network of active
neurons (non-zero output) is {\em sparsely} connected. It is sometimes argued 
that sparse networks have desirable properties;  at least sparse representations\index{representation!sparse} of a classification problem tend to be easier to learn because they are more likely to be linearly separable (Section \ref{sec:ct}). Figure \figref{C7S2Sparse} illustrates that for a given input pattern\index{input pattern}, only a certain fraction of hidden neurons \index{neuron!active} is active. For these
neurons the computation is linear, yet different input patterns give
different sets of \index{neuron!active} active neurons. The product in Equation (\ref{eq:maJ}) acquires a particularly simple structure:
the matrices $\ma D^{({p})}$ are diagonal with 0/1 entries. But while the weight matrices are independent initially, they become correlated as the training
proceeds. Also the $\ma D^{({p})}$-matrices develop correlations: 
which elements vanish depends on which pattern is clamped to the input terminals.

A hidden layer with only one or very few active neurons might 
act as a {\em bottleneck}\index{bottleneck} preventing efficient backpropagation of 
output errors\index{output error}  which could in principle slow down 
\index{training!slow down}training. For  the examples given in Ref.~\cite{GlorotBordesBengio2011},
this does not occur.
To describe information propagation through a network with ReLU neurons, one should compute the
probability that a given number of singular values of the matrix $\ma J_{{p}}$ vanish (Section~\ref{sec:vg}).
\begin{figure}[t]
\centering
\begin{overpic}[scale=\myFigureScale]{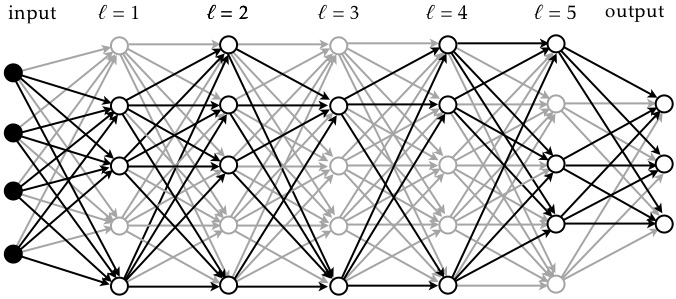}
\end{overpic}
\caption{\figlab{C7S2Sparse} Sparse network of active neurons with ReLU activation functions. The black paths correspond to {\em active}\index{neuron!active} neurons with positive local fields\index{local field}.}
\end{figure}

The ReLU function is unbounded for large positive local fields. \index{local field}
Therefore, the vanishing-gradient problem\index{vanishing gradient} (Section~\ref{sec:vg})
is thought to be less severe in networks made of rectified linear units. 
However, since the ReLU function does not saturate, the weights tend to increase. 
Glorot {\em et al.} \cite{GlorotBordesBengio2011} suggested
to use $L_1$-regularisation\index{regularisation!L1 regularisation} (Section \ref{sec:weightdecay}) to make sure
that the weights do not grow.\index{regularisation!L1 regularisation}

Finally, using ReLU functions \index{ReLU function} instead of sigmoid functions speeds up the \index{training!speed up}training,
\index{activation function!sigmoid}
because the ReLU function  has piecewise constant derivatives. Such functions are faster to evaluate than non-linear activation functions and their derivatives.
\index{rectified linear unit|)}
\index{ReLU function|)}

\section{Residual networks}
\label{sec:resnets}
One way of reducing the vanishing-gradient problem is to introduce 
short cuts, connections 
that skip layers \cite{He2015}.
Empirical evidence shows that networks with such short cuts are easier to train than standard multilayer perceptrons.
The likely reason is that the vanishing-gradient problem is less severe in networks with short cuts, because error\index{error} propagation in such networks is determined by the matrix product with the smallest number of factors.

This Section explains how to train networks with short cuts \cite{resnet}.
\begin{figure}
\centering
\begin{overpic}[scale=\myFigureScale]{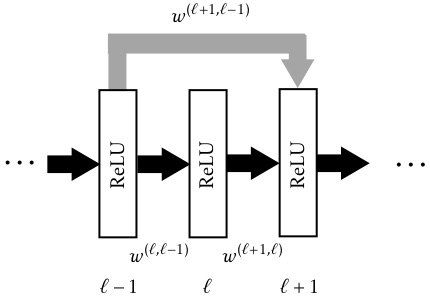}
\end{overpic}
\caption{\figlab{C7S6ResNet} Schematic illustration of a network with 
a short cut that skips one layer (gray arrow). After Fig.~1 from Ref.~\cite{resnet}.
}
\end{figure}
The layout is illustrated schematically 
in Figure \figref{C7S6ResNet}. Black arrows stand for 
usual feed-forward connections, and the gray arrow indicates
a connection that skips a layer.  The notation in  Figure \figref{C7S6ResNet}
differs somewhat from that of Algorithm
\ref{bp_algorithm}. The weights
from layer $\ell-1$ to $\ell$ are denoted
by $w_{jk}^{(\ell,\ell-1)}$, and those
from layer $\ell-1$ to $\ell+1$ by $w_{ij}^{(\ell+1,\ell-1)}$
(gray arrow in Figure \figref{C7S6ResNet}). Note that the superscripts are ordered in
the same way as the subscripts: the {\em right} index refers to the layer on the {\em left}.
According to Figure~\figref{C7S6ResNet}, neuron $j$ in layer $\ell+1$ computes
\begin{equation}
V_j^{(\ell+1)}
= 
g\Big(\sum_k w_{jk}^{(\ell+1,\ell)} V_k^{(\ell)}-\theta_j^{(\ell+1)}+
\sum_n w_{jn}^{(\ell+1,\ell-1)}  V_n^{(\ell-1)}\Big)\,.
\end{equation}
As usual, the argument of the activation function is the \index{local field}local field $b_j^{(\ell+1)}$.
The weights of all connections are trained in the usual fashion, by stochastic gradient descent\index{gradient descent!stochastic}. 
\begin{figure}
\centering
\begin{overpic}[scale=\myFigureScale]{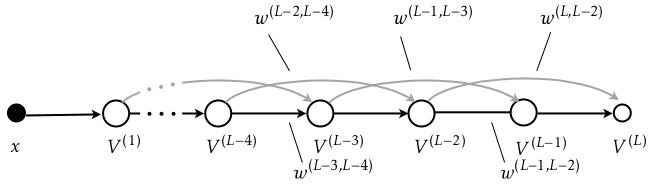}
\end{overpic}
\caption{\figlab{C7S6BackPropagation} Chain of neurons with short cuts
(gray arrows) that skip single neurons.}
\end{figure}

To illustrate the structure of the resulting formulae, consider a chain of neurons, just one neuron per layer, with short cuts that skip one neuron (Figure \figref{C7S6BackPropagation}).  
We calculate the weight increments using Equations (\ref{eq:w_update_general}) and (\ref{eq:delta_gradients}). The recursion (\ref{eq:delta_rec_2}) 
applies only to standard feed-forward networks without skipping layers. In order to determine how to update the weights for the network
shown in Figure \figref{C7S6BackPropagation}, we need to evaluate the gradients $\partial V^{(L)}/\partial V^{(\ell)}$.
To begin with, consider the learning rule for $w^{(L,L-1)}$. Using Equations (\ref{eq:w_update_general}) and (\ref{eq:delta_gradients}) 
one finds
\begin{equation}
\delta w^{(L,L-1)} = \eta \delta^{(L)} V^{(L-1)}\quad\mbox{with}\quad
\delta^{(L)} = (t-V^{(L)}) g'(b^{(L)})\,,
\end{equation}
as in Algorithm \ref{bp_algorithm}.
In the same way one obtains
\begin{equation}
\delta w^{(L,L-2)} = \eta \delta^{(L)} V^{(L-2)}\quad\mbox{with}\quad
\delta^{(L)} = (t-V^{(L)}) g'(b^{(L)})\,.
\end{equation}
Now consider the learning rule for $w^{(L-1,L-2)}$. Using $\partial V^{(L)}/\partial V^{(L-1)} = g'(b^{(L)}) w^{(L,L-1)}$ gives
\begin{equation}
\delta w^{(L-1,L-2)} = \eta \delta^{(L-1)} V^{(L-2)}\quad\mbox{with}\quad
\delta^{(L-1)} = \delta^{(L)} w^{(L,L-1)} g'(b^{(L-1)})\,,
\end{equation}
as before.
But the update for $w^{(L-2,L-3)}$ is different, because now the short cuts come into play. 
The connection from layer $L-2$ to $L$ gives rise to an 
extra term:
\begin{align}
\frac{\partial V^{(L)}}{\partial V^{(L-2)}} = \frac{\partial V^{(L)}}{\partial V^{(L-1)}}\frac{\partial V^{(L-1)}}{\partial V^{(L-2)}}+
g'(b^{(L)}) w^{(L,L-2)}\,.
\end{align}
Evaluating the  partial derivatives yields:
\begin{align}
\nonumber
\delta w^{(L-2,L-3)} = \eta \delta^{(L-2)} V^{(L-3)}\quad\mbox{with}\quad
\delta^{(L-2)} &= \delta^{(L-1)}  w^{(L-1,L-2)}g'(b^{(L-2})\\
&+\textcolor{black}{\delta^{(L)} w^{(L,L-2)}g'(b^{(L-2)})}\,.
\end{align}
Iterating further in this way, one finds the following error-backpropagation rule:
\index{error!backpropagation}
\begin{equation}
\delta^{(\ell-1)} = \delta^{(\ell)}  w^{(\ell,\ell-1)}g'(b^{(\ell-1)})+\delta^{(\ell+1)} w^{(\ell+1,\ell-1)}g'(b^{(\ell-1)})
\end{equation}
with $w^{(\ell+1,\ell-1)}=0$ for $\ell \geq L-1$.
for $\ell=L,L-1,\ldots$. The first term is the same as 
in the error recursion in Algorithm \ref{bp_algorithm}.
The second term is due to the skipping connections. 
These connections reduce the vanishing-gradient problem. To see this, note that we can write
the error $\delta^{(\ell)}$ as
\begin{equation}
\label{eq:delta_path}
\delta^{(\ell)} = \delta^{(L)}\sum_{\ell_1,\ell_2,\ldots,\ell_n}
w^{(L,\ell_n)} g'(b^{(\ell_{n})})\cdots w^{(\ell_2,\ell_{1})} g'(b^{(\ell_1)}) w^{(\ell_1,\ell)} g'(b^{(\ell)})
\end{equation}
where the sum is over all paths $L> \ell_n >\ell_{n-1}>\cdots> \ell_1> \ell$ back through the network.
The structure of the general formula, for networks with more than only one neuron per layer, is analogous
to Equation (\ref{eq:delta_path}).
According to this equation, the smallest errors, or gradients, in networks with many layers are dominated by the product corresponding to the path with the smallest number of steps (factors).
Therefore short cuts tend to increase small gradients.

Finally, the network described in Ref.~\cite{He2015} used unit weights for the skipping connections. In this case, the local field of $V_j^{(\ell+1)}$
takes the form
\begin{equation}
\label{eq:residual_net}
b_j^{(\ell+1)} = \sum_k w^{(\ell+1,\ell)}_{jk} V_k^{(\ell)}-\theta_j^{(\ell+1)}+V_j^{(\ell-1)}\equiv F+V_j^{(\ell-1)}\,,
\end{equation}
assuming that the hidden layers have the same number of neurons.
Here $F$ is a residual contribution to the local field (when $F=0$, the inputs
$V_j^{(\ell-1)}$ are passed right through to $b_j^{(\ell+1)}$).
Therefore such networks are called {\em residual networks} \cite{He2015}.
But note that the networks described in Ref.~\cite{He2015} use convolution layers (Section~\ref{sec:cl81}).
\index{residual network|textbf|)}

\section{Outputs and energy functions}
\index{energy function|(}
\label{sec:ocf}
Up to now we discussed networks that have the same activation functions for all neurons in all layers, either sigmoid or $\tanh$ activation functions
\index{activation function!sigmoid}\index{activation function!tanh}
[Equation (\ref{eq:activation_functions})], or ReLU functions 
(Sections \ref{sec:other:models} and  \ref{sec:ReLU}).
In the output layer,  one often uses neurons with a different activation function, so-called {\em softmax}\index{softmax} outputs:
\begin{equation}
\label{eq:softmax}
O_i = \frac{ {\rm e}^{\alpha b_i^{(L)}} }{ \sum_{k=1}^M {\rm e}^{\alpha b_k^{(L)}} }\,.
\end{equation}
Here $b_i^{(L)} = \sum_j w_{ij}^{(L)} V_j^{(L-1)}-\theta_i^{(L)}$ 
are the \index{local field}local fields in the output layer. 
In the limit $\alpha\to\infty$, we see that $O_i=\delta_{ii_0}$ 
where $i_0$ is the index of the winning\index{winning neuron}
output neuron\index{output neuron}, the one with the largest value $b_i^{(L)}$
(Chapter \ref{ch:uhl}).
For $\alpha=1$, Equation (\ref{eq:softmax}) is a {\em soft} version of this maximum criterion, thus the name {\em softmax}. We set $\alpha$ to unity from now on.

Two important properties of softmax outputs are, first,
that $0 \leq O_i \leq 1$. Second, the values of the outputs sum to unity, 
\begin{equation}
\sum_{i=1}^MO_i = 1\,.
\end{equation}
Therefore the outputs of softmax units can be interpreted as probabilities. 
Consider classification problems where the inputs must
be assigned to one of $M$ classes. In this case, the output $O_i^{(\mu)}$
of softmax unit $i$ is assumed to represent the probability that the input $\ve x\tomu$ 
is in class $i$ (in terms of the \index{target}targets: $t_i\tomu=1$ while $t_k\tomu=0$ for $k\neq i$). If $O_{i}^{(\mu)}\approx 1$, we assume that the network is quite certain that input $\ve x^{(\mu)}$ is in class $i$. On the other hand, if all $O_k^{(\mu)} \approx M^{-1}$, we interpret the network output as uncertain. But note that neural networks may fail like humans sometimes do: their output can be very certain yet wrong (Section \ref{sec:s8}).

Softmax units are used in conjunction with a different energy
function, 
\begin{equation}
\label{eq:loglikelihood}
H = -\sum_{i\mu} t_i\tomu\log O_i\tomu\,.
\end{equation}
Here and in the following $\log$ stands for the natural logarithm. 
The function (\ref{eq:loglikelihood}) is minimal when $O_i\tomu = t_i\tomu$ (Exercise 7.5). 
To find the correct backpropagation formula for the energy function (\ref{eq:loglikelihood}), we need
to evaluate
\begin{align}
\frac{\partial H}{\partial w_{mn}}
= -\sum_{i\mu} \frac{t_i\tomu}{O_i\tomu} \frac{\partial O_i\tomu}{\partial w_{mn}}\,.
\end{align}
Here the labels denoting the output layer were omitted,
and in the following equations the index $\mu$ that
refers to the input pattern is dropped as well. Using the identities
\begin{equation}
\frac{\partial O_i}{\partial b_{l}} 
=O_i(\delta_{il}-O_l)\quad\mbox{and}\quad
\frac{\partial b_l}{\partial w_{mn}} = \delta_{lm}V_n\,,
\end{equation}
one obtains
\begin{equation}
\frac{\partial O_i}{\partial w_{mn}} = \sum_l 
\frac{\partial O_i}{\partial b_l} 
\frac{\partial b_l}{\partial w_{mn}}
= O_i  (\delta_{im}-O_m)  V_n\,.
\end{equation}
So
\begin{equation}
\label{eq:dwll}
\delta w_{mn} = -\eta \frac{\partial H}{\partial w_{mn}}
= \eta \sum_{i\mu} t_i\tomu (\delta_{im}-O_m\tomu)V_n\tomu
= \eta\sum_\mu(t_m\tomu-O_m\tomu)V_n\tomu\,,
\end{equation}
since $\sum_{i=1}^M t_i\tomu=1$ for the type of classification problem
where each input belongs to precisely one class.
The corresponding learning rule for the \index{threshold}thresholds reads
\begin{equation}
\label{eq:dthetall}
\delta\theta_m = -\eta \frac{\partial H}{\partial \theta_{m}}
= -\eta\sum_\mu (t_m\tomu-O_m\tomu)\,.
\end{equation}
Equations (\ref{eq:dwll}) and (\ref{eq:dthetall}) highlight a further
advantage of softmax output neurons\index{output neuron} (apart from the fact that they
allow the output to be interpreted in terms of probabilities). 
The weight and threshold increments for the output layer
derived in Section \chref{C6Chapter} 
[Equations \eqnref{C6S2SynapticWeightsOutput} and (\ref{eq:Theta_output})]
contain factors of derivatives $g'(B_m\tomu)$. As noted earlier, these
derivatives tend to zero when the activation function saturates, slowing
down the learning. But here the rate at which the neuron learns
is simply proportional to the output error\index{output error}, $(t_m\tomu-O_m\tomu)$, without any possibly small factor $g'(b)$.
Softmax units are normally only used in the output layer, because the learning speedup is coupled to the use of the energy function (\ref{eq:loglikelihood}), and because it is customary to avoid dependencies between the neurons within a hidden layer.

There is an alternative form of the energy function that is very similar
to the above, but works with sigmoid activation functions and 0/1 \index{target}targets.
\index{activation function!sigmoid}
Instead of Equation (\ref{eq:loglikelihood}) one chooses:
\begin{equation}
\label{eq:Hcee}
H =- \sum_{i\mu}{\Big[}t_i\tomu \log O_i\tomu + (1-t_i\tomu)\log(1-O_i\tomu){\Big]}\,,
\end{equation}
with $O_i = \sigma(b_i)$, $i=1,\ldots,M$, and where $\sigma$ denotes the sigmoid function (\ref{eq:sigmoid}).
To compute the \index{weight increment}weight increments, we apply the chain rule:
\begin{align}
\frac{\partial H}{\partial w_{mn}} &= 
-\sum_{i\mu} \Bigg( \frac{t_i\tomu}{O_i\tomu} -\frac{1-t_i\tomu}{1-O_i\tomu} \Bigg)\frac{\partial O_l}{\partial w_{mn}}
= -\sum_{i\mu}  \frac{t_i\tomu-O_i\tomu}{O_i\tomu(1-O_i\tomu)} \frac{\partial O_l}{\partial w_{mn}}\,.
\end{align}
Using Equation (\ref{eq:dsigma}) we obtain
\begin{equation}
\delta w_{mn} = \eta \sum_\mu (t_m\tomu-O_m\tomu)V_n\tomu\,,
\end{equation}
identical to Equation (\ref{eq:dwll}). The \index{threshold}thresholds 
are adjusted in an analogous fashion, Equation (\ref{eq:dthetall}).
But now the interpretation of the outputs is slightly different, since
the values of the softmax units in the output layers sum to unity, while
those with sigmoid activation functions do not. In either
case one can use the definition (\ref{eq:Cey}) for the classification error\index{classification error}.

To conclude this Section, we briefly discuss the meaning of the energy
functions (\ref{eq:loglikelihood}) and (\ref{eq:Hcee}).
In Chapter  \chref{C6Chapter} we saw how deep neural networks\index{network}\index{deep network} are trained to fit input-output functions (Section \ref{sec:hmhl}) by minimising the quadratic energy function (\ref{eq:H3}). This reminds of regression analysis in
mathematical statistics, where the predictive accuracy of a model is improved
by minimising the sum over the squared errors. \index{error!squared}
Now consider the energy function (\ref{eq:Hcee}) for a single sigmoid output
with \index{target}targets $t=0$ and $t=1$. In this case, the network
output is interpreted as the probability $O^{(\mu)} = \mbox{Prob}(t^{(\mu)}=1|\ve x^{(\mu)})$ 
of observing $t^{(\mu)}=1$.
\index{likelihood}
The corresponding likelihood is the joint probability of observing the outcomes $t^{(\mu)}$
for $p$ independent inputs $\ve x^{(\mu)}$:
\begin{equation}
\mathscr{L}
=\prod_{\mu=1}^p \big({O^{(\mu)}}\big)^{t^{(\mu)}} \big(1-O^{(\mu)}\big)^{1-t^{(\mu)}}\,,
\end{equation}
under the model determined by the weights and thresholds of the network.
\index{log likelihood}
Minimising the negative log-likelihood $-\mathscr{L}$ (Section \ref{sec:bm1}) corresponds
to minimising (\ref{eq:Hcee}). This is just binary logistic
\index{regression!logistic, binary} 
regression \cite{kleinbaum2008applied} to predict a binary outcome $t=0$ or $1$.
 The case $M>1$ corresponds to a \index{regression!multivariate} multivariate regression problem \cite{kleinbaum2008applied} with
$M$ possibly correlated outcome variables $t_1,\ldots,t_M$.

When the \index{target}targets describe $M$ mutually exclusive categorical outcomes\index{categorical outcome},  $t_i=0,1$ with $\sum_{i=1}^Mt_i=1$, the softmax output $O_i$ is interpreted as the probability of observing $t_i=1$. An example is the problem of classifying hand-written digits (Section \ref{sec:MNIST}). Training the network then corresponds to \index{regression!multinomial} multinomial regression \cite{kleinbaum2008applied} with the log-likelihood  (\ref{eq:loglikelihood}). 
Note that Equation (\ref{eq:Hcee}), for $M=1$, is equivalent to (\ref{eq:loglikelihood}) for $M=2$, because $O_2=1-O_1$ and $t_2=1-t_1$.
At any rate, these remarks motivate why the energy functions 
 (\ref{eq:loglikelihood}) and (\ref{eq:Hcee}) 
are sometimes called log-likelihoods. Equation (\ref{eq:Hcee})
is also referred to as {\em cross entropy}\index{cross entropy|textbf},
because it has the same for as the cross entropy \cite{Murphy} characterising
the difference between two Bernoulli distributions: the network output
$O_i$, and the \index{target}target $t_i$.
\index{distribution!Bernoulli}
\index{energy function|)}

\section{Regularisation}
\index{regularisation|(}
\label{sec:reg}
Deeper networks have more neurons, so the problem of overfitting\index{overfitting} 
(Figure \figref{C6S3CT10}) tends to be more severe for deeper networks\index{network}\index{deep network}. 
{\em Regularisation}\index{regularisation} schemes limit the tendency to overfit. 
Apart from  cross validation \index{cross validation} (Section \ref{sec:crossvalidation}), a number of other regularisation schemes have proved useful for deep networks: {\em weight decay}\index{weight!decay}, {\em pruning}\index{pruning}, {\em drop out}\index{drop out}, \index{regularisation|textbf}
{\em expansion of  the training set}\index{training set!expansion of}, and {\em batch normalisation}
\index{batch normalisation}. This Section summarises the most important aspects of these methods.
\index{regularisation|)}

\subsection{Weight decay}
\index{weight!decay|(}
\label{sec:weightdecay}
Recall Figure \figref{C5S4XORWithHiddenGraph} which shows a solution of the classification problem defined by the Boolean XOR function.\index{XOR function}  
\index{Boolean function}
In the solution illustrated in this Figure, all weights equal $\pm 1$, and also the thresholds are of order unity. If one uses the backpropagation algorithm to find
a solution to this problem, one may find that the weights continue to grow during training. As mentioned above, this can be problematic because
it may imply that
the \index{local field}local fields become so large
that the activation functions saturate.\index{activation function!saturation}
Then
\index{training!slow down}training slows down, as explained in 
Section~\ref{sec:vg}.

To prevent the weights from growing, one can reduce them by some factor during training, either at each iteration or in regular intervals,
    $w_{ij} \rightarrow (1-\varepsilon) w_{ij}$ for $0 < \varepsilon < 1$, or
\begin{equation}
\delta\! w_{mn} = -\varepsilon w_{mn}\quad\mbox{for}\quad 0 < \varepsilon < 1\,.
\end{equation}
This is achieved by adding a term to the energy function:\index{energy function}
\begin{equation}
\label{eq:modifiedH}
    H = \underbrace{\frac{1}{2} \sum_{i\mu}\Big(t_{i}^{(\mu)} - O_{i}^{(\mu)}\Big)^{2}}_{\equiv H_{0}} + \frac{\gamma}{2} \sum_{ij}w_{ij}^{2}\,.
\end{equation}
Gradient descent on $H$ gives:\index{gradient descent}
\begin{equation}
\begin{split}
\label{eq:dwL2}
    \delta\! w_{mn} = -\eta \frac{\partial H_{0}}{\partial w_{mn}} - \varepsilon w_{mn}
\end{split}
\end{equation}
with $\varepsilon = \eta\gamma$. One can include a corresponding term for the \index{threshold}thresholds. The scheme summarised here
is sometimes called {\em $L_2$-regularisation}\index{regularisation!L2 regularisation|textbf}.
An alternative scheme is {\em $L_1$-regularisation}\index{regularisation!L1 regularisation|textbf}. It amounts to
\begin{equation}
\label{eq:modifiedH2}
    H = \frac{1}{2} \sum_{i\mu}\Big(t_{i}^{(\mu)} - O_{i}^{(\mu)}\Big)^2 + \frac{\gamma}{2} \sum_{ij}|w_{ij}|\,.
\end{equation}
This gives the learning rule
\begin{equation}
\label{eq:dwL1}
\delta\! w_{mn} =  -\eta \frac{\partial H_{0}}{\partial w_{mn}} - \varepsilon \mbox{sgn}(w_{mn})\,.
\end{equation}
The  discontinuity of the learning rule at $w_{mn}=0$ is cured by defining $\mbox{sgn}(0)=0$.
Comparing Equations (\ref{eq:dwL2}) and (\ref{eq:dwL1}), we see that $L_1$-regularisation puts more weights to zero, compared with the $L_2$-scheme \cite{Nielsen}.

An alternative to these two methods is {\em max-norm regularisation}\index{regularisation!max norm regularisation|textbf}
\cite{Srivastava}, where the weights are constrained to remain smaller than a given constant: $|w_{ij}| \leq c$.
If a $|w_{ij}|$ exceeds the positive constant $c$, then $w_{ij}$ is rescaled so that 
$|w_{ij}| = c$.

\index{weight!decay}
These weight-decay schemes are referred to as {\em regularisation}\index{regularisation} schemes because
they tend to help against overfitting\index{overfitting}. 
How does this work? Weight decay adds a constraint to the problem of minimising the energy function. The result is a compromise \cite{Nielsen} between a small value of $H$ and small weight  values.  
The idea is that a network with smaller weights is more robust to the effect of noise. When the weights
are small, then small changes in some of the patterns do not give a substantially different training result.
When the network has large weights, by contrast, it may happen that small changes in the input
yield significant differences in the training result that are difficult to generalise\index{generalisation}.
\index{weight!decay|)}

\subsection{Pruning}
\index{pruning|textbf|(}
\label{sec:pruning}
The term {\em pruning} \index{pruning} refers to removing unnecessary weights or neurons from the network, to improve its efficiency. The simplest approach is {\em weight elimination}\index{weight!elimination}
by {\em weight decay}\index{weight!decay} \cite{HansonPratt1989}. Weights that tend to remain very close to zero during \index{training}training are removed by setting them to zero and not updating them anymore. Neurons that have zero weights for all incoming connections are effectively removed ({\em pruned}). 
Pruning is a regularisation method: by removing
unnecessary weights, one reduces the risk of overfitting\index{overfitting}. As opposed to drop out (Section~\ref{sec:dropout}), where \index{hidden neuron} hidden neurons are only temporarily ignored, pruning
refers to permanently removing hidden neurons. The idea is to train
a large network, and then to prune a large fraction of neurons to
obtain a much smaller network. It is usually found that such pruned networks
generalise better than small networks that were trained without pruning.
Up to $90\%$ of the hidden neurons can be removed in some cases. In general, pruning 
is an excellent way to create efficient classifiers for real-time applications.

An efficient pruning algorithm is based on the idea 
to remove  weights in such a way that the effect upon the energy function is as small as possible \cite{HassibiStork1993}. 
The idea is to find the optimal weight, to remove it, and to change the other weights in such a way that the energy function increases as little as possible. The algorithm works as follows.
Assume that
the network was trained, so that it reached a (local) minimum of the energy function $H$.  
One expands
 the energy function around this minimum. To second order, the expansion of $H$ reads:
\begin{equation}
H = H_{\rm min}+\tfrac{1}{2} \delta\! \ve w \cdot \ma M \delta\! \ve w
\quad +\quad\mbox{higher orders in}\quad  \delta\! \ve w\,.
\end{equation}
The term linear in $\delta \!\ve w$ vanishes because we expand
around a \index{local minimum}local minimum. The matrix $\ma M$
is the {\em Hessian}\index{matrix!Hessian|textbf}, the matrix
of second derivatives of the energy function.

For the next step it is convenient to adopt the following
notation \cite{HassibiStork1993}.  One groups all weights in the network 
into a long weight vector $\ve w$ (as opposed
to grouping them into a \index{weight!matrix}weight matrix $\ma W$ as we
did in Chapter \ref{ch:dhn}). A particular component $w_q$ 
is extracted from the  vector $\ve w$ as follows:
\begin{equation}
w_q = \hat{\textbf e}_q\cdot \ve w\quad\text{where}
\quad \hat{\textbf e}_q=\begin{bmatrix}\vdots\\1\\\vdots\end{bmatrix}
\raisebox{0cm}{$\leftarrow q$}\,.
\end{equation}
Here $\hat{\textbf e}_q$ is the
Cartesian unit vector in the direction
$q$, with  components $[\hat{\textbf e}_{q}]_j=\delta_{qj}$. In this notation, the
elements of $\ma M$ are $M_{pq} = \partial^2 H/\partial \!w_p \partial \!w_q$.
Now, eliminating the weight  $w_{q}$ amounts to setting
\begin{equation}
\label{eq:q_constr}
\delta\! w_{q} =-w_q\,.
\end{equation}
To minimise the damage 
to the network we should eliminate the weight that has least effect upon $H$, changing the other weights at the same time so that $H$ increases as little as possible (Figure \ref{fig:C6S3Pruning}).
This is achieved by minimising
\begin{equation}
\min_{q} \min_{\delta\!\small \ve w}\{\tfrac{1}{2}  \delta\! \ve w\cdot\ma  M\delta \!\ve w\}
 \quad\mbox{subject to the constraint}\quad
\hat{\textbf e}_q \cdot \delta\!  \ve w + w_q=0\,.
\end{equation}
\begin{figure}[t]
\centering
\begin{overpic}[scale=\myFigureScale]{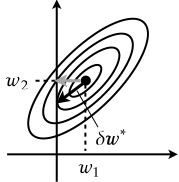}
\end{overpic}
\caption{
\figlab{C6S3Pruning} Pruning algorithm (schematic). 
The minimum of $H$ is located at $[w_1,w_2]^{\sf T}$. The contours of the quadratic approximation to $H$ are represented as solid black lines.     
 The weight change $\delta \!\ve w = [-w_1,0]^{\sf T}$ (gray arrow) 
leads to a smaller increase in $H$ than 
$\delta \!\ve w = [0,-w_2]^{\sf T}$. The black arrow represents the optimal 
$\delta\!\ve w_{\!q}^\ast$ which leads to an even smaller increase in $H$.
}
\end{figure}
The constant term $H_{\rm min}$ was dropped because it does not matter. Now we first minimise $H$ w.r.t. $\delta \!\ve w$, for a given value
of $q$. The linear constraint is incorporated using a {\em Lagrange multiplier}
\index{Lagrange multiplier} as in Section \ref{sec:pprid},
to form the {\em Lagrangian}\index{Lagrangian}              
\begin{equation}
\label{eq:Lagrangian}
\mathscr{L}=\tfrac{1}{2} \delta \!\ve w \cdot \ma M \delta \!\ve w+ \lambda (\hat{\textbf e}_q \cdot \delta \! \ve w + w_q)\,.
\end{equation}
A necessary condition for a minimum $[\delta\!\ve w,\lambda]$ satisfying the constraint is 
\begin{equation}
\frac{\partial\mathscr{L}}{\partial \delta\!\ve w}=\ma M \delta\!\ve w+\lambda \hat{\textbf e}_q=0
\quad\mbox{and}\quad
\frac{\partial\mathscr{L}}{\partial \lambda}=\hat{\textbf e}_q \cdot \delta\!  \ve w + w_q=0\,.
\end{equation}
We denote the solution of these Equations by $\delta \!\ve w^\ast$ and $\lambda^\ast$.
It is obtained by solving the linear system
\begin{equation}
\label{eq:lin_sys}
\begin{bmatrix}
\ma M & \hat{\textbf e}_q\\
\hat{\textbf e}_q^{\sf\small T} & 0
\end{bmatrix}
\begin{bmatrix}
\delta\!\ve w^\ast \\ \lambda^\ast
\end{bmatrix}
= 
\begin{bmatrix}
0 \\ -w_q
\end{bmatrix}\,.
\end{equation}
If $\ma M$ is invertible\index{matrix!inverse}, then the top rows of Eq.~(\ref{eq:lin_sys}) give
\begin{equation}
\delta\!\ve w^\ast = -\ma M^{-1} \hat{\textbf e}_q\lambda^\ast\,.
\end{equation}
Inserting this result into $\hat{\textbf e}_q^{\sf T} \delta\!\ve w^\ast + w_q=0$
we find 
\begin{equation}
\label{eq:wast}
\delta\!\ve w^\ast =-\ma M^{-1} \hat{\textbf e}_q w_q(\hat{\textbf e}_q^{\sf \small T}\ma M^{-1}\hat{\textbf e}_q)^{-1}
\mbox{and}\quad \lambda^\ast = w_q(\hat{\textbf e}_q^{\sf \small T}\ma M^{-1}\hat{\textbf e}_q)^{-1}\,.
\end{equation}
We  see that $\hat{\textbf e}_q\cdot \delta\! \ve w^\ast = -w_q$, so that the weight $w_q$ is eliminated.
The other weights are also changed (black arrow in Figure \ref{fig:C6S3Pruning}).
The final step is to find the optimal $q$ by minimising
\begin{equation}
\label{eq:Last}
\mathscr{L}(\delta\!\ve w^\ast,\lambda^\ast;q) = \frac{1}{2} w_q^2 (\hat{\textbf e}_q^{\sf \small T}\ma M^{-1}\hat{\textbf e}_q)^{-1}\,.
\end{equation}
The \index{matrix!Hessian}Hessian of the energy function is expensive to evaluate, 
and so is the inverse of this matrix. Usually one
resorts to an approximate expression for $\ma M^{-1}$ \cite{HassibiStork1993}.
One possibility is to set the off-diagonal elements of $\ma M$ to zero \cite{LeCun1990}.
But in this case the other weights are not adjusted, because  
$\hat{\textbf e}_{q'} \cdot \delta\!\ve w_q^\ast = 0$ for $q'\neq q$ if $\ma M$ is diagonal.
In this case it is necessary to retrain the network after weight elimination.
\index{weight!elimination}

The algorithm is summarised in Algorithm \ref{prune}.
It succeeds better than elimination by weight decay\index{weight!decay}
 in removing the unnecessary weights in the network \cite{HassibiStork1993}.
Weight decay eliminates the smallest weights. 
One obtains weight elimination of the
smallest weights  by substituting $\ma M=\ma I$ in the algorithm
described above [Equation (\ref{eq:Last})]. Since
small weights are often needed to achieve a small \index{training!error}
training error,
this is usually not a good approximation.
\begin{algorithm}[b]
\caption{\label{prune} pruning least important weight}
\begin{algorithmic}
\STATE train the network to reach $H_{\rm min}$;
\STATE compute $\ma M^{-1}$ approximately; \label{inverse}
\STATE determine $q^\ast$ as the value of $q$ for which $\mathscr{L}(\delta\!\ve w^\ast,\lambda^\ast;q)$ is minimal;
\IF{ $\mathscr{L}(\delta\!\ve w^\ast,\lambda^\ast;q^\ast)\ll H_{\rm min}$ }
\STATE adjust all weights using $\delta\!\ve w = -w_{q^\ast} \ma M^{-1} \hat{\textbf e}_{q^\ast}(\hat{\textbf e}_{q^\ast}^{\sf \small T}\ma M^{-1}\hat{\textbf e}_{q^\ast})^{-1}$;\label{update}
\STATE goto \ref{inverse};
\ELSE
\STATE end; \label{end}
\ENDIF
\end{algorithmic}
\end{algorithm}

To illustrate the effect of pruning for neural networks with hidden layers, consider
the XOR function\index{XOR function}. Recall that it
can be represented by a hidden layer with two neurons (Figure \figref{C5S4XORWithHiddenGraph}).
For random \index{weight!initial}initial weights, backpropagation takes a long time to find a valid solution,
and networks with many more \index{hidden neuron} hidden neurons tend to perform much better \cite{FrankleCarbin}.
The numerical experiments in Ref.~\cite{FrankleCarbin} indicate that 
with two hidden neurons, only about 49\% of the networks learned the task in 10 000 training steps of
stochastic gradient descent, and networks with more neurons in the hidden layer learn more easily  
(98.5 \% for $n=10$ hidden neurons). 
The data in Ref.~\cite{FrankleCarbin} also shows that 
pruned networks, initially trained with $n=10$ hidden neurons, still show excellent 
training success (83.3 \% if only $n=2$ hidden neurons remain). The networks were pruned
iteratively during training, removing the neurons with the largest average magnitude.
After \index{training}training, the weights and threshold were reset to their initial values, the values before training began.

One can draw three conclusions from the numerical experiments described in Ref.~\cite{FrankleCarbin}.
First, \index{pruning!iterative}iterative pruning during training singles out neurons in the hidden layer that had initial weights and thresholds\index{weight!initial}\index{threshold!initial}
resulting in the correct decision boundaries\index{decision boundary} ({\em lottery-ticket} effect \cite{FrankleCarbin}). 
Second, the pruned network with two hidden neurons has much better training
success than the network that was trained with only two hidden neurons. Third, despite pruning more than 50\% of the hidden neurons,
the network with $n=4$ hidden neurons performs almost as well as the one with $n=10$ hidden neurons (97.9 \% training success).
When \index{training!deep networks}training deep networks it is common to start with many neurons in the hidden layers, and to prune
up to 90\% of them. This results in small trained networks that can classify efficiently and reliably.
\index{pruning|)}

\subsection{Drop out}
\index{drop out|(}
\label{sec:dropout}
In this regularisation scheme, some hidden neurons are ignored during \index{training}training. 
In each step of the \index{training!algorithm}training algorithm (for each \index{mini batch}mini batch, or for each
individual pattern), one disregards at random a fraction $q$ of neurons
from each hidden layer by setting their outputs to zero, and by
updating only the weights and thresholds of the remaining neurons.
For the next step in the \index{training!algorithm}training algorithm, the 
ignored neurons at put back, and  another set of hidden neurons is 
disregarded. 
Once the training is completed, all hidden neurons are activated,
but their outputs are multiplied by ${1-q}$, to ensure that the
local fields are independent of $q$, on average.

This method is motivated by noting that the performance of machine-learning algorithms is usually improved by combining the results of several learning attempts \cite{Nielsen,Srivastava}, for instance
by separately training networks with different layouts, and averaging over their outputs. However, for deep networks this is computationally very expensive. Drop out is an attempt to achieve the same goal more efficiently. The idea is that dropout corresponds to effectively 
training a large number of different networks. If there are $k$ hidden neurons,
then there are $2^k$ different combinations of neurons that are turned on or off. The hope is that
the network learns more robust features of the input data in this way, and that
this reduces overfitting\index{overfitting}. In practice the method is applied
 together with max-norm regularisation\index{regularisation!max norm regularisation} (Section \ref{sec:weightdecay}).
\index{drop out|)}

\subsection{Expanding the training set}
\index{training set!expansion of|(}
If one trains a network with a fixed number of hidden neurons on larger training sets, one observes
that the network generalises with higher accuracy (better classification success). The reason
is that overfitting\index{overfitting} is reduced when the training set is larger. Thus,  
a way of avoiding overfitting is to {\em expand} or {\em augment} the training set\index{training set!expansion of|textbf}\index{data set augmentation}. 

It is sometimes argued that the recent success of deep neural networks in image recognition and 
object recognition\index{object recognition} is in large part due to larger training sets. One example is \href{www.image-net.org}{ImageNet}\index{imagenet}, a database of
more than 10$^7$ hand-classified images, into more than 20 000 categories \cite{imagenet}. 

It is expensive to augment training sets because
it requires manual annotation. An alternative is to expand a training
set  {\em artificially}. For digit recognition (Figure \figref{C2S1Digits}), for example, one could create more input patterns by by shifting, rotating, and shearing the digits, or by adding noise.
\index{training set!expansion of|)}

\subsection{Batch normalisation}
\index{batch normalisation|textbf|(}
\label{sec:bn}
Batch normalisation \cite{IoffeSzegedy2015} can significantly speed up the 
\index{training!deep networks}training of deep networks with backpropagation. 
The idea is to shift and normalise the input data for each hidden layer, not only for the 
input patterns\index{input pattern!normalisation} (Section \ref{sec:pprid}).
This is done separately for each \index{mini batch}mini batch, and for each component of the inputs 
into the given layer (Algorithm \ref{alg:bn}). 
Denoting the states of the neurons 
feeding into the layer in question by
$V_j^{(\mu)}$, $j=1,\ldots,j_{\rm max}$, one calculates the average and variance over each \index{mini batch}mini batch
\begin{align}
\overline{V}_j &= \frac{1}{m_B}\sum_{\mu=1}^{m_B} V_j^{(\mu)}\quad\mbox{and}\quad \sigma_B^2 =\frac{1}{m_B}\sum_{\mu=1}^{m_B} (V_j^{(\mu)}-\overline{V}_j)^2\,,
\end{align}
subtracts the mean from the $V_j^{(\mu)}$,  and divides by $\sqrt{\sigma_B^2+\epsilon}$.
The parameter $\epsilon>0$ is added to the 
denominator to avoid division by zero when $\sigma^2_B$ evaluates to zero.
There are two additional parameters in Algorithm \ref{alg:bn}, $\gamma_j$ and $\beta_j$.
They are are learnt by backpropagation, just like 
the weights and thresholds. In general the new parameters 
are allowed to differ from layer to layer, $\gamma_j^{(\ell)}$ and $\beta_j^{(\ell)}$. 

Batch normalisation was originally motivated by arguing that it reduces possible covariate shifts\index{covariate shift} faced
by \index{hidden neuron} hidden neurons in layer $\ell$: as the parameters of the neurons in the preceding layer $\ell-1$ change, their outputs shift
thus  forcing the neurons in layer $\ell$ to adapt. However in Ref.~\cite{Santurkar2018} it was argued that batch normalisation
does not reduce the internal covariate shift, but that it speeds up the training by effectively smoothing
the \index{energy landscape}energy landscape. 

Batch normalisation helps to combat the {\em vanishing-gradient problem}\index{vanishing gradient} because it prevents local 
fields of hidden neurons to grow. This makes it possible to use sigmoid functions\index{activation function!sigmoid}  in deep networks, because the distribution of inputs remains normalised. 
It is sometimes argued that batch normalisation has a regularising effect, and it has been suggested
\cite{IoffeSzegedy2015} that batch normalisation can replace drop out\index{drop out} (Section \ref{sec:dropout}).
It is also argued that batch normalisation may help the network to generalise better, in particular if each \index{mini batch}mini batch contains randomly picked inputs. Then batch normalisation corresponds to randomly transforming the inputs to each hidden neuron (by the randomly changing means and variances).  This may help to make the learning more robust. There is no theory that proves either of these claims, but it is an empirical fact that batch normalisation often speeds up the training.\index{training}

\begin{algorithm}[t]
\caption{\label{alg:bn} batch normalisation}
\begin{algorithmic}
\FOR {$j=1,\ldots,j_{\rm max}$}
   \STATE calculate mean $\overline{V}_j \leftarrow \frac{1}{m_B}\sum_{\mu=1}^{m_B} V_j^{(\mu)}$
   \STATE calculate variance $\sigma_B^2 \leftarrow \frac{1}{m_B}\sum_{\mu=1}^{m_B} (V_j^{(\mu)}-\overline{V}_j)^2$
   \STATE normalise $\hat V_j^{(\mu)} \leftarrow (V_j^{(\mu)}-\overline{V}_j)/\sqrt{\sigma_B^2+\epsilon}$
   \STATE calculate outputs as: $g(\gamma_j \hat V_j^{(\mu)} + \beta_j)$
   \ENDFOR
\end{algorithmic}
\end{algorithm}
\index{batch normalisation|)}

\section{Summary}
Neural networks with many layers of hidden neurons are called deep networks.  Error backpropagation in deep networks suffers from the va\-ni\-shing\--gradient problem. It can be reduced by using ReLU units, by initialising the weights in certain ways, and with networks containing connections that skip layers. Yet vanishing\index{vanishing gradient} or exploding gradients \index{exploding gradient} remain a fundamental difficulty, slowing learning down in the initial phase of training.  \index{training!slow down}
Nevertheless, deep neural networks have become immensely successful in object recognition\index{object recognition}, outperforming other algorithms significantly.  

Since deep networks contain many free parameters, deep networks tend to overfit the training data, so that the networks must be regularised.  Apart
from cross validation\index{cross validation}, there are other ways of regularising the problem: weight decay\index{weight!decay}, drop out\index{drop out}, pruning\index{pruning}, and data-set augmentation\index{data set augmentation}.

\section{Further reading}
\label{sec:further_reading_7}
Deep networks suffer from \index{catastrophic forgetting}{\em catastrophic forgetting}: when we train a network on a new \index{input distribution}input distribution that is quite different from the one the network was originally trained on, then the network tends to forget what it learned initially. A good starting point for further reading is Ref.~\cite{Kirkpatrick2016}.

The stochastic-gradient descent algorithm (with or without minibatches) samples
the input-data distribution uniformly randomly. As mentioned in Section
\ref{sec:pprid}, it may be advantageous to sample those inputs more frequently that
initially cause larger output errors\index{output error}. More generally, the algorithm may use other criteria to choose certain input data more often, with the goal to speed up
learning. It may even suggest how to augment a given training set most efficiently, by asking to specifically
label certain types of input data ({\em active learning})\index{active learning} \cite{Settles2009}.

Another  question concerns the structure of the \index{energy landscape}energy landscape for multilayer perceptrons. \index{multi layer perceptron}
It seems that \index{local minimum}local minima are perhaps less important for deep networks than for Hopfield networksindex{Hopfield network}, because the energy functions of deep networks tend to have more saddle points than minima \cite{Choromanska2014}, just like Gaussian random functions \cite{Fyodorov}. A recent study explores the relation between the multilayer layout of the perceptron network and the properties of the \index{energy landscape}energy landscape \cite{becker2020geometry}.

Finally, training tends to work best when all input patterns appear with roughly the same frequency in the training set. Unlike humans, neural networks may struggle with rare input patterns. Special techniques, however, allow networks to recognise rare patterns, by comparison with features of the input distribution that are well represented \index{learning!few shot}({\em few-shot} learning \cite{wang2020generalizing}\index{learning!few shot}).
Standard algorithms for few-shot learning use elements of unsupervised learning (Chapter \ref{ch:uhl}).

\vfill\eject

\chapter{Convolutional networks} 
\label{chapter:con_net}
\begin{figure}
\centering
\begin{overpic}[scale=\myFigureScale]{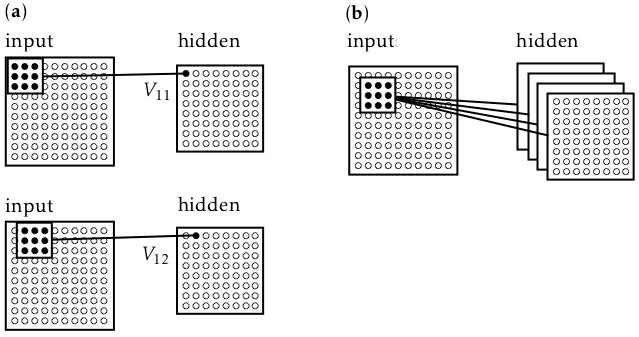}
\end{overpic}
\caption{\figlab{C7S2ConvNet} ({\bf a}) 
Feature map\index{feature map}, kernel\index{kernel}, 
and receptive field\index{receptive field} (schematic). A feature map\index{feature map|textbf}  (the $8\times 8$ array of \index{hidden neuron} hidden neurons) is obtained by translating a kernel\index{kernel} (filter\index{filter}), here with a $3\times 3$ receptive field\index{receptive field} over the input image, here a $10\times 10$ array of pixels.  ({\bf b}) A convolution layer\index{convolution layer} consists of a number of feature maps, each corresponding to a given kernel that detects a certain feature in parts of the input image.  After a figure in Ref.~\cite{Nielsen}.  }
\end{figure}
Convolutional networks \index{convolutional network} have been around since the 1980's. They became widely used
after Krizhevsky {\em et al.} \cite{Krizhevsky2012} won the \index{imagenet}ImageNet challenge
(Section \ref{sec:dlor}) with a convolutional net.
One reason for the recent success of convolutional networks is that they have fewer connections
than fully connected networks with the same number of neurons. This has two
advantages. Firstly, such networks are obviously cheaper to train. Secondly, as pointed out above,
reducing the number of connections regularises the network, it reduces the risk of overfitting\index{overfitting}. 

Convolutional neural networks are designed for object recognition\index{object recognition} and 
image   classification. They take images
as inputs (Figure \figref{C7S1Iris}), not just a list of attributes (Figure \figref{C5S1Iris}).
Convolutional networks have important properties in common with networks of neurons in
the visual cortex of the human brain \cite{Goodfellow}. First, there is a spatial array\index{input!array of input terminals} of input terminals\index{input terminal}. For image analysis this is the two-dimensional array of bits shown in Figure \figref{C7S2ConvNet}({\bf a}). Second, neurons 
are designed to detect local features of the image (such as edges or corners for instance).
The maps learned by such neurons, from inputs to output, are referred to as {\em feature maps}\index{feature map}.
Since these features occur in different parts of the image, one uses the same {\em kernel}\index{kernel} 
(or {\em filter}\index{filter})
for different parts of the image, always with the same weights and thresholds.
Since these kernels are local, and since they act in a 
translational-invariant way\index{translation invariance}, the number of
neurons from the two-dimensional input array is greatly reduced, compared with  fully connected
networks. Feature maps are obtained by convolution\index{convolution} of the kernel with the input image.  Therefore, layers consisting of a number of feature maps corresponding
to different kernels are also referred to as {\em convolution layers}\index{convolution layer}, Figure \figref{C7S2ConvNet}({\bf b}).

Convolutional networks can have many convolution layers. The idea
is that the additional layers can learn more abstract features
(Section \ref{sec:fr8}). Apart from feature maps, convolutional networks contain other types of layers.  
{\em Pooling layers}\index{pooling layer} perform local averages over the output of the convolution layers, to speed up learning by reducing the number of variables. 
Convolutional networks may also contain fully connected layers.\index{layer!fully connected}

\section{Convolution layers}
\label{sec:cl81}
\begin{figure}[t]
\centering
\begin{overpic}[scale=\myFigureScale]{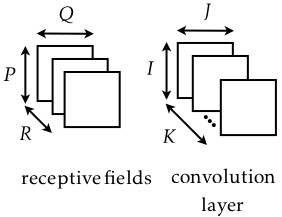}
\end{overpic}
\caption{\label{fig:conv_tensors} Illustration of summation in Equation (\ref{eq:tensormult}).
Each feature map has a receptive field of dimension $P\times Q \times R$. 
There are $K$ feature maps, each of dimension $I\times J$.}
\end{figure}

Figure \figref{C7S2ConvNet}({\bf a}) illustrates how a feature map is obtained by convolution 
of the input image with a kernel\index{kernel} which reads a $3\times 3$ part of the input image \cite{Nielsen}. 
In analogy with the terminology used in neuroscience, this $3\times 3$ array is called the 
{\em local receptive field}\index{receptive field|textbf} of the kernel. 
The outputs of the kernel from different parts of the input image make up the feature map\index{feature map}, 
here an $8\times 8$ array of hidden neurons: neuron $V_{11}$ connects to the $3\times 3$ area
in the upper left-hand corner of the input image.
Neuron $V_{12}$ connects to a shifted
area, as illustrated in Figure \figref{C7S2ConvNet}({\bf a}), and so forth. 
Since the input has $10\times 10$ pixels, the dimension
of the feature map is $8 \times 8$ in this example.
The important point is that the neurons $V_{11}$ and $V_{12}$, and all other neurons in this convolution layer, \index{convolution layer|textbf}
share their weights and the threshold. In the example shown in Figure \figref{C7S2ConvNet}({\bf a}) there are thus only nine independent weights, and one \index{threshold}threshold. Since the different hidden neurons share
\index{hidden neuron}
weights and thresholds, their computation rule is a discrete {\em convolution}\index{convolution} \cite{Goodfellow}:
\begin{equation}
\label{eq:feature_map}
V_{ij} = g\Big(\sum_{p=1}^3\sum_{q=1}^3 w_{pq}x_{p+i-1,q+j-1}-\theta\Big)\,.
\end{equation}
In Figure \figref{C7S2ConvNet}({\bf a}) the local receptive field is shifted by one pixel at a time. Sometimes it is useful
to use a larger {\em stride}\index{stride} $[s_1,s_2]$, 
to shift the \index{receptive field} receptive field by $s_1$ pixels horizontally and by $s_2$
pixels vertically. Also, the local
receptive regions need not have size $3\times 3$. 
If we assume that their size is $Q\times P$, and that $s_1=s_2=s$, 
the rule (\ref{eq:feature_map}) takes the form
\begin{equation}
V_{ij} = g\Big(\sum_{p=1}^P\sum_{q=1}^Q w_{pq}x_{p+s(i-1),q+s(j-1)}-\theta\Big)\,.  
\end{equation}
Figure \figref{C7S2ConvNet}({\bf a}) depicts a two-dimensional input array. For colour images there are three colour {\em channels}\index{colour channel},
in this case the input array is three-dimensional, and the input bits are labeled by three indices: two for position and the last one for colour, $x_{pqr}$. 
Usually one connects several feature maps with different kernels
to the input layer, as shown in Figure \figref{C7S2ConvNet}({\bf b}). The different
kernels detect different features of the input image, one detects edges for example, and another one detects corners, and so forth.
To account for these extra dimensions, one groups weights (and thresholds) into higher-dimensional arrays ({\em tensors}\index{tensor}). The convolution
takes the form:
\begin{equation}
\label{eq:tensormult}
V_{ijk} 
= g\Big(\sum_{p=1}^P\sum_{q=1}^Q \sum_{r=1}^R
w_{pqrk}
x_{p+s(i-1),q+s(j-1),r}-\theta_{k}\Big)
\end{equation}
(see Figure~\ref{fig:conv_tensors}). All neurons in a given convolution layer have the same \index{threshold}threshold. 
The software package {\em TensorFlow} \cite{tensorflow} is designed to efficiently perform tensor
operations as in Equation (\ref{eq:tensormult}).

If one couples several convolution layers together, the number of neurons in these layers decreases 
as one moves to the right. To avoid this, one can {\em pad}\index{padding} 
the image (and the convolution layers) by adding rows and columns of bits set to zero \cite{Goodfellow}. 
In Figure  \figref{C7S2ConvNet}({\bf a}), for example, one obtains a convolution
layer of the same dimension as the original image by adding one column each on the left-hand and right-hand sides
of the image, as well as two rows, one at the bottom and one at the top.
In general, the numbers of rows and columns need not be equal, 
so the amount of padding is specified by four numbers, $[p_1,p_2,p_3,p_4]$.

Convolution layers are trained with backpropagation. Consider the simplest case, Equation (\ref{eq:feature_map}). 
As usual, we use the chain rule to evaluate the gradients:
\begin{equation}
\frac{\partial V_{ij}}{\partial w_{mn}} =  g'(b_{ij}) \frac{\partial b_{ij}}{\partial w_{mn}}
\end{equation}
with local field \index{local field} $b_{ij}=\sum_{pq} w_{pq}x_{p+i-1,q+j-1}-\theta$.
The derivative of $b_{ij}$ is evaluated by applying rule (\ref{eq:Kronecker_w_1}):
\begin{align}
 \frac{\partial b_{ij}}{\partial w_{mn}}& = \sum_{pq} \delta_{mp} \delta_{nq} x_{p+i-1, q+j -1}
\end{align}
In this way one can train networks with several stacked convolution layers too. It is important to keep track
of the summation boundaries. To that end it helps to pad out the image and the convolution layers, so that the upper bounds remain the same in different layers.

Details aside, the fundamental principle of feature maps is that the map
is applied in the same form to different parts of the image ({\em translational invariance})\index{translation invariance|textbf}. In this way, each weight in a given feature map
is trained on different parts of the image. This effectively increases 
the \index{training set}training set for the feature map and combats overfitting\index{overfitting}.

\begin{figure}[t]
\centering
\begin{overpic}[scale=\myFigureScale]{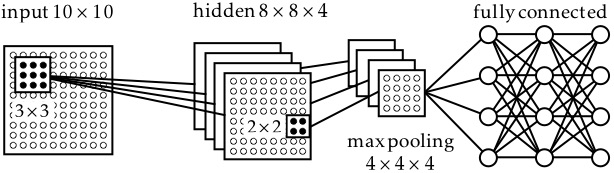}
\end{overpic}
\caption{\figlab{C7S2ConvNet1} Layout of a convolutional neural network for 
object recognition\index{object recognition} and image classification (schematic). The inputs are stored in a $10\times 10$ array. They feed into a convolution layer with four different feature maps with $3\times 3$ kernels, stride $[1,1]$, and zero padding.  Each convolution layer connects to a $2\times 2$ max-pooling layer
\index{pooling layer},
with stride $[2,2]$ and zero padding.
 Between these and the output layer are two fully connected
hidden layers.\index{layer!fully connected} After a figure in Ref.~\cite{Nielsen}. }
\end{figure}

\section{Pooling layers}
\index{pooling layer|(}
Pooling layers process the output of convolution layers. A neuron in a pooling layer takes the outputs of several neighbouring feature maps and compresses their outputs into a single number \cite{Nielsen}. There are no weights or \index{threshold}thresholds associated with the pooling layers. \index{pooling layer}{\em Max-pooling units},
\index{pooling layer!max pooling layer}
for example, take the maximum over several nearby feature-map outputs. Instead,
one may compute the root-mean square of the map values ($L_2$-{\em pooling})\index{pooling layer!L2 pooling layer}. 
Just as for convolution layers, we need to specify stride\index{stride} and padding \index{padding} for pooling layers.
Other ways of pooling are discussed in Ref.~\cite{Goodfellow}.

Usually several feature maps are connected to the input. Pooling is performed 
independently for each feature map \cite{Nielsen}. The network layout looks
like the one shown schematically in Figure \figref{C7S2ConvNet1}.  In this Figure, the pooling layers connect to a number of fully connected hidden layers\index{hidden layer} that feed into the output neurons\index{output neuron}. There are
as many output neurons\index{output neuron}  as there are classes to be recognised. This layout is similar to the layout used by Krizhesvky {\em et al.} \cite{Krizhevsky2012} in the \index{imagenet}ImageNet challenge  (Section \ref{sec:dlor}).
\index{pooling layer|)}

\section{Learning to read handwritten digits}
\index{MNIST data set|(}
\index{hidden neuron|(}
\index{hidden layer|(}
\label{sec:MNIST}
\begin{figure}[bt]
        \centering 
\begin{overpic}[scale=0.3]{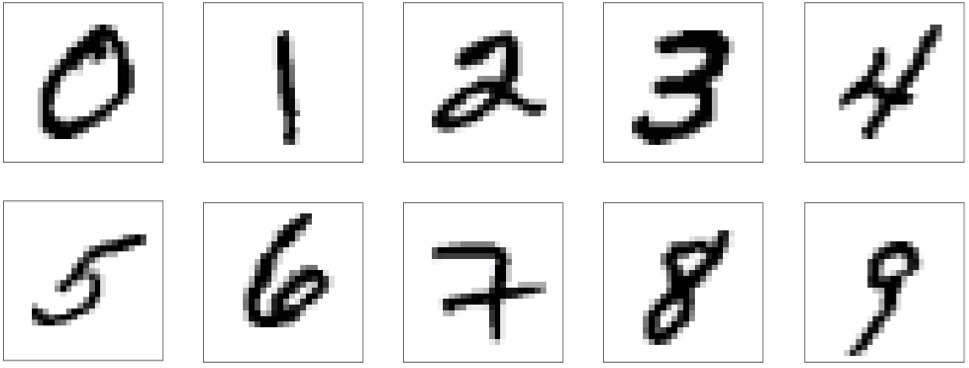} 
\end{overpic}
        \caption{  \figlab{C7S4MNIST} Examples of digits from the \href{http://yann.lecun.com/exdb/mnist/}{MNIST} data set of handwritten digits \cite{MNIST}.  The images were produced using \href{https://se.mathworks.com/products/matlab.html}{MATLAB}. 
Copyright for the data set: Y. LeCun and C. Cortes. }
\end{figure}
Figure \figref{C7S4MNIST} shows patterns from the \href{http://yann.lecun.com/exdb/mnist/}{MNIST} data set of handwritten digits \index{MNIST data set|textbf}\cite{MNIST}. The data set derives from a data set compiled by the National Institute of Standards and Technology (NIST), of digits handwritten by high-school students and employees of the United States Census Bureau.  The data  contains 60 000 images of digits, 
each with $28\times 28$ pixels, and a {\em test set}\index{test set|textbf} of 10 000 digits. The images  are grayscale with 8-bit resolution, so each pixel contains a value ranging from 0 to 255. The images in the database were preprocessed. \index{input!preprocessing}
The procedure is described on the \href{http://yann.lecun.com/exdb/mnist/}{MNIST} home page.  Each original binary image from the National Institute of Standards and Technology was  represented as a 20$\times$20 gray-scale image, preserving the aspect ratio of the digit.  The resulting image was placed in a 28$\times$28 image so that the centre-of-mass of the image coincided with its geometrical centre. 
These preprocessing steps improve the performance of the learning algorithm.

The goal of this Section is to show how the principles described up to
now allow neural networks to learn the \href{http://yann.lecun.com/exdb/mnist/}{MNIST} data
with low  classification error\index{classification error}, following Ref.~\cite{Nielsen}. As described in Chapter \chref{C6Chapter}, one divides the data set into a {\em training set}\index{training set}
and a {\em validation set}\index{validation!set}, here with 50 000 digits and 10 000 digits, respectively \cite{Nielsen}. The validation set is used for \index{cross validation}cross validation. The test \index{test data set} data set allows to measure the \index{classification error}classification error after training. For this purpose one must use a data set that was not involved in the training.
As described in Section \ref{sec:pprid}, the inputs are preprocessed further by subtracting the mean image averaged over the whole training set from each input image [Equation (\ref{eq:subtract_mean})].

To find good  parameter values and network layouts is one of the main difficulties
when training a neural network, and it usually requires a fair deal of experimenting. There are recipes for finding certain parameters \cite{Smith15}, but the general approach is still trial and error \cite{Nielsen}.
Consider first a network with one hidden layer with ReLU activation functions (Section \ref{sec:ReLU}),
and a softmax output layer (Section \ref{sec:ocf}) with ten outputs $O_i$ and 
energy function\index{energy function} (\ref{eq:loglikelihood}). Output $O_i$ is interpreted as the probability that the pattern fed to the network falls into category $i$.
The networks are trained with stochastic gradient descent with \index{momentum}momentum, Equation (\ref{eq:inertia}). 
The learning rate\index{learning rate} is set to $\eta=0.001$, and the \index{momentum!constant}momentum constant to $\alpha=0.9$.
The mini-batch size [Equation (\ref{eq:minibatch})] equals 8192.
\index{cross validation}Cross validation and early stopping is implemented as follows:
during \index{training}training, the algorithm keeps track of the smallest validation error observed so far. Training stops when the validation error becomes larger than the minimum for a
specified number of times, equal to $5$ in this case.

\index{classification!accuracy|(}
\begin{figure}[t]
\centering
\begin{overpic}[scale=\myFigureScale]{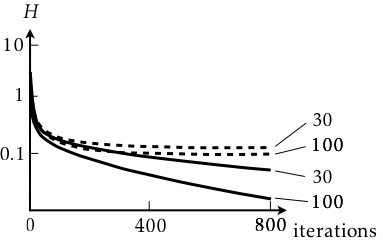}
\end{overpic}
\caption{\figlab{C7S4H}
Energy functions for the \href{http://yann.lecun.com/exdb/mnist/}{MNIST} 
\index{training set!MNIST}training set (solid lines)
and for the \index{validation!set}validation set (dashed lines) for a fully connected hidden layer with $30$ neurons, and for a similar algorithm, but with $100$  neurons
in the hidden layer. The data was smoothed and the plot is schematic. The $x$-axis shows iterations.
One iteration corresponds to feeding one minibatch of patterns. One epoch
consists of $50000/8192\approx 6$ iterations. Schematic, based on simulations
performed by Oleksandr Balabanov.}
\end{figure}
\begin{figure}
\centering
\begin{overpic}[scale=\myFigureScale]{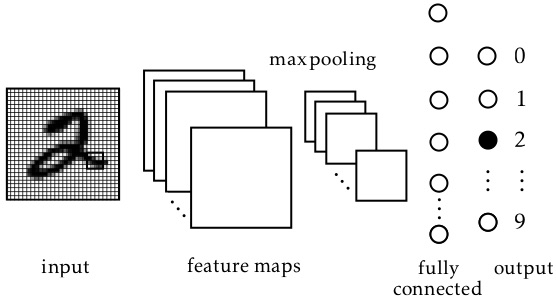}
\end{overpic}
\caption{\label{fig:convMNIST} Convolutional network that classifies the handwritten digits in the \href{http://yann.lecun.com/exdb/mnist/}{MNIST} data set (schematic).}
\end{figure}
\begin{figure}[b]
\centering
\begin{overpic}[scale=\myFigureScale]{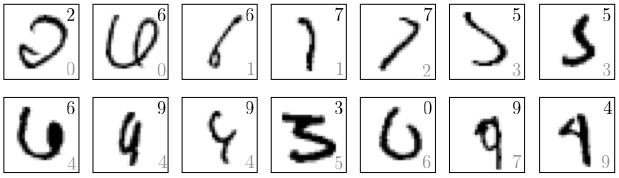}
\end{overpic}
\caption{\figlab{MNISTErrors} Some hand-written digits from the \href{http://yann.lecun.com/exdb/mnist/}{MNIST} \index{test set}test set, misclassified by a convolutional network that achieved an overall classification accuracy of 98\%. Target (top right),  network output (bottom right). 
Data from Oleksandr Balabanov. After a figure in Ref.~\cite{Nielsen}, see also Fig.~2({\bf c}) in Ref.~\cite{MNIST}.}
\end{figure}
Figure \figref{C7S4H} shows how the \index{training!energy}training and the 
\index{validation!energy}validation energies
decrease during training, for networks with $30$ and $100$ hidden neurons \cite{Nielsen}. 
One epoch\index{epoch}  corresponds to
applying $p$ patterns or $p/m_{\rm B}=50000/8192$ iterations (Section \ref{sec:chainrule}).
The energies
are a little lower for the network with $100$ hidden neurons. But one observes
overfitting\index{overfitting} in both cases: after many training steps
the validation energy is much higher than the \index{training!energy}training energy. 
Early stopping  caused the training of the larger network to abort after $135$ epochs, this corresponds to $824$ iterations.
The resulting classification accuracy is  about
97.2\%  for the network with 100 hidden neurons.

It is difficult to increase the classification accuracy by adding more hidden layers, most likely because
the network overfits the data (Section \ref{sec:crossvalidation}). This problem becomes more acute as one adds more hidden neurons. 
The tendency of the network to overfit is reduced by regularisation (Section \ref{sec:reg}). 
For the network with one hidden layer with 100 ReLU neurons, $L_2$-regularisation\index{regularisation!L2 regularisation} improves
the classification accuracy to almost 98\%.

Convolutional networks can be optimised to yield higher classification accuracies \index{classification!accuracy}
than those quoted above. A convolutional network with one convolution layer with 20 feature maps, a max-pooling layer\index{pooling layer}, and a fully connected hidden layer\index{layer!fully connected}\index{hidden layer} with 100 
ReLU neurons, similar to the network shown schematically in Figure \figref{convMNIST},
gives classification accuracy only slightly 
above 98\% after training for 60 epochs. 
Adding a second convolution layer and batch normalisation (Section \ref{sec:bn})
gives a classification accuracy is 98.99\% after 30 epochs (this layout is similar to 
a layout described in MathWorks \cite{MathWorks}). 
The accuracy can be improved further by tuning parameters and network layout, and by using \index{network!ensemble of networks|textbf} ensembles of convolutional neural networks \cite{MNIST}. The best classification accuracy found in this way is 99.77\% \cite{Schmidhuber}. 
Several of the  \href{http://yann.lecun.com/exdb/mnist/}{MNIST} digits are difficult to classify for humans too (Figure \figref{MNISTErrors}), 
so we conclude that convolutional networks really work very well. 

The above examples show also that it takes much experimenting to find the right parameters and network layout, as well as long training times to reach the best classification accuracies. It could be argued that one reaches a stage of {\em diminishing returns}\index{diminishing returns} as the classification
error falls below a fraction of a percent.
\index{classification!accuracy|)}
\index{hidden layer|)}
\index{hidden neuron|)}

\section{Coping with deformations of the input distribution}
\label{sec:pp}
How well does a  \href{http://yann.lecun.com/exdb/mnist/}{MNIST}-trained convolutional network 
classify your own hand-written digits?  
Figure \ref{fig:C7S4PreprocessedMNIST}({\bf a}) shows examples of digits drawn by colleagues
at the University of Gothenburg, 
preprocessed in the same way as the  \href{http://yann.lecun.com/exdb/mnist/}{MNIST} data.
\begin{figure}[t]
 \centering
\begin{overpic}[scale=\myFigureScale]{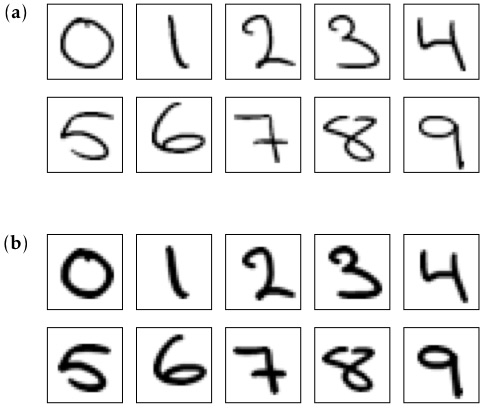}
\end{overpic}
\caption{\figlab{C7S4PreprocessedMNIST} ({\bf a}) Non-MNIST hand-written digits, preprocessed like the  \href{http://yann.lecun.com/exdb/mnist/}{MNIST} digits. {\bf b}) Same digits, except that the thickness of the stroke was normalised (see text). Data from Oleksandr Balabanov.}
\end{figure}
Using a \href{http://yann.lecun.com/exdb/mnist/}{MNIST}-trained convolutional 
network on these digits yields
a classification accuracy of about 90\%, substantially lower than the \index{classification error}classification errors
quoted in the previous Section. 

A possible cause  is that
the digits in Figure \figref{C7S4PreprocessedMNIST}({\bf a})  have a more slender stroke
than those in Figure \figref{C7S4MNIST}.
It was suggested in Ref.~\cite{preprocessing} that
differences in line thickness can confuse algorithms designed to read hand-written text \cite{Kozielski2012}.  There are different methods for normalising the line thickness of hand-written text. Applying the method proposed in Ref.~\cite{Kozielski2012} to our digits results in Figure  \figref{C7S4PreprocessedMNIST}({\bf b}). The
algorithm has a free parameter, $T$, 
that specifies the line thickness. In 
Figure  \figref{C7S4PreprocessedMNIST}({\bf b}) it was taken to be $T=10$, close to the average line thickness
of the \href{http://yann.lecun.com/exdb/mnist/}{MNIST} digits, which is
approximately $T\approx 9.7$.
If we run a \href{http://yann.lecun.com/exdb/mnist/}{MNIST}-trained convolutional network on a data set of 60 digits with normalised line thickness, it fails on only two digits. This corresponds to a classification accuracy of roughly $97\%$, not so bad -- yet not as good as the best results
in Section \ref{sec:MNIST}. 
But note that this estimate of the classification accuracy is
not very precise, because the test set had only 60 digits. To obtain a better  estimate, more test digits are needed. 

A question is of course whether
there are  other significant differences between our non-MNIST hand-written digits and those in the \href{http://yann.lecun.com/exdb/mnist/}{MNIST} data. 
At any rate, the results of this Section raise a point of fundamental importance. We have seen that convolutional networks can be trained to represent a distribution of input patterns\index{input pattern!distribution}  with very high accuracy.  But the network may not work as well on
a data set with a  slightly different input distribution, perhaps because the patterns were preprocessed differently, 
or because they were slightly deformed in other ways.
\index{MNIST data set|)}

\section{Deep learning for object recognition}
\index{object recognition|(}
\label{sec:dlor}
Deep learning\index{deep learning} has become so popular in the last few years because deep convolutional networks are good at recognising objects in images. Figure \figref{C7S5YoloIphone} shows a frame from a movie taken by a data-collection vehicle. A convolutional network was trained to recognise objects, and to localise them in the image by means of 
bounding boxes around the objects.
\begin{figure}[t]
\centering
        \begin{overpic}[scale=0.375]{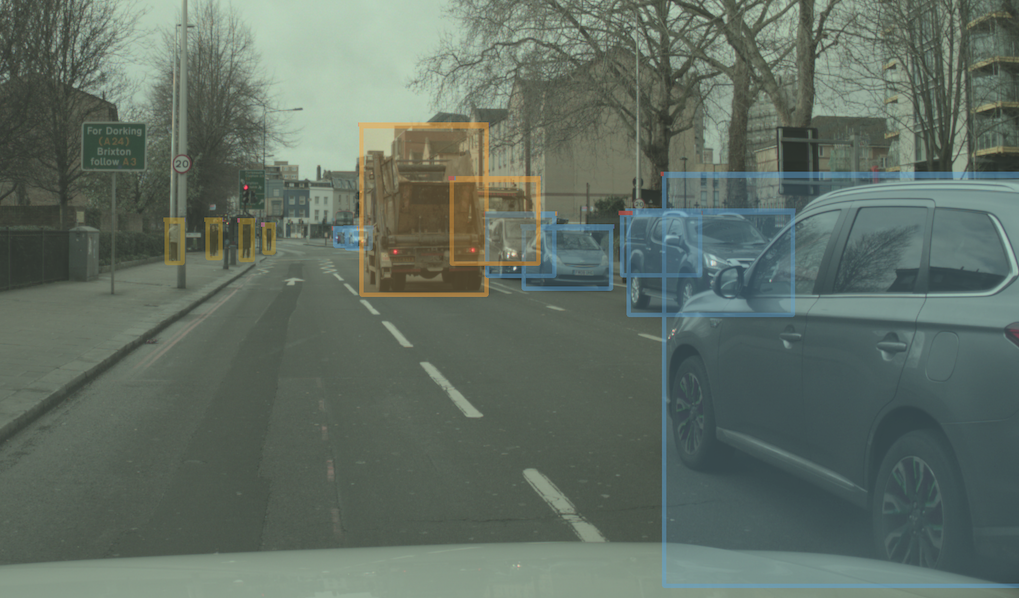}
        \end{overpic}
        \caption{\figlab{C7S5YoloIphone} 
Object recognition using a deep convolutional network. Shown is a frame 
from a movie recorded by a data-collection vehicle of the company \href{https://zenseact.com}{Zenseact}. The neural net recognises pedestrians, cars, and lorries, and localises them in the image by bounding boxes. 
Copyright \copyright \mbox{} Zenseact AB 2020. Reproduced with permission.}
\end{figure}

Convolutional networks excel at this task, as demonstrated by the  \index{imagenet}{\em ImageNet large-scale visual recognition challenge} (ILSVRC) \cite{Russakovsky2015}, a competition for object recognition and localisation in images, based upon the \href{www.image-net.org}{ImageNet}  \index{imagenet}database \cite{imagenet}. The challenge is based on a subset of  \index{imagenet}ImageNet. The \index{training set!image net}training set contains more than $10^6$ images manually classified into one of 1000 classes. There are approximately 1000 images for each class. The validation set\index{validation!set} contains 50 000 images. 

The ILSVRC challenge consists of several tasks. One task is {\em image classification}\index{image classification}, to list the object classes found in the image. A common measure for accuracy is the so-called {\em top-5 error}\index{top 5 error} for this classification task\index{classification!task}. The algorithm lists the five object classes it with     the
highest softmax outputs.
The result is defined to be correct if the \index{annotation}annotated class is among these five.  The error equals the fraction of incorrectly classified images. Why does one not simply judge whether the most probable class is the correct one?  The reason is that the images in the  \index{imagenet}ImageNet database are  \index{annotation}annotated by a single-class identifier. Often this is not unique. The image in Figure \figref{C7S5YoloIphone}, for example, shows not only a car but also trees, yet the image is  \index{annotation}annotated with the class label {\em car}. 
The resulting classification ambiguity is reduced by considering
the top five softmax outputs, and checking whether
the  \index{annotation}annotated class is among them.
\begin{figure}[t]
\centering
        \begin{overpic}[scale=\myFigureScale]{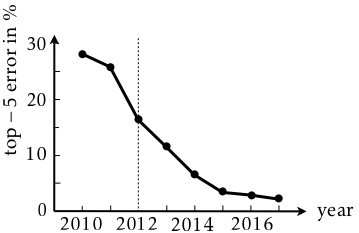}
        \end{overpic}
        \caption{\figlab{C7S5Top5} Smallest \index{classification error}classification error for the  \index{imagenet}ImageNet challenge \cite{Russakovsky2015}.
The data up to 2014 comes from Ref. \cite{Russakovsky2015}.
The data for 2015 comes from Ref.~\cite{He2015},
for 2016 from Ref.~\cite{CNNa},
and for 2017 from Ref.~\cite{HuShenSun2018}.
From 2012 onwards the smallest error was achieved by convolutional neural networks.
After Figure~1.12 in Goodfellow {\em et al.} \cite{Goodfellow}.
}
\end{figure}

The tasks in the ILSVRC challenge are significantly more difficult than the digit recognition described in Section \ref{sec:MNIST}.
One reason is that the  \index{imagenet}ImageNet classes are organised into a deep hierarchy of
subclasses. This results in highly specific sub classes that can be difficult to tell apart. The algorithm must
be very sensitive to small differences between similar sub classes. We say that the algorithm must have high {\em inter-class variability} \cite{Seif2018}.
Different images in the same sub class, on the other hand, may look quite different. The algorithm should nevertheless recognise them
as similar, belonging to the same class. In other words, the algorithm should have small {\em intra-class variability} \cite{Seif2018}. 

Since 2012, algorithms based on deep convolutional networks won the ILSVRC challenge. 
Figure  \figref{C7S5Top5} shows that the error has significantly decreased until 2017, the last year of the challenge in the form 
described above. We saw in previous  Sections that deep networks are difficult to train. So how can these algorithms work so well?
It is generally argued that the recent success of deep convolutional networks is mainly due to three factors.

\begin{figure}
\centering
\begin{overpic}[scale=\myFigureScale]{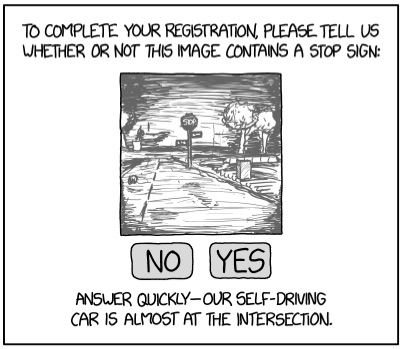}
\end{overpic}
\caption{\figlab{C7S5XKCD} Reproduced from \href{https://xkcd.com/1897}{xkcd.com/1897} under the creative commons attribution-noncommercial 2.5 license.}
\end{figure}

First, there are now much larger and better  \index{annotation}annotated \index{training set}training sets available.  \index{imagenet}ImageNet is an example. Excellent training data
is now recognised as one of the most important factors. Companies developing software for self-driving cars and
systems that help to avoid accidents understand that good training sets are indispensable. 
At the same time, it is a challenge to create high-quality training data, because one must {\em manually} collect and  \index{annotation}annotate the data (Figure \figref{C7S5XKCD}). This is costly, also because it is important to have as large data sets as possible, in order to reduce overfitting\index{overfitting}. In addition one must aim for a large variability in the collected data.
Second, the hardware is much better today. Deep networks are nowadays implemented on single or multiple \index{GPU}GPUs. There are even 
dedicated chips for this purpose.
Third, improved regularisation \index{regularisation} techniques (Section \ref{sec:reg}) help to fight overfitting\index{overfitting}, and skipping connections
(Section~\ref{sec:resnets}) render the networks
less susceptible to the vanishing-gradient problem\index{vanishing gradient} (Section \ref{sec:vg}).

The winning algorithm for 2012 was based on a network with five convolution layers and three fully connected layers, 
using {drop out}\index{drop out}, ReLU\index{ReLU function} activation functions, and data-set augmentation\index{data set augmentation} \cite{Krizhevsky2012}. The algorithm was implemented on \index{GPU}GPU processors.  The 2013 ILSVRC challenge was also won by a convolutional network \cite{GoogLeNet}, with 22 layers. Nevertheless, the network has substantially fewer free parameters (weights and thresholds) than the 2012 network: 4$\times$10$^6$ instead of 60 $\times$ 10$^6$. In 2015, the winning algorithm \cite{He2015} had 152 layers. One significant new element in the layout was the idea to allow connections that skip layers (Section \ref{sec:resnets}).  
The best algorithms in 2016 \cite{CUImage} and 2017 \cite{HuShenSun2018} 
used \index{network!ensemble of networks}ensembles of convolutional networks, where the classification is based on the ensemble average of the outputs.
\index{object recognition|)}

\section{Summary}
\label{sec:s8}
Convolutional networks can be trained to recognise objects in images with high accuracy. 
An advantage of convolutional networks is that they have fewer weights than fully connected networks with the same number of neurons, and that the weights of a given feature map are trained on different parts of the input images, effectively increasing the size of the \index{training set!size of}training set. This helps against overfitting. Another view is that the hidden neurons are forced to agree on a particular choice of weights, they must compromise. This yields a more robust training result.

It is sometimes stated that convolutional networks are now better than humans, in that they recognise objects with lower \index{classification error}classification errors than humans \cite{Guardian}.  This and similar statements refer to an experiment showing that the human \index{classification error}classification error in recognising objects in the  \index{imagenet}ImageNet database is about 5.1\% \cite{HumanError}, worse than the most recent convolutional neural-network algorithms (Figure \figref{C7S5Top5}).

This notion is not unproblematic, for several reasons. To begin with, the article  \cite{Guardian} refers to the 2015 ILSVRC competition, 
where the top scores were quite similar, and it has been debated whether interpreting the rules of the competition in different ways allowed competitors to gain an advantage.
Second, and more importantly, it is clear that these algorithms learn in quite a different way from humans. The algorithms can detect local
features, but since these convolutional networks rely on translational invariance\index{translation invariance},
they do not easily understand global features, and can mistake a leopard-patterned sofa for a leopard \cite{leopardsofa}.  It may help to include more leopard-patterned sofas in the training set, but the essential difficulty remains:  translational invariance imposes constraints on what convolutional networks can learn \cite{leopardsofa}.  More fundamentally one may argue that humans learn differently, by abstraction instead of going through very large training sets. 

We have also seen that convolutional networks are sensitive to small changes in the input data. 
Convolutional networks excel at learning the properties of a given \index{input distribution}input distribution, but they may have difficulties in recognising patterns sampled from a slightly different distribution, even if the two distributions appear to be very similar to the human eye.  Note also that this problem cannot be solved by \index{cross validation}cross validation, because training \index{training set}and \index{validation!set}validation sets are drawn from the same \index{input distribution}input distribution, but here we are concerned with what happens when the network is applied to a \index{input distribution}input distribution different from the one it was trained on.  

Here is another example illustrating this point: the authors of Ref.~\cite{Geirhos2018} trained a convolutional network on perturbed grayscale images from the  \index{imagenet}ImageNet data base, adding a little bit of noise independently to each pixel \index{white noise}({\em white noise}) before training. This network failed to recognise images that were weakly perturbed in a different way, by setting a small number of pixels to white or black. But when we look at the images we have no difficulties seeing through the noise.

Refs.~\cite{Szegedy,Nguyen} illustrate intriguing failures of convolutional networks \cite{Nielsen}. Sze\-ge\-dy {\em et al.} \cite{Szegedy} demonstrate that the way convolutional networks partition input space\index{input space} can lead to unexpected  results. The authors
took an image that the network classifies correctly with high confidence, and  perturbed it slightly. The perturbation was not random, but specifically designed to push the input pattern over a decision boundary.
The difference between the original and perturbed images ({\em adversarial images})\index{adversarial images} is undetectable to the human eye, yet the network misclassifies the perturbed image with high confidence \cite{Szegedy}. This reflects the fact  that decision boundaries\index{decision boundary}  are always close in 
high-dimensional input space\index{input space}.

Figure 1 in Ref.~\cite{Nguyen} shows images that are completely unrecognisable to the human eye. Yet a convolutional network classifies these images with high confidence. 
This illustrates that there is no telling what a network may do if the input is far away from the \index{training!distribution}training distribution.  Unfortunately the network can sometimes be highly confident yet wrong.                
Nevertheless, despite these problems, deep convolutional networks have enjoyed tremendous success in image classification during the past years, and they have found widespread use in industry and science.

Finally, the fundamental mechanisms of deep learning are quite well understood, but many open questions remain. It is fair to say that
the theory
of deep learning\index{deep learning} has somewhat lagged behind the associated practical successes, although some progress has been made in recent years.

\section{Further reading}
\label{sec:fr8}
The online book of Nielsen \cite{Nielsen} is an excellent introduction to convolutional neural networks, 
and guides the reader through all the steps required to program a convolutional network to recognise
hand-written digits. Nielsen's chapter {\em Deep Learning} \cite{Nielsen} is the main source for Section~\ref{sec:MNIST}.

What do the hidden layers in a convolutional layer actually compute? Feature maps that are directly coupled to the inputs detect local features, such as edges or corners. Yet it is unclear precisely how hidden convolutional layers help the network to learn. To which input features do the neurons of a certain hidden layer react most strongly? Input patterns chosen to maximise the outputs
of neurons in a given layer \cite{Yosinski,Graetz} reveal intricate geometric structures that defy straightforward interpretation. An example is shown on the
cover of this book, see
also Exercise 8.7.
\label{optimalpatterns}

It has been suggested that more general models, normally used for natural-language processing, 
may outperform convolutional nets in image-processing tasks when there is enough data \cite{doso}. An advantage is that these models do not rely on translational invariance, unlike convolutional networks.

\vfill\eject

\chapter{Supervised recurrent networks}
\label{ch:rn}
\index{supervised learning}
The layout of the perceptrons\index{perceptron} analysed in the previous Chapters is special. All connections are one way, and only to the layer to the right, 
so that the update rule for the $i$-th neuron in layer $\ell$ becomes, for example, 
\index{update rule}
\begin{equation}
\label{eq:update_recurrent}
V_i^{(\ell)} = g\Big (\sum_j w_{ij}^{(\ell)} V_j^{(\ell-1)}-\theta_i^{(\ell)}\Big)\,.
\end{equation}
The backpropagation\index{backpropagation} algorithm relies
on this {\em feed-forward}\index{feed forward network} layout. 
It means that the derivatives $\partial V_j^{(\ell-1)}/\partial w_{mn}^{(\ell)}$ vanish. This ensures that the outputs are nested functions of the inputs, which in turn implies the simple iterative structure of the backpropagation algorithm (Chapter \chref{C6Chapter}).

In some cases it is necessary or convenient to use networks that do not have this simple layout. The \index{Hopfield network}Hopfield networks discussed in part \ref{part:hopfield} are examples where all connections are symmetric.  More general networks may have a feed-forward layout with {\em feedbacks}\index{feedback|textbf}, as shown in  Figure \figref{C8S1RecBP}.
Such networks are called {\em recurrent networks}\index{recurrent network}.
There are many different ways in which the feedbacks can act: from the output layer 
to hidden neurons\index{hidden neuron} for example, or there could be connections between the neurons in a given layer. Neurons $3$ and $4$ in Figure \figref{C8S1RecBP} are
output neurons\index{output neuron}, they are associated with \index{target}targets just as in Chapters \chref{C5Chapter} to \ref{chap:dl}.
The layout of recurrent networks is very general, but because of the feedbacks we must consider how such networks can be trained.

Unlike multilayer perceptrons that   
represent an input-to-output mapping in terms of nested activation
functions, recurrent networks are used as
{\em dynamical networks}\index{recurrent network|textbf}, 
where the iteration index $t$ replaces the layer index $\ell$:
\begin{equation}
\label{eq:dyndis}
V_i(t) = g\Big(\sum_j w_{ij}^{(vv)} V_j(t-1) + \sum_k w_{ik}^{(vx)}x_k-\theta^{(v)}_i\Big)\quad\mbox{for}\quad t=1,2,\ldots\,.
\end{equation}
See Figure \figref{C8S1RecBP} for the definition of the different weights, 
and the parameters $\theta_i^{(v)}$ are \index{threshold}thresholds. 
Equation (\ref{eq:dyndis}) is analogous to the deterministic McCulloch-Pitts
dynamics of Hopfield networks\index{Hopfield network} and Boltzmann machines\index{Boltzmann machine} [c.f. Equation (\ref{eq:g})]. As in the case of Hopfield networks
(Exercise~2.10), one may also consider
a continuous network dynamics\index{network!dynamics}:
\index{update rule!continuous}
\begin{equation}
\label{eq:dync}
\tau \frac{{\rm d}V_i}{{\rm d}t} = -V_i + g\Big(\sum_j w_{ij}^{(vv)} V_j(t) + \sum_k w_{ik}^{(vx)}x_k-\theta^{(v)}_i\Big)\,,
\end{equation}
with time constant $\tau$. \index{time constant} We shall see in a moment why it is convenient
to assume that the dynamics is continuous in $t$, as in Equation (\ref{eq:dync}).

Recurrent networks can learn in different ways. One possibility is
to use a \index{training set}training set of pairs $[\ve x^{(\mu)},\ve y^{(\mu)}]$ with $\mu=1,\ldots,p$.
To avoid confusion with the iteration index $t$, the \index{target}targets are denoted by $y$ in this Chapter.
One feeds a pattern from this set and runs the dynamics (\ref{eq:dyndis})
or (\ref{eq:dync}) for the given $\ve x\tomu$ until it reaches a steady state\index{steady state} $\ve V^\ast$
(if this does not happen, the training fails). Then one adjusts the weights
by one gradient-descent step using the energy function\index{energy function}
\begin{equation}
\label{eq:Hrecurrent}
    H = \frac{1}{2} \sum_{k} (E_{k}^\ast)^{2} \quad  \text{where } E_{k}^\ast = \begin{cases} y_k^{(\mu)}-V_k^\ast ~~ &\text{if } V_{k} \text{ is an output neuron\index{output neuron},} \\ 0 &\text{otherwise.} \end{cases}
\end{equation}
The asterisk in this Equation indicates that
all variables are evaluated in the steady state, at $\ve V=\ve V^\ast$.
Iterating these steps, one feeds another pattern $\ve x^{(\mu)}$, finds the steady state $\ve V^\ast$, adjusts the weights and so forth. 
Instead of defining the energy function\index{energy function} in terms of the mean-squared output errors\index{output error}, one could also use the negative log-likelihood function (\ref{eq:Hcee})\index{log likelihood}.
One continues to iterate these steps until the steady-state outputs yield the correct \index{target}targets for all 
input patterns. \index{input pattern}
This is reminiscent of the algorithms discussed in Chapters \chref{C5Chapter} to \ref{chap:dl}. 
\begin{figure}[bt!]
        \centering
        \begin{overpic}[scale=\myFigureScale]{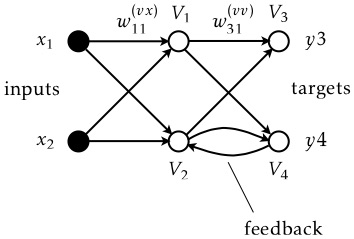}
        \end{overpic}
        \caption{\figlab{C8S1RecBP} Network with a feedback connection\index{feedback}.
Neurons $1$ and $2$ are \index{hidden neuron} hidden neurons. The weights from the input $x_k$ to
the neurons $V_i$ are denoted by $w_{ik}^{(vx)}$, the weight from
neuron $V_j$ to neuron $V_i$  is $w_{ij}^{(vv)}$.
Neurons $3$ and $4$ are output neurons\index{output neuron},
with prescribed \index{target}target values $y_i$. To avoid confusion with the iteration index
$t$, the \index{target}targets are denoted by $y$ in this Chapter.}
    \end{figure}

Another possibility is that inputs and \index{target}targets change as functions of time $t$ while the network dynamics runs. 
This allows to solve {\em temporal classification tasks}.\index{classification task, temporal} 
The network is trained on a set of input sequences  $\ve x(t)$ and corresponding target sequences $\ve y(t)$.
In this way, recurrent networks can translate 
written text or recognise speech.  
The network can be trained by unfolding its  dynamics in time as explained
in Section \ref{sec:bptt}, although this
algorithm suffers from the vanishing-gradient problem discussed in Chapter \ref{chap:dl}.
\index{vanishing gradient}

\section{Recurrent backpropagation}
\index{recurrent backpropagation|(}
\label{sec:rbp}
This Section summarises  how to generalise Algorithm \ref{bp_algorithm} to recurrent networks with feedback connections.\index{feedback}
Recall the recurrent network shown in Figure \figref{C8S1RecBP}. 
The neurons $V_{i}$  have 
smooth activation functions, and they are connected by
weights $w_{ij}^{(vv)}$.
Several neurons may be linked to inputs $x_{k}\tomu$,
with weights $w_{ik}^{(vx)}$.
Other neurons are output units\index{output unit} with associated \index{target}target values $y_i^{(\mu)}$.

One takes the dynamics to be continuous in time, Equation (\ref{eq:dync}), and assumes
that $\ve V(t)$ runs into a steady state\index{steady state}
\begin{equation}
\label{eq:steadystate}
\ve V(t) \to \ve V^\ast\quad\mbox{so that}\quad
    \frac{{\rm d}V^\ast_{i}}{{\rm d}t} = 0\,.
\end{equation}
Equation (\ref{eq:dync}) implies
\begin{equation}
    V_{i}^\ast = g\Big( \sum_{j}w_{ij}^{(vv)}V_{j}^\ast + \sum_k w_{ik}^{(vx)}x_k-\theta_i^{(v)}\Big)\,,
    \eqnlab{C12S2FixedPoint}
\end{equation}
and it is assumed that $\ve V^\ast$ is a {\em linearly stable} steady state\index{steady state} of the dynamics (\ref{eq:dync}), so that small perturbations 
$\delta\ve V$ away from $\ve V^\ast$ decay with time. 

The synchronous 
discrete dynamics (\ref{eq:dyndis}) can exhibit undesirable stable periodic solutions \cite{ott2002chaos}, as mentioned in Section \ref{sec:other:models}. This is a reason for using the continuous dynamics  (\ref{eq:dync}), yet 
\index{convergence!recurrent network}convergence to the steady state is not guaranteed in this case either.

Equation  \eqnref{C12S2FixedPoint} is a non-linear self-consistent condition
for the components of $\ve V^\ast$, in general difficult to solve. However, if the steady state $\ve V^\ast$ is stable,  we can use the 
dynamics (\ref{eq:dync}) to automatically pick out the steady-state solution $\ve V^\ast$.  
This solution depends on the pattern $\ve x^{(\mu)}$. Note that the superscript $(\mu)$ is left out in Equations (\ref{eq:steadystate}) and \eqnref{C12S2FixedPoint} and also in the 
remainder of this Section.

The goal is to find weights so that the outputs give the correct \index{target}target values in the steady state\index{steady state}.
To this end one uses gradient descent on the energy function\index{energy function} (\ref{eq:Hrecurrent}).  
Following Ref.~\cite{hertz1991introduction},  consider first how to adjust the weights $w_{ij}^{(vv)}$:
\begin{equation}
    \delta \!w_{mn}^{(vv)} = -\eta \frac{\partial H}{\partial w_{mn}^{(vv)}} 
= \eta \sum_{k} E_{k}^\ast \frac{\partial V_{k}^\ast}{\partial w_{mn}^{(vv)}}\,.
    \eqnlab{C12S2GradientDescent}
\end{equation}
One calculates the gradients of $\ve V^\ast$ by differentiating Equation \eqnref{C12S2FixedPoint}:
\begin{align}
    \frac{\partial V_{i}^\ast}{\partial w_{mn}^{(vv)}} 
=g'(b_{i}^\ast) \frac{\partial b_i^\ast}{\partial w_{mn}^{(vv)}}
= g'(b_{i}^\ast) \big( \delta_{im} V_{n}^\ast + \sum_{j} w_{ij}^{(vv)} \frac{\partial V_{j}^\ast}{\partial w_{mn}^{(vv)}}  \big)\,.
\label{eq:Vgradients}
\end{align}
Here $b_i^\ast = \sum_{j} w_{ij}^{(vv)} V_{j}^\ast + \sum_k w_{ik}^{(vx)} x_{k}-\theta_i^{(v)}$ 
is the local field in the steady state.  Equation (\ref{eq:Vgradients}) is a self-consistent equation for the gradient, as opposed to the explicit expressions we found in Chapter \chref{C6Chapter}. The reason for the difference
is that the recurrent network has feedbacks. 

Since Equation (\ref{eq:Vgradients}) is linear in the 
gradients, it can be solved by matrix inversion, at least formally. 
In terms of the  matrix $\ma L$ with elements
\begin{align}
    L_{ij} = \delta_{ij} - g'(b_{i}^\ast) w_{ij}^{(vv)}\,,
\end{align}
Equation (\ref{eq:Vgradients}) can be written as
\begin{equation}
\sum_{j} L_{ij} \frac{\partial V_{j}^\ast}{\partial w_{mn}^{(vv)}} = \delta_{im} g'(b_{i}^\ast) V_{n}^\ast     \,.
\end{equation}
If $\ma L$  is invertible, one applies $\sum_i \left(\ma L^{-1}\right)_{ki}$ to 
both sides. Using the fact that $\sum_i \left(\ma L^{-1}\right)_{ki}
L_{ij}= \delta_{kj}$  one finds:
\begin{equation}
\label{eq:sol_r1}
    \frac{\partial V_{k}^\ast}{\partial w_{mn}^{(vv)}} 
= \left(\ma L^{-1}\right)_{km} 
g'(b_{m}^\ast) V_{n}^\ast\,.
\end{equation}
Inserting this result into \eqnref{C12S2GradientDescent} one
obtains:
\begin{equation}
\label{eq:reclear}
    \delta \!w_{mn}^{(vv)} = \eta \sum_{k} E_{k}^\ast \,\left(\ma L^{-1}\right)_{km} g'(b_{m}^\ast)
V_{n}^\ast\,.
\end{equation}
This learning rule\index{learning rule} can be written in the form of the 
backpropagation rule \eqnref{C6S2GeneralRule} by introducing
the error\index{error}
\begin{equation}
\label{eq:Delta_m}
    \Delta_{m}^\ast = g'(b_{m}^\ast) \sum_{k} E_{k}^\ast \,\left(\ma L^{-1}\right)_{km}\,.
\end{equation}
Then the learning rule (\ref{eq:reclear}) takes the form
\begin{equation}
\label{eq:rlr}  
    \delta \!w_{mn}^{(vv)} = \eta \Delta_{m}^\ast V_{n}^\ast\,.
\end{equation}
If there are no recurrent connections, then $L_{ij} = \delta_{ij}$. In this case
Equation (\ref{eq:Delta_m}) reduces to the standard expression (\ref{eq:def_output_error}), Exercise 9.1.

The  learning rule for the weights $w_{mn}^{(vx)}$ is derived in an analogous fashion. The result is:
\begin{equation}
\label{eq:wvcrbp}
\delta\! w_{mn}^{(vx)} = \eta \Delta_m^\ast x_n\,.
\end{equation}
The learning rules (\ref{eq:rlr}) and (\ref{eq:wvcrbp}) are well-defined only if 
the matrix $\ma L$ is invertible.\index{matrix!inverse} Otherwise the solution \index{matrix!inverse}    (\ref{eq:sol_r1}) does not exist. Also, matrix inversion is an expensive operation.
As described in Chapter \chref{C5Chapter}, one can try to avoid the problem by finding the inverse iteratively. The trick \cite{hertz1991introduction} is to write down a dynamical equation for $\Delta_i$ that has a steady state\index{steady state} at the solution of Equation (\ref{eq:Delta_m}):
\begin{equation}
\label{eq:Delta_dyn}
    \tau \tfrac{{\rm d}}{{\rm d}t} \Delta_{j}= - \Delta_{j}  + g'(b_{j}^\ast) E_{j}^\ast+ \sum_{i} \Delta _i 
w_{ij}^{(vv)} g'(b_{j}^\ast) \,.
\end{equation}
It is left as an exercise (Exercise 9.2) to verify that the dynamics
(\ref{eq:Delta_dyn}) has a steady state satisfying Equation (\ref{eq:Delta_m}).
Equation (\ref{eq:Delta_dyn}) is written in a form to 
stress that (\ref{eq:Delta_dyn}) and (\ref{eq:dync})  exhibit the same {\em duality}\index{duality} as  Algorithm \ref{bp_algorithm}, between forward propagation of states of neurons and backpropagation of errors\index{error!backpropagation}. The sum in Equation (\ref{eq:Delta_dyn}) has the same form as the recursion for the errors in Algorithm \ref{bp_algorithm}, except that there are no layer indices $\ell$ here.  

Equation (\ref{eq:Delta_dyn}) admits
the steady state (\ref{eq:Delta_m}). But does $\Delta_i(t)$ converge to $\Delta_i^\ast$? For \index{convergence!recurrent backpropagation}convergence it is necessary that the 
steady state is linearly stable.
Whether or not this is the case is determined by {\em linear stability analysis}
\index{stability, linear|textbf} \cite{strogatz2000chaos}. One asks: does a small deviation from the steady state increase or decrease
under Equation (\ref{eq:Delta_dyn})? 
In other words, if one writes 
\begin{align}
    \ve V(t) & = \ve V^{*} + \delta\! \ve V(t) \quad\mbox{and}\quad
    \ve \Delta(t)  = \ve \Delta^{*} + \delta\! \ve \Delta(t)\,,
\end{align}
do $\delta\! \ve V(t)$ and $ \delta\! \ve \Delta(t)$ grow in magnitude?
To answer this question, one
inserts this {\em ansatz} into (\ref{eq:dync}) and (\ref{eq:Delta_dyn}),
and linearises:
\begin{subequations}
\begin{align}
\label{eq:ev_delta_1}
    \tau \tfrac{\rm d}{{\rm d}t} \delta\! V_{i}  = - ~ \delta\! V_{i} + g'(b_{i}^\ast) \sum_{j} w_{ij}^{(vv)} \delta\! V_{j} &\approx -\sum_{j}L_{ij} \delta\! V_{j} \,,\\
    \tau  \tfrac{\rm d}{{\rm d}t} \delta\!\Delta_{j} = -\delta\! \Delta_{j} 
+ \sum_{i}\delta\!\Delta_{i} w_{ij}^{(vv)} g'(b_{j}^\ast)  
&\approx -\sum_{i} \delta\! \Delta_{i} g'(b_i^\ast)L_{ij}/g'(b_j^\ast)\,.
\end{align}
\end{subequations}
Equation (\ref{eq:ev_delta_1}) shows: whether or not 
the norm of $\delta \ve V(t)$ grows is determined by the eigenvalues
of the matrix $\ma L$. We say that $\ve V^{\ast}$  is a linearly stable steady state of Equation (\ref{eq:dync}) if all eigenvalues of $\ma L$ have
negative real parts. In this case $|\delta \ve V(t)|\to 0$. If 
at least one eigenvalue has a positive real part then $|\delta\!\ve V|$ grows.
In this case we say that $\ve V^{\ast}$ is linearly unstable.
Since the matrix with elements 
$g'(b_i^\ast)L_{ij}/g'(b_j^\ast)$ has the same eigenvalues as $\ma L$, 
$\ve \Delta^{\ast}$ is a stable steady state of (\ref{eq:Delta_dyn})  if  $\ve V^{\ast}$ is a stable steady state  of (\ref{eq:dync}).  If the steady states are unstable, the algorithm does not converge.

In summary, recurrent backpropagation\index{recurrent backpropagation|textbf} is analogous to backpropagation  
(Algorithm \ref{bp_algorithm}) for layered feed-forward networks, 
save for two differences. First, the non-linear network dynamics\index{network!dynamics}  is no longer a
simple input-to-output mapping with  nested activation functions, but a non-linear dynamics that may (or may not) converge to a steady state.
Second, the feedbacks give rise to linear self-consistent equations for the 
\index{feedback}
steady-state gradients $\partial V_j^\ast/\partial w_{mn}$, which can be viewed
as steady-state conditions for a dual dynamics of the errors.\index{error}

The main conclusion of this Section is that \index{convergence!recurrent backpropagation}convergence of the \index{training set}training is not guaranteed if the network has feedback connections
(for a layered feed-forward network without feedbacks, 
recurrent backpropagation simplifies to stochastic gradient descent, Algorithm \ref{bp_algorithm}, see Exercise 9.1). This explains why 
stochastic gradient descent is used mostly for multi-layer networks with feed-forward layouts. The algorithm tends to fail for 
networks with feedbacks. However, it is possible to get rid of the 
feedbacks in recurrent networks by unfolding the dynamics in time.
\index{unfolding in time}
This is described in the next Section.
\index{recurrent backpropagation|)}

\section{Backpropagation through time}
\index{backpropagation!through time|(}
\label{sec:bptt}
Recurrent networks can be used to learn sequential inputs\index{input!sequential}, as in speech recognition\index{speech recognition} and
machine translation\index{machine translation}. The training set\index{training set} consists of time sequences $[\ve x(t),\ve y(t)]$ of
inputs and \index{target}targets. The network is trained on the sequences and learns
to predict the \index{target}targets. In this context the layout differs from the one 
described in the previous Section. There are two main differences. Firstly, the inputs and \index{target}targets
depend on $t$,  and one uses a discrete-time update rule\index{update rule!discrete}. Secondly, separate output neurons\index{output neuron}   
$O_i(t)$ are added to the layout. The update rule takes the form
\begin{subequations}
\label{eq:dyn_rnn_2}
\begin{align}
\label{eq:dyn_rnn_2_a}
V_i(t) &= g\Big(\sum_j w_{ij}^{(vv)} V_j(t-1) + \sum_k w_{ik}^{(vx)}x_k(t)-\theta_i^{(v)}\Big)\,,\\
O_i(t) &= g\Big(\sum_j w_{ij}^{(ov)} V_j(t) -\theta_i^{(o)}\Big)\,.
\label{eq:dyn_rnn_2_b}
\end{align}
\end{subequations}
The activation function of the output neurons\index{output neuron} $O_i$ can be different from that of the 
hidden neurons\index{hidden neuron}  $V_j$.\index{activation function}
One possibility is to use the softmax function for the outputs \cite{Sutskever,Lipton}\index{softmax}. For the hidden neurons one often uses tanh activations.
\index{activation function!tanh}
\begin{figure}[bt!]
 \centering
 \begin{overpic}[scale=\myFigureScale]{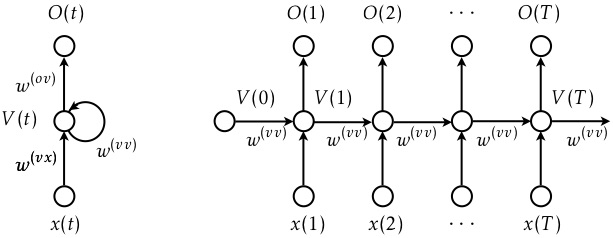}
\end{overpic}
\caption{\figlab{C8S2RNNunfold} Left: recurrent network with
one input terminal\index{input terminal}, one hidden neuron, and one output neuron\index{output neuron}. 
Right: same network but unfolded in time.\index{unfolding in time} 
The weights $w^{(vv)}$
remain unchanged as drawn, also the weights $w^{(vx)}$ and $w^{(ov)}$ remain unchanged (not drawn). After Figures~7 and 8 in Ref.~\cite{Lipton}.}
\end{figure}

To train recurrent networks with time-dependent inputs and
with the dynamics  (\ref{eq:dyn_rnn_2}), one uses
{\em backpropagation through time}\index{backpropagation!through time}. The idea is to unfold the network 
in time to get rid of the feedbacks. \index{feedback}
The price paid is that one obtains large networks in this way, with as many copies of the original neurons as there are time steps. 

The procedure is illustrated in  Figure \figref{C8S2RNNunfold}  for a recurrent network
with one hidden neuron\index{hidden neuron}, one input terminal\index{input terminal}, and one output neuron\index{output neuron}. The unfolded network has $T$ inputs and outputs. 
It can be trained in the usual way with 
{\em stochastic gradient descent}\index{stochastic gradient descent}. 
The \index{error!backpropagation} errors are calculated using backpropagation as in Algorithm \ref{bp_algorithm},
but here the error is propagated back in time, not from layer to layer.
The energy function is the squared error summed over all time steps
\index{energy function}
\begin{equation}
\label{eq:H_bptt}
H = \frac{1}{2} \sum_{t=1}^T E_t^2\quad\mbox{with}\quad E_t = y_t-O_t\,.
\end{equation}
One could use the negative log-likelihood function (\ref{eq:loglikelihood}),
but here we use the squared \index{output error} output-error function~(\ref{eq:H_bptt}).
There is only one hidden neuron in our example, and the inputs and outputs are
also one-dimensional. Here and in the following we write the time argument as a subscript, $O_t$ instead of $O(t)$ and so forth, because there is no risk of 
confusing the time index with other subscripts. 

Consider first how to adjust the weight $w^{(vv)}$. Gradient descent 
(\ref{eq:HReta0}) yields 
\begin{subequations}
\begin{equation}
\delta\!w^{(vv)}  
= \eta\sum_{t=1}^T E_t \frac{\partial O_t}{\partial w^{(vv)}} 
= \eta \sum_{t=1}^T \Delta_t  w^{(ov)} \frac{\partial V_t}{\partial w^{(vv)}}\,.
\label{eq:delta_w_t}
\end{equation} 
\end{subequations}
Here 
\begin{equation}
\Delta_t=E_t g'(B_t)
\end{equation}
is an output error,     \index{output error} and
$B_t=w^{(ov)}V_{t-1}-\theta^{(o)}$ is the \index{local field}local field of the output neuron\index{output neuron} at time $t$ [Equation (\ref{eq:dyn_rnn_2})]. Equation (\ref{eq:delta_w_t}) 
is similar to the learning rule for recurrent backpropagation,
Equations  \eqnref{C12S2GradientDescent} and (\ref{eq:Vgradients}),
but the derivative $\partial V_t/\partial w^{(vv)}$
is evaluated differently.  Equation (\ref{eq:dyn_rnn_2_a}) yields the recursion
\begin{equation}
\label{eq:Vg2}
\frac{\partial V_t}{\partial w^{(vv)}} =
g'(b_t) \,
\Big(V_{t-1}+w^{(vv)} \frac{\partial V_{t-1}}{\partial w^{(vv)}}\Big)
\end{equation}
for $t\geq 1$.
Since $\partial V_0/\partial w^{(vv)}=0$,
Equation (\ref{eq:Vg2}) implies:
\begin{align*}
\frac{\partial V_1}{\partial w^{(vv)}} &= g'(b_1) V_0\,,\\
\frac{\partial V_2}{\partial w^{(vv)}} &= g'(b_2) V_1+ g'(b_2) w^{(vv)} g'(b_1) V_0\,,\\
\frac{\partial V_3}{\partial w^{(vv)}} 
&=g'(b_3) V_2 + g'(b_3) w^{(vv)} g'(b_2) V_1 + g'(b_3)w^{(vv)}g'(b_2) w^{(vv)} g'(b_1) V_0\\
&\vdots\\
\frac{\partial V_{T-1}}{\partial w^{(vv)}} &= g'(b_{T-1}) V_{T-2} + g'(b_{T-1}) w^{(vv)} g'(b_{T-2}) V_{T-3}+\ldots\\
\frac{\partial V_T}{\partial w^{(vv)}} &= g'(b_T) V_{T-1} + g'(b_{T}) w^{(vv)} g'(b_{T-1}) V_{T-2}+\ldots
\end{align*}
Equation (\ref{eq:delta_w_t}) says that we must sum over $t$. Regrouping the terms in this sum yields:
\begin{align}
\nonumber
 & \Delta_1\frac{\partial V_1}{\partial w^{(vv)}} 
+ \Delta_2 \frac{\partial V_2}{\partial w^{(vv)}} 
+ \Delta_3 \frac{\partial V_3}{\partial w^{(vv)}}  + \hdots\\
&= [\Delta_1 g'(b_1) + \Delta_2 g'(b_2) w^{(vv)} g'(b_1) + \Delta_3 g'(b_3)w^{(vv)}g'(b_2) w^{(vv)} g'(b_1)+\ldots]V_0 \nonumber\\
&+[\Delta_2 g'(b_2) + \Delta_3 g'(b_3) w^{(vv)} g'(b_2) + \Delta_4 g'(b_4) w^{(vv)}g'(b_3) w^{(vv)} g'(b_2)+\ldots] V_1\nonumber \\
&+[\Delta_3 g'(b_3) + \Delta_4 g'(b_4) w^{(vv)} g'(b_3) + \Delta_5 g'(b_5) w^{(vv)} g'(b_4) w^{(vv)} g'(b_3)+\ldots] V_2 \nonumber\\
&\vdots\nonumber\\
&+ [\Delta_{T-1} g'(b_{T-1}) + \Delta_{T}g'(b_{T})w^{(vv)} g'(b_{T-1})] V_{T-2}\nonumber\\
&+ [\Delta_T g'(b_T)] V_{T-1}\,.\nonumber
\end{align}
To write the learning rule\index{learning rule} in the usual form, we define {\em errors}\index{error} $\delta_t$ recursively:
\begin{align}
\label{eq:delta_t_recursion}
\delta_{t} = 
\begin{cases}
\Delta_T  w^{(ov)} g'(b_T)\quad & \mbox{for $t=T$,}\\
\Delta_{t} w^{(ov)} g'(b_{t}) + \delta_{t+1}w^{(vv)} g'(b_{t}) & \mbox{for $0 < t< T$.}
\end{cases}
\end{align}
Then the learning rule (\ref{eq:delta_w_t}) takes the form
\begin{equation}
\label{eq:delta_wvv}
\delta w^{(vv)} = \eta \sum_{t=1}^T \delta_t V_{t-1}\,,
\end{equation}
just like Equation   \eqnref{C6S2Delta}, or like the error recursion
in Algorithm \ref{bp_algorithm}.
The factor $w^{(vv)}g'(b_{t-1})$ in the recursion (\ref{eq:delta_t_recursion})  gives 
rise to a product of many such factors in $\delta_t$ when $T$ is large, exactly as described
in Section \ref{sec:vg} for multilayer perceptrons. 
This means that the \index{training!recurrent networks}training of recurrent networks 
suffers from {\em unstable gradients}\index{unstable gradient},
as backpropagation of multilayer perceptrons does: if the factors $|w^{(vv)} g'(b_p)|$ are smaller than unity, then the errors $\delta_t$ become very small when $t$ becomes small ({\em vanishing-gradient problem}\index{vanishing gradient}).
This means that the early states of the hidden neuron\index{hidden neuron} no longer contribute to the learning, causing the network to forget what it has learned about early inputs. When $|w^{(vv)} g'(b_p)|>1$, on the other hand, exploding gradients make learning impossible.\index{exploding gradient}
In summary, {\em unstable gradients}\index{unstable gradient} in recurrent neural networks
occur much in the same way as in multilayer perceptrons. The resulting
difficulties for \index{training!recurrent networks}training recurrent neural networks are discussed in more detail in 
Section \ref{sec:LSTM}, see also Ref.~\cite{Pascanu2012}.

A slight variation of the above algorithm ({\em truncated backpropagation through time}) suffers less
from the exploding-gradient problem. The idea is that the exploding gradients are tamed by truncating the memory. 
This is achieved by limiting the \index{error!backpropagation} error propagation backwards in time, errors are computed back to $T-\tau$ and not further, where $\tau$ is the truncation time \cite{Haykin}.  Naturally this implies that 
\index{time correlation}long-time correlations cannot be learnt.

The learning rules for the weights $w^{(vx)}$ are obtained in a similar fashion.
Equation (\ref{eq:dyn_rnn_2_a}) yields the recursion 
\begin{equation}
\frac{\partial V_t}{\partial w^{(vx)}} =
g'(b_t) \,
\Big(x_{t}+w^{(vv)} \frac{\partial V_{t-1}}{\partial w^{(vx)}}\Big)\,.
\end{equation}
This looks just like Equation (\ref{eq:Vg2}), except that $V_{t-1}$ is replaced by $x_t$. As a consequence we have
\begin{equation}
\label{eq:delta_wvx}
\delta w^{(vx)} = \eta \sum_{t=1}^T \delta_t x_t\,.
\end{equation}
The learning rule  for  $w^{(ov)}$ is simpler to derive.
From Equation (\ref{eq:dyn_rnn_2_b}) we find by differentiation w.r.t. $w^{(ov)}$:
\begin{align}
\label{eq:delta_wox}
\delta w^{(ov)} &= \eta \sum_{t=1}^T E_t g'(B_t) V_t = \eta \sum_{t=1}^T \Delta_t V_t\,.
\end{align}
How are the \index{threshold}thresholds $\theta^{(v)}$ adjusted? Going through the above derivation we see that we must replace 
$V_{t-1}$ in Equation (\ref{eq:delta_wvv}) by $-1$. It works
in the same way for the output threshold. 
\begin{algorithm}[p]
\caption{\label{alg:bptt} backpropagation through time\index{backpropagation!through time|textbf}}
\begin{algorithmic}
\STATE initialise  weights $w_{mn}^{(vv)}$, $w_{mn}^{(vx)}$, $w_{mn}^{(ov)}$ and thresholds $\theta_m^{(v)}$, $\theta_m^{(o)}$;
\FOR {$\tau = 1,\ldots,\tau_{\rm max}$}
   \STATE choose input sequence $\ve x(1),\ldots,\ve x(T)$;
   \STATE initialise $V_j(0)=0$;
   \FOR {$t=1,\ldots,T$}
      \STATE  propagate forward:
      \STATE $b_i(t) \leftarrow \sum_j w_{ij}^{(vv)} V_j(t-1) + \sum_k w_{ik}^{(vx)}x_k(t)-\theta_i^{(v)}$ and $V_{i}(t) \leftarrow  g[b_i(t)]$;
      \STATE compute outputs:
      \STATE $B_i(t) \leftarrow  \sum_j w_{ij}^{(ov)} V_j(t) -\theta_i^{(o)}$ and $O_i(t) \leftarrow  g[B_i(t)]$;
   \ENDFOR
   \STATE compute errors for $t=T$ (targets $y_i$):
   \STATE $\Delta_i(T)\leftarrow [y_i-O_i(T)] g'[B_i(T)]$ and $\delta_{j}(T) \leftarrow  \sum_i \Delta_i(T) w_{ij}^{(ov)} g'[b_j(T)]$;
   \FOR{$t =T,\ldots,2$}
       \STATE propagate backwards:  $\Delta_i(t)= [y_i-O_i(t)]  g'[B_i(t)]$ and
   \STATE $\delta_{j}{(t-1)} \leftarrow  \sum_i \Delta^{(t)}_i w_{ij}^{(ov)} g'(b_j^{(t)}) +
\sum_i \delta_i^{(t+1)} w_{ij}^{(vv)} g'(b_j^{(t)})$;
     \ENDFOR
   \STATE $\delta w_{mn}^{(vv)}=0$, $\delta w_{mn}^{(vx)}=0$, $\delta w_{mn}^{(ov)}=0$, $\delta\theta^{(v)}=0$, $\delta\theta^{(o)}=0$;
\FOR{$t=1,\ldots,T$}
       \STATE $\delta w_{mn}^{(vv)}=\delta w_{mn}^{(vv)}+ \eta \delta_{m}(t) V_{n}(t-1)$;
       \STATE $\delta w_{mn}^{(vx)}=\delta w_{mn}^{(vx)}+ \eta \delta_{m}(t) x_{n}(t)$;
       \STATE $\delta w_{mn}^{(ov)}=\delta w_{mn}^{(ov)}+ \eta \Delta_m(t) V_n(t)$;
       \STATE $\delta \theta_{m}^{(v)}=\delta \theta_{m}^{(v)}- \eta \delta_m(t)$;
       \STATE $\delta \theta_{m}^{(o)}=\delta \theta_{m}^{(o)}- \eta \Delta_m(t)$;
   \ENDFOR
\STATE adjust weights and thresholds: $w_{mn}^{(vv)}=w_{mn}^{(vv)}+\delta w_{mn}^{(vv)},\ldots$;
\ENDFOR
\end{algorithmic}
\end{algorithm}

In order to keep the formulae simple, we derived the algorithm for a single hidden neuron\index{hidden neuron}, a single output neuron\index{output neuron}, and one-component inputs, so that we could leave out the indices referring to different
hidden neurons\index{hidden neuron}, and different input and  output components. 
If we consider several hidden and output neurons\index{output neuron}
and multi-dimensional inputs,
the structure of the Equations remains exactly the same, except for a number of extra sums
over those indices:
\begin{align}
\label{eq:delta_with_indices}
\delta\!w_{mn}^{(vv)} &= \eta \sum_{t=1}^T\delta_{m}^{(t)} V_{n}^{(t-1)}\\
\delta_{j}^{(t)} &= \begin{cases}
\sum_i \Delta^{(T)}_i w_{ij}^{(ov)} g'(b_j^{(T)}) \quad &\mbox{for $t=T$,}\\
\sum_i \Delta^{(t)}_i w_{ij}^{(ov)} g'(b_j^{(t)}) +
\sum_i \delta_i^{(t+1)} w_{ij}^{(vv)} g'(b_j^{(t)}) \quad &\mbox{for $0< t <T$.}
\nonumber
\end{cases}
\end{align}
The second term in the recursion for $\delta_j^{(t)}$ is analogous
to the error recursion in of Algorithm \ref{bp_algorithm}.
The time index $t$ here plays the role of the layer index $\ell$ in Algorithm \ref{bp_algorithm}.
A difference is that the weights in Equation (\ref{eq:delta_with_indices}) are the same for all time steps.
The algorithm is summarised in Algorithm \ref{alg:bptt}.

In conclusion we see that backpropagation through time for recurrent networks 
is similar to backpropagation for multilayer perceptrons. After the recurrent network is  unfolded to get rid of the feedback connections, it can be trained by backpropagation. The time index $t$ takes the role of the layer index $\ell$. Backpropagation through time is the standard approach for \index{training!recurrent networks}training recurrent networks, despite the fact that it suffers from the vanishing-gradient problem. The next Section describes how improvements to the layout make it possible to more
efficiently train  recurrent networks.
\index{backpropagation!through time|)}

\section{Vanishing gradients}
\index{vanishing gradient|(}
\label{sec:LSTM}
Hoch\-rei\-ter and Schmidhuber \cite{Hochreiter1997} suggested to replace the 
hidden neurons\index{hidden neuron} of the recurrent network with computation units that are specially designed to reduce the vanishing-gradient problem. The method is referred to as {\em long-short-term memory}\index{long short term memory} 
(LSTM).  The basic ingredient is the same as in {\em residual networks}\index{residual network} (Section \ref{sec:resnets}): short cuts reduce the vanishing-gradient problem. For our purposes we can think of LSTMs as  units that replace the hidden neurons. For a detailed description of LSTMs see Ref.~\cite{olah2015understanding}.

\begin{figure}[bt]
\centering
\begin{overpic}[scale=\myFigureScale]{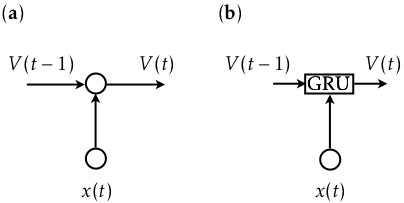}
\end{overpic}
\caption{\label{fig:GRU} Gated recurrent unit\index{unit!gated recurrent}.
({\bf a}) The symbol refers to the standard recursion (\ref{eq:dyn_rnn_2_a}) for the hidden variable,
as in the right panel of Figure \figref{C8S2RNNunfold}. ({\bf b})
To combat the vanishing-gradient problem, the standard unit is replaced by a gated recurrent
unit (\ref{eq:GRU})\index{unit!gated recurrent}.}
\end{figure}
{\em Gated recurrent units}\index{unit!gated recurrent|textbf}~\cite{cho2014learning} serve the same purpose as LSTMs, and they function
in a similar way. It has been argued that LSTMs outperform gated recurrent units for certain tasks, 
but since they are simpler than LSTMs, the remainder of this Section focuses on gated recurrent units.
As illustrated in Figure \ref{fig:GRU}, these units replace the hidden neurons of a recurrent neural network 
[with update rule (\ref{eq:dyn_rnn_2_a})] by a new rule:
\index{update rule}
\begin{subequations}
\label{eq:GRU}
\begin{align}
z_m(t) &= \sigma\big(\sum_k w_{mk}^{(zx)} x_k(t) + \sum_jw_{mj}^{(zv)} V_j(t-1)\big)\,,\\
r_n(t) &= \sigma\big(\sum_k w_{nk}^{(rx)} x_k(t) + \sum_jw_{nj}^{(rv)} V_j(t-1)\big)\,,\\
h_i(t) &= g\big(\sum_k w_{ik}^{(hx)} x_k(t)+ \sum_j w_{ij}^{(hv)} r_j(t) V_j(t-1)\big)\,,\\
V_i(t) &= [1-z_i(t)]h_i(t)+z_i(t)V_i(t-1)\,.
\end{align}
\end{subequations}
The first two Equations are referred to as {\em gates}\index{gate} because they regulate
how the values of the hidden state variables $V_i$ are passed through the unit.
Here $\sigma(b)$ is the sigmoid function (\ref{eq:sigmoid}).\index{activation function!sigmoid}
If $z_m(t)=0$  for all $m$, and $r_n(t)=1$ for all $n$, Equation (\ref{eq:GRU}) coincides
with the standard update rule  (\ref{eq:dyn_rnn_2_a}), save for the \index{threshold}thresholds
which were left out in Equation (\ref{eq:GRU}). As explained above, the 
resulting recurrent network suffers from the vanishing-gradient problem. This means that 
states in the past history, $\ve V(0), \ve V(1),\ldots,$ have little effect upon 
the present state $\ve V(t)$ for $t\gg 1$. The recurrent network forgets
early inputs, so that it cannot learn from them.  

If by contrast $z_m(t)=1$ for all $m$, then the input is passed right through the unit.
Since $\partial V_i(t)/\partial V_j(t-1)=\delta_{ij}$ in this case, the gradients do not decrease as the dynamics explores
the history. However, since $\ve V(t-1)=\ve V(t)$, the recurrent network reproduces
previous states. This is analogous to skipping layers in a residual network,
although comparison with Equation (\ref{eq:residual_net}) reveals some differences in detail.

For the recurrent network to learn in a meaningful way from past inputs, the weights 
in Equation (\ref{eq:GRU}) (and the \index{threshold}thresholds) are adjusted so that the gated recurrent unit operates 
between these two extreme limits. This is achieved by including the weights and thresholds
of the gated recurrent unit in the gradient-descent minimisation 
\index{gradient descent} of the energy function (\ref{eq:H_bptt}).
The learning rules for the weights (and thresholds) are calculated in the same as before. Using Equation (\ref{eq:dyn_rnn_2_b})
one has:
\begin{equation}
\delta\!w_{mn}^{(ab)} = \eta \sum_{t=1}^T\sum_i \Delta_i(t)  w_{ij}^{(ov)} \frac{\partial V_j(t)}{\partial w_{mn}^{(ab)}}\,,
\end{equation}
where $w^{(ab)}$ stands for $w^{(zx)}, w^{(zv)}, w^{(rx)},\ldots$. The 
derivatives $\partial V_j(t)/\partial w_{mn}^{(ab)}$ are evaluated using 
the chain rule and Equations (\ref{eq:GRU}). 

It is instructive to inspect the values of $z_i(t)$ and $r_i(t)$ when the recurrent 
network operates
after \index{training}training. Suppose that a unit assumes small values of 
$z_{i}(t)$ and $r_{j}(t)$. 
This means that the update of the state variable $V_i(t)$ is determined entirely by
the instantaneous inputs $x_k(t)$. Since the unit does not refer to the past history of the hidden-state
variables, it truncates the dynamical memory. In the opposite limit, when $z_i(t)\approx r_i(t)\approx 1$,
the unit can contribute to building up long-term dynamical memory.
These arguments suggest that a unit with just one gate may achieve the same goal \cite{heck2017simplified}:
\begin{subequations}
\label{eq:GRU_mod}
\begin{align}
z_m(t) &= \sigma\big(\sum_k w_{mk}^{(zx)} x_k(t) + \sum_jw_{mj}^{(zv)} V_j(t-1)\big)\,,\\
h_i(t) &= g\big(\sum_k w_{ik}^{(hx)} x_k(t)+ \sum_j w_{ij}^{(hv)} z_j(t) V_j(t-1)\big)\,,\\
V_i(t) &= (1-z_i)h_i(t)+z_i(t)V_i(t-1)\,.
\end{align}
\end{subequations}
This unit is easier to train because it has fewer parameters than the standard gated recurrent unit (\ref{eq:GRU}).
Yet, the additional parameters in Equation (\ref{eq:GRU}) may help to represent and exploit correlations on different time scales.
LSTMs have even more parameters. How this tradeoff between ease of training and accurate representation of \index{time correlation}time correlations works out may well depend on the problem at hand. In the following Section we describe recurrent networks
with LSTM units, following Refs.~\cite{Sutskever,Lipton}.
\index{vanishing gradient|)}

\section{Recurrent networks for machine translation}
\begin{figure}[p]
\centering
\begin{overpic}[scale=\myFigureScale,angle=90]{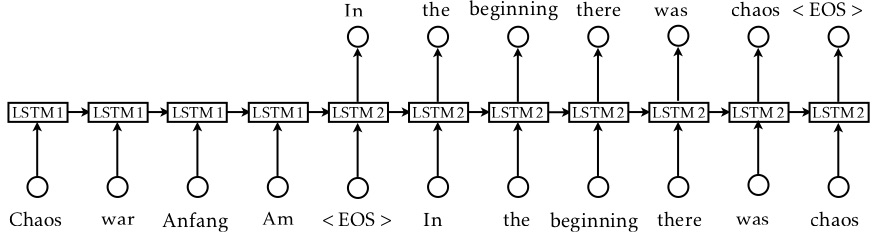}
\end{overpic}
\caption{\figlab{C8S3Translate} Schematic illustration of unfolded recurrent network for machine translation. 
The rectangular boxes represent the hidden states in the form
of long-short-term memory \index{long short term memory}(LSTM) units, see Section \secref{LSTM}. 
Sutskever {\em et al.} \cite{Sutskever} found that the network
translates much better if the \index{machine translation!sentence}sentence is read in reverse order,\index{machine translation}
from the end. The tag <EOS> denotes the \index{machine translation!end of sentence tag}end-of-sentence tag. \index{machine translation!end of sentence tag} Here it
denotes the beginning of the \index{machine translation!sentence}sentence. After Figure~1 in Ref.~\cite{Sutskever}.}
\end{figure}
Recurrent networks are used for machine translation \cite{Sutskever,Lipton}. How does this
work?  The networks are trained using backpropagation through time. The vanishing-gradient problem 
is dealt with by using LSTMs (Section \ref{sec:LSTM}).

How are the network inputs and outputs coded?  For machine translation one respresents all words in a given 
dictionary in terms of a code. The conceptually simplest code\index{machine translation!code}
is one where 100\ldots represents the first word in the dictionary,
010\ldots the second word, and so forth. The drawback of this scheme is that it does not account for the fact that two given words might be more or less closely related to each other.  Other encoding schemes are described in Ref.~\cite{Lipton}.

Each input to the recurrent network is a vector with as many
components as there are words in the dictionary. A \index{machine translation!sentence}sentence
corresponds to a sequence $\ve x_1,\ve x_2,\ldots,\ve x_T$. 
Each \index{machine translation!sentence}sentence ends with an \index{machine translation!end of sentence tag}end-of-sentence tag, <EOS>.
This tag tells the network when the input sentence ends. This is necessary because the number of words per sentence is not fixed. Now suppose that a possible translation reads $\ve x_1^\prime, \ve x_2^\prime,\ldots,\ve x_{T'}^\prime$. The task of the network is to determine the probability
$p(\ve x_1^\prime,\ldots,\ve x_{T'}^\prime|\ve x_1,\ldots\ve x_T)$ that the translation is correct. The idea is to estimate this probability recursively as
\begin{equation}
\label{eq:933}
p(\ve x_1^\prime,\ldots,\ve x_{T'}^\prime|\ve x_1,\ldots\ve x_T)
=\prod_{t=1}^{T'} p(\ve x_t^\prime|\ve x_1^\prime,\ldots,\ve x_{t-1}^\prime)\,.
\end{equation}
Sutskever {\em et al.} \cite{Sutskever} describe how to achieve this with a recurrent network with two hidden LSTMs. The network uses softmax outputs $\ve O_{\!t}$, where $j$-th component of $\ve O_{\!t}$ is interpreted as the probability that the $j$-th component of $\ve x_t'$ is the correct word at position $t$ in the translated sentence. As shown in Figure \ref{fig:C8S3Translate}, the first LSTM processes the input sentence
$\ve x_1,\ldots,\ve x_T$, encoding its contents in the hidden states. When the <EOS> tag appears, the second LSTM takes over, using the information
encoded by the first LSTM as an input. The second LSTM recursively outputs the translated sentence word by word, using Equation (\ref{eq:933}).

There is a large number of recent papers on machine translation with recurrent neural networks. Most studies    are based on the \index{training!algorithm}training algorithm described in Section \ref{sec:bptt}, backpropagation through time. 
The different approaches mainly differ in their network layouts. Google's machine translation system uses a deep network with 
several layers of hidden units arranged in a bidrectional layout \cite{Wu2016}.\index{deep network}
In such bidirectional networks, different hidden units are unfolded forward as well as backwards in time, as shown schematically in Figure \figref{C8S3BiDirectional}.\index{unfolding in time}
\begin{figure}[t]
\centering
\begin{overpic}[scale=\myFigureScale]{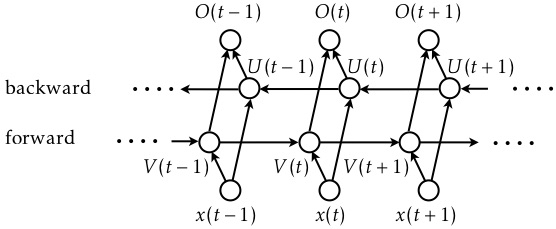}
\end{overpic}
\caption{\figlab{C8S3BiDirectional} Schematic illustration of a bidirectional recurrent network
\index{recurrent network!bidirectional}. The network consists of two hidden neurons, $U(t)$ and $V(t)$, that are unfolded in
different ways. 
\index{recurrent network}
After Figure~12 in Ref.~\cite{Lipton}.}
\end{figure}
For several hidden and output neurons and multidimensional inputs,
the bidirectional network has the dynamics
\begin{align}
\nonumber             
V_i(t) &= g\Big(\sum_j w_{ij}^{(vv)} V_j(t-1) + \sum_k w_{ik}^{(vx)}x_k(t)-\theta^{(v)}_i\Big)\,,\\
\label{eq:dyn_rnn_3}
U_i(t) &= g\Big(\sum_j w_{ij}^{(uu)} U_j(t+1) + \sum_k w_{ik}^{(ux)}x_k(t)-\theta^{(u)}_i\Big)\,,\\
O_i(t) &= g\Big(\sum_j w_{ij}^{(ov)} V_j(t) + \sum_j w_{ij}^{(ou)} U_j(t)-\theta^{(o)}_i\Big)\,.
\nonumber
\end{align}
It is natural to use bidirectional networks for machine translation because correlations go either way in a \index{machine translation!sentence}sentence,
forward and backwards. In German, for example, the finite verb form is usually at the end of the \index{machine translation!sentence}sentence.

Different schemes for scoring the accuracy of a translation are described by Lipton {\em et al.} \cite{Lipton}.
One difficulty is that there are often several different valid translations of a given \index{machine translation!sentence}sentence, and the score
must compare the machine translation with all of them. More recent papers on machine translation usually use the so-called BLEU score to evaluate the translation accuracy.
The acronym stands for {\em bilingual evaluation understudy}\index{machine translation!bilingual evaluation understudy}. The scheme was proposed
by Papieni {\em et al.} \cite{Papieni2002}. It is argued to evaluate the accuracy of a translation
not too  differently from humans.

\section{Reservoir computing}
\index{reservoir computing|(}
\label{sec:srcomp}
\begin{figure}[t]
 \centering
 \begin{overpic}[scale=\myFigureScale]{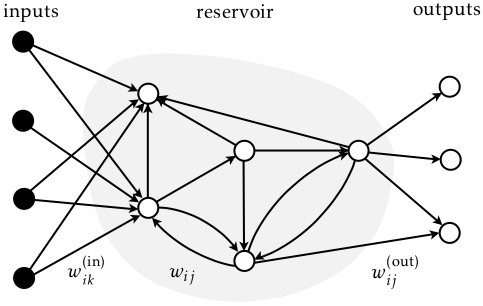}
\end{overpic}
\caption{\label{fig:reservoir} Reservoir computing (schematic). Not all connections are drawn.
There can be connections from all inputs to all neurons in the reservoir (gray), and from
all reservoir neurons to all output neurons\index{output neuron}.} 
\end{figure}
An alternative to backpropagation through time  for 
recurrent networks is {\em reservoir computing}\index{reservoir computing|textbf} \cite{Lukosevicius}.
This method has been used with success to predict chaotic dynamics \cite{pathak2018model,jaeger04harnessing} and rare transitions in stochastic bi-stable systems \cite{lim2019predicting}. 

Consider input data in the form of a \index{time series} time series $\ve x(0),\ldots\ve x(T-1)$ 
of $N$-dimensional vectors $\ve x(t)$, and a corresponding series
of $M$-dimensional \index{target}targets $\ve y(t)$. The goal is to train the recurrent
network so that its outputs $\ve O(t)$ approximate the \index{target}targets as precisely as possible, by minimising the \index{energy function} energy function $H = \tfrac{1}{2}\sum_{t=\tau}^{T-1} \sum_{i=1}^{{M}}[E_i(t)]^2$, where $E_i(t) =y_i(t)-O_i(t)$ is the output error, and\index{output error}
$\tau$ represents an initial transient that is disregarded.\index{transient}

Figure \ref{fig:reservoir} shows the layout for this task. There are
$N$ input terminals\index{input terminal}. They are connected with weights
\index{hidden neuron}\index{recurrent network}
$w_{jk}^{({\rm in})}$ to a reservoir of hidden neurons\index{hidden neuron} with state variables $r_j(t)$. The reservoir is linked to $M$ linear output units\index{unit!linear} $O_i(t)$ with weights  $w_{ij}^{({\rm out})}$. The reservoir itself is a large recurrent network with weights $w_{ij}$. The update rule\index{update rule} is similar to Equation (\ref{eq:dyn_rnn_2}).
There are many different versions that differ in detail 
\cite{lukosevicius2012practical}. 
One possibility is \cite{lim2019predicting}
\begin{subequations}
\label{eq:RCD}
\begin{align}
\label{eq:res_dyn}
r_i(t+1) &=  g\Big(\sum_j w_{ij} r_j(t) +\sum_{k=1}^N w^{({\rm in})}_{ik} x_k(t)\Big)\,,\\
O_i(t+1) &= \sum_{{j}} w_{ij}^{({\rm out})} r_j(t+1)\,,
\end{align}
\end{subequations}
for $t=0,\ldots,T-1$ with initial conditions $r_j(0)$.

The main difference to the \index{training!algorithm}training algorithms described in the previous Sections of this Chapter is that the input weights $w_{jk}^{({\rm in})}$ and the reservoir weights $w_{jk}$ are randomly initialised and 
then kept constant. Only the output weights $w_{jk}^{({\rm out})}$ are trained.  The idea is that the dynamics of a sufficiently large reservoir finds nonlinear, high-dimensional representations of the input data \cite{Lukosevicius}, \index{representation!nonlinear}\index{representation!high dimensional}
not unlike sparse\index{sparse} representations\index{representation!sparse} of binary classification problems\index{classification!problem, binary}
 embedded in a high-dimensional space
that become linearly separable in this way (Section \ref{sec:ct}).  

In addition, and this is a difference to the problem described in 
Section \ref{sec:ct}, the reservoir is  a dynamical memory. 
\index{memory, dynamical}
This requires that the reservoir states faithfully 
represent the input sequence: similar
input sequences should yield similar reservoir activations, provided one iterates it long enough.
However, for random weights the recurrent reservoir dynamics can be chaotic \cite{strogatz2000chaos}. 
In this case, the state of the reservoir after many iterations bears no relation to the input sequence. 
To avoid this, one requires that the reservoir dynamics is linearly stable. 
\index{stability, linear}
Linearising\index{dynamics!linearised} the reservoir dynamics (\ref{eq:res_dyn}) gives 
\begin{equation}
\label{eq:ls19}
\delta\ve r(t+1) = \ma D(t+1)\ma W \delta \ve r(t)\,,
\end{equation}
where $\ma D(t+1)$ is a diagonal matrix with entries $D_{ii}(t+1) = g'[b_i(t+1)]${, and where $b_i(t+1)= \sum_j w_{ij} r_j(t)+\sum_k w_{ik}^{({\rm in})} x_k(t)$.} Whether or not $\delta \ve r$ grows is then determined by 
the singular values of $\ma J_t = \ma D({t})\ma W \ma D({t-1}) \ma W \ma D(1) \ma W$,
as in Section \ref{sec:vg}. The singular values of $\ma J_t$ are denoted  by $\Lambda_1({n}) \geq \Lambda_2({n})\geq \cdots$. At large times, when driven with a stationary input series, the maximal Lyapunov exponent  $\lambda_1=\lim_{{t}\to\infty}{t}^{-1}\log \Lambda_1({t})$
must be negative to ensure that the reservoir dynamics is stable:
\begin{equation}
\label{eq:echo}
\lambda_1 {<} 0 \,.
\end{equation}
Sometimes the stability criterion is quoted in terms
of the maximal eigenvalue of $\ma W$.
If one uses tanh activation-functions and if the local\index{activation function!tanh} 
fields $b_i(t)$ remain small, then the diagonal elements of $\ma D{(t)}$ remain close to unity.
In this case the stability condition for the reservoir dynamics is given by the weight
matrix $\ma W$ alone.   In general the singular values of $\ma W$ are different from its eigenvalues, but what matters here is that the maximal singular value of $\ma W^t$ approaches ${\rm e}^{t\lambda |\nu_1|}$, where $\nu_1$ is the eigenvalue of $\ma W$ with largest modulus (Exercise 9.8).

For inputs with long time correlations\index{time correlation}, the reservoir must not decay too quickly, so that it can represent the dynamical correlations in the input sequence. There is no precise mathematical theory that says how to optimise the reservoir. In practice one adjusts the maximal Lyapunov exponent by trial and error. Its optimal value depends on the properties of the input series, for instance on its time correlations. 

There are many different recipes for how to set up a reservoir. 
Usually the reservoir is sparse\index{reservoir!sparse}, with only a small fraction of weights non-zero.
The elements of the resulting weight matrix $\ma W$ are rescaled to adjust $\lambda_1$ \cite{jaeger04harnessing}.
The  \index{weight!matrix}weight matrix $\ma W^{({\rm in})}$ is commonly taken to be a full matrix, and its elements are 
drawn from the same distribution as those of the reservoir. Lukosevicius~\cite{lukosevicius2012practical} 
gives a practical overview over different schemes for setting up reservoir computers.  

For time-series prediction, one trains the network on\index{time series!prediction of} an input series $\ve x(0),\ldots,$ $\ve x(T-1)$ with \index{target}targets $\ve y(t) = \ve x(t)$. After training,\index{training}
one continues to iterate the network dynamics \index{network!dynamics}
with inputs $\ve x(T+k)=\ve O(T+k)$ to predict $\ve x(T+k+1)$, 
for $k=0,1,2,\ldots$. Using Equation (\ref{eq:RCD}), we see that
the reservoir dynamics takes the form
$r_i(t+1) =  g\big(\sum_j \big[w_{ij} +\sum_{k=1}^{N} 
w^{({\rm in})}_{ik} w^{({\rm out})}_{kj} ] r_j(t)\big)$ during the prediction phase.
In order to represent complex spatio-temporal patterns, Pathak {\em et al.}~\cite{pathak2018model}\index{pattern!spatio temporal}
found  it necessary to employ several parallel reservoirs\index{reservoir!parallel reservoirs}.  Lim {\em et al.} \cite{lim2019predicting} used a chain of reservoirs\index{reservoir!chain of}, replacing Equation (\ref{eq:res_dyn}) by a set of nested update rules.\index{update rule!nested}

Tanaka {\em et al.} \cite{tanaka2019recent} describe different physical implementations of reservoir computers, based on electronic RC-circuits, optical cavities or resonators, spin-torque oscillators, or mechanical devices.
\index{reservoir computing|)}

\section{Summary} 
It is sometimes said that recurrent networks learn {\em dynamical systems}, while multilayer perceptrons\index{dynamical system} learn {\em input-output maps}. This emphasises a difference in how these networks are usually used, but
we should bear in mind that they are trained in similar ways, by backpropagation. Neither is it given that the tasks must differ: recurrent networks are also used to learn time-independent data. It is true, however, that tools from {\em dynamical-systems theory}\index{dynamical system} help to analyse the dynamics of recurrent networks \cite{Doya1993,Pascanu2012}.\index{dynamical system}

Recurrent neural networks can be trained by stochastic gradient descent after unfolding the network in time, to get rid of feedback connections. This algorithm suffers from the vanishing-gradient problem. To overcome this difficulty, the hidden neurons in the recurrent network are replaced by composite units that are trained to sometimes act as residual connections, passing the signal right through, and sometimes as non-linear units that can learn correlations in a meaningful way.  There are different versions, long-short-term memory units and gated recurrent units. They all work in similar ways.  Succesful layouts for machine translation use deep bidirectional networks with layers of LSTMs. 

An alternative scheme is reservoir computing\index{reservoir computing}, where a large reservoir of hidden neurons is used to represent correlations in the input data, and a set of linear output units\index{unit!linear} is trained to learn the original sequence from such representations. The idea is that it is easier to learn intricate features of an input sequence from a high-dimensional, sparse representation of the  data.  

\section{Further reading}
The \index{training!recurrent networks}training of recurrent networks is discussed in Chapter 15 of Ref.~\cite{Haykin},
see also Refs.~\cite{williams1995back,doya1995recurrent}.
Recurrent backpropagation\index{recurrent backpropagation} is described by Hertz, Krogh and Palmer \cite{hertz1991introduction}, for a very similar  network layout.
How LSTMs combat the vanishing-gradient problem is  explained in Ref.~\cite{olah2015understanding}.
For a recent review of recurrent neural networks, see Ref.~\cite{Lipton}. 
This \href{http://karpathy.github.io/2015/05/21/rnn-effectiveness/}{webpage} \cite{Krnn} gives a very enthusiastic overview about what recurrent networks can do. 
A more pessimistic view is expressed in this \href{https://towardsdatascience.com/the-fall-of-rnn-lstm-2d1594c74ce0}{blog}.
For a review of reservoir computing, see Ref.~\cite{Lukosevicius}.
\vfill\eject

\cleardoublepage
\part{Learning without labels}
\label{part:unsupervised}
Chapters \chref{C5Chapter} to \ref{ch:rn} describe supervised learning
\index{supervised learning} of labeled data
with neural networks. The network is trained 
to reproduce the correct labels (targets) for each input pattern.\index{label}\index{target}
\index{input pattern}
The analysis of unlabeled data requires different methods. 
Machine learning can be applied
with success to large data sets of
 high-dimensional unlabeled data. The machine can for instance 
mark patterns that are typical for the given distribution, or
detect outliers. Other tasks are to detect similarity\index{similarity}, to find clusters\index{cluster|textbf} in the data (Figure~\ref{fig:C9S1SU}), and to determine non-linear, low-dimensional representations of high-dimensional data. More recently, such {\em unsupervised} learning
algorithms have been
used to generate synthetic data, patterns that resemble those \index{data, synthetic}
in a certain data set. One possible application is data-set augmentation\index{data set augmentation}
for supervised learning. 

Learning without labels is called {\em unsupervised learning}
\index{unsupervised learning|textbf}, 
because there are no \index{target}targets that tell the network whether it has learnt correctly or not. There is no obvious function to fit, or dynamics to learn. Instead the network organises  the input data in relevant 
ways. 
This requires {\em redundancy}\index{redundancy} in the input data.
It is sometimes said that unsupervised learning
corresponds to learning without a teacher, implying 
that the network itself discovers suitable ways of organising the input data. This is inaccurate, because unsupervised networks usually operate with a pre-determined learning rule\index{learning rule},
like Hopfield networks.

Part \ref{part:unsupervised} of this book is organised as follows.
Chapter \ref{ch:uhl} describes unsupervised-learning algorithms,
starting with unsupervised Hebbian learning to detect
familiarity\index{familiarity}  and similarity of input patterns\index{input pattern!similarity}\index{input pattern!familiarity} (Sections \ref{sec:Ojar} and \ref{sec:cp}). Related algorithms can be used
to find low-dimensional non-linear projections
of high-dimensional input data (self-organising maps, Section \ref{sec:som}). 
In Section \ref{sec:Kmeans}, these algorithms are compared and contrasted with
a standard unsupervised clustering algorithm, $K$-means clustering.
Section \ref{sec:rbf} introduces radial basis-function networks, they learn using a hybrid algorithm with supervised and unsupervised elements.
Section \ref{sec:ae} explains  how to use layered feed-forward networks for unsupervised learning. 

Chapter \ref{ch:rl} deals with learning tasks that lie
in between supervised and unsupervised learning,
problems where
the machine receives partial feedback on its performance in the shape 
of a penalty or a reward. Such tasks can be solved by 
{\em reinforcement-learning} algorithms\index{reinforcement learning}
that allow a neural network or more generally an agent to learn to reproduce
outputs that tend to give positive rewards.  
Several
algorithms for reinforcement learning are described, the associative reward-penalty algorithm (Section \ref{sec:arp}),
temporal difference learning (Section \ref{sec:TDL}), and $Q$-learning (Section \ref{sec:ql}).  
The $Q$-learning algorithm is illustrated by demonstrating how it allows two players to learn to compete in the board
game tic-tac-toe. 

\chapter{Unsupervised learning}   
\index{unsupervised learning|(}
\begin{figure}[bt!]
  \centering
    \begin{overpic}[scale=\myFigureScale]{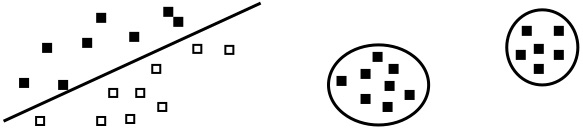} 
    \end{overpic}
    \caption{\figlab{C9S1SU} Supervised learning finds decision boundaries
for labeled data, like in the binary classification problem \index{classification!problem, binary}  shown on the left.
Unsupervised learning can find \index{cluster}clusters in the input data (right).}
\end{figure}
\label{ch:uhl}
\section{Oja's rule}
\index{Oja's rule|(}
\label{sec:Ojar}
\begin{figure}[bt!]
  \centering
    \begin{overpic}[scale=\myFigureScale]{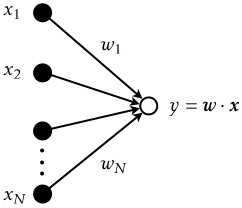}
    \end{overpic}
    \caption{\figlab{C9S1OneLinearOutput} Neural net
for unsupervised Hebbian learning, with a single linear output unit\index{unit!linear}
that has weight vector $\ve w$. The network output is denoted by $y$ in this Chapter.}
\end{figure}
A simple example for an unsupervised-learning algorithm
uses a single McCulloch-Pitts neuron with linear activation function 
(Figure \figref{C9S1OneLinearOutput}). The neuron computes\footnote{In this Chapter we follow a common convention \cite{hertz1991introduction} and denote the output of unsupervised-learning algorithms by $y$.}
 $y=\ve w\cdot\ve x$ with weight vector
$\ve w = [w_{1},\ldots, w_{N}]^{\sf T}$.
Now consider a distribution
$\Pd(\ve x)$ of input patterns\index{input pattern!distribution|textbf}
$\ve x = [x_{1},\ldots, x_{N}]^{\sf T}$
with continuous-valued components $x_{i}$. Patterns are drawn from this distribution at random and fed one after another to the net.  For each pattern $\ve x$,
the weights $\ve w$ are adjusted as follows:
\begin{equation}
\label{eq:HRus}
    \ve w' = \ve w+\delta\!\ve w\quad\mbox{with}\quad
\delta\!\ve w = \eta y \ve x\,.
\end{equation}
This rule is also called {\em Hebbian unsupervised learning} rule \cite{hertz1991introduction}, because it is reminiscent of Hebb's rule\index{Hebb's rule}
(Chapter \ref{ch:dhn}).
As usual, $\eta$ is the learning rate. 

What can this rule learn about the \index{input distribution}input distribution $\Pd(\ve x)$? 
Since we keep adding multiples of the pattern vectors $\ve x$
to the weights (just as described in Section \ref{sec:ila}), the magnitude of the output $|y|$ becomes the larger
the more often the input pattern occurs in the distribution $\Pd(\ve x)$. 
So the most familiar pattern produces the largest output.
In this way the network can detect how {\em familiar}\index{familiarity} certain input patterns are.

A problem is that the components of the weight vector continue to grow as we keep adding. 
This means that the simple Hebbian learning rule\index{Hebb's rule} (\ref{eq:HRus})
does not converge to a steady state\index{steady state}. To analyse learning outcomes we want the learning to 
converge.  This is achieved by 
adding a weight-decay term (Section \ref{sec:weightdecay}) with coefficient proportional to $y^2$ to Equation (\ref{eq:HRus}):
\begin{equation}
\label{eq:Ojas_rule}
    \delta\!\ve w= \eta y(\ve x - y\ve w)\,.
\end{equation}
Making use of  $y = \ve w\cdot\ve x=\ve w^{\sf\small T}\ve x =\ve x^{\sf\small T}\ve w$, Equation (\ref{eq:Ojas_rule}) can be rewritten in the following form:
\begin{equation}
\label{eq:Ojas_rule_2}
    \delta\!\ve w 
= \eta \big\{\ve x\ve x^{\sf \small T} \ve w-[\ve w\cdot(\ve x\ve x^{\sf\small T})\ve w]\ve w\big\}\,.
\end{equation}
This learning rule is called {\em Oja's rule}\index{Oja's rule|textbf} \cite{Oja1982}.
Equation (\ref{eq:Ojas_rule_2}) ensures that $\ve w$ remains normalised. 
To see why, consider an analogy: a vector $\ve q$ that obeys the differential equation
\begin{equation}
\label{eq:Aw}   
\tfrac{{\rm d}}{{\rm d}t}\ve q = \ma A(t)\ve q.
\end{equation}
For a general matrix $\ma A(t)$, the norm $|\ve q|$ may increase
or decrease, depending on the singular values of $\ma A$.  We can ensure that $\ve q$ remains normalised by adding a term to Equation (\ref{eq:Aw}):
\begin{equation}
\label{eq:Aw2}   
\tfrac{{\rm d}}{{\rm d}t}\ve w = \ma A(t)\ve w
-[\ve w\cdot\ma A(t)\ve w]\ve w\,.
\end{equation}
The vector $\ve w$ turns in the same way as $\ve q$,
and if we set $|\ve w|=1$ initially, then $\ve w$ remains
normalised, $\ve w = \ve q/|\ve q|$ (Exercise~10.1).
Equation (\ref{eq:Aw2})
describes the dynamics of the normalised orientation vector of a small rod in turbulence \cite{Wilkinson2009}, where $\ma A(t)$ is the matrix of fluid-velocity gradients. 

Returning to Equation (\ref{eq:Ojas_rule}), we note that the dynamics of (\ref{eq:Ojas_rule}) and (\ref{eq:Aw2}) is the same in the limit
of small learning rates $\eta$. Therefore we conclude
that $\ve w$ remains normalised under (\ref{eq:Ojas_rule_2}) when the learning
rate is small enough. Oja's algorithm is summarised in Algorithm \ref{alg:oja}. One draws a pattern $\ve x$
from the distribution $\Pd(\ve x)$ of input patterns\index{input pattern!distribution}, applies it to the network, and updates
the weights as prescribed in Equation (\ref{eq:Ojas_rule}). This is repeated many times.
In the following we denote the average over $T$ input patterns as
$\langle \cdots \rangle = \tfrac{1}{T}\sum_{t=1}^T\cdots$. 
\begin{algorithm}[t]
\caption{\label{alg:oja} Oja's rule}
\begin{algorithmic}
\STATE initialise weights randomly;
\FOR {$t=1,\ldots,T$}
\STATE  draw an input pattern $\ve x$ from $\Pd(\ve x)$;
\STATE  adjust all weights using $\delta\!\ve w= \eta y(\ve x - y\ve w)$;
\ENDFOR
\end{algorithmic}
\end{algorithm}

While the rule (\ref{eq:HRus}) does not have a steady state, Oja's rule (\ref{eq:Ojas_rule_2}) does.
For zero-mean input data, its steady state $\ve w^\ast$ corresponds to the 
principal component\index{principal component}  of the input data, as illustrated in Figure \figref{OjaPCA}. 
This can be seen by analysing the
steady-state condition\index{steady state}
\begin{equation}
\label{eq:wast0}
    0=\langle \delta\!\ve w \rangle_{\small \ve w^{*}}\,.
\end{equation}
Here $\langle \cdots  \rangle_{\small \ve w^{*}}$ is an average over iterations of the learning
rule (\ref{eq:Ojas_rule_2}) at fixed $\ve w^\ast$, the steady state\index{steady state}. 
Equation (\ref{eq:wast0}) says that  the \index{weight increment}weight increments $\delta\!\ve w$ must
average to zero in the steady state, to ensure that the weights neither
grow nor decrease in the long run.
Equation (\ref{eq:wast0}) is a condition upon $\ve w^\ast$. 
Using the learning rule\index{learning rule} (\ref{eq:Ojas_rule_2}),
it can be written as:
\begin{equation}
\label{eq:wast1}
 0 = \ma C' \ve w^{*} -(\ve w^{*}\cdot \ma C' \ve w^{*})\ve w^{*}\quad\mbox{with}\quad \ma C' = \langle \ve x \ve x^{\small\sf T}\rangle\,.
\end{equation}
Equation (\ref{eq:wast1}) shows that $\ve w^{*}$ 
must be an \index{eigenvector}eigenvector of the matrix\footnote{For zero-mean input data, $\ma C'$ equals the data-covariance matrix\index{covariance matrix}, Equation (\ref{eq:dcm}).}
 $\ma C'$, normalised to unity, $|\ve w^\ast|=1$. But which one?
\begin{figure}[t]
\centering
\begin{overpic}[scale=\myFigureScale]{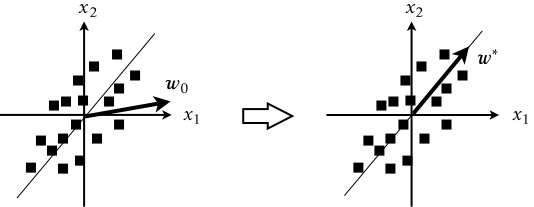}
    \end{overpic}
\caption{\figlab{OjaPCA} Oja's rule finds the principal component\index{principal component}  of zero-mean data (schematic). The initial weight
vector is $\ve w_0$, the steady-state weight vector is $\ve w^\ast$.}
\end{figure}

We denote the \index{eigenvector}eigenvectors and \index{eigenvalue}eigenvalues of $\ma C'$
by $\ve u_\alpha$ and $\lambda_\alpha$, and investigate the stability of $\ve w^\ast=\ve u_\alpha$ for different values of $\alpha$
by linear stability analysis\index{stability, linear}, just as in Section \ref{sec:rbp}.
To this end, consider a small perturbation $\dw_{\!t}$ away from $\ve w^\ast=\ve u_\alpha$:
\begin{equation}
    \ve w_{\!t} = \ve u_\alpha + \dw_{\!t}\,.
\end{equation}
A difference to the analysis in Section \ref{sec:rbp}
is that the dynamics is discrete in time. The perturbation at the next time step, $\dw_{t+1}$,  is defined by $\ve w_{t+1}=\ve u_\alpha+ \dw_{t+1}$. 
A second difference is that the sequence of weight increments depends on the randomly chosen input patterns. In order to determine the linear stability\index{stability, linear} one should
iterate and then linearise the dynamics (\ref{eq:Ojas_rule_2}), to see whether $\dw_{\!t}$ grows or not. However,
in the limit of small learning rate it is sufficient to average over $\ve x$ before iterating (Exercise 10.4). To linear order 
in $\dw_{\!t}$ one finds:
\begin{equation}
\label{eq:dw10}
\dw_{\!t+1} \approx  \dw_{\!t}+\eta\Big[\ma C' \dw_{\!t} - 2 \ve u_\alpha (\ve u_\alpha \cdot\ma C'{\dw_{\! t}}) - (\ve u_\alpha\cdot \ma C\ve u_\alpha)\ve \dw_{\!t}\Big] = \ma M^{(\alpha)}\dw_{\!t}\,,
\end{equation}
where the last equality sign defines the matrix $\ma M^{(\alpha)}$.
The steady state $\ve w^\ast =\ve u_\alpha$ is linearly stable if all \index{eigenvalue}eigenvalues of $\ma M^{(\alpha)}$ have real parts 
with magnitudes smaller
than unity.\footnote{For time-continuous dynamics (Section \ref{sec:rbp}), linear stability is ensured when all \index{eigenvalue}eigenvalues have negative real parts, for discrete dynamics their magnitudes must all be smaller than unity \cite{strogatz2000chaos}.}
To determine the \index{eigenvalue}eigenvalues of $\ma M^{(\alpha)}$, we use the fact that
$\ma M^{(\alpha)}$ has the same \index{eigenvector}eigenvectors as $\ma C'$. Since $\ma C'$ is symmetric, these  \index{eigenvector}eigenvectors
form an \index{basis!orthonormal}orthonormal basis, $\ve u_\alpha\cdot\ve u_\beta=\delta_{\alpha\beta}$. As a consequence,
the \index{eigenvalue}eigenvalues of $\ma M^{(\alpha)}$ are simply given by
\begin{equation}
\label{eq:ev_M}
\Lambda_\beta^{(\alpha)}=\ve u_\beta\cdot\ma M^{(\alpha)}\ve u_\beta=1+\eta [(\lambda_{\beta} - \lambda_{\alpha}) - 2\lambda_{\alpha}\delta_{\alpha\beta}]\,.
\end{equation}
Since $\ma C'$ is a positive-semidefinite matrix (its \index{eigenvalue}eigenvalues $\lambda_\alpha$
cannot be negative), Equation (\ref{eq:ev_M}) shows that there are \index{eigenvalue}eigenvalues
with $|\Lambda_\beta^{(\alpha)}|>1$ unless $\ve w^\ast$ is the leading  \index{eigenvector}eigenvector of $\ma C'$, 
the one corresponding to its largest \index{eigenvalue}eigenvalue.
This means that 
Algorithm \ref{alg:oja} finds the principal component of zero-mean data,
and it also implies that the algorithm  maximises $\langle y^2\rangle$ over all $\ve w$ with $|\ve w|=1$, see 
Section~\ref{sec:pprid}.  Note that $\langle y\rangle = 0$ for zero-mean input data.

Now consider inputs with non-zero mean.
In this case Algorithm \ref{alg:oja}  still finds the 
maximal-\index{eigenvalue!maximal}eigenvalue direction of $\ma C'$.  But for inputs with non-zero mean, this direction is different from the principal direction. Figure 
\figref{C9S1OjaExample}  illustrates this difference.
\begin{figure}[t]
  \centering
    \begin{overpic}[scale=\myFigureScale]{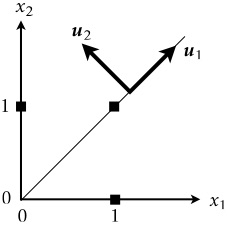}
    \end{overpic}
    \caption{\figlab{C9S1OjaExample} Input data with non-zero mean. Algorithm \ref{alg:oja} converges
to $\ve u_1$, but the principal direction is $\ve u_2$.}
\end{figure}
The Figure shows three data points in a two-dimensional input plane. 
The elements of $\ma C'=\langle \ve x\ve x^{\sf\small T}\rangle $ are
\begin{equation}
\ma C' = \frac{1}{3}\begin{bmatrix}2& 1\\1& 2\end{bmatrix}\,,
\end{equation}
with \index{eigenvalue}eigenvalues and  \index{eigenvector}eigenvectors
\begin{align} 
\lambda_{1} = 1\,,  \quad \ve u_{1} = \frac{1}{\sqrt{2}}\begin{bmatrix} 1 \\ 1 \end{bmatrix}
\quad\mbox{and}\quad  \lambda_{2} = \tfrac{1}{3}\,,  \quad\ve u_{2} = \frac{1}{\sqrt{2}}\begin{bmatrix} -1 \\ 1 \end{bmatrix} \,.
\end{align}
So the the maximal-\index{eigenvalue!maximal}eigenvalue direction of $\ma C'$ is $\ve u_1$. To compute the principal direction
of the data we must determine the data-covariance  matrix\index{covariance matrix} $\ma C$, Equation (\ref{eq:dcm}). Its \index{direction!maximal eigenvalue}maximal-eigenvalue direction is $\ve u_2$, and this is the  principal component\index{principal component}  of the data shown in Figure \figref{C9S1OjaExample}.

Oja's rule can be generalised to determine $M$ principal components
\index{principal component}
of zero-mean input data using $M$ output neurons  \index{output neuron}
that compute $ y_{i} = \ve w_{i}\cdot \ve x$ for $i=1,\ldots, M$:
\begin{equation}
\label{eq:OMr}
    \delta \!w_{ij} = \eta y_{i}\Big(x_{j} - \sum_{k=1}^{M}y_{k}w_{kj}\Big)
\end{equation}
This is called Oja's M-rule\index{Oja's rule!Oja's M rule} \cite{hertz1991introduction}. For $M=1$, Equation (\ref{eq:OMr}) simplifies to Oja's rule. 
\index{Oja's rule|)}

\section{Competitive learning}     
\index{competitive learning|(}
\label{sec:cp}
Oja's $M$-rule (\ref{eq:OMr}) results in neurons that are activated simultaneously. Any input usually causes several outputs to assume non-zero values $y_i\neq 0$ at the same time. In Sections \ref{sec:rbm} and \ref{sec:hmhl}
we encountered the notion of a winning neuron\index{winning neuron} where the weights are trained in such a way that each pattern activates only a single neuron, and different patterns activate different winning neurons. This allows to represent a distribution of input patterns with a
neural network. \index{input pattern!distribution}

Unsupervised learning algorithms can categorise or \index{cluster}cluster input data in this way: similar inputs are classified to belong to the same category, and activate the same winning neuron.
This is called {\em competitive learning}\index{competitive learning} \cite{hertz1991introduction}.
Figure \ref{fig:C9S1Circle}({\bf a}) shows an example, input patterns on the unit circle that  \index{cluster}cluster into two distinct clusters.
\begin{figure}[t]
  \centering
    \begin{overpic}[scale=\myFigureScale]{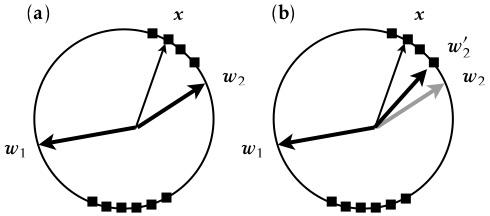}
    \end{overpic}
    \caption{\figlab{C9S1Circle} Detection of  \index{cluster}clusters by unsupervised learning.
({\bf a}) Distribution of input patterns on the unit circle and two unit-length weight vectors initialised to random angles.
 The winning neuron for pattern $\ve x$ is
the one with weight vector $\ve w_2$.  ({\bf b}) 
Updating $\ve w_2'=\ve w_2+\delta \ve w$ 
moves this weight vector closer to $\ve x$.}
\end{figure}
\begin{algorithm}[b]
\caption{\label{cl} competitive learning (Figure~\figref{C9S1Circle})}
\begin{algorithmic}
\STATE initialise weights to vectors with random angles and norm $|\ve w_i|=1$;
\FOR {$t=1,\ldots,T$}
\STATE  draw a pattern $\ve x$ from  $\Pd(\ve x)$;
\STATE  find the winning neuron $i_0$ (smallest angle between $\ve w_{\!i_0}$ and $\ve x$);
\STATE  adjust only the weight of the winning neuron $\delta \!\ve w_{\!i_0} = \eta (\ve x-\ve w_{\!i_0}) $;
\ENDFOR
\end{algorithmic}
\end{algorithm}
The idea is to find weight vectors $\ve w_i$ that point into the direction of the  \index{cluster}clusters.
To this end we take $M$ linear output units\index{unit!linear} $i$ with weight vectors $\ve w_i$, $i=1,\ldots,M$.
We feed a pattern $\ve x$ from the distribution $\Pd(\ve x)$ and {\em define} the {\em winning neuron}\index{winning neuron|textbf} $i_0$ as the one that has minimal angle between its weight and the pattern vector $\ve x$.
This is illustrated in Figure \ref{fig:C9S1Circle}({\bf b}), 
where $i_0=2$. Then only this weight vector
is updated by adding a little bit of the difference
$\ve x -\ve w_{i_0}$ between the pattern vector and the weight
of the winning neuron. The other weights remain unchanged:
\begin{equation}
\label{eq:winning_w}
\delta\!\ve w_i = 
\begin{cases} \eta (\ve x-\ve w_i) \quad\text{for}\quad i=i_{0}(\ve x,\ve w_1\ldots\ve w_M)\,, \\
        0 \qquad \hspace*{14mm}\text{otherwise} \,.\end{cases}
\end{equation}
In other words, only the winning neuron\index{winning neuron} is updated, $\ve w_{\!i_0}' = \ve w_{i_0} + \delta\!\ve w_{i_0}$. Equation (\ref{eq:winning_w}) is called {\em competitive-learning} rule\index{competitive learning|textbf}. 

The learning rule (\ref{eq:winning_w})  has the following
geometrical interpretation: the weight of the winning neuron is drawn towards the pattern $\ve x$. Upon iterating (\ref{eq:winning_w}), the weight vectors are  drawn to {\em clusters}\index{cluster} of inputs. 
If the input patterns are normalised as in Figure \figref{C9S1Circle},
\index{input pattern!normalised}
the weights end up normalised on average, even though
$|\ve w_{i_0}|=1$ does not imply  that
$|\ve w_{i_0} + \delta\ve w_{i_0}|=1$, in general.
The algorithm for competitive learning
is summarised in Algorithm \ref{cl}. 
When weight and input vectors are normalised, then the winning neuron 
$i_0$ is the one with the largest scalar product $\ve w_{i}\cdot\ve x$.
For linear output units\index{unit!linear} $y_i= \ve w_{i}\cdot\ve x$ 
(Figure \figref{C9S1OneLinearOutput}) this is simply the unit with the largest output.
Equivalently, the winning neuron\index{winning neuron} is the one with the
smallest \index{distance} distance $|\ve w_{i}-\ve x|$.
Output units with $\ve w_{i}$ that are very far away from any pattern may never be updated
({\em dead units}). There are several strategies to avoid this \cite{hertz1991introduction}.
One possibility is to initialise the weights to directions found in the inputs.
Also, how to choose the number of weight vectors is a matter of trial and error. Clearly it is better to start with too many rather than too few.

Finally, consider the relation between the competitive learning rule (\ref{eq:winning_w}) and Oja's rule\index{Oja's rule} (\ref{eq:OMr}). If we define
\begin{equation}
y_i = \delta_{ii_0} = \begin{cases} 1 \quad \text{for} ~ i=i_{0}\,,\\
     0 \quad \text{otherwise.} \end{cases}
\end{equation}
then the rule (\ref{eq:winning_w}) can be written in the form of Oja's $M$-rule:
\begin{equation}
    \delta \!w_{ij} = \eta y_{i} \bigg( x_{j} - \sum_{k=1}^{M} y_{k}w_{kj}\bigg)\,.
    \eqnlab{C10S1UpdateRule2}
\end{equation}
Equation  \eqnref{C10S1UpdateRule2} is reminiscent of Hebb's rule\index{Hebb's rule} (Chapter \ref{ch:dhn}) with weight decay.
\index{competitive learning|)}

\begin{figure}[t]
  \centering
    \begin{overpic}[scale=\myFigureScale]{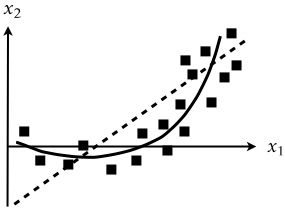}
    \end{overpic}
 \caption{\figlab{NonlinearPCA}
Principal-component analysis (Section~\ref{sec:pprid})
finds the linear principal direction  (dashed line) 
of the data ($\blacksquare$). 
A self-organising map can instead find the {\em principal manifold}\index{principal manifold|textbf} (solid line), a non-linear 
approximation to the data.}
\end{figure}
\section{Self-organising maps} 
\label{sec:som}
In order to analyse high-dimensional data it is often useful to map the high-dimensional 
input patterns\index{input pattern!high dimensional} to 
a low-dimensional output space\index{output space}, to obtain a low-dimensional representation of the \index{input distribution}input distribution. Principal-component analysis (Section \ref{sec:pprid}) does just that. However, it does not
necessarily preserve distance\index{distance}. To visualise  \index{cluster}clusters or other arrangements of the input patterns, 
similar patterns or patterns that are close in input space\index{input space} should be mapped to nearby points in output space\index{output space}, and patterns that are far apart should be mapped to outputs that are far from each other.
Maps that achieve this are called   {\em semantic} or {\em topgraphic} maps.\index{map!topographic}\index{map!semantic} 

Moreover, principal-component analysis is a linear method. As explained
in Section \ref{sec:pprid}, it projects the data to the space spanned by the leading  \index{eigenvector!leading}eigenvectors of the \index{matrix!correlation}correlation matrix. In many cases, however, the data may not lie in a linear subspace, as illustrated in Figure~\figref{NonlinearPCA}. In order to project the data onto the non-linear {\em principal manifold}\index{principal manifold} (solid line), a non-linear map is needed. 

In neuroscience, the term topographic map refers to the relation between the spatial arrangement of \index{stimulus}stimuli and the activation patterns in certain parts of the mammalian brain. Similar patterns of visual \index{stimulus!visual}stimuli on the retina, for instance, activate close-by regions in the visual cortex \index{cortex!visual} \cite{weliky1996systematic}. 
Other cognitive \index{stimulus!cognitive}stimuli, auditory and sensory, are mapped in analogous ways.
The complex neural networks in the mammalian \index{cortex!cerebral} cortex 
contain large numbers of 
such maps, arranged in a hierarchical fashion. They represent local \index{stimulus!local}stimuli in terms of spatially localised neural activation. How did this complex structure arise? One possibility is that the mappings are coded in the genetic sequence, that the connections are hard wired, so to speak. 
However, it is observed that such maps can change over time \cite{Kohonen2013}, leading to the hypothesis that 
they are learned, and that the our DNA merely encodes a set of fairly simple {\em learning rules}\index{learning rule}. 

This motivated Kohonen \cite{Kohonen2013,KohonenBook} and others to propose and analyse learning rules for topographic maps. The term {\em self-organising map}\index{map!self organising|textbf} \cite{kohonen1990self,Martin2009} emphasises that the mapping develops in response to the \index{stimulus}stimuli it maps, that it learns in
an unsupervised fashion.
\begin{figure}[bt!]
  \centering
    \begin{overpic}[scale=\myFigureScale]{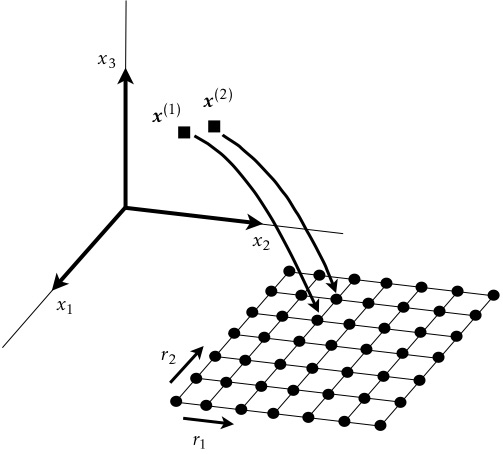}
    \end{overpic}
    \caption{\figlab{C9S2TopographicMap} Kohonen's self-organising map. If patterns
$\ve x^{(1)}$ and $\ve x^{(2)}$ are close in input space\index{input space}, then the two patterns
activate neighbouring winning neurons\index{winning neuron} in the output array 
(with coordinates $\ve r = [r_1,r_2]^{\sf T}$). Often the dimension of the output array is much lower than that of input space\index{input space}.}
\end{figure}
Kohonen's model for a non-linear self-organising map relies on an ordered array of output neurons\index{output neuron}, as illustrated in Figure~\figref{C9S2TopographicMap}. The map learns to activate nearby output neurons\index{output neuron} for similar inputs.
This is achieved using a competitive learning rule\index{competitive learning}, similar to the learning rule (\ref{eq:winning_w}) described in the previous Section. In order to represent the proximity or similarity of inputs, the rule is endowed with the notion of distance\index{distance!in output array} in the output array, by
updating not only the winning neuron, but also those that are neighbours in the output array.
To this end one replaces the competitive-learning rule (\ref{eq:winning_w}) by
\begin{equation}
\label{eq:Kohonen_update}
\delta\!\ve w_i = \eta h(i,i_{0}) (\ve x-\ve w_i)\,,
\end{equation}
where $i_{0}$
is the index of 
the winning neuron, the one with weight vector closest to the input $\ve x$.
The {\em neighbourhood function}\index{neighbourhood function|textbf} function $h(i,i_0)$
depends on the distance\index{distance!in output array|textbf} of the neurons $i$ and $i_0$ in the output array. The neighbourhood function has a maximum at $i=i_0$ and decreases as the distance\index{distance!in output array} between $i$ and $i_0$ increases.
One possibility is to assign decreasing values to $h(i,i_0)$ for nearest neighbours,
next-nearest neighbours, and so forth (Figure \ref{fig:C10NN}). 
\begin{figure}
\centering
\begin{overpic}[scale=\myFigureScale]{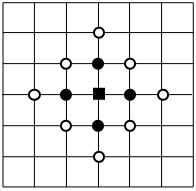}
\end{overpic}
\caption{\label{fig:C10NN} Nearest neighbours ($\bullet$) and next-nearest neighbours ($\circ$) to the neuron at the centre ($\blacksquare$) of the output array.}
\end{figure}
Another possibility is to use a Gaussian function of
the \index{Euclidean distance}Euclidean distance\index{distance!in output array} $|\ve r_i - \ve r_{i_0}|$ in the output array 
\cite{hertz1991introduction}:
\begin{equation}
\label{eq:lambda_kohonen}
      h(i,i_{0}) = \exp\Big(-\tfrac{1}{2\sigma^2}|\ve r_{i} - \ve r_{i_{0}}|^{2}\Big)\,.
\end{equation}
Here $\ve r_i$ is the position of neuron $i$ in the output array (Figure \ref{fig:C9S2TopographicMap}). Different normalisations of the Gaussian \cite{Haykin} can be subsumed in different learning rates. 

Kohonen's rule has two parameters: the learning rate $\eta$,  and
the width $\sigma$ of the neighbourhood function.
\index{neighbourhood function}
Usually one adjusts these parameters as the learning proceeds. Typically one begins with large values for $\eta$ and $\sigma$ ({\em ordering phase}\index{ordering phase}), and then reduces these
parameters  as the elastic network evolves ({\em convergence phase})\index{convergence!phase}:
quickly at first and then in smaller steps, until the algorithm converges \cite{Kohonen2013,hertz1991introduction,Haykin}.

According to Equations (\ref{eq:Kohonen_update}) and (\ref{eq:lambda_kohonen}),  
similar patterns activate nearby neurons in output space\index{output space}, and their
weight vectors change in similar ways.
Kohonen's rule drags the winning weight vector $\ve w_{\!i_{0}}$ towards $\ve x$,
just as the competitive learning rule \index{competitive learning} (\ref{eq:winning_w}), but it also drags the neighbouring weight vectors along. 
Figure \figref{C9S3NetEx} illustrates a geometrical interpretation
of Kohonen's rule \cite{kohonen1990self}. 
We can think of the weight vectors as pointing to the nodes
of an {\em elastic net}\index{elastic net} that has the same layout as the output array. As one feeds patterns from the \index{input distribution}input distribution, 
the weights are updated, causing the nodes of the network to move. 
This changes the shape of the elastic network until it resembles the shape defined by the distribution of input patterns.\index{input pattern!distribution}
Figure~\ref{fig:NonlinearPCA} shows another example where the dimensionality of the output array (one-dimensional) is lower than
that of the input space\index{input space} (two-dimensional). The algorithm finds a non-linear approximation to the data, the {\em principal manifold}\index{principal manifold}.  As opposed to the principal direction in principal-component analysis, the principal manifold need not be linear. Therefore it can approximate the data more precisely, leading to a smaller residual variance (Exercise 10.7). 

In summary, Kohonen's algorithm learns by distributing the weight vectors of the output neurons\index{output neuron} to reflect the distribution of input patterns. In general this works well, but problems
occur at the boundaries.
Why this happens is quite clear (Figure \figref{C9S3NetEx}): 
since the density of patterns
outside the parallelogram is low, the elastic network cannot be drawn very close to the boundary.
To analyse how the boundaries affect learning for Kohonen's rule, consider the steady-state \index{steady state}
condition
\begin{equation}
\label{eq:kss}
       \langle \delta \!\ve w_{i} \rangle = \frac{\eta}{T}\sum_{t=1}^T h(i,i_{0})\left( \ve x^{(t)} - \ve w_{i}^\ast \right) = 0\,.
\end{equation}
This condition is more complicated than it looks at first sight, because $i_0$ depends on the weights and on the patterns, as mentioned above.  The steady-state condition (\ref{eq:kss}) is very difficult to analyse in general. One of the reasons is that global geometric information is difficult to learn. It is usually much easier to learn local structures. This is particularly true in the {\em continuum limit}\index{continuum limit} where we can analyse local learning progress using Taylor expansions.
\begin{figure}[bt]
\centering
        \begin{overpic}[scale=\myFigureScale]{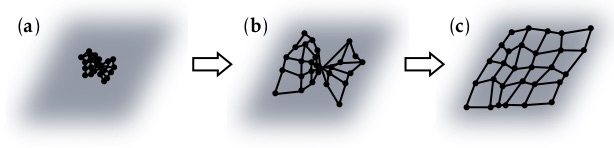}
         \end{overpic}
    \caption{\figlab{C9S3NetEx} Learning a
distribution $\Pd(\ve x)$ (gray) of two-dimensional real-valued inputs $\ve x$
with Kohonen's algorithm.
Illustration of the dynamics of the self-organising map
in terms of an {\em elastic net}\index{elastic net}.
({\bf a}) Initial condition. ({\bf b}) Intermediate stage. ({\bf c}) In the
steady-state the elastic network resembles the shape defined
by the \index{input distribution}input distribution $\Pd(\ve x)$. }
\end{figure}

The analysis of condition (\ref{eq:kss}) in the continuum limit
\index{continuum limit}  
is due to Ritter and Schulten \cite{ritter1986stationary}, and
it is described in detail by 
Hertz, Krogh, and Palmer \cite{hertz1991introduction}.
One assumes that there is a very dense network of output neurons\index{output neuron}, so that one can approximate
    $i \to \ve r$, $i_{0} \to \ve r_{0}$, $\ve w_{i} \to \ve w(\ve r)$, $h(i,i_{0})  \to h\big(\ve r - \ve r_{0}(\ve x)\big)$,
and $\tfrac{1}{T}\sum_{t} \to \int {\rm d}\ve x \Pd(\ve x)$.
In this continuum limit\index{continuum limit|textbf}, Equation (\ref{eq:kss}) reads
\begin{equation}
\label{eq:kssc}
       \int\! {\rm d}\ve x\, \Pd(\ve x) 
\,h\big(\ve r - \ve r_{0}(\ve x)\big)\,\big[\ve x - \ve w^\ast(\ve r)\big] = 0\,.
\end{equation}
This is a condition for the steady-state learning outcome, 
the function $\ve w^\ast(\ve r)$. 

In the continuum limit the position $\ve r_{0}(\ve x)$ of the 
winning neuron in the output array for pattern $\ve x$  is given by
\begin{equation}
\label{eq:winning}
\ve w^\ast(\ve r_0) = \ve x\,.
\end{equation}
We use this relation to write Equation (\ref{eq:kssc}) as:
\begin{equation}
\label{eq:kssc1}
       \int\! {\rm d}\ve x\, \Pd(\ve x)
\,h\big(\ve r - \ve r_{0}(\ve x)\big)\,\big[\ve w^\ast(\ve r_0(\ve x)) - \ve w^\ast(\ve r)\big] = 0\,.
\end{equation}
Equation (\ref{eq:winning}) defines a mapping $\ve r_0(\ve x)$ from input space\index{input space}  to output space\index{output space}, the self-organising map\index{map!self organising} (Figure \figref{C9S2TopographicMap}).  Assuming that this mapping is one-to-one, we change integration variable from $\ve x$ to 
$\ve r_0$:
\begin{equation}
\label{eq:kssc2}
       \int\! {\rm d}\ve r_0\, |\mbox{det}\ma J |
 \,Q(\ve r_0) \,h(\ve r-\ve r_0)\,\big[\ve w^\ast(\ve r_0) - \ve w^\ast(\ve r)\big] = 0\,,
\end{equation}
where  $Q(\ve r_0)\equiv  \Pd\big(\ve x(\ve r_0)\big)$, and where the determinant
represents the volume element of the variable transformation. Using Equation (\ref{eq:winning}), the
Jacobian $\ma J$ of the transformation has elements
\begin{equation}
J_{ij} = \frac{\partial w_i(\ve r_0)}{\partial r_j}\,.
\end{equation}
The neighbourhood function is sharply peaked at $\ve r=\ve r_0$,
\index{neighbourhood function}
and this makes it possible to evaluate the steady-state condition (\ref{eq:kssc2}) approximately,
\begin{figure}
  \centering
   \begin{overpic}[scale=\myFigureScale]{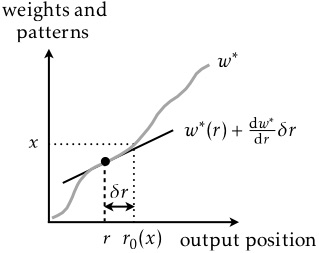}
    \end{overpic}
    \caption{\figlab{C9S3Neighbourx} In order to find out how 
the steady-state map $w^{\ast}(r)$ varies near $r$ (gray line),
one expands $w^\ast$ in $\delta\!r$ around $r$, 
$w^\ast(r+\delta r) = w^\ast(r)+\tfrac{{\rm d}w^\ast}{{\rm d}r}\,\delta\!r +
 \tfrac{1}{2}  \tfrac{{\rm d}^2 w^\ast}{{\rm d}r^2} \delta\!r^2+\ldots$.
}
\end{figure}
expanding the integrand in $\delta\ve r=\ve r_0-\ve r$, assuming that $\ve w^\ast(\ve r)$ is
a smooth function. 
This is illustrated in Figure 
\figref{C9S3Neighbourx}, for one-dimensional
inputs and outputs. We consider this special case not only to simplify the notation, but also because it is one of the few cases that 
admit mathematical analysis (Exercise 10.9). Expanding $w^\ast(r+\delta r)$ as shown in Figure \figref{C9S3Neighbourx} yields
\begin{align}
\label{eq:newc}
    w^\ast(r+\delta r)-w^\ast(r) = \tfrac{{\rm d}}{{\rm d}r} w^\ast(r)\delta\!r + \tfrac{1}{2} \tfrac{{\rm d}^2}{{\rm d}r^2}w^\ast(r) 
\delta\!r^{2} + \ldots\,.
\end{align} 
The other factors in Equation (\ref{eq:kssc2}) are expanded 
in a similar way:
\begin{subequations}
\begin{align}
    J(r+\delta r) &= \tfrac{{\rm d}w^\ast}{{\rm d}r} + \tfrac{{\rm d}^2w^\ast}{{\rm d}r^2}\delta r+\ldots  \,,\\
    Q(r+\delta r) &= \Pd(w^\ast) + 
\delta\!r \tfrac{{\rm d}w^\ast}{{\rm d}r} \tfrac{\rm d}{{\rm d}w} \Pd({w^\ast})\,.
\end{align}
\end{subequations}
Inserting these expressions
into Equation (\ref{eq:kssc}), discarding terms of order higher than $\delta r^2$,
and changing the integration variable to $\delta r$,  one finds
\begin{align}
    0 &= w'[\tfrac{3}{2}w''\Pd(w) + (w')^{2}\tfrac{\rm d}{{\rm d}w} \Pd(w)]\int_{-\infty}^\infty\hspace*{-4mm}{\rm  d}\delta\!r \,\delta\!r^{2} h(\delta\!r)\,,
\label{eq:wast4}
\end{align}
where we introduced the short-hand notation $w' = \tfrac{{\rm d}}{{\rm d}r} w^\ast(r)$, dropped the asterisk, 
and used that the neighbourhood function (\ref{eq:lambda_kohonen}) is symmetric, $h(-\delta r)= h(\delta r)$.
\index{neighbourhood function}
Since the integral in Equation (\ref{eq:wast4}) is non-zero, we must either have
 \begin{equation}
     w'=0\quad\mbox{or}\quad  \frac{3}{2}w''\Pd(w) + (w')^{2}
\tfrac{\rm d}{{\rm d}w} \Pd(w) = 0\,.
 \end{equation}
The first solution can be excluded because it corresponds to a singular weight distribution
that does not contain any geometrical information about the \index{input distribution}input distribution $\Pd$. The second solution gives 
\begin{equation}  
\eqnlab{C16S1first} 
\frac{w''}{w'}=-\frac{2}{3}\frac{w'\tfrac{\rm d}{{\rm d}w} \Pd(w)}{\Pd(w)}\,.  \end{equation}
In other words, $\tfrac{\rm d}{{\rm d}r}\log|w'| = -\tfrac{2}{3}\tfrac{\rm d}{{\rm d}r}\log \Pd(w)$,
and this means that $|w'| \propto [\Pd(w)]^{-\frac{2}{3}}$.
The density of output weights can be computed as
\begin{equation}
\varrho(w) = \int\!{\rm d}r \delta[w-w^\ast(r)]\,,
\end{equation}
where $\delta(w)$ is the Dirac  $\delta$-function \cite{jackson1999classical}.
Changing variables in the $\delta$-function
\begin{equation} 
\delta[w-w^\ast(r)] =  \sum_{j|w=w^\ast(r_j)} \frac{1}{|w'|} \delta(r-r_j)\,,
\end{equation}
and assuming that the function $w^\ast(r)$ is one-to-one, one finds
\begin{equation}
\label{eq:final_Kohonen}
\varrho(w) = \frac{1}{|w'|} = [\Pd(w)]^{\frac{2}{3}}\,.
\end{equation}
This tells us that the self-organising map  learns the \index{input distribution}input distribution in the following way: the distribution
of output weights in the steady state\index{steady state} reflects the distribution of 
input patterns. Equation (\ref{eq:final_Kohonen})\index{input pattern!distribution}
shows that the two distributions are not equal (equality would have been a perfect outcome). 
The distribution of weights is instead proportional to $\big[\Pd(w)\big]^{\frac{2}{3}}$. Little is known in higher dimensions, but the general
idea is that the elastic network has difficulties reaching the corners and edges of the domain where the \index{input distribution}input distribution is non-zero.

The output of a self-organising map can be interpreted in different ways.
For a low-dimensional inputs and outputs, one can simply plot the
map $w^\ast(\ve r)$, as in Figure \figref{C9S2TopographicMap}.  
Dense regions of weights point to regions in input space\index{input space} with a high
density of inputs. 
\begin{figure}[t]
\begin{overpic}[scale=0.5]{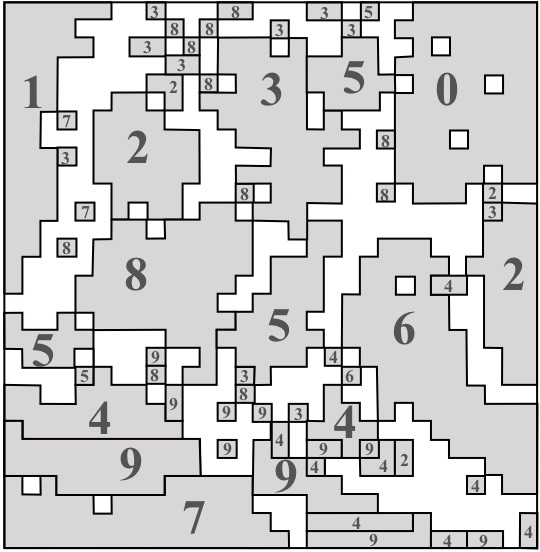}
\end{overpic}
\caption{\label{fig:C9KohonenMNIST} \index{MNIST data set}Clustering of hand-written digits (\href{http://yann.lecun.com/exdb/mnist/}{MNIST} data set) with a 
self-organising map
with a $30\times 30$ output array. In the shaded regions the outputs are quite certain:
here the winning neurons are activated by the indicated digit in 80\% of the cases.
The white regions correspond to outputs where the majority digit appears in less than 80\% of the cases,
or to outputs that are never activated, or only once. Schematic, based on simulations
performed by Juan Diego Arango.}
\end{figure}
Often the output dimension is taken to be much lower than 
the dimension of input space\index{input space}. In this case the self-organising map
performs non-linear {\em dimensionality reduction}\index{dimensionality reduction},
and it can be used to find  \index{cluster}clusters in high-dimensional input data~\cite{Snyder}. The analysis proceeds in two steps. First, one runs Kohonen's algorithm until the map has converged to a steady state. Second, one feeds
all inputs into the net, and for each input one determines the location of the winning neuron in the output array. The spatial activation patterns 
in the output array represent  \index{cluster}clusters of similar inputs. This is illustrated
in Figure~\ref{fig:C9KohonenMNIST}, which shows how a self-organising map represents handwritten digits
from the \href{http://yann.lecun.com/exdb/mnist/}{MNIST} \index{MNIST data set}data set. To reveal the semantic map, 
the Figure labels  \index{cluster}clusters of outputs that correspond to the same digits (as determined by the labels\index{label} in the \index{training set}training set). We see that the self-organising  map groups the same digits together, but it has some difficulty
distinguishing the digits 3 and 8, and also 4 and 9.

\section{$K$-means clustering}
\index{cluster|(}
\label{sec:Kmeans}
\begin{figure}[t]
    \centering
\begin{overpic}[scale=\myFigureScale]{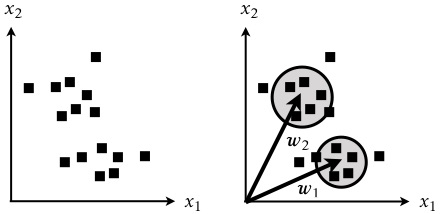}
\end{overpic}
\caption{\label{fig:kmeans} Schematic illustration of the $K$-means clustering algorithm
with two weight vectors $\ve w_1$ and $\ve w_2$. The radii of the disks equal $s_1$
and $s_2$, Equation (\ref{eq:sigmak}).}
\end{figure}

Sections \ref{sec:cp}  and  \ref{sec:som} described different ways of finding clusters
in input data. In particular, it was shown 
 how self-organising maps can find clusters in high-dimensional input data, and represent them in a low-dimensional, non-linear projection.   $K$-means clustering\index{K means clustering} \cite{Haykin} is an alternative unsupervised-learning algorithm for finding clusters in the input data.  Let us compare and contrast this algorithm with  Kohonen's self-organising map.
The goal is to cluster input patterns\index{input pattern!cluster} $\ve x^{(\mu)}$, $\mu=1,\ldots,p$ into $K$ clusters. 
Usually $K$ is much smaller than the number of inputs, $p$, and than the input dimension  $N$. 

A solution of the clustering task is a mapping $k(\mu)$ that associates
each input $\ve x^{(\mu)}$ with one of the clusters $k=1,\ldots,K$. The function $k(\mu)$ is determined
by minimising the energy function
\index{energy function}
\begin{equation}
H(\ve w_1,\ldots,\ve w_K) 
= \frac{1}{2} \sum_{k=1}^K \Big(\!\!\sum_{\mu\,|\,k(\mu)\!=\!k} |\ve x^{(\mu)} -\ve w_k|^2\Big)\,.
\end{equation}
The second sum is over all values of $\mu$ that satisfy $k(\mu)=k$.
The vector $\ve w_k$ becomes the average of all pattern vectors in cluster $k$, and 
the expression in the parentheses is the variance associated with this cluster:
\begin{equation}
\label{eq:sigmak}
\sigma_k^2 = \sum_{\mu\,|\,k(\mu)\!=\!k} |\ve x^{(\mu)} -\ve w_k|^2\,.
\end{equation}
In other words, $H$ measures the sum of the cluster variances $\sigma^2_k$. 
A solution to the clustering problem corresponds to a \index{local minimum}local minimum of $H$. 
To determine the  cluster vectors $\ve w_k$ and the corresponding variances $\sigma_k^2$, one starts
from an initial guess for $k(\mu)$. For each cluster, one begins by adjusting $\ve w_k$ to minimise
the cluster variance:
\begin{equation}
\argmin_{\ve w_k}= \sum_{\mu|k(\mu)\!=\!k} |\ve x^{(\mu)} -\ve w_k|^2\,.
\end{equation}
 In a second step, one optimises the encoding function
\begin{equation}
k(\mu) = \argmin_{1 \leq k\leq K} |\ve x^{(\mu)} -\ve w_k|^2\,,
\end{equation}
given the vectors $\ve w_k$.
These steps are repeated until a satisfactory solution is found. The solution is not unique, 
usually the algorithm converges to a \index{local minimum}local minimum of $H$. In practice one should try different random initialisations 
to find the best local minimum.

All three algorithms, competitive learning\index{competitive learning}, the self-organising map, and $K$-means clustering move weight vectors
towards clusters in input space\index{input space}. A difference between the self-organising  map and the other two algorithms is that
\index{neighbourhood function}
the self-organising map uses a neighbourhood function (so that similar inputs activate close-by neurons in the output array), and updates their weight vectors in similar fashion. In this way, a self-organising map with a large output array can find a smooth parameterisation of the principal manifold.  If we shrink the neighbourhood function to the centre point in Figure \ref{fig:C10NN}, all geometric information is lost, and the self-organising map becomes equivalent to competitive learning (Algorithm \ref{cl}). 
Essentially, competitive learning and $K$-means clustering are sequential and batch \index{batch!learning} versions of the same algorithm \cite{hertz1991introduction}. So the self-organising map becomes equivalent to $K$-means clustering when the neighbourhood range tends to zero.
\index{cluster|)}

\section{Radial basis functions}      
\index{radial basis function|(}
\label{sec:rbf}
Problems that are not linearly separable\index{linear separability} can be solved by perceptrons with hidden layers,
\index{hidden layer}
as we saw in Chapter \chref{C5Chapter}.  Figure \ref{fig:sns}({\bf b}), for example, shows a piecewise linear decision boundary
parameterised by hidden neurons.\index{hidden neuron}

Separability  can also be achieved by a non-linear transformation of input 
space\index{input space}.
Figure \figref{C10S1XORFunction} shows how the XOR problem can be transformed into a linearly separable problem 
by the transformation
\begin{equation}
\label{eq:utrans}
\rbf_1(\ve x) = (x_2-x_1)^2-\tfrac{1}{2}\quad\mbox{and}\quad \rbf_2(\ve x) = x_2\,.
\end{equation}
The Figure shows the non-separable problem in the $x_1$-$x_2$ plane, and in the new coordinates $\rbf_1$ and $\rbf_2$.
\begin{figure}[t]
    \centering
\begin{overpic}[scale=\myFigureScale]{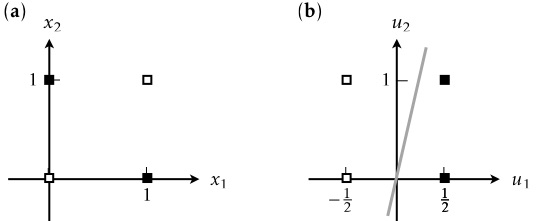}
    \end{overpic}
\caption{\figlab{C10S1XORFunction} Linear separation of the XOR function by the \index{map!non linear}non-linear mapping (\ref{eq:utrans}). ({\bf a}) In the input plane the problem is not linearly separable. ({\bf b})  In the $\rbf_1$-$\rbf_2$ plane the problem is homogeneously linearly separable.}
\end{figure}
The problem is homogeneously 
linearly separable in the $\rbf_1$-$\rbf_2$ plane. We can solve it by a single McCulloch-Pitts neuron with
weights $\ve W$ and zero \index{threshold}threshold, parameterising the decision boundary as $\ve W\cdot \ve \rbf(\ve x) =0$.

It is even better to map the patterns (non-linearly)\index{map!non linear} to a space of higher dimension, because Cover's theorem (Section \ref{sec:ct}) says
that it is easier to separate the patterns there:
consider a set $\ve u(\ve x)=[u_1(\ve x),\ldots,u_m(\ve x)]^{\sf T}$ of $m$ polynomial functions of finite order that \index{embedding}embed $N$-dimensional input space\index{input space} in an $m$-dimensional space. 
Then the probability that a problem with $p$ points $\ve x^{(\mu)}$ in $N$-dimensional input space\index{input space} is separable by a polynomial decision boundary is given by $P(p,m)$ [Equation (\ref{eq:Ppm})] \cite{Cover,Haykin}. Note that this probability is independent of the dimension $N$ of input space.

The question is of course how to find the \index{map!non linear}non-linear
mapping $\ve \rbf(\ve x)$. One possibility is to use {\em radial basis functions}\index{radial basis function}. 
The idea is to parameterise the
functions $\rbf_j(\ve x)$ in terms of weight vectors $\ve w_j$, and to 
use an unsupervised-learning algorithm to find weights that separate the input data. 
A common choice \cite{Haykin} is to use 
radial basis functions\index{radial basis function|textbf} of the form:
\begin{equation} 
\label{eq:gj}
\rbf_{j}(\ve x) = \exp\Big(-\frac{1}{2s_{j}^{2}} \left|\ve x - \ve w_{j}\right|^{2}\Big) \,.
\end{equation}
Note that these functions are not of the finite-order polynomial form that was assumed above. So strictly speaking we cannot invoke Cover's theorem.  In practice the mapping $\rbf_{j}(\ve x)$ works nevertheless quite well.
The parameters $s_j$ parameterise the {\em widths} of the radial basis functions. 
In the simplest version of the algorithm they are set to unity. 
Hertz, Krogh, and Palmer \cite{hertz1991introduction} discuss radial basis-function networks with {\em normalised} radial basis functions\index{radial basis 
function!normalised}
\begin{equation}
\rbf_{j}(\ve x) =
\frac{\exp\big(-\frac{1}{2s_{j}^{2}} \left|\ve x - \ve w_{j}\right|^{2}\big)}
{\sum_{k=1}^m \exp\big(-\frac{1}{2s_{k}^{2}} \left|\ve x - \ve w_{k}\right|^{2}\big)}\,.
\end{equation}
Other choices for radial basis functions are given by Haykin \cite{Haykin}. 
\begin{figure}[bt!]
  \centering
    \begin{overpic}[scale=\myFigureScale]{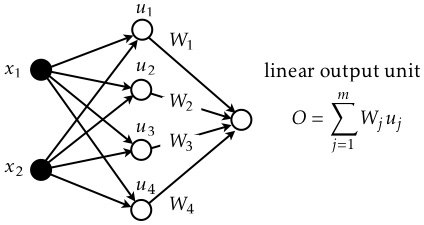}
    \end{overpic}
   \centering \caption{\figlab{C10S3RadialBasisNetwork}
Radial basis-function network for $N=2$ inputs
and $m=4$ radial basis functions (\ref{eq:gj}). The output neuron\index{output neuron} \index{unit!linear}has a linear activation function\index{activation function!linear}, weights $\ve W$, and zero \index{threshold}threshold. }
\end{figure}

\begin{figure}[bt]
    \centering
\begin{overpic}[scale=\myFigureScale]{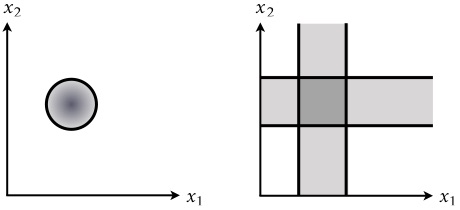}
    \end{overpic}
\caption{\figlab{C10S2RbfPerceptronComparison} Comparison between radial-basis function network and perceptron. Left:
the output of a radial basis function is localised in input space\index{input space}. 
Right: to achieve a localised output with sigmoid units one needs
\index{activation function!sigmoid}
two hidden layers (Figure \ref{fig:C7S1BasisFunction2D}).}
\end{figure}

Figure \figref{C10S3RadialBasisNetwork} shows a radial basis-function network for $N=2$ and $m=4$. The four neurons in the hidden layer stand
for the four radial basis functions (\ref{eq:gj}) that map the inputs
to four-dimensional $\ve \rbf$-space. The network looks like a perceptron
(Chapter \chref{C5Chapter}). But here the hidden layers work in a different 
way.
Perceptrons have hidden McCulloch-Pitts neurons that compute {\em non-local} outputs $\sigma(\ve w_j\cdot\ve x-\theta)$.  The output of radial basis functions $\rbf_j(\ve x)$, by contrast, is {\em localised} in input space\index{input space} [Figure \ref{fig:C10S2RbfPerceptronComparison} (left)]. We saw in Section \ref{sec:hmhl}
how to make localised basis functions out of McCulloch-Pitts neurons
with sigmoid activation functions $\sigma(b)$, but
\index{activation function!sigmoid}
one needs two hidden layers to do that [Figure \ref{fig:C10S2RbfPerceptronComparison} (right)]. 

Radial basis functions produce localised outputs with a single hidden layer, they divide up input space\index{input space} into localised regions, each corresponding to one radial basis function. Imagine for a moment that 
we have as many radial basis functions as input patterns. In this case we
can simply take $\ve w_\nu = \ve x^{(\nu)}$ for $\nu=1,\ldots,p$.
The linear output in Figure~\figref{C10S3RadialBasisNetwork} computes
$O^{(\mu)} = \ve W\cdot \ve u(\ve x^{(\mu)})$, and so the 
classification problem in $\rbf$-space takes the form
\begin{equation}
\label{eq:Umatrix}
\sum_{\mu=1}^p \ve W_\mu U_{\mu\nu} = t^{(\nu)}
\end{equation}
with  $U_{\mu\nu} = 
\rbf_\nu(\ve x^{(\mu)})$. 
If all patterns are pairwise different, $\ve x^{(\mu)}\neq \ve x^{(\nu)}$
for $\mu\neq \nu$, then the matrix $\ma U$ is invertible \cite{Haykin}. 
In this case the solution of the classification problem reads
\begin{align}
W_\mu = \sum_{\nu=1}^p  t^{(\nu)} [\ma U^{-1}]_{\nu\mu}\,,
\end{align}
where $\ma U$ is the symmetric $p\times p$  matrix with elements $U_{\mu\nu}$. 

In practice one can get away with fewer radial basis functions by choosing
their weights to point in the directions of \index{cluster}clusters of input data. 
To this end one uses
unsupervised competitive learning\index{competitive learning} (Algorithm \ref{rb}), where the index $j_0$ of the {\em winning neuron}
\index{winning neuron} is defined to be the one with largest $u_j$.
How are the widths $s_j$ determined?  
The width $s_j$ of radial basis function $u_j(\ve x)$
is taken to be equal to the minimum distance\index{distance}
between $\ve w_j$ and the centers of the surrounding radial basis
functions. Once weights and widths of the radial
basis functions are found, the weights of the output neuron\index{output neuron} are determined 
by minimising
\begin{equation}
H = \frac{1}{2} \sum_{\mu}\left(t^{(\mu)}-O^{(\mu)}\right)^{2}
\end{equation}
with respect to $\ve W$.  
This works even if $\ma U$ is not invertible.
An approximate solution can be obtained by stochastic gradient descent
on $H$, keeping the parameters of the radial basis functions fixed.
Cover's theorem indicates that the problem is more likely to be
separable if the \index{embedding!dimension}embedding dimension $m$ is higher. 
\begin{algorithm}[t]
\caption{\label{rb} radial basis functions}
\begin{algorithmic}
\STATE initialise the weights $w_{jk}$ independently randomly from $[-1,1]$;
\STATE set all widths to $s_j=0$;
\FOR {$t=1,\ldots,T$}
\STATE feed randomly chosen pattern $\ve x^{(\mu)}$;
\STATE determine winning neuron $j_{0}$: $\rbf_{j_{0}} \geq \rbf_{j} \quad \text{for all values of $j$}$;
\STATE update widths:  $s_{j} = \min_{j\neq k} |\ve w_{j} - \ve w_{k}|$;
\STATE update only winning neuron: $ \delta\!\ve w_{j_0} =  \eta (\ve x^{(\mu)} -\ve w_{j_0})$;
\ENDFOR
\end{algorithmic}
\end{algorithm}

Radial basis-function networks are similar to the perceptrons described
in Chapters \chref{C5Chapter} to \ref{chap:dl}, in that they are feed-forward
networks designed to solve classification problems. 
A fundamental difference is that the parameters of the radial basis functions are determined by unsupervised learning, whereas perceptrons are trained using supervised learning for {\em all} units.
While McCulloch-Pitts neurons compute weights to minimise their output from given targets, the radial basis functions compute weights by maximising the output  $\rbf_{j}$ as a function of $j$.  The algorithm for finding the weights of the radial basis functions is summarised in Algorithm \ref{rb}.  Further, as opposed to the deep networks from Chapter \ref{chap:dl}, radial basis-function networks have only one hidden layer, and a linear output neuron\index{output neuron}.  
In summary, radial basis-function networks learn using a {\em hybrid} scheme: unsupervised learning for the parameters of the radial basis functions, and supervised learning for the weights of the output neuron.\index{output neuron} 
\index{radial basis function|)}

\section{Autoencoders}
\index{hidden layer|(}
\index{autoencoder|(}
\label{sec:ae}
Multi-layer perceptrons, layered feed-forward networks, 
were developed for supervised learning, as described in 
Part \ref{part:supervised}. Such layouts can also be used for unsupervised learning. Examples are {\em autoencoders}\index{autoencoder} and {\em generative adversarial networks}\index{generative adversarial networks}.
\begin{figure}[t]
  \centering
    \begin{overpic}[scale=\myFigureScale]{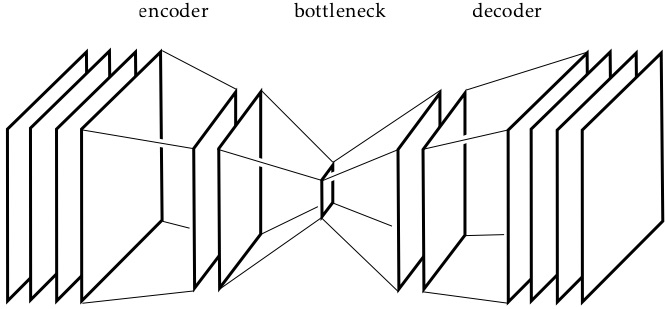}
    \end{overpic}
    \caption{\figlab{C9S6AE} Autoencoder (schematic). Both \index{encoder}encoder and \index{decoder}decoder
consist of a number of fully connected or convolutional layers (depicted as squares). \index{autoencoder|textbf}
In the layout shown, the \index{bottleneck}bottleneck consists of a layer with
very few neurons. Sparse autoencoders have bottlenecks with many neurons, but only few are activated.}
\end{figure}

Autoencoders\index{autoencoder} employ layered feed-forward networks for unsupervised learning of an unlabeled data set of input patterns\index{input pattern}, using the inputs as targets, $\ve t^{(\mu)} = \ve x^{(\mu)}$. The layout is illustrated
in Figure \ref{fig:C9S6AE}. The network consists of two main parts, an 
{\em encoder}\index{encoder|textbf} (on the left), and a 
{\em decoder}\index{decoder|textbf} (on the right). 
The \index{encoder}encoder consists for instance of several fully connected or convolutional
layers and maps the inputs to a \index{layer!bottleneck}\index{bottleneck}bottleneck layer with a small number $M$ of neurons,
significantly smaller than the input dimension, $M \ll N$.
We denote the states of the bottleneck neurons by $z_j$. The \index{encoder}encoder corresponds to a non-linear mapping
$\ve z = \ve f_{\!\rm e}(\ve x)$. The \index{decoder}decoder
maps the \index{bottleneck}bottleneck (or {\em latent}\index{latent variable}) variables back to the inputs, $\ve x=\ve f_{\!\rm d}(\ve z)$.
One adjusts weights and \index{threshold}thresholds by backpropagation until the network learns to approximate
 the inputs as
\begin{equation}
\label{eq:ae17}
\ve x = \ve f_{\!\rm d}[\ve f_{\!\rm e}(\ve x)]\,.
\end{equation}
The energy function reads:
\index{energy function}
\begin{equation}
\label{eq:ae18}
H = \frac{1}{2} \sum_{\mu} \big|\ve x^{(\mu)} -\ve f_{\!\rm d}[\ve f_{\!\rm e}(\ve x^{(\mu)})]\big|^2\,,
\end{equation}
where $|\ve x|^2 = \ve x^{\sf T}\ve x$.  In other words, the autoencoder learns the identity function. 
The point
is that the identity is represented in terms of two non-linear functions,
the \index{encoder}encoder $\ve f_{\!\rm e}$ and the \index{decoder}decoder $\ve f_{\!\rm d}$. 
While the identity function is trivial, the encoding and decoding functions need not be. 
The \index{bottleneck}bottleneck ensures that the network does not simply learn $\ve f_{\!\rm e}(\ve x) = \ve f_{\!\rm d}(\ve x)  = \ve x$.

The latent variables $\ve z$ may encode interesting properties of the input patterns\index{input pattern!encoding}. If the number of neurons is much smaller than the number of pattern bits, as indicated by the term \index{bottleneck}bottleneck, the \index{encoder}encoder is a low-dimensional ({\em compressed}) representation
of the input data. In this way autoencoders can perform non-linear dimensionality reduction\index{dimensionality reduction}, like self-organising maps (Section \ref{sec:som}). If both \index{encoder}encoder and \index{decoder}decoder are linear functions with zero thresholds, then $H = \frac{1}{2} \sum_{\mu} |\ve x^{(\mu)} -\ma W_{\rm d}\ma W_{\rm e}\ve x^{(\mu)})|^2$. In this case,
$z_1(\ve x),\ldots, z_M(\ve x)$ are simply the first $M$ principal components
\index{principal component}   
of zero-mean input data  \cite{bourlard1988auto} 
(Exercise 10.14).

{\em Sparse autoencoders} \cite{ng2011sparse} have a large number of neurons in the \index{bottleneck}bottleneck, possibly more than the number of pattern bits. But only a small number of \index{neuron!bottleneck}\index{bottleneck}bottleneck neurons are allowed to be active at the same time. The idea is that sparse representations\index{sparse} of input data are more robust than dense ones, and generalise more reliably. At least high-dimensional but sparse representations of binary classification problems \index{classification!problem, binary}  are more likely to be linearly separable (Section \ref{sec:ct}). 
There are different ways of enforcing sparsity, for instance using \index{regularisation!L1 regularisation}\index{regularisation!L2 regularisation} $L_1$- or $L_2$-regularisation (Section \ref{sec:weightdecay}).  
An alternative \cite{ng2011sparse} is to ensure that the 
average activation of each \index{bottleneck}bottleneck
neuron with sigmoid activation function,
\index{activation function!sigmoid}
\begin{equation}
a_j = \frac{1}{p} \sum_{\mu=1}^p \sigma(b_j^{(\mu)})\,,
\end{equation}
remains small. This is achieved by adding the term 
\begin{equation}
\lambda \sum_j a\log \frac{a}{a_j} + (1-a)\log\frac{1-a}{1-a_j}
\end{equation}
\index{energy function}
to the energy function, with \index{Lagrange multiplier}Lagrange multiplier $\lambda$ (Section \ref{sec:pprid}).  This term penalises any deviation of $a_j$ from $a_j = a \ll 1$, where $a>0$ is a sparsity parameter.
Each term in the sum is non-negative and vanishes
when $a_j=a$ for all $j$, because each term can be interpreted as 
the Kullback-Leibler divergence\index{Kullback Leibler divergence} (Section \ref{sec:bm1}) between two Bernoulli distributions with parameters $a$ and $a_j$.

{\em Variational autoencoders}~\cite{kingma2014auto,doersch2016tutorial,rezende2014stochastic} have layouts similar to the one shown schematically in Figure~\ref{fig:C9S6AE}, but their purpose is quite different. 
Variational autoencoders are generative models\index{generative model} (Section \ref{sec:rbm}): just like restricted Boltzmann machines\index{Boltzmann machine} they approximate
a data distribution of inputs $\Pd(\ve x)$, and allow to sample from it.
As an example consider the  \href{http://yann.lecun.com/exdb/mnist/}{MNIST} 
data set of handwritten digits. The patterns define a data distribution that encodes the properties of the digits in terms of covariances and higher-order correlations. The question is how to generate new digits from this distribution, different from those in the data set, yet with their defining properties. 
In other words, how can a machine learn to generate images that look like
handwritten digits?

The idea of variational autoencoders\index{autoencoder}  is to represent the data distribution in terms of a Gaussian distribution $\Platent(\ve z)$ of 
latent variables\index{latent variable} $\ve z$, using the fact that one can approximate
any given data distribution $\Pd(\ve x)$ in terms of $\Platent(\ve z)$ by a suitable non-linear transformation $\ve f(\ve z)$ \cite{doersch2016tutorial}. 
Variational autoencoders are trained not unlike
neural networks, but an essential difference to the algorithms described in Part \ref{part:supervised} is that 
variational autoencoders learn probabilities rather than deterministic input-output mappings.

Given the Gaussian distribution $\Platent(\ve z)$ of the \index{latent variable}latent variables, the goal is to maximise
the log-likelihood (Section \ref{sec:rbm}):
\begin{equation}
\mathscr{L}=\log P(\ve x) = \log \!\int{\rm d}\ve z \,\pG(\ve x|\ve z) \Platent(\ve z)\,.
\end{equation}
Here $\pG(\ve x|\ve z)$ is the probability to generate $\ve x$ given $\ve z$. 
In the simplest case, this distribution 
is assumed to be Gaussian with mean $\ve \mu_P(\ve z) = \ve f (\ve z)$, and with correlation
matrix $\ma C_P{(\ve z)}$.
The \index{decoder}decoder represents these functions in terms of a multilayer perceptron
or a convolutional neural network. Weights and \index{threshold}\index{weight}thresholds are determined to maximise $\mathscr{L}$ by gradient ascent. 
To this end we must find an efficient way
of computing $\mathscr{L}$ and its gradients. One possibility is Monte-Carlo sampling (Section \ref{sec:mcs}), 
but this is not very efficient because most values of $\ve z$ drawn from $\Platent(\ve z)$ result in unlikely patterns $\ve x$, with only negligible contributions to $\mathscr{L}$. To get around this problem, one needs to know which values of $\ve z$ are likely to produce a given pattern $\ve x$. 
The idea is to learn a second approximate distribution $Q(\ve z|\ve x)$ of $\ve z$ given $\ve x$.
We can think of $Q(\ve z|\ve x)$ as an \index{encoder}encoder. So $Q(\ve x|\ve z)$ corresponds
to the encoder \index{encoder}  $\ve f_{\!\rm e}$    discussed above, while $P(\ve z|\ve x)$ corresponds to the decoder
$\ve f_{\!\rm d}$. An important difference is that $P$ and $Q$ are probabilities, not deterministic functions.

To determine a good approximation $Q(\ve z|\ve x)$, we minimise the difference between
$Q(\ve z|\ve x)$ and the unknown exact distribution $P(\ve z|\ve x)$,
\begin{equation}
\DKL[Q(\ve z{|}\ve x),P(\ve z|\ve x)] = \langle \log Q(\ve z{|}\ve x)-\log P(\ve z|\ve x)\rangle_Q\,.
\end{equation}
A first trick is to rewrite this expression using 
Bayes' theorem \cite{MathewsWalker}, 
\begin{equation}
P(\ve z|\ve x) = \pG(\ve x|\ve z)\Platent(\ve z)/P(\ve x)\,.
\end{equation}
This gives:
\begin{equation}
\label{eq:target}
\mathscr{L}-\DKL [Q(\ve z|\ve x)|P(\ve z|\ve x)] = \langle \log P(\ve x|\ve z) 
-\DKL [Q(\ve z|\ve x)|\Platent(\ve z)]\rangle_Q\,.
\end{equation}
The second trick is to note that the l.h.s of Equation (\ref{eq:target}) is a suitable target function to maximise. We want to maximise $\mathscr{L}$ subject to the constraint that the unknown function $Q(\ve z|\ve x)$ approximates the probability of $\ve z$ encoding the pattern $\ve x$.
Usually one takes $Q(\ve z|\ve x)$ to be a Gaussian with mean $\ve \mu_Q$ and 
correlation matrix $\ma C_Q$. 
The task is then to determine the functions $\ve \mu_P{(\ve z)}, \ma C_P{(\ve z)}, \ve \mu_Q{(\ve z)}$, and 
$\ma C_Q{(\ve z)}$ by adjusting the weights of two neural networks, encoder and decoder.

This task, maximising the r.h.s. of Equation (\ref{eq:target}),
is not as straightforward as it may seem, because the target function (\ref{eq:target}) involves
an average over  the distribution $Q$ which in turn depends on the weights. The question is how to move
the derivative $\partial /\partial_{w_{mn}}$ inside the average $\langle \cdots \rangle_Q$, to obtain an unbiased expression
for the weight updates $\delta w_{mn}$. In other words, the goal is to ensure that the average weight increments are proportional
to the gradients of the target function (\ref{eq:target}).
A related
problem occurs when training binary stochastic neurons (Section \ref{sec:arp}). Similarities
and differences are described in Ref.~\cite{jankowiak}.

One solution is to use {\em stochastic backpropagation}\index{stochastic backpropagation} \cite{rezende2014stochastic}. 
In its simplest form, this algorithm makes use of a relation for the gradient of the average of
  a test function $F(\ve z)$ with a Gaussian probability $Q(\ve z;w_{ij})$ that depends on the weights $w_{ij}$:
\begin{equation}
\label{eq:sbp}
\frac{\partial }{\partial w_{mn}} \langle F(\ve z)\rangle_Q
 = \langle \ve b \cdot \tfrac{\partial }{\partial w_{mn}} \ve \mu_Q
+ \frac{1}{2} \mbox{tr} \ma A \tfrac{\partial }{\partial w_{mn}}\ma C\rangle_Q\,.
\end{equation}
Here $\ve b$ and $\ma A$ are the gradient and the Hessian of the function $F(\ve z)$. 
A challenge is that these derivatives tend to be difficult to compute reliably. Suitable
approximations are described in Ref.~\cite{rezende2014stochastic}. The expression inside the average in Equation (\ref{eq:sbp}) is the unbiased weight increment. Iterating the learning rule allows one to 
determine the parameters of $P(\ve x|\ve z)$ and $Q(\ve z|\ve x)$. This allows tio efficiently sample $\ve x$ by 
sampling the latent variables and then applying the decoder.

{\em Generative adversarial networks} \cite{goodfellow2014adversarial} are generative models \index{generative model} based on learning rules similar to that described above for variational autoencoders, but there are some differences in detail.
Generative adversarial networks consist of two multilayer perceptrons, a generator and a discriminator.
The generative network produces new outputs from a given data distribution ({\em fakes}), and 
the task of the discriminator is to classify these outputs into two classes: real or fake data.
Generator and discriminator are trained together. The  weights of the generator are adjusted to maximise the \index{classification error}classification error of the discriminator, while those of the discriminator are trained to minimise this error
\cite{rocca2019understanding}.
\index{autoencoder|)}
\index{hidden layer|)}

\section{Summary}
The unsupervised-learning algorithms described in Sections \ref{sec:Ojar} and \ref{sec:cp} are based on Hebb's rule.  These algorithms can learn different features of unlabeled input data: they can detect the familiarity of inputs, perform principal-component analysis\index{principal component}, and identify \index{cluster|(}clusters in the input data.  

Self-organising  maps also rely on Hebb's rule. An important difference is that the outputs are arranged in an array, and that output neurons\index{output neuron} that are close-by in the output array are updated in similar ways. Self-organising maps can therefore represent topographic and semantic maps, where close-by or similar inputs are mapped to nearby outputs. When the dimension of the output array is much lower than the input dimension, self-organising maps perform non-linear dimensional reduction.  

Radial basis-function networks are classifiers, just like multilayer perceptrons. Their output neurons\index{output neuron} are trained in the same way, using labeled input data. However, the decision boundaries of radial basis-function networks are polynomial functions (not just hyperplanes), and their parameters are determined by unsupervised learning.  

Autoencoders are multilayer perceptrons. They can learn to encode non-linear features of unlabeled input data by using the input patterns as targets.  Finally, generative adversarial networks do not require labeled inputs, so they can be considered unsupervised-learning machines.  They are used to generate synthetic data in order to expand \index{training set}training sets for supervised learning, and pose an ethical dilemma because they can be used to generate {\em deep fakes}\index{deep fake} \cite{deepfake}, manipulated videos where someone's facial expression and speech are replaced by another person's.  

In summary, the simple algorithms described in this Chapter provide a proof of concept: how machines can learn without labels. 

\section{Further reading}
The primary source for Sections \ref{sec:Ojar} and \ref{sec:cp} is the book by Hertz, Krogh, and Palmer \cite{hertz1991introduction}.
A good reference for self-organising maps is Kohonen's book \cite{KohonenBook}.
Radial basis-function networks are discussed by Haykin
in Chapter 5 of his book \cite{Haykin}. 
It has been argued that radial-basis function networks do not generalise as well as perceptrons do \cite{Wettschereck1992}. To solve this problem, Poggio and Girosi \cite{Poggio1989} suggested to determine the parameters $\ve w_j$ of the radial basis function by supervised learning\index{supervised learning}, using stochastic gradient descent.  

Autoencoders can generate non-linear, low-dimensional representations of an \index{input distribution}input distribution. The relation to principal
component analysis is discussed in Refs.~\cite{bourlard1988auto,bourlard2000auto}.

The recommended introduction to variational autoencoders is the tutorial by Doersch \cite{doersch2016tutorial}. He also mentions that the underlying mathematics for variational autoencoders is similar to that of 
Helmholtz machines\index{Helmholtz machine} (Section \ref{sec:fr4}), 
although the two machines learn in quite different ways.

Variational autoencoders are used for a number of different purposes. Ref.~\cite{pourkamali2019effectiveness} suggests to employ a variational autoencoder for active learning\index{active learning} (Section \ref{sec:further_reading_7}). The idea is to represent the \index{input distribution}input distribution in terms of lower-dimensional \index{latent variable}latent variables, and to use $K$-means clustering (Section \ref{sec:Kmeans}) to identify groups of patterns that should be labeled. Variational autoencoders have also
been used for outlier detection \cite{eduardo2019robust} and language generation \cite{li2020optimus}.

\vfill\eject

\chapter{Reinforcement learning}
\label{ch:rl}

Supervised learning\index{supervised learning} requires labeled data, where each input comes with a target the network is supposed to learn.  Unsupervised learning\index{unsupervised learning}, by contrast, does not require labeled data.  {\em Reinforcement learning}
\index{reinforcement learning|textbf}
lies  between these extremes.  The term reinforcement\index{reinforcement} describes the principle of learning by means of a {\em reward function}. This function assigns {\em penalties}\index{penalty} 
or {\em rewards}\index{reward} to the network output, depending on how the output
relates to the learning goal.
For a neural network with a vector of outputs,
the reward function\index{reward!function|textbf} could be
\begin{equation}
\label{eq:reward_function}
   r=
\begin{cases}
    +1& \text{reward if all outputs correct\,,}\\
    -1& \text{penalty otherwise\,.}
\end{cases}
\end{equation}
The goal is to learn to produce
outputs that receive a reward more frequently than those
that trigger a penalty. We say that rewarded outputs are {\em reinforced}\index{reinforcement}.
The feedback\index{feedback|textbf} may be random, given by a distribution initially unknown to the network. 

The reward function reflects the learning goal\index{learning!goal}.The 
\index{training}training process as well as the learning outcome depend crucially on this function. Suppose one replaces
the reward function (\ref{eq:reward_function}) by the more lenient alternative: $r=1$ if at least one output is correct, and $r=-1$ otherwise. Naturally this leads to more errors, possibly not a good idea if the goal is to teach a robot to fly.

One distinguishes two different types of reinforcement problems, {\em associative}\index{reinforcement learning!associative task} and {\em non-associative}\index{reinforcement learning!non associative task} tasks \cite{williams1992simple}.
An example for a non-associative task is the $N$-armed bandit problem\index{N armed bandit problem} \cite{sutton1998reinforcement}. Imagine $N$ slot machines with different reward distributions\index{reward!distribution}, initially unknown to the player. Given a finite amount of money, the question is in which order to play the machines so as to maximise the overall profit. The dilemma is whether
to stick with a machine that yields a good reward, or whether to try out other machines that may yield a low reward initially, but could give much higher rewards eventually ({\em exploit-versus-explore} dilemma)\index{exploitation}\index{exploration}.
In this type of problem, the player receives only the reinforcement signal\index{reinforcement!signal}, no other inputs. 
In associative tasks, by contrast, the agent receives inputs, or  {\em stimuli}\index{stimulus},
and it should learn to {\em associate} with each stimulus the output that yields the highest reward. 
Such tasks occur for instance in behavioural psychology\index{psychology, behavioural}, where the problem is
to discriminate between different stimuli, and to associate the right behaviour with each stimulus.

\begin{figure}[t]
\centering
\begin{overpic}[scale=\myFigureScale]{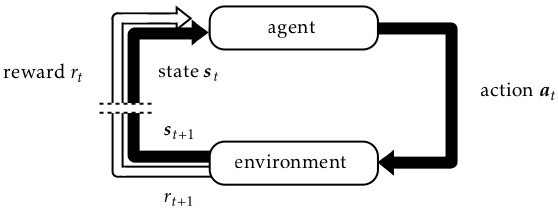}
\end{overpic}
\caption{\label{fig:C11Reinforcement1} Sequential decision process (schematic).
After Figure~3.1 in Ref.~\cite{sutton1998reinforcement}.}
\end{figure}

In general, such associative tasks can be described as {\em sequential decision processes}\index{decision process, sequential} (Figure \ref{fig:C11Reinforcement1}),
where an {\em agent}\index{agent} explores a sequence of states $\ve s_0, \ve s_1,\ve s_2,\ldots$  through a sequence of actions $\ve a_0,\ve a_1,\ve a_2,\ldots$. Consider for instance a motile microorganism in the turbulent ocean that should swim to the water surface as quickly as possible \cite{colabrese2017flow}. It 
determines its state by observing the local environment. The microorganism might measure local strain and vorticity of the flow. The environment provides a reinforcement signal\index{reinforcement!signal} (the distance\index{distance!to surface} to the surface for example), and the organism determines which action to take, given its state and the \index{reinforcement!signal}reinforcement signal. Should it turn, stop to swim, or accelerate? 
The organism learns to associate actions with certain states that maximise the reward. This sounds quite similar to associating optimal outputs with stimuli. A conceptual difference is that the action of the agent modifies the environment: its actions take it to a different place in the turbulent flow, with different vorticity and different strain. A second point is
that the reward to an action may not be immediate. In this case the challenge is to credit actions
that optimise the expected future reward, given the information collected so far. This is the {\em credit-assignment} problem\index{credit assignment problem} \cite{minsky1961steps}.

There are  two different kinds of associative tasks: {\em continuous}\index{reinforcement learning!continuous task|textbf} and
 {\em episodic}\index{reinforcement learning!episodic task|textbf} ones.
In continuous tasks, the intertwined sequences of states and actions have no natural end, so that one must either terminate the sequence in an {\em ad-hoc} fashion or introduce a weighting factor to ensure that the expected future reward remains finite.  In episodic tasks, by contrast, the learning is divided into episodes\index{episode|textbf} that terminate after a finite number of steps. An example is to learn a strategy for winning a board game.
In this case, each \index{episode}episode corresponds to a round of the game, and the reward is incurred at the end of each episode. The number of steps per round, the \index{episode!length}episode length $T$, may vary from round to round. In order to estimate the expected reward one usually needs many \index{episode}episodes.  \index{reward!expected}
A second, very simple example is the stimulus problem described above, where the states (stimuli) are independent from the actions.  Each \index{episode}episode consists of only one step, so that $T=1$. In response to a randomly chosen state $\ve s_0$, the agent learns to perform the action $\ve a_0$ that maximises the immediate reward\index{reward!immediate|textbf}. This can be achieved
by the \index{associative reward penalty algorithm}associative reward-penalty algorithm (Section \ref{sec:arp}). It uses stochastic neurons with weights that are trained by gradient ascent\index{gradient ascent} to maximise the expected immediate reward. 

To estimate the expected future reward\index{reward!future} when $T>1$, one must use a different method, usually
{\em temporal difference learning}\index{learning!temporal difference}. It allows
to estimate the expected future reward, after $T$ steps,
by breaking up the learning into time steps $t=1,\ldots,T$. The idea is
that it is better to adjust the prediction of a future reward as one iterates, 
rather than waiting for $T$ iterations before updating the prediction. 
In temporal difference learning one expresses  the reward at time $T$ in terms of differences at time steps $t+1$ and $t$ \cite{sutton1988learning}. 

Temporal difference learning builds up a lookup table\index{lookup table} that summarises the best actions for each state, the {\em $Q$-table}\index{Q learning!Q table|textbf} $Q(\ve s,\ve a)$. Given $N_{\rm s}$ states $\ve s$ and $N_{\rm a}$ possible actions $\ve a$, $Q(\ve s,\ve a)$ 
is a $N_{\rm s}\times N_{\rm a}$ table. Its elements contain the expected future reward for each state-action pair.  To implement temporal difference learning, one must adopt a {\em policy}\index{policy} or {\em strategy}\index{strategy}. It specifies for each state which action is taken. In total there are $N_{\rm a}^{N_{\rm s}}$ possibilities of assigning actions to states.  When there are many states and actions it quickly becomes impossible to determine the best strategy by simple sampling, because there are too many 
to consider.  

The advantage of temporal difference learning and related algorithms is that they do not rely on the evaluation of all possible policies. Instead, the policy is updated using iterated estimates of the $Q$ table. We write
$Q_t$ for the estimate at time step $t$.
There are different ways of deriving a policy from a $Q$-table.
The greedy policy \index{policy!greedy|textbf} is a {\em deterministic}\index{policy!deterministic} policy, it corresponds to choosing the action that corresponds to the largest $Q$-element in a given row of the $Q$-table:  $\ve a=\mbox{argmax}_{\footnotesize \ve a'} Q_t(\ve s, \ve a')$.\index{argmax} This policy maximises the current estimate of the future reward. 

{\em Stochastic}\index{policy!stochastic} policies are often better, in 
particular for non-stationary\index{environment!non-stationary} or stochastic environments\index{environment!stochastic}, 
because they allow the agent to explore potentially better alternatives. Also, for a deterministic environment\index{environment!deterministic}, a deterministic policy may lead to cycles\index{cycle}. This can be avoided with a stochastic policy\index{stochastic policy}. 
One example for a stochastic policy is the $\varepsilon$-greedy policy\index{policy!$\varepsilon$ greedy|textbf}. With probability $1-\varepsilon$, it chooses the greedy action\index{action!greedy} $\ve a=\mbox{argmax}_{\footnotesize \ve a'} Q_t(\ve s, \ve a')$, but 
with a small probability $\varepsilon$ it takes a suboptimal action\index{action!suboptimal}. Another example is the softmax policy\index{policy!softmax}, 
where argmax is replaced by the softmax function (Section \ref{sec:ocf}). The softmax policy can handle actions described
in terms of continuous variables.  

In general the policy can change as the algorithm is iterated. A  common choice
is to reduce the parameter $\varepsilon$ in the $\varepsilon$-greedy policy 
as one iterates,
so that the algorithm converges to the optimal deterministic policy.\index{policy!deterministic}

{\em $Q$-learning}\index{Q learning} is an approximation to 
the temporal difference algorithm. In $Q$-learning, the $Q$-table\index{Q learning!Q table}\index{Q learning} is updated
assuming that the agent always follows the greedy policy, even though it might actually follow a different policy. 
$Q$-learning allows agents to learn to play strategic games  \cite{alphago}. A simple example 
is the game of tic-tac-toe\index{tic tac toe} (Section \ref{sec:ql}). Games such as chess or go require to keep track of a very large number of states, so large that $Q$-learning in its simplest form becomes impractical.
An alternative is to represent the state-action mapping by a deep neural network\index{deep network} \cite{mnih2015human}.

\section{Associative reward-penalty algorithm}
\index{associative reward penalty algorithm|(}
\label{sec:arp}
The associative reward-penalty algorithm\index{associative reward penalty algorithm} 
uses {\em stochastic neurons}\index{neuron!stochastic} that  are trained to maximise the average immediate reward. 
\index{reward!immediate}
In Chapters  \chref{C5Chapter} to \ref{ch:rn} the output neurons
were deterministic functions of their inputs. 
For reinforcement learning, by contrast, it is better to use stochastic neurons.  
The idea is the same as in Chapters \ref{sec:shn} and \ref{chapter:so}:  stochastic neurons can explore a wider range of possible states which may in the end lead to a better solution. The state $y_i$ of neuron $i$ is given by the {\em stochastic update rule}  \eqnref{C3S2StochasticRule}:
\index{update rule!stochastic}
\begin{equation}
 \eqnlab{C11S1StochasticRule}
   y_{i}=
\begin{cases}
    +1& \text{with probability }\pr(b_{i})\,,\\
    -1& \text{with probability }1-\pr(b_{i})\,,
\end{cases}
\end{equation}
where $b_i = \ve w_{i}\cdot \ve x$ is the \index{local field}local field (no thresholds), and $\pr(b) = (1+e^{-2\beta b})^{-1}$. Recall that the 
parameter $\beta^{-1}$ is the noise level\index{noise level}.  Since the outputs can assume only two values, $y_i=\pm 1$, Equation~\eqnref{C11S1StochasticRule} describes a {\em binary} stochastic neuron.\index{neuron!stochastic|textbf}

\begin{figure}[t]
\begin{tabular}{lll}
\hline\hline
\small $\prew$  & \small $y\!=\!-1$ & \small $y\!=\!+1$\\\hline
\small ${\ve x^{(1)}} =[1,0]^{\sf T}$ & \small 0.6 & \small 0.8\phantom{${A^A}^B$}\\
\small ${\ve x^{(2)}} = [1,1]^{\sf T}$ & \small 0.3 & \small 0.1\\
\hline\hline
\end{tabular}\\[5mm]
\begin{overpic}[scale=\myFigureScale]{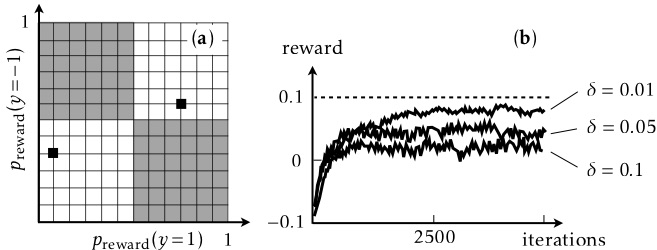}
\end{overpic}
\caption{\label{fig:C12ARP} Conditioning by reward  \cite{barto1985learning}. 
A stochastic neuron responds to stimuli\index{stimulus} ${\ve x^{(1)}}$ and ${\ve x^{(2)}}$
with different outputs,  $y=\pm 1$ and receives the reward (\ref{eq:reward_function}): $r=+1$ with 
probability $\prew(\ve x,y)$, and $r=-1$ with probability
$1-\prew(\ve x,y)$. The goal is to always respond with
the output that maximises the expected reward.
Table: \index{reward!distribution|textbf}reward distribution.
Panel ({\bf a}): contingency space\index{contingency space} \cite{barto1985learning} of the problem,
representing each input $\ve x$ in a plane with coordinates  $\prew(\ve x,+1)$ and $\prew(\ve x,-1)$.
 ({\bf b}) Reward versus iteration number of the \index{associative reward penalty algorithm}associative reward-penalty rule 
(\ref{eq:RLdelta2}). Schematic, based on simulations by Phillip Graefensteiner averaged
over $100$ independent realisations. 
} 
\end{figure}
To illustrate the \index{associative reward penalty algorithm}associative reward-penalty algorithm for a single binary stochastic neuron, consider
an agent experiencing different stimuli\index{stimulus} $\ve x$ drawn with equal probability from a distribution of inputs. 
Upon receiving stimulus $\ve x$, the stochastic neuron outputs either $y=1$  or
$y=-1$. Given $\ve x$ and $y$, the environment provides a stochastic reward\index{reward!stochastic} $r(\ve x,y)=\pm 1$ drawn from a {\em reward distribution}\index{reward!distribution|textbf} $\prew(\ve x,y)$:
\begin{equation}
r(\ve x,y) = \begin{cases} +1 & \mbox{with probability $\prew(\ve x,y)$},\\
-1 & \mbox{with probability $1-\prew(\ve x,y)$.}
\end{cases}
\end{equation}
The goal is to adjust the weights so that the neuron produces outputs that are rewarded with high probability. 
Figure \ref{fig:C12ARP}({\bf a}) shows an example with just two stimuli, 
${\ve x^{(1)}} = [1,0]^{\sf T}$ and ${\ve x^{(2)}} = [1,1]^{\sf T}$.  
The numerical values of $\prew(\ve x,y)$ indicate that the expected reward is maximised
when the neuron outputs $y=1$ in response to ${\ve x^{(1)}}$, and $y=-1$ in response to ${\ve x^{(2)}}$. 
Since ${\ve x^{(1)}}$ and ${\ve x^{(2)}}$ occur with equal probability, the maximal expected reward\index{reward!maximal} is
\begin{equation}
\label{eq:rmax}
 r_{\rm max} =\tfrac{1}{2}\big[\langle r({\ve x^{(1)}},+1)\rangle_{\rm reward}+\langle r({\ve x^{(2)}},-1)\rangle_{\rm reward}\big]
=0.1\,.
\end{equation}
Here we used $\langle r(\ve x,y)\rangle_{\rm reward}=\prew(\ve x,y)-[1-\prew(\ve x,y)]$, as well as the numerical values for $\prew(\ve x,y)$ given in Figure \ref{fig:C12ARP}({\bf a}).

Figure~\ref{fig:C12ARP} ({\bf b})  shows the {\em contingency space}\index{contingency space|textbf}  \cite{barto1985learning} of the problem, representing the inputs $\ve x$ in a plane with coordinates  $\prew(\ve x,+1)$ and $\prew(\ve x,-1)$. It is easier to learn to associate the correct output with inputs that lie in the shaded regions where $\prew(\ve x,+1)>\tfrac{1}{2}$ and $\prew(\ve x,-1)< \tfrac{1}{2}$, or vice versa. In this case one can solve the problem by fixing $y=+1$ and sampling $\prew(\ve x,+1)$ for all $\ve x$. If $\prew(\ve x,+1)>\tfrac{1}{2}$ then $y=+1$ is the optimal output for $\ve x$, otherwise it is $y=-1$. This strategy cannot be used outside the shaded region. For example, if both \index{reward!probability|textbf} reward probabilities are larger than one half, it is necessary to sample both  $\prew(\ve x,-1)$ and  $\prew(\ve x,+1)$ sufficiently often in order to determine which one is 
larger: one must find {\em the greater of two goods} according to Barto \cite{barto1985learning}. 
This illustrates the fundamental dilemma of reinforcement learning: an output
that appears at first to yield a high reward may not be the optimal one in the long run. To find the optimal output it is necessary to estimate both \index{reward!probability}reward probabilities precisely. This means that one must try all possible outputs frequently, not only the one that appears to be optimal at the moment.

To derive a learning rule\index{learning rule} we need a cost function.\index{cost function}
One possibility is to use the average of the immediate reward for a given stimulus $\ve x$,\index{reward!immediate}
\begin{equation}
\label{eq:r_av}
\langle r\rangle  =
\sum_{y_1=\pm 1,\ldots,y_M=\pm 1} \big\langle r(\ve x,\ve y) P(\ve y| \ve x)\big\rangle_{\rm reward}\,.
\end{equation}
Here $\langle \cdots\rangle_{\rm reward}$ is an average
over the response of the environment,
determined by the reward distribution $\prew(\ve x,\ve y)$. It is assumed that the reward distribution 
is stationary.
Furthermore,
\begin{equation}
P(\ve y| \ve x) =
\prod_{i=1}^M\begin{cases}
\pr(b_i) &\mbox{for} \quad y_i =1\,,\\
1-\pr(b_i)&\mbox{for} \quad y_i =-1\end{cases}
\end{equation}
is the probability that the network produces the output $\ve y = [y_1,\ldots,y_M]^{\sf T}$ given
the \index{local field}local field $b_{i} = \sum_{j}w_{ij}x_{j}$.

To find the maximum of $\langle r\rangle$ one uses gradient ascent\index{gradient ascent} on $\langle r\rangle$, analogous to maximising the log-likelihood for Boltzmann machines (Section \ref{sec:bm1}), and to gradient descent on the energy function for perceptrons in supervised learning (Chapter \chref{C6Chapter}). 
\index{energy function}
The gradient is computed by applying the chain rule\index{chain rule}, as usual.
The calculation is similar to the one for Boltzmann machines (Chapter \ref{chapter:so}).
After some algebra (Exercise~{11.2}) one finds
for given $\ve x$ and $\ve y$ that the derivative of $P(\ve y| \ve x)$ with
respect to $w_{mn}$ equals  $P(\ve y| \ve x)\, \beta [y_m- \tanh(\beta b_m)]x_n$.
We conclude that
\begin{equation}
 \frac{\partial \langle r\rangle}{\partial w_{mn}} = 
\beta  \big\langle  r(\ve x,\ve y) \big[y_{m}-\tanh(\beta b_m)\big] \big\rangle x_{n}
\label{eq:RLdelta0av}
\end{equation}
with $b_m = \sum_j w_{mj} x_j$, as before. The average is taken over the output of the network and over the reward distribution,
just as in Equation (\ref{eq:r_av}).

Now we seek a learning rule that increases the expected immediate reward $\langle r\rangle$. 
In other words, we require that the \index{weight increment}weight increment $\delta\!w_{mn}$ is an unbiased\index{unbiased estimator
estimator (Section \ref{sec:ae})} of the gradient of the expected immediate reward \cite{williams1992simple}:\index{reward!immediate}
\begin{equation}
\langle \delta\! w_{mn} \rangle = \eta \frac{\partial \langle r\rangle}{\partial w_{mn}}\,.
\end{equation}
Comparison with Equation (\ref{eq:RLdelta0av}) leads to:
\begin{equation}
\delta w_{mn} = \alpha r [y_m-\mbox{tanh}(\beta b_m) ]x_n\,,
\label{eq:RLdelta0}
\end{equation}
with $\alpha = \eta\beta$.
This learning rule belongs
to a set of more general rules derived by Williams \cite{williams1992simple}.
It is plausible that the rule (\ref{eq:RLdelta0}) converges to a steady state, because
the \index{weight increment}weight increments approach zero as the network 
learns to produce the output $\max_{\small\ve y}\{\prew(\ve x,\ve y)\}$, independent of $\ve y$, so that $\ve y-\langle \ve y\rangle$ averages to zero. But there is no proof of convergence.\index{convergence!proof}

An alternative is the {\em associative reward-penalty}\index{rule!associative reward penalty} rule \cite{barto1985learning}:
\begin{equation}
   \delta \!w_{mn}= \alpha
\begin{cases}\phantom{-\delta}\,
    \big[y_{m}-\tanh(\beta b_m)\big] x_{n} &\text{for}\quad r=+1\,,\\
    -\delta \big[y_{m}+\tanh(\beta b_m)\big] x_{n} &\text{for}\quad r=-1\,,
\end{cases}
\label{eq:RLdelta2}
\end{equation}
with  $0 < \delta \ll  1$.  For  $r=1$, the learning rules (\ref{eq:RLdelta0}) and (\ref{eq:RLdelta2})
give the same \index{weight increment}weight increment, but for $r=-1$ the increments are different.
With rule (\ref{eq:RLdelta2}) the agent learns primarily from positive feedback\index{feedback}. One advantage
of this asymmetric rule\index{learning rule} is that it can be proven to converge in the limit of 
$\delta\to 0$ \cite{barto1985learning}.
In general, however, the \index{convergence!slow}convergence becomes quite slow when $\delta$ is small.
Figure~\ref{fig:C12ARP}({\bf c}) shows simulation results for the immediate
reward, averaged over $100$ independent realisations of the learning process,  
versus the iteration number of the rule (\ref{eq:RLdelta2}).
We see that the average immediate reward approaches a steady state. The 
steady-state average of the immediate reward is smaller than $r_{\rm max}=0.1$,
but as expected it approaches $r_{\rm max}$ as $\delta$ decreases.

The averaged learning curves\index{learning!curve|textbf} still exhibit substantial fluctuations\index{fluctuations}. They reflect
significant variations within and between individual realisations.
Furthermore, the \index{convergence!proof}convergence proof assumes that the input patterns\index{input pattern!linearly independent} are linearly independent. This means that the number of patterns cannot exceed the input dimension $N$. Associative reinforcement problems with linearly dependent inputs can be solved 
by embedding\index{embedding} the input patterns in a higher-dimensional input space\index{input space} (Section \ref{sec:ct}).

The \index{associative reward penalty algorithm}associative reward-penalty rule illustrates how an agent can use a \index{reinforcement!signal}reinforcement signal to maximise the expected immediate reward. The algorithm could for instance be  a model for how an animal learns to respond in different ways to different stimuli. 

Yet there are  many problems where the reward is not immediate. When we play chess, 
the reward comes at the end of the game, for example $r=+1$ if we won, $r=-1$ if we lost, and $r=0$
if the game ended in a draw.
More generally, an agent navigating a complex environment should not only consider immediate rewards, but also how a certain  action may affect possible future rewards. One way of estimating future rewards for such tasks is temporal difference learning, discussed next.
\index{associative reward penalty algorithm|)}

\section{Temporal difference learning}
\label{sec:TDL}
How does temporal difference learning\index{learning!temporal difference|textbf}  allow an agent to optimise its expected future reward?  
For an episodic task\index{reinforcement learning!episodic task}, given an \index{episode}episode with $T$ steps, the agent visits the finite sequence of states  $\ve s_0,\ldots,\ve s_{T-1}$, and collects the rewards $r_1,\ldots, r_T$. The future reward\index{reward!future|textbf}  is defined as 
\begin{equation} \label{eq:edfr0} R_t= \sum_{\tau=t}^{T-1} r_{\tau+1}\,.  
\end{equation} 
Continuous tasks\index{reinforcement learning!continuous task}, by contrast, do not have defined end points. Since the sum in (\ref{eq:edfr0}) might diverge as $T\to\infty$, it is customary to introduce a weighting factor $0 \leq \gamma \leq 1$ in the sum over rewards: 
\begin{equation} \label{eq:edfr1} R_t= \sum_{\tau=t}^{\infty} \gamma^{\tau-t} r_{\tau+1}\,.  
\end{equation} 
The weighting factor\index{factor, weighting}  reduces the contribution of the far future to the estimate.  
Smaller values of $\gamma$  give more weight to the immediate future,
and the limit $\gamma \to 0^+$ corresponds to $R_t = r_{t+1}$. The sum in Equation (\ref{eq:edfr1}) is called  {\em future discounted reward}\index{reward!discounted}. 

We use a neural network with input $\ve s_t$ to estimate  $R_t$.  In general, the network output is a non-linear function of the inputs, parameterised by weights that could be arranged into several layers of hidden neurons\index{hidden neuron} \index{hidden layer}
(Part \ref{part:supervised}).  The simplest choice is to use a single 
linear unit\index{unit!linear}, just as in Equation \eqnref{C5S5Output}:
\begin{equation}
\label{eq:linear_model}
O(\ve s_t) = \ve w\cdot \ve s_t\,.
\end{equation}
The components $w_j$ of the weight vector $\ve w$ are determined so that the network output $O(\ve s_t)$ approximates $R_t$.  
This can be achieved by minimising the energy function \index{energy function}
\begin{equation}
\label{eq:Htdl}
H = \tfrac{1}{2}\sum_{t=0}^{T-1} [R_t-O(\ve s_t)]^2
\end{equation}
using gradient descent\index{gradient descent}. The corresponding learning rule reads:     
\begin{equation}
\label{eq:gdr2}
\delta w_m = \alpha \sum_{t=0}^{T-1} [R_t-O(\ve s_t)]\, \frac{\partial O}{\partial w_m}\,.
\end{equation}
\begin{figure}[t]
\centering
\begin{overpic}[scale=\myFigureScale]{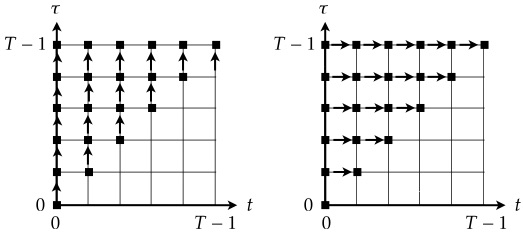}
\end{overpic}
\caption{\label{fig:double_sum} The double
sum in Equation (\ref{eq:ds_11}) extends over the terms indicated in black. 
The corresponding terms can be summed in two ways, as illustrated in the two panels, for $T=6$.}
\end{figure}
The idea of temporal difference learning \cite{sutton1988learning} is to express the error $R_t-O(\ve s_t)$ as
a sum of temporal differences:\index{difference, temporal} 
\begin{align}
R_t -O(\ve s_t) 
= \sum_{\tau=t}^{T-1} [r_{\tau+1}+O(\ve s_{\tau+1})-O(\ve s_\tau)]\,,
\end{align}
where $O(\ve s_T)$ is defined to be zero, $O(\ve s_T)\equiv 0$.
Using the gradient-descent rule (\ref{eq:gdr2}) 
one obtains
\begin{align}
\label{eq:ds_11}
\delta\!\ve w = \alpha \sum_{t=0}^{T-1}\sum_{\tau=t}^{T-1}
 [r_{\tau+1}+O(\ve s_{\tau+1})-O(\ve s_\tau)] \ve s_t\,.
\end{align}
The terms in this double sum 
can be summed in a different way, as 
illustrated in Figure \ref{fig:double_sum}:
\begin{align}
\delta\!\ve w = \alpha \sum_{\tau=0}^{T-1}\sum_{t=0}^{\tau}
 [r_{\tau+1}+O(\ve s_{\tau+1})-O(\ve s_\tau)] \ve s_t\,.
\end{align}
Exchanging the summation variables and introducing
a weighting factor $0\leq \lambda \leq 1$ gives:
\begin{align}
\delta\ve w= \alpha\sum_{t=0}^{T-1}
[r_{t+1}+O(\ve s_{t+1})-O(\ve s_t)]  \sum_{\tau=0}^t \lambda^{t-\tau} \ve s_\tau\,.
\end{align}
The purpose of the weighting factor\index{factor, weighting} is to reduce the weight of past states in the sum \cite{Szepesvari2010}.
Alternatively one may update $\ve w$ (and hence $O$) at each time step, with increment \cite{sutton1988learning}
\begin{equation}
\label{eq:TDdeltaw}
\delta \!\ve w_{\!t} = \alpha [r_{t+1}+O({\ve w_t,}\ve s_{t+1})-O({\ve w_t,}\ve s_t)] \sum_{\tau=0}^{t} \lambda^{t-\tau}
\ve s_\tau\,.
\end{equation}
This is the temporal-difference learning rule, also called TD($\lambda$) 
\index{learning rule!TD($\lambda$)}
\cite{Szepesvari2010}.
Temporal difference learning allows a machine to learn the board game backgammon \cite{TDGammon,mcclelland2015explorations}, using a deep layered network\index{network!layered}\index{deep network} and backpropagation\index{backpropagation} (Section \ref{sec:chainrule}) to determine the weights.
\begin{figure}[t]
\centering
\begin{overpic}[scale=\myFigureScale]{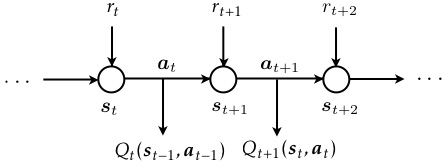}
\end{overpic}
\caption{\label{fig:TD} Sequence of states $\ve s_t$ in sequential reinforcement learning\index{reinforcement learning!sequential}. The action $\ve a_t$ leads from $\ve s_t$ to $\ve s_{t+1}$ where the agent receives reinforcement $r_{t+1}$. The $Q$-table with elements $Q(\ve s_t,\ve a_t)$ estimates the future discounted reward.   }
\end{figure}

The rule TD($0$)\index{learning rule!TD($0$)}  is similar to the learning rule \eqnref{C6S2SynapticWeightsOutput} with target $r_{t+1}+O({\ve w_t,}\ve s_{t+1})$. It allows to learn one-step prediction of the \index{time series!prediction of}time series.  \index{prediction, one step}
Using Equation (\ref{eq:linear_model}), we see that the TD($0$)-learning rule corresponds to the following learning rule for 
the output $O$:
\begin{align}
\label{eq:TDP}
O_{t+1}(\ve s_t) = O_t(\ve s_t)+ \alpha[r_{t+1}+O_t(\ve s_{t+1})-O_t(\ve s_t)]\,.
\end{align}
The subscript $t$ in  $O_t$ emphasises that the output function is updated iteratively .
The learning rule (\ref{eq:TDP}) applies to estimating the future reward (\ref{eq:edfr0}) for episodic tasks. If the environment is stationary, one may average over many consecutive \index{episode}episodes, using the final weights from \index{episode}episode $k$ 
as initial weight values for episode $k+1$.
For continuous tasks, the corresponding rule for estimating the future discounted  reward (\ref{eq:edfr1}) reads:
\begin{align}
\label{eq:TDPgamma}
O_{t+1}(\ve s_t) = O_t(\ve s_t)+ \alpha[r_{t+1}+\gamma O_t(\ve s_{t+1})-O_t(\ve s_t)]\,.
\end{align}
Returning to the problem outlined in the beginning of this Chapter, consider an agent exploring a complex environment. The task might be to get from location $A$ to location $B$ as quickly as possible, or expending as little energy as possible. At time $t$ the agent is at position $\ve x_t$ with velocity $\ve v_t$. These variables as well as the local state of the environment\index{environment}  are summarised in the state vector\index{state!vector} $\ve s_t$. Given $\ve s_t$, the agent can act in certain ways: it might for example slow down, speed up, or turn. These possible actions are summarised in a vector $\ve a_t$. At each time step, the agent takes the action $\ve a_t$ that optimises the expected future discounted reward (\ref{eq:edfr1}), given its present state $\ve s_t$.  The estimated expected future reward for any state-action  pair\index{state!state action pair} is summarised in a table: the {\em $Q$-table}\index{Q learning!Q table} with elements $Q_{t}(\ve s_t,\ve a_t)$ is the analogue of $O_t(\ve s_t)$. Different rows of the $Q$-table correspond to different states, and different columns to different actions. \index{action}
The TD($0$) rule for the $Q$ table reads:
\begin{align}
\label{eq:sarsa}
Q_{t+1}(\ve s_t, \ve a_t)&= Q_{t}(\ve s_t, \ve a_t)
+\alpha_t \big[r_{t+1} +\gamma Q_t(\ve s_{t+1}, \ve a_{t+1})-Q_t(\ve s_t, \ve a_t)\big]\,.
\end{align}
This algorithm is called SARSA, because one needs $\ve s_t, \ve a_t, r_{t+1}, \ve s_{t+1}$,\index{learning rule!SARSA}
and $\ve a_{t+1}$ to update the $Q$-table (Figure \ref{fig:TD}). 
A difficulty with the rule  
(\ref{eq:sarsa}) is that it depends not only on the present state-action pair  $[\ve s_t,\ve a_t]$,
but also on the next action $\ve a_{t+1}$, and thus indirectly upon the policy. Sometimes this is indicated by writing
$Q_\pi$ for the $Q$-table given policy\index{policy} $\pi$.

\begin{algorithm}[t]
\caption{\label{q_learning}\index{Q learning|textbf} $Q$-learning for episodic task with the $\varepsilon$-greedy policy}
\begin{algorithmic}
\STATE initialise  $Q$;
\FOR {$k = 1,\ldots, K$}
\STATE initialise  $\ve s_0$;
\FOR {$t=0,\ldots,T_k-1$}
   \STATE choose $\ve a_t$ from $Q(\ve a_t, \ve s_t)$ according to $\varepsilon$-greedy policy;
   \STATE compute $\ve s_{t+1}$ and record $r_{t+1}$;
   \STATE update $Q(\ve s_{t}, \ve a_{t}) \leftarrow
   Q(\ve s_{t}, \ve a_{t}) + \alpha [r_{t+1} + \max_{\footnotesize\ve a}
Q(\ve s_{t+1}, \ve a)-Q(\ve s_t, \ve a_t)]$;
   \ENDFOR
\ENDFOR
\end{algorithmic}
\end{algorithm}
\section{$Q$-learning}
\label{sec:ql}
The $Q$-learning rule\index{Q learning|(} \cite{watkins1989} is an approximation to Eq.~(\ref{eq:sarsa}) that 
does not depend on  $\ve a_{t+1}$.  Instead
one assumes that the next action, $\ve a_{t+1}$, is the optimal one:
\begin{align}
\label{eq:qlearning}
 Q_{t+1}(\ve s_t, \ve a_t)&= Q_{t}(\ve s_t, \ve a_t) + \alpha_t \big[r_{t+1} 
+\gamma \max_{\footnotesize \ve a} Q_t(\ve s_{t+1}, \ve a)-Q_t(\ve s_t, \ve a_t)\big]\,,
\end{align}
regardless of the policy that is currently followed. 
Although Equation (\ref{eq:qlearning}) does not refer to any policy, 
the learning outcome nevertheless depends on it, because the policy determines the sequence of states and actions $[\ve s_t,\ve a_t]$.
For the greedy policy,\index{policy!greedy}
Eq.~(\ref{eq:qlearning})  is equivalent to (\ref{eq:sarsa}),
but in general the two algorithms differ, and \index{convergence!SARSA}\index{convergence!Q learning}converge to different solutions.
While the $Q$-table converges to the expected future reward of the greedy policy in $Q$-learning, 
for SARSA it converges to the expected future reward corresponding to the policy used in \index{training}training.
This can be an advantage when performance during
training is important, for example when training an expensive robot that should not crash too often, 
or for a small bird that learns flying by doing.  $Q$-learning is simpler, and can be used for problems 
where the final strategy based on $\mbox{argmax}_{\ve a'} Q(\ve s,\ve a')$ matters, but where
the reward during training is less important. Examples are board games where only the quality of the final
strategy counts, not how often one loses during training. In summary, $Q$-learning is simpler, and if one takes $\varepsilon\to 0$ during training, it yields the optimal strategy as well as SARSA, but it may give lower rewards during \index{training}training.

The $Q$-learning algorithm is summarised in Algorithm \ref{q_learning}. 
Usually one sets the initial entries in the $Q$-table to large positive values ({\em optimistic} initialisation),
\index{initialisation!optimistic of Q table} because this prompts the agent to explore many different actions, at least in the beginning.
If the agent is in state $\ve s_t$, it chooses the action $\ve a_t$ 
from $Q_{t}(\ve s_t,\ve a_t)$ according to the given policy. For
the $\varepsilon$-greedy policy, for example, 
the agent picks a random action from the corresponding row\index{policy!$\varepsilon$ greedy|(}
of the $Q$-table with probability $\varepsilon$.
With  probability $1-\varepsilon$
it chooses the action $\ve a_t$ that yields the largest\footnote{If several elements in the relevant row have the same maximal value then any one of them is chosen with equal probability.} $Q_{t}(\ve s_t,\ve a_t)$ given $\ve s_t$.
The choice of action $\ve a_t$ determines the next state $\ve s_{t+1}$, and this in turn allows to update the $Q$-table: given the new state $\ve s_{t+1}$ resulting from the action $\ve a_t$, 
one updates $Q_{t}(\ve s_t,\ve a_t)$ using Equation~(\ref{eq:qlearning}).
For episodic tasks one puts $\gamma=1$ in (\ref{eq:qlearning}), and one averages over many \index{episode}episodes using the outcome $Q_{T_k}$ from episode $k$ as initial condition for the $Q$-table for episode $k+1$.
Each new episode can start with a new initial state $\ve s_0$. It helps the exploration process if $\ve s_0$ is one
of the states that are rarely visited by the learning algorithm.

When the sequence $\ve s_0,\ve s_1, \ve s_2,\ldots$ is a Markov chain\index{Markov chain} (Section \ref{sec:mcs}), then the $Q$-learning algorithm can be shown \cite{watkins1992q,Szepesvari2010} to converge \index{convergence!Q learning}
if one uses a time-dependent learning rate $\alpha_t$\index{learning rate} that satisfies 
\begin{equation} \sum_{t=0}^\infty\alpha_t=\infty\quad\mbox{and}\quad\sum_{t=0}^\infty \alpha_t^2 < \infty\,.  
\end{equation} 
Often $Q$-learning is implemented in combination with the $\varepsilon$-greedy policy. This policy shares
an important  property with the \index{associative reward penalty algorithm}associative reward-penalty algorithm with stochastic neurons: 
stochasticity allows for a wider range of responses, some of which may turn out beneficial in the long run. When $\varepsilon$ is very small, the agent picks the action that appears optimal. As a consequence, suboptimal $Q$-elements are sampled less frequently and are therefore subject to larger errors.  Therefore it is advantageous to begin with a relatively large value of $\varepsilon$.
It is customary to decrease $\varepsilon$ as the algorithm is iterated, because this accelerates \index{convergence!Q learning}convergence to the greedy policy.

It is important to bear in mind that the learning outcome depends on the reward function, as mentioned
above. 
In general it is a good idea to analyse how the optimal strategy changes as one varies the reward function. Sometimes we are faced with the inverse problem:  
consider how a microorganism\index{microorganism} swimming in the turbulent ocean responds to different stimuli. How was this behaviour shaped by genetic evolution\index{evolution, genetic}? Which quantity was optimised? Is it most important to reduce the
energy cost for propulsion\index{propulsion}? Or is it more important to avoid predation\index{predation}?

Another challenge is to determine suitable states and actions. An agent navigating a complex environment may have a continuous range of positions and velocities, and may experience continuous-valued signals from the environment. To represent the corresponding states in a $Q$-table it is necessary to discretise. To this end one must determine suitable ranges and resolutions of these variables, and for the actions. If there are too many states and actions, $Q$-learning becomes inefficient.
This is referred to as the {\em curse of dimensionality}\index{curse of dimensionality} \cite{bellman1957dynamic,sutton1998reinforcement}.

Let us see how $Q$-learning  works for a very simple example, for
the associative task\index{reinforcement learning!associative task} described in Fig.~\ref{fig:C12ARP}. 
One \index{episode}episode corresponds to computing the output of the neuron given its initial
state, so $T=1$. There is no sequence of states, and the task
is to estimate the immediate reward\index{reward!immediate}. In this case the learning rule 
(\ref{eq:qlearning}) simplifies to
\begin{align}
\label{eq:QLsimple}
\delta Q(\ve s, a)&= \alpha\big[ r(\ve s,a) -Q(\ve s, a)\big]\,.
\end{align}
Since each episode consists only of a single time step, we dropped the subscript $t$. Also, the term $\max_{\footnotesize \ve a} Q(\ve s_{t+1}, \ve a)$ from Equation (\ref{eq:qlearning})
does not appear in (\ref{eq:QLsimple}) since $Q$ estimates the immediate reward. 
There are only two states in this problem, $\ve s={\ve x^{(1)}}$ and $\ve s={\ve x^{(2)}}$,
and the possible actions are $a=\pm 1$.
In other words,  $N_{\rm s}=N_{\rm a}=2$ in this case.
In each round, one of the states is chosen randomly, with equal probability.
The action is determined from the current estimate of the immediate reward as
${\rm argmax}_a Q(\ve s,a)$ with probability $1-\varepsilon$, and uniformly randomly otherwise.
These steps are  iterated
over many iterations (\index{episode}episodes), using the outcome of episode $k$ as initial condition
for episode $k+1$.
The rule (\ref{eq:QLsimple}) describes exponential relaxation\index{relaxation, exponential} to the target\index{target} for small learning rate $\alpha$. 
In this limit, Equation (\ref{eq:QLsimple}) is approximated by the stochastic differential equation \index{differential equation, stochastic}
\begin{align} \label{eq:deq}
\tfrac{\rm d}{{\rm d}{k}}Q (\ve s,a)= \alpha\, f_\varepsilon(\ve s,a)\,[ r(\ve s,a) -Q_{k}(\ve s, a)]\,,
\end{align}
where $f_\varepsilon(\ve s,a)$ is the stationary frequency with which the state-action pair\index{state!state action pair} $[\ve s,a]$ is visited using the $\varepsilon$-greedy policy:
\begin{align}
\label{eq:feps}
f_\varepsilon(\ve s,a) =\frac{1}{N_{\rm s}} 
\begin{cases} 
1-\varepsilon +\tfrac{\varepsilon}{N_{\rm a}} & \mbox{if $a = \mbox{argmax}_{a'}Q_{k}(\ve s,a')$}\,,\\ 
\tfrac{\varepsilon}{N_{\rm a}} & \mbox{otherwise.}
\end{cases} 
\end{align}
The frequency is normalised to unity, $1=\sum_{s,a} f_\varepsilon(\ve s,a)$.
Averaging the solution 
of Equation (\ref{eq:deq}) with initial condition
 $Q_0(\ve s,a)=1$ over the reward distribution gives:\index{reward!distribution}
\begin{equation}
\label{eq:Qsol11}
\langle Q_{k}(\ve s,a)\rangle = {\rm exp}[-\alpha f_\varepsilon(\ve s,a) {k}]
+ \alpha f_\varepsilon (\ve s,a) \int_0^{{k}}\!{\rm d}{{k}'}\, \langle r(\ve s,a)\rangle 
{\rm exp}[f_\varepsilon(\ve s,a) \alpha ({k}'-{k})] \,.
\end{equation}
\begin{figure}
\centering
\begin{overpic}[scale=\myFigureScale]{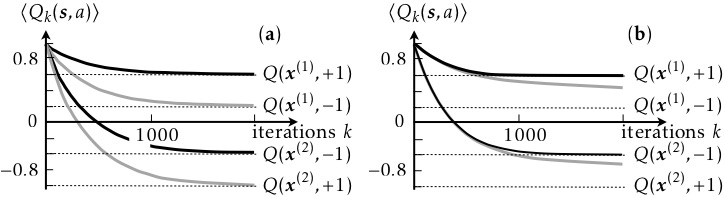}
\end{overpic}
\caption{\label{fig:C12QL} 
$Q$-learning for the task described in Figure~\ref{fig:C12ARP}.
({\bf a}) 
Entries of the $Q$-table versus the number of iterations of Equation (\ref{eq:QLsimple}) for $\alpha=0.01$ and $\varepsilon=1$.  ({\bf b}) Same, but for $\varepsilon = 0.05$.  Schematic, based on simulations by Navid Mousavi, averaged over $5000$ independent realisations of the learning curve\index{learning!curve}.  } 
\end{figure}
For $\varepsilon >0$, $Q_t(\ve s,\ve a)$ \index{convergence!Q learning}converges on average to 
\begin{align}
\begin{bmatrix}
Q^\ast({\ve x^{(1)}},-1)& Q^\ast({\ve x^{(1)}},+1)  \\
Q^\ast({\ve x^{(2)}},-1)& Q^\ast({\ve x^{(2)}},+1)
\end{bmatrix}
=
\begin{bmatrix}
0.2&0.6   \\
-0.4&-0.8
\end{bmatrix}\,.
\end{align}
Here we used that $\langle r(\ve x,y)\rangle = 2\prew(\ve x,y)-1$, as well the \index{reward!probability}reward probabilities in Figure~\ref{fig:C12ARP}. Figure \ref{fig:C12QL} illustrates how the rate of convergence\index{convergence!rate} depends on the value of the parameter $\varepsilon$. For $\varepsilon=1$, all state-action pairs are visited and evaluated equally often, independently of the present elements of the $Q$-table. 
Equations (\ref{eq:feps}) and (\ref{eq:Qsol11}) show that all elements of the $Q$-table converge to their\index{steady state}
steady-state values at the same rate, equal to $\tfrac{\alpha}{4}$.
For small values of $\varepsilon$, by contrast, 
the algorithm tends to take optimal actions, $\mbox{argmax}_{a'}Q_{k}(\ve s,a')$. Therefore it
finds the optimal elements of the $Q$-table more quickly, at the rate $\tfrac{\alpha}{2}$.
However, the other elements \index{convergence!slow}converge much more slowly. 
Initially, the theory (\ref{eq:Qsol11}) does not apply because optimal and suboptimal $Q$-elements
in each row of the $Q$-table are not well separated. As a result the decay rates of all elements are similar
at first. But once optimal and suboptimal elements are significantly different, the suboptimal
ones decay at the rate $\varepsilon\alpha/4$, as predicted by the theory.

This example illustrates the strength of $Q$-learning with the $\varepsilon$-greedy policy.\index{policy!$\varepsilon$ greedy}
For small values of $\varepsilon$, the algorithm tends to converge to the optimal strategy
more quickly than a brute-force algorithm that visits every state-action pair equally often.
Equation (\ref{eq:feps}) shows that this advantage becomes larger for larger
values of $N_{\rm a}$. The price one pays is that the suboptimal entries of the $Q$-table converge \index{convergence!slow}
more slowly.  

The gain is even more significant for episodic tasks with $T>1$ steps. 
In this case the learning rule for  the $Q$ table depends
on $\mbox{argmax}_{\footnotesize \ve a'} Q_{t}(\ve s, \ve a')$. 
As mentioned above, there are $N_{\rm a}^{N_{\rm s}}$  ways in which 
the largest elements can be distributed over the rows of the $Q$-table. To find the optimal deterministic
strategy by complete enumeration\index{enumeration, complete}, 
one must 
run a large number of \index{episode}episodes
for each of the $N_{\rm a}^{N_{\rm s}}$ possibilities,
to find out which one gives the largest reward.
This is impractical 
when either $N_{\rm a}$ or $N_{\rm s}$ or both become too large. 
$Q$-learning with a small value of $\varepsilon$, by contrast, 
tends to explore actions that appear to yield the largest 
expected future reward given the current estimates of the $Q$-values. 
Using experience in this way allows the algorithm to simultaneously
improve its policy and the $Q$-values towards optimality.            
As a consequence, $Q$-learning
can find optimal, or at least good strategies 
when complete enumeration of all possibilities fails.

\begin{figure}
\mbox{}
\begin{overpic}[scale=\myFigureScale]{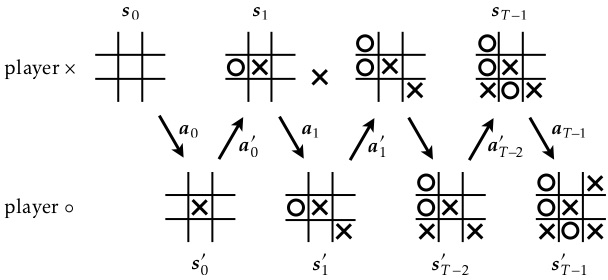}
\end{overpic}
\caption{\label{fig:C12TTT} Tic-tac-toe. Two players,  $\times$ and $\large \circ$, take turns in placing a piece on an empty field of $3\times 3$ board. The goal is to be the first to complete a row, column, or diagonal consisting of three of one's own pieces. In the example shown, player $\times$ starts and ends up winning the game. 
The states encountered by player $\times$ are denoted by $\ve s_t$, those encountered by player $\circ$ by $\ve s_t'$. Their actions are denoted by $\ve a_t$ and $\ve a_t'$.}
\end{figure}

A second example for $Q$-learning is illustrated in Figure \ref{fig:C12TTT}, the board game 
tic-tac-toe\index{tic tac toe|textbf|(}. It is a very simple game where two players take turns in placing their pieces
on a $3\times 3$ board. The player who manages to first obtain three pieces in a row, column, or diagonal wins and receives the reward $r=+1$. A draw gives $r=0$, and the player receives $r=-1$ when the round is lost.
The goal is to win as often as possible, to maximise the expected future reward. However, there is 
a strategy for both players to ensure that they do not lose. If both players
try to maximise their expected future reward, then they end up following this strategy. As a consequence, every game must end in a draw \cite{crowley1993flexible}. As a result, the game is quite boring.

Nevertheless it is instructive to ask how the players can learn to find this strategy using $Q$-learning with the $\varepsilon$-greedy policy. 
\index{policy!$\varepsilon$ greedy}
To this end we let two agents
play many rounds against each other. 
The \index{state!space}state space is the collection of all board configurations. 
Player $\times$ starts, and thus always sees a board with an even number of pieces, while the number of pieces is odd for player~$\circ$. Since the players encounter different sets of states, each must keep track of their own $Q$-table. The task is episodic, and the number $T$ of steps may vary from round to round. Feedback is only obtained at the end of each round. 

We use Equation (\ref{eq:qlearning}) with a constant learning rate $\alpha$.
We can set $\gamma=1$ since the number of steps in each round is finite.
The $Q$-table\index{Q learning!Q table} is a  $2\times n$ table where each entry is a $3\times 3$ array. Here $n$ is the number of states the player encountered so far. The first row lists the states, each a $3\times 3$ array  with entries $-1$ ($\circ$), $1$ ($\times$), or $0$ (empty). The second row contains the $Q$-values. Given a certain state of the board, a player can play a piece on any empty field. The corresponding estimate of the expected future reward is stored in the corresponding $3\times 3$ array in the second row. Since one cannot
place a piece onto an occupied field, the corresponding entries in the
$Q$-table  are assigned NaN. 

During a round, the $Q$-tables of the players are updated in turns, always the one of the player who places a piece. After a round, the $Q$-tables of both players are updated. During the first round, the elements of 
$Q$-tables encountered are initialised to zero and remain zero. 
The first change to $Q$ occurs in the last step of this round. If player $\times$ wins, for example,
the element $Q_{T-1}(\ve s_{T-1},a_{T-1})$ corresponding to the state-action pair that led to the final state $\ve s_{T-1}'$ is updated for the winning player, and $Q_{T-2}(\ve s_{T-2}',\ve a_{T-2}')$  is set to $-1$ for player $\circ$.

Both players follow the $\varepsilon$-greedy policy
\index{policy!$\varepsilon$ greedy|)}. With probability $1-\varepsilon$ they take the optimal move (if the maximal $Q$-element in the relevant row is degenerate, then one of the maximal elements is chosen randomly). With probability $\varepsilon$, a random action is chosen.
As the players continue to play rounds against each other, the rewards spread to other
elements of $Q$. Suppose that the state $\ve s_{T-1}$ is encountered once more,
the one that allowed player $\times$ to win the first round with $\ve a_{T-1}$. Then the term ${\rm max}_{\scriptstyle a}Q_{T-1}(\ve s_{T-1},\ve a)$ causes a $Q$-element for the previous state to change, the one from which $\ve s_{T-1}$ was reached the second time.
However as times goes on, this process slows down, because later updates are multiplied with higher
powers of the learning rate\index{learning rate} $\alpha$. Also, if the opponent lost in the previous round, it will try different actions that may block winning moves for the other player.
\begin{figure}\centering
\begin{overpic}[scale=\myFigureScale]{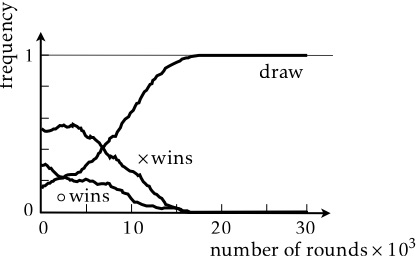}
\end{overpic}
\caption{\label{fig:C12TTTconv} Learning curves\index{learning!curve} for two players learning
to play tic-tac-toe with $Q$-learning and the $\varepsilon$-greedy policy.
Shown are the frequencies that the game ends in a draw, that player $\times$
wins, and that player $\circ$ wins. 
Similar curves are obtained using a learning rate $\alpha=0.1$. The parameter $\varepsilon$ was equal to unity
for the first $10^4$ rounds, and then decreased by a factor of $0.9$ after each 100 rounds,
and averaging each curve over a running window of $30$ rounds. Schematic, based on simulations performed by Navid Mousavi.}
\end{figure}

Figure \ref{fig:C12TTTconv} 
illustrates how the players learn, after playing many rounds against each other. 
Since both players try to maximise their expected future reward in the steady state\index{steady state} of the $Q$-learning algorithm, all games
end in a draw in this case.
The corresponding $Q$-tables contain the strategies\index{strategy} each player should adopt to maximise their reward.
Suppose player $\circ$ places the first piece as shown below:
\begin{equation}
\label{eq:Q2}
\begin{overpic}[scale=0.35]{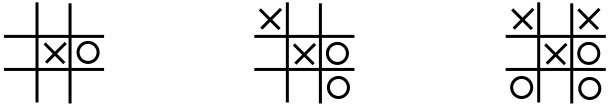}
\put(-35,8){\footnotesize board}
\put(-35,-13){\footnotesize $Q$-table}
\end{overpic}
\mbox{}\hspace*{-7.9cm}
\raisebox{-1cm}{$
\begin{bsmallmatrix}
1.00 & 1.00 &1.00\\[0.1cm]
0.34 & {\rm NaN}  & {\rm NaN}\\[1mm]
1.00 & 1.00 & 1.00
\end{bsmallmatrix}
\quad\quad
\begin{bsmallmatrix}
{\rm NaN} &  -0.57 & 1.00\\[1mm]
-0.69 &  {\rm NaN} & {\rm NaN}\\[1mm]
-0.59 & -0.73  &  {\rm NaN}
\end{bsmallmatrix}
\qquad
\begin{bsmallmatrix}
{\rm NaN} & 1.00 & {\rm NaN}\\[0.1cm]
-0.86 & {\rm NaN} & {\rm NaN}\\[1mm]
{\rm NaN} & 0.027 &{\rm NaN}
\end{bsmallmatrix}\,.$}
\end{equation}
How does this game continue?
There are several different ways in which player $\times$ may try to win.
The left $Q$-table in Equation (\ref{eq:Q2})
shows that 
one possibility is to place the piece in the top or bottom row, because this creates the opportunity of creating a {\em bridge}\index{bridge} in the next move, a configuration that cannot be blocked by the opponent, allowing $\times$ to win. 
The right $Q$-table shows that player $\times$ could still lose or end up with a draw if he makes the wrong move. The corresponding $Q$-entries have not quite converged to $-1$ and $0$, respectively. $Q$-entries corresponding to suboptimal states are not estimated as precisely because
they are visited less frequently. Here $\varepsilon=0.3$ was chosen quite large. 
Smaller values of $\varepsilon$ give even less accurate estimates for the 
suboptimal $Q$-elements  
compared with those in Equation (\ref{eq:Q2}), after \index{training}training for the same number of rounds.

As pointed out above, the learning outcome depends on the reward function\index{reward!function}. 
If one increases reward for winning, to $r=+2$ for instance, the optimal strategy appears 
to be to take turns in winning. The same learning outcome is
 expected if one imposes a penalty for a draw, $r=+1$ (win), $r=-1$ (draw, lose), 
Exercise 11.8.
More examples of reinforcement-learning problems in robotics and in the natural sciences are described in Ref.~\cite{cichos2020machine}.\index{robotics}

The $Q$-learning algorithm described above is quite efficient
when the number of states and actions is not too large. For very large $Q$-tables, the
algorithm becomes quite  slow. In this case it may be more efficient to replace
the $Q$-table by an approximate $Q$-function that maps states to actions. As explained
in Section \ref{sec:TDL}, one can use a neural network to represent the $Q$-function \cite{mnih2015human}
({\em deep reinforcement learning})\index{reinforcement learning!deep}.
An application of this method is \href{https://deepmind.com/research/alphago/}{AlphaGo}, a machine-learning algorithm that learnt to play the game of go \cite{alphago}.  The $Q$-function is represented in terms of a convolutional neural 
network\index{convolutional network}. This makes it possible to use a variant of $Q$-learning despite the fact that the number of states is enormous. 

In recent years many proof-of-principle studies have demonstrated the possibilities of reinforcement learning in a wide range of scientific problems. Recent advances in deep reinforcement learning hold promise for the future,
for real-world control problems in the engineering sciences.

\section{Summary}
Reinforcement learning lies between unsupervised learning (Chapter \ref{ch:uhl}) and supervised learning (Chapters  \chref{C5Chapter} to \ref{ch:rn}). In reinforcement learning, there are no labeled data sets. Instead, the neural network or agent learns through feedback from the environment in the shape of a reward or a penalty. The goal is to find a strategy that maximises the expected reward.  Reinforcement learning is applied in a wide range of fields, from psychology to mechanical engineering, using a large variety of algorithms.  The \index{associative reward penalty algorithm}associative reward-penalty algorithm and many versions of temporal difference learning were originally formulated using neural networks.  $Q$-learning is an approximation to temporal difference learning for sequential decision processes. In its simplest form it does not rely on neural networks.  However, when the number of states and actions is large this algorithm becomes slow. 
In this case it may be  more efficient to approximate the $Q$-function by a neural network.

\section{Further reading}
The standard reference for reinforcement learning is {\em Reinforcement learning: an introduction} by Sutton and Barto \cite{sutton1998reinforcement}.
The original reference for the convergence\index{convergence!Q learning} of the $Q$-learning algorithm is Ref.~\cite{watkins1992q}. 
A more mathematical introduction to reinforcement learning is given in Ref.~\cite{Szepesvari2010}. Examples for reinforcement learning in statistical and non-linear
physics are summarised in Ref.~\cite{cichos2020machine}.

An open question is when and how \index{symmetry}{\em symmetries}\index{Q learning!symmetry} can be exploited to simplify a reinforcement problem. For a small microorganism learning to navigate a turbulent flow, some aspects are discussed in Ref.~\cite{Lihao}, but little is known in general. 
Another open question concerns the \index{convergence!Q learning}convergence of the $Q$-learning algorithm. Convergence to the optimal policy is assured if the sequence of states is a Markov chain\index{Markov chain}. However, most real-world problems are not Markovian, so that \index{convergence!Q learning}convergence is not guaranteed. The algorithm appears to perform well nevertheless (a recurring theme in this book), but it is an open question under which circumstances
it may fail.\index{Q learning|)}

\vfill\eject

\cleardoublepage
\nobibliography*

\cleardoublepage
{\em Congratulations, you reached the end of this book. This is the end of the first episode. You are rewarded by +1.  Please multiply everything you have learned by $\alpha=0.01$ and add to your knowledge. Then start reading from the first page to start a new episode.}
\end{document}